\documentclass[10pt,twocolumn,letterpaper]{article}

\usepackage[pagenumbers]{cvpr} 

\definecolor{cvprblue}{rgb}{0.21,0.49,0.74}
\usepackage[pagebackref,breaklinks,colorlinks,allcolors=cvprblue]{hyperref}
\usepackage[accsupp]{axessibility}
\usepackage{tcolorbox} 
\usepackage{enumitem} 
\usepackage{xcolor} 
\usepackage{adjustbox}

\def\paperID{31} 
\def\confName{CVPR}
\def\confYear{2026}

\title{ProductConsistency: Improving Product Identity Preservation in Instruction-Based Image Editing via SFT and RL}

\author{
Mukund Khanna\\
Fractal Analytics\\
{\tt\small mukund.khanna@fractal.ai}
\and
Raj Singh Yadav\\
Fractal Analytics\\
{\tt\small raj.yadav@fractal.ai}
\and
Kunal Singh\\
Fractal Analytics\\
{\tt\small kunal.singh@fractal.ai}
}

\begin{document}
\maketitle
\begin{abstract}
Recent advances in instruction-based image editing have enabled models to perform complex visual edits from natural language instructions. However, in product-centric scenarios where preserving product features, branding, and textual elements are critical, current open and closed source models often struggle to maintain this fine-grained object identity. This issue is further compounded by the lack of datasets for instruction-based product image editing with text fidelity constraints, leaving it largely treated as an implicit capability of instruction-based image editing models.

In this work, we introduce the \textbf{ProductConsistency} dataset which is designed to improve product-centric image editing. Our approach includes a supervised fine-tuning (SFT) dataset of 87k samples for product editing, a reinforcement learning (RL) dataset with 869 unique product images, and a new benchmark dataset, the \textbf{ProductConsistency Benchmark}, to allow rigorous and standardized evaluation of editing models. To guide RL training, we propose a \textbf{Cyclic Consistency reward} that enforces semantic preservation of product identity by using caption similarity between the original product description and captions generated from the edited image. We fine-tune both Qwen-Image-Edit-2511 and Flux.1-Kontext-dev using our dataset and demonstrate consistent improvements over baseline models in OCR and Perceptual metrics, and MLLM-based evaluations as well, indicating stronger product consistency, text rendering, and overall visual quality; with the Qwen-Image-Edit-2511 model achieving a \textbf{5× reduction in the character error rate}. The code and pipeline is available at \href{https://anonymous.4open.science/r/ProductConsistency-6FCC/README.md}{code}.
\end{abstract}    
\section{Introduction}
\label{sec:intro}

Diffusion models have made substantial progress in text-to-image generation and, more recently, instruction-based image editing. Systems have evolved from early mask-based pipelines to instruction based frameworks that enable more fine-grained control. The Initial diffusion-based editors were derived directly from their text-to-image counterparts and relied on inversion\cite{dong2023prompt_inversion,li2023stylediffusion_inversion,mokady2023null_inversion} and spatial masking for localized edits\cite{1rombach2022highdiffusionmodel,ju2024brushnet,brooks2023instructpix2pix}. Subsequent approaches introduced instruction-guided image editing, both with and without explicit spatial masks \cite{ju2024brushnet,brooks2023instructpix2pix,blackforestlabs_flux1_fill_dev_2024_blx}, significantly improving usability and generalization. More recently, the integration of vision language models and transformer-based architectures \cite{wu2025qwen,cai2025hidream,blackforestlabs_flux1_fill_dev_2024_blx,liu2025step1x} has further improved instruction following, global structure preservation, and overall visual quality. As a result, these models are increasingly being incorporated into real-world workflows that demand high-quality image output.

\begin{figure*}[t]
  \centering
  \renewcommand{\arraystretch}{1.0}
  \setlength{\tabcolsep}{1pt}

  \begin{tabular}{cccc}
    \toprule
    \textbf{Input Image} & \textbf{HiDream-E1-1} & \textbf{Qwen-Image-Edit-2511} & \textbf{NanoBanana} \\
    \midrule

    \includegraphics[width=0.24\linewidth]{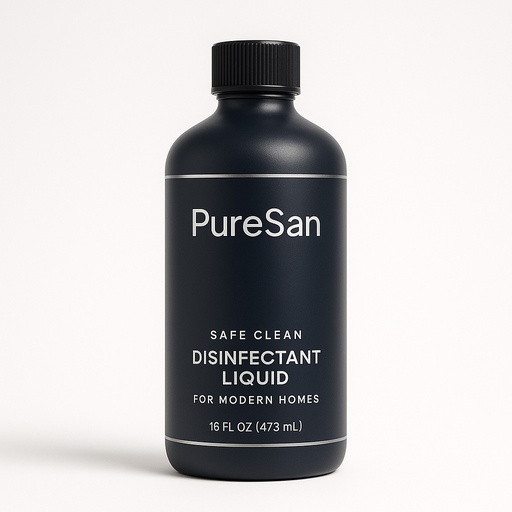} &
    \includegraphics[width=0.24\linewidth]{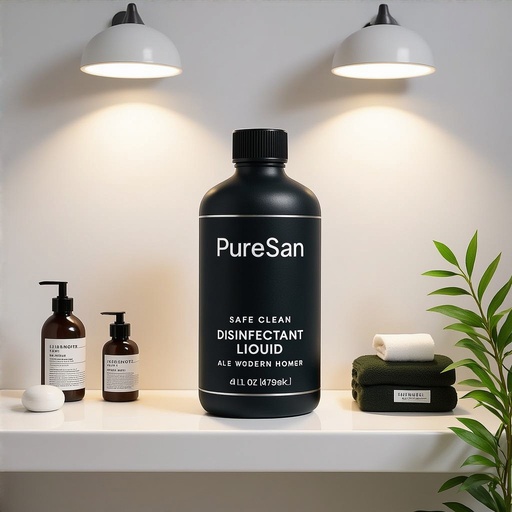} &
    \includegraphics[width=0.24\linewidth]{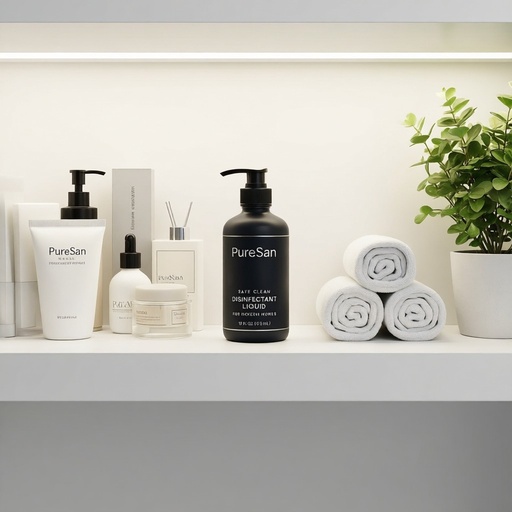} &
    \includegraphics[width=0.24\linewidth]{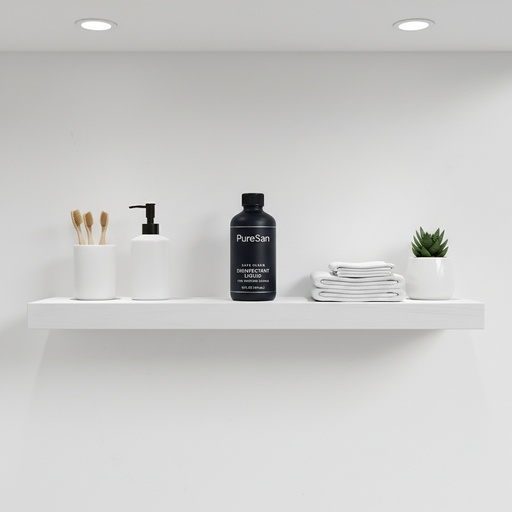} \\

    \includegraphics[width=0.24\linewidth]{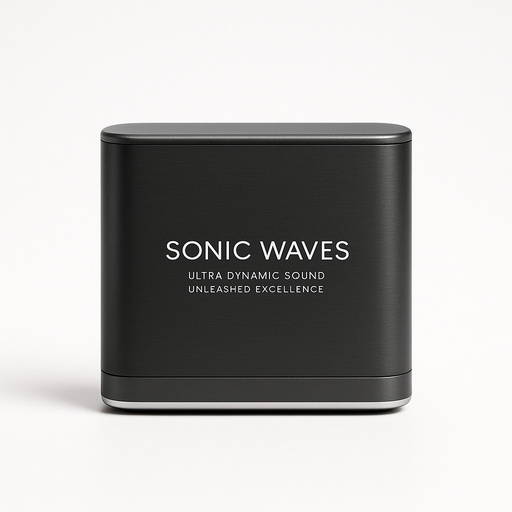} &
    \includegraphics[width=0.24\linewidth]{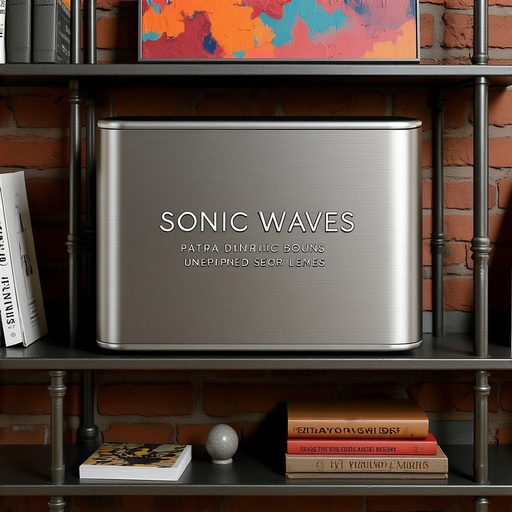} &
    \includegraphics[width=0.24\linewidth]{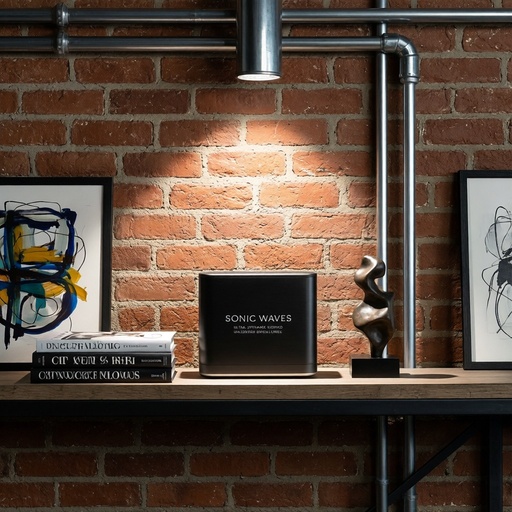} &
    \includegraphics[width=0.24\linewidth]{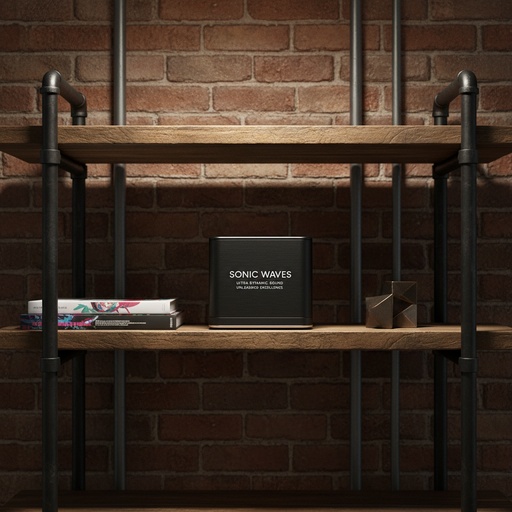} \\

    \includegraphics[width=0.24\linewidth]{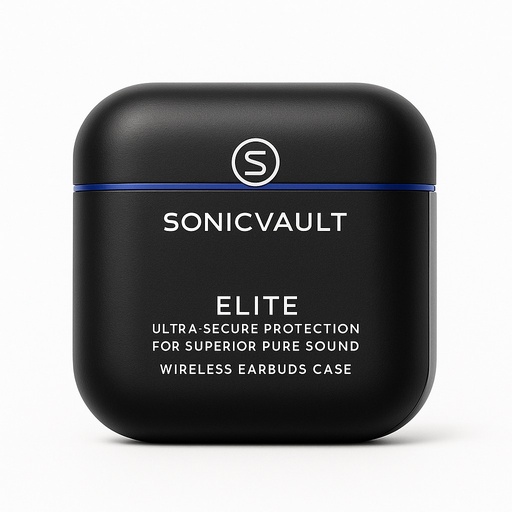} &
    \includegraphics[width=0.24\linewidth]{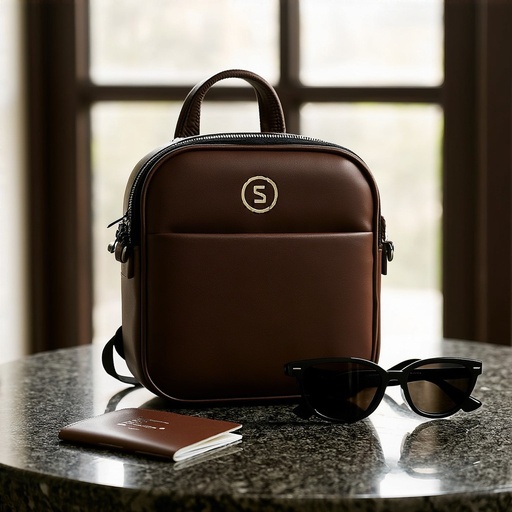} &
    \includegraphics[width=0.24\linewidth]{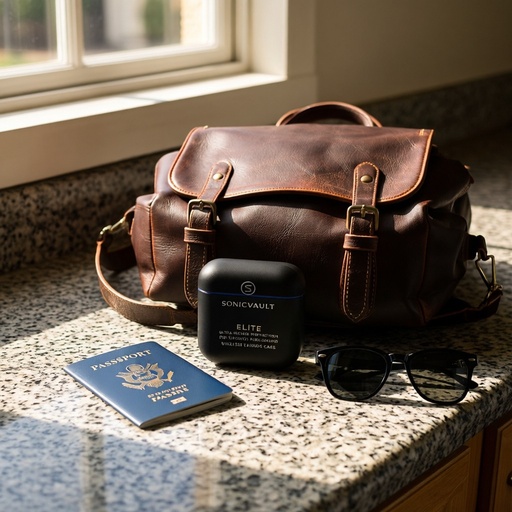} &
    \includegraphics[width=0.24\linewidth]{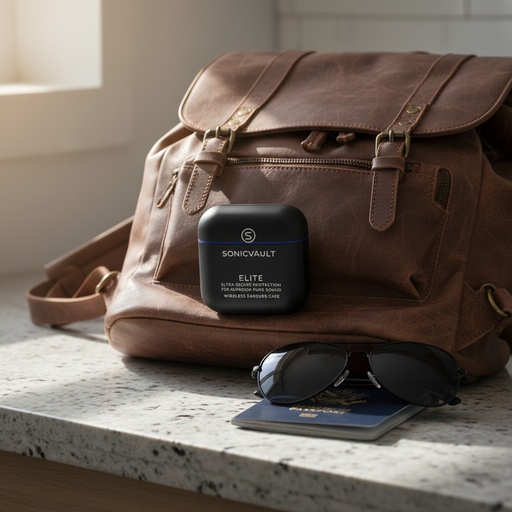} \\

    \bottomrule
  \end{tabular}

    \caption{Qualitative comparison showing that HiDream‑E1‑1, Qwen‑Image‑Edit-2511, and Nano Banana all struggle with product consistency and accurate text rendering across different product categories. Common failure modes include altered product shapes, incorrect or distorted text, extra hallucinated text, altered colors, and inconsistent branding. The prompts used for the edits from top to bottom are: (a) Display the bottle on a minimalist white bathroom shelf among neatly arranged personal care items, illuminated by bright, even overhead lighting with soft shadows emphasizing surface textures; subtle greenery from a nearby plant maintains a clean and sophisticated aesthetic. (b) Feature the speaker on an industrial-style bookshelf in a chic loft setting with an exposed brick wall and metal piping that provide a textured backdrop; abstract art books and a small sculpture flank the speaker, while directional spotlighting highlights its brushed metal finish to balance urban and artistic elements. (c) Showcase the case in a travel setting placed on a dark leather backpack on a granite countertop; include a passport and sunglasses as supporting props, with soft morning light filtering through a window to produce gentle highlights and shadows, conveying a chic, ready-for-adventure mood.}
  \label{fig:hidream_qwen_grid}
\end{figure*}

Despite these advances, current image editing models remain ill-suited for production settings that require strict visual correctness. In domains such as advertising, e-commerce, and product marketing, edited images must be perfect: even minor errors in branding elements, product geometry, logos, or generated text render an image unusable. As we can see in Figure \ref{fig:hidream_qwen_grid}, existing diffusion-based editors frequently struggle to preserve such fine-grained features during editing. They often distort or hallucinate text on objects, subtly alter brand logos, or modify product geometry when performing otherwise reasonable edits. Even closed source models struggle with this task and often render text with spelling errors and/or hallucinated text. A key reason is the absence of product-centric datasets and benchmarks that explicitly target brand consistency and text consistency during image editing, making it difficult to study, diagnose, or systematically improve these failure modes.

Recent works have made progress on related problems, but do not directly address product and brand consistency. Text-focused generative models \cite{wu2025qwen,labs2025flux} improve the legibility of newly generated text, while reasoning-based editors \cite{li2025editthinker,li2025reflect,zou2025beyond_MURE,qu2025replan,yin2025reasonedit,zhang2025r_genie} introduce multi-step planning and reflection to better follow complex instructions. However, neither line of work explicitly targets the preservation of existing on-object text or brand identity during editing. Reference-guided and adapter-based methods \cite{zhang2023adding_controlnet,li2024controlnet++,mou2024t2i_adapter,ye2023ip,zhang2025easycontrol,qin2023unicontrol} improve global visual consistency by conditioning on auxiliary inputs, but they still struggle with pixel-level fidelity and textual consistency on real product images, as they enforce spatial constraints between input and output. Reinforcement learning approaches  \cite{he2025tempflowgrpo,liu2025flowgrpo,li2025mixgrpo} align models with human preferences using learned reward models, yet these reward models are designed for general-purpose instruction following and aesthetic quality, and do not directly capture product consistency. As a result, current models frequently alter or hallucinate branding elements and text when applied to realistic product scenes. Existing benchmarks ignore rendered text, or focus on document-style layouts rather than photographed objects, leaving the most critical failure modes for product editing largely unmeasured \cite{collins2022abo,bai2020products,liang2025evaluation_hfpc}.

\begin{figure*}[t]
\centering

\begin{minipage}{0.85\textwidth}
\centering
\includegraphics[width=\linewidth]{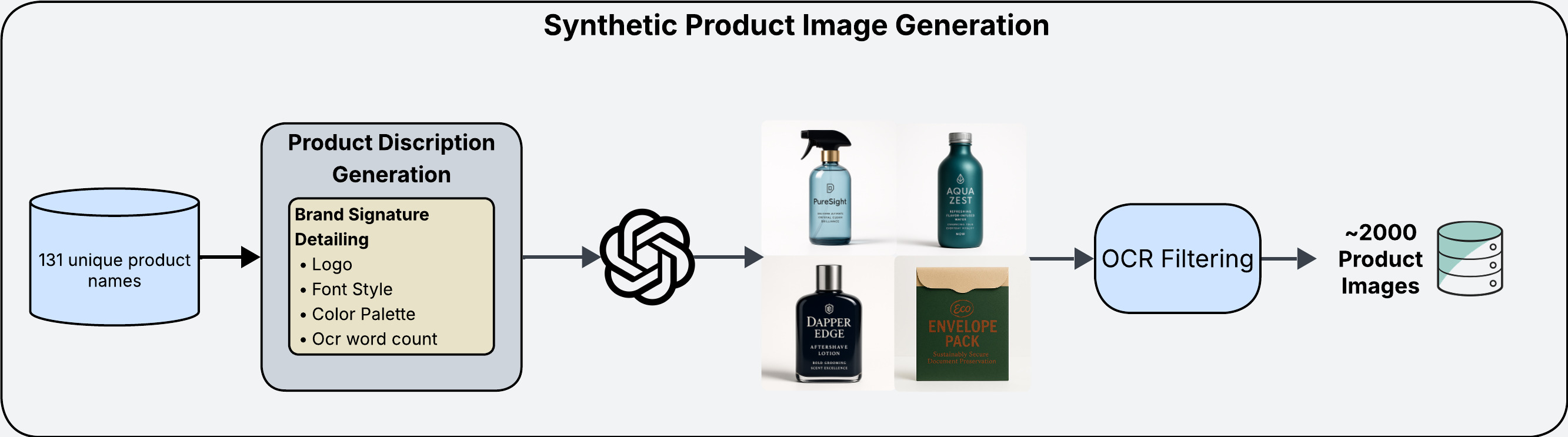}

{\footnotesize (a) Synthetic product image generation.}
\end{minipage}

\vspace{8pt}

\begin{minipage}{0.85\textwidth}
\centering
\includegraphics[width=\linewidth]{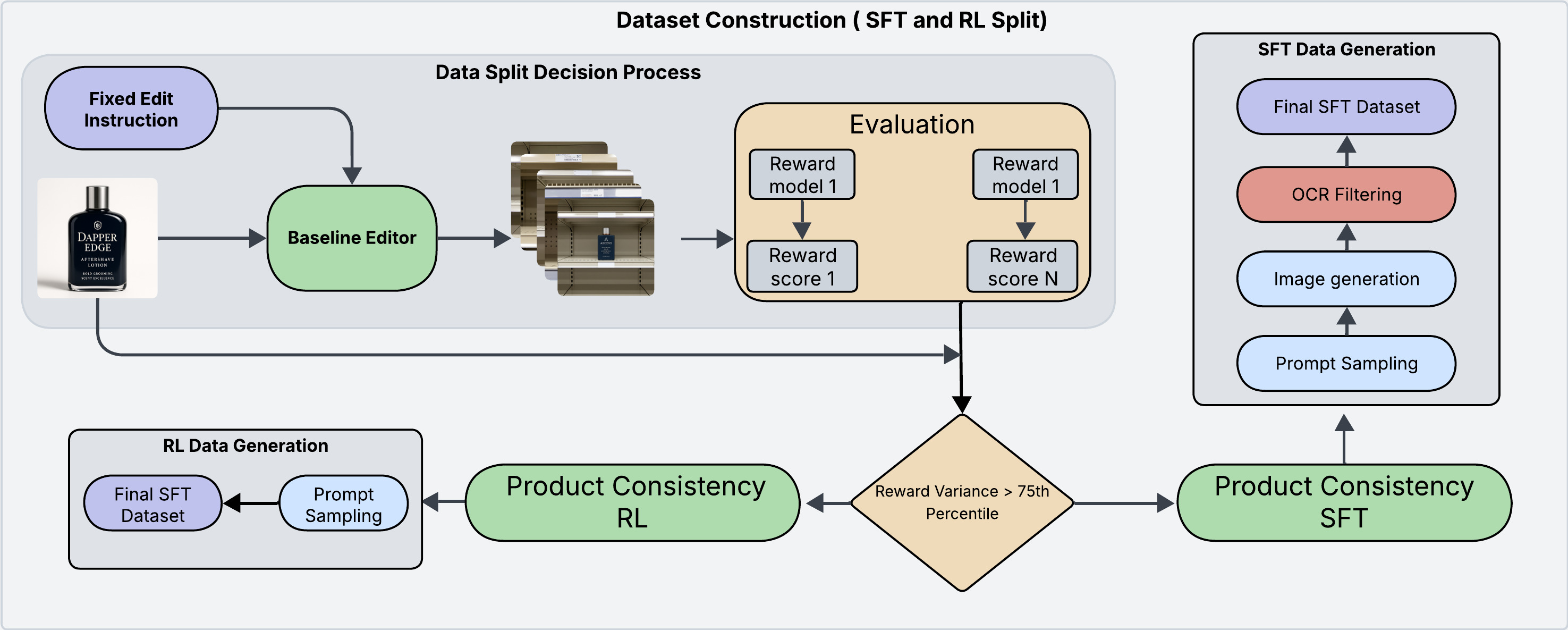}

{\footnotesize (b) Data splitting into SFT and RL sets.}
\end{minipage}

\caption{
Overview of the ProductConsistency dataset construction pipeline.
(a) Synthetic product image generation with unique branding and rendered text.
(b) The generated data is split into SFT and RL subsets, after which images are generated for SFT supervision and prompts are sampled for both training sets.
}
\label{fig:pipeline}
\end{figure*}  

To address this gap, we introduce the \textbf{ProductConsistency Dataset}, a new dataset, benchmark, and training framework designed specifically to study and improve product and brand consistency in instruction-based image editing. We present a fully automated pipeline for generating high-quality synthetic product images with unique brand identities and verifiable rendered text, enabling controlled supervision at scale. Using this pipeline, we create the ProductConsistency dataset, which we use to train two image editing models and demonstrate consistent improvements across multiple quantitative metrics. To enable rigorous evaluation, we also release the ProductConsistency benchmark, consisting of 174 product images that span 8 product categories, paired with five distinct editing prompts per product for a total of 870 evaluation samples. We evaluated both open-source and closed-source models on this benchmark and show that models trained on our ProductConsistency-SFT and ProductConsistency-RL dataset achieve improvements across all reported metrics, including Seg CLIP-I, Seg DINO-I, and OCR character error rate. In addition, we leverage closed-source large language models as automated judges to assess generated images along three axes: product consistency, OCR fidelity, and overall aesthetics. For our RL framework, we introduce the Cyclic Consistency reward that uses caption similarity as a proxy for product similarity. We explore product-aware reward functions that explicitly target instruction adherence, semantic product identity, visual consistency, and text fidelity, and demonstrate that fine-tuning a strong open-source editor under this framework yields substantial gains in product consistency. \textbf{Our contributions are summarized as follows:}

\begin{itemize}
\item \textbf{ProductConsistency Dataset and Generation Pipeline.} We introduce a fully automated pipeline for generating synthetic product images with unique brand identities and verifiable rendered text, enabling scalable supervision for product-centric instruction-based image editing.

\item \textbf{ProductConsistency Benchmark.} We release a human-verified benchmark consisting of 174 product images across 8 categories paired with five unique editing prompts each (870 evaluation samples), designed to rigorously evaluate product identity preservation, text fidelity, and visual consistency.

\item \textbf{Product-Aware Training with Cyclic Consistency Rewards.} We propose a Cyclic Consistency reward that aligns generated captions with original product descriptions while incorporating OCR-based rewards to improve textual fidelity, yielding consistent improvements across multiple automated and MLLM-based evaluation metrics across multiple open-source models.
\end{itemize}

\section{Related work}
\textbf{Image Editing}: Diffusion models have revolutionized image editing. Early editing models \cite{podell2023sdxl, 1rombach2022highdiffusionmodel} were initialized from pretrained text-to-image models with the first layers modified to use the mask and reference image as input, or they used Mask based editing with text-to-image \cite{podell2023sdxl} backbones with the latent being initialized using the masked image. InstructPix2Pix\cite{brooks2023instructpix2pix} trained a conditional diffusion model on synthetic pairs of input images and instructions that allowed mask-free instruction-based edits. Mask guided approaches such as \cite{blackforestlabs_flux1_fill_dev_2024_blx, ju2024brushnet, li2024brushedit},  generally use a designated edit mask along with text prompts to produce targeted, high‑quality modifications.
Other approaches like ControlNet\cite{zhang2023adding_controlnet} introduced a special adapter for different types of control. UniControl\cite{qin2023unicontrol} unified these multiple control adapters into a single adapter using MoE. Similarly, other adapter based methods \cite{li2024controlnet++, mou2024t2i_adapter, ye2023ip, zhang2025easycontrol} added additional control branches to the base model. These methods improved the fidelity of the reference inputs. However, finding the balance between precise control and flexible editing remained a challenge. 

Adding MLLMs as backbones for joint textual and visual processing further improved editing quality. Step-1X-edit\cite{liu2025step1x} jointly processed textual and visual inputs using an MLLM and used its intermediate outputs as conditional inputs to the diffusion model. Similarly, OmniGen\cite{xiao2025omnigen}, OmniGen2\cite{wu2025omnigen2} used newer architectures by coupling vision and language representation. These approaches largely succeed in preserving the global structure and following the instructions. However, they still falter at pixel-level consistency and struggle with more finer details like Text rendering, brand representation, logos etc. More recent models like Qwen-Image-Edit\cite{wu2025qwen}, Flux.1-Kontext-dev\cite{labs2025flux}, Hidream-E1-1\cite{cai2025hidream} are multi billion parameter generative models trained with flow-matching loss and optimized for text rendering. Despite these advances, even these models struggle with real world product images which often have multiple sections of text with different font sizes, colors, stylized elements, and they struggle to maintain fine-grained pixel-level consistency, which is critical for product imagery where branding elements are non-negotiable. In particular, small and densely packed text on product surfaces remains a persistent failure mode. Reasoning based approaches such as \cite{li2025editthinker, li2025reflect, qu2025replan, yin2025reasonedit, zhang2025r_genie, zou2025beyond_MURE} combine reasoning with reflection, instruction grounding, and multi-step editing to further improve the reasoning capabilities of reasoning models. However, these reasoning-driven methods only address complex-instruction following, and do not explicitly target preservation of existing on-object text.

\textbf{Reward models and datasets}: Reinforcement Learning has been extensively applied to align LLMs and MLLMs with end rewards\cite{guo2025deepseekr1}, and also for text-to-image models \cite{he2025tempflowgrpo,liu2025flowgrpo,li2025mixgrpo}, with many existing reward models for feedback \cite{xu2023imagereward,thon2025laionaesthetics,wu2023human_hps2,ma2025hpsv3,cho2022fineclipmodel}. However, extending the RL framework to editing models has been challenging due to the lack of good reward models for instruction based image-editing. InstructRL4Pix\cite{li2024instructrl4pix} fine-tunes a diffusion editor via PPO, using a score based on the alignment between the attention-maps of the edited and target objects as a proxy reward. In UniWorld-V2\cite{li2025uniworld} the output logits of a frozen MLLM act as an implicit reward. More recently, large human aligned reward models have been developed for instruction based editing. EditReward\cite{wu2025editreward} and Editscore\cite{luo2025editscore} are VLM-based reward models, and RL training with these reward models has shown that strong base editors improve dramatically, whereas generic VLMs were ineffective. However, no reward models have been designed explicitly for product consistency.

Another challenge in training models for product consistency is the limited availability of suitable open-source datasets. The ABO (Amazon-Berkeley Objects) dataset~\cite{collins2022abo} provides images of 147,702 products. However, many images exhibit substantial visual variability, including multiple items within a frame, unconventional viewing angles, cluttered arrangements, and product text that is partially obscured or difficult to read.  Similarly, Products-10k~\cite{bai2020products} contains ten thousand images across different SKUs, but is primarily designed for product recognition tasks and shares similar limitations. As a result, these datasets are not well suited for training image editing models that must preserve fine-grained product attributes, particularly textual elements, which are often difficult even for humans to read in these images. The HFPC-44K\cite{liang2025evaluation_hfpc} dataset consists of 44,244 real product images paired with outputs produced by AI-driven background inpainting, each manually annotated as “good” or “bad”. However, HFPC does not explicitly evaluate on-object text fidelity.

\section{Methodology}

This work addresses the lack of product- and text-consistent instruction-based image editing by introducing the ProductConsistency Dataset and benchmark, and a scalable synthetic data generation pipeline. We do SFT and RL finetuning using our dataset and our proposed Cyclic Consistency Reward. Our methodology is fully automated and designed to expose and correct failure modes related to brand identity and text fidelity on the object. Using the proposed datasets, we fine-tune two open-source image editing models and demonstrate significant improvements in product consistency under advertisement-style edits. The data construction pipeline is depicted in Figure~\ref{fig:pipeline}

\subsection{Synthetic Product Image Generation}

We begin by defining 131 unique product items in 8 categories that naturally contain visible textual elements, such as packaged food, beverages, cosmetics, household items, etc. For each item, GPT-o3-mini is prompted  using a fixed system prompt with in-context examples to generate diverse product descriptions. The system prompt is available in Figure \ref{fig:system_prompt_prompt_generation}. The MLLM is prompted to generate images with a varying number of words to be rendered on the product (5-12 words).  Before generating each description, the model is instructed to create a brand signature consisting of a brand name, color palette, font style, and packaging tone. This step enforces a coherent visual identity and enables the generation of realistic branded products that closely resemble real-world items and adds diversity to the product images. This process yields approximately 3,000 unique product prompts, each containing between 5 and 12 words that need to be rendered directly on the product, along with a unique brand identity.

For each product prompt, we generate the multiple corresponding images using GPT-Image-1 high\cite{openai2025gptimage1}. Since generative models frequently introduce errors in rendered text, we apply OCR-based filtering to ensure text correctness. Let $T_{\text{gt}}$ denote the ground-truth text specified in the prompt and $T_{\text{ocr}}$ the text detected by an OCR model. An image is retained only if $T_{\text{ocr}} = T_{\text{gt}}$. All images with missing characters, hallucinated text, or incorrect spellings are discarded. This filtering stage produces a high-precision set of images of 2002 synthetic products with verifiable and legible text.

\subsection{Construction of SFT and RL Training Sets}

\textbf{SFT and RL data split}: To analyze the robustness of existing image editing models and to split the training data for supervised fine-tuning and reinforcement learning, we apply a fixed, simple edit instruction to each filtered product image: \emph{``Put this product inside an empty supermarket shelf (inside a shelf bay) at eye level, close-up shot, front view.''} For every input image, five edited outputs are generated using a baseline editor with different seeds.

\begin{table}[t]
  \centering
  \resizebox{0.95\linewidth}{!}{
  \begin{tabular}{@{}lccc@{}}
    \toprule
    \textbf{Category} 
    & \textbf{SFT} 
    & \textbf{RL} 
    & \textbf{Benchmark} \\
    \midrule
    Electronics            & 10,172 (11.7\%) & 1,700 (19.6\%) & 120 (13.8\%) \\
    Personal Care          & 11,958 (13.7\%) & 1,460 (16.8\%) & 140 (16.1\%) \\
    Food \& Snacks         & 13,911 (15.9\%) &   810 (9.3\%)  & 120 (13.8\%) \\
    Beverages              & 12,459 (14.3\%) &   900 (10.4\%) & 115 (13.2\%) \\
    Cooking \& Kitchen     & 11,662 (13.4\%) &   950 (10.9\%) &  90 (10.3\%) \\
    Stationery             & 10,919 (12.5\%) & 1,050 (12.1\%) & 100 (11.5\%) \\
    Household              &  7,235 (8.3\%)  & 1,040 (12.0\%) & 110 (12.6\%) \\
    Health \& Supplements  &  8,926 (10.2\%) &   780 (9.0\%)  &  75 (8.6\%) \\
    \midrule
    \textbf{Total}         & \textbf{87,242 } 
                           & \textbf{8,690} 
                           & \textbf{870} \\
    \bottomrule
  \end{tabular}
  }
\caption{
    \textbf{Category Distribution (SFT, RL \& Benchmark sets).}
    The SFT dataset (87,242 samples) has a fairly even distribution across all eight categories, with Food \& Snacks and Beverages being the largest groups. The RL set (8,690 samples) contains more samples in Electronics and Personal Care. The benchmark (870 samples) follows a similar pattern, helping maintain consistent category coverage during evaluation.
}
  \label{tab:category_distribution}
\end{table}

Each edited image is evaluated using two complementary metrics. First, we compute an alignment score using the EditReward model, which evaluates prompt following and serves as a proxy for soft product-consistency, conditioned on the input image, the edited image, and the edit instruction. Second, we measure text preservation using OCR-based consistency. OCR is performed with Qwen3-VL-2B, and a per-word character error rate (CER) is computed by greedy matching between ground-truth and detected words. Let $\{o_i\}_{i=1}^{N}$ denote the set of OCR-detected words and $\{g_j\}_{j=1}^{M}$ the set of ground-truth words. For each detected word $o_i$, we greedily match it to the unmatched ground-truth word $g_j$ that minimizes the normalized Levenshtein distance:
\[
\mathrm{CER}(o_i, g_j) = \frac{d_{\text{lev}}(o_i, g_j)}{|g_j|}.
\]
The final OCR error score is computed as the sum of the CER values over all matched pairs:
\[
\mathrm{CER}_{\text{total}} = \sum_{(o_i, g_j) \in \mathcal{M}} \mathrm{CER}(o_i, g_j),
\]
where $\mathcal{M}$ denotes the set of greedy word matches.

For each product image, we compute the variance of both the EditReward score and the CER score across all outputs. Images that fall above the 75th percentile in either reward's variance distribution are assigned to the reinforcement learning dataset, as they correspond to unstable or failure-prone cases where existing models struggle to preserve product fidelity, but sometimes manage to get better outputs; thus, the benefits from potential reward optimization would be higher. The remaining images are assigned to the supervised fine-tuning dataset. This procedure results in 1,133 unique product images for the ProductConsistency-SFT dataset and 869 images for the ProductConsistency-RL dataset.

\textbf{Edit instruction generation}: To simulate real-world advertisement scenarios, we define multiple categories of advertisement styles, including studio shots, lifestyle scenes, outdoor settings, and festive environments. For each category, a language model generates 10-15 unique edit prompts, resulting in a total of 220 prompts. To introduce visual diversity, we vary background settings, lighting conditions, color tones, weather, and time of day across the prompts in each category. For supervised fine-tuning, each product image is paired with 100 randomly sampled prompts. For reinforcement learning, each image is randomly paired with 10 prompts.

\begin{figure*}[t]
\centering
\includegraphics[width=0.2\textwidth]{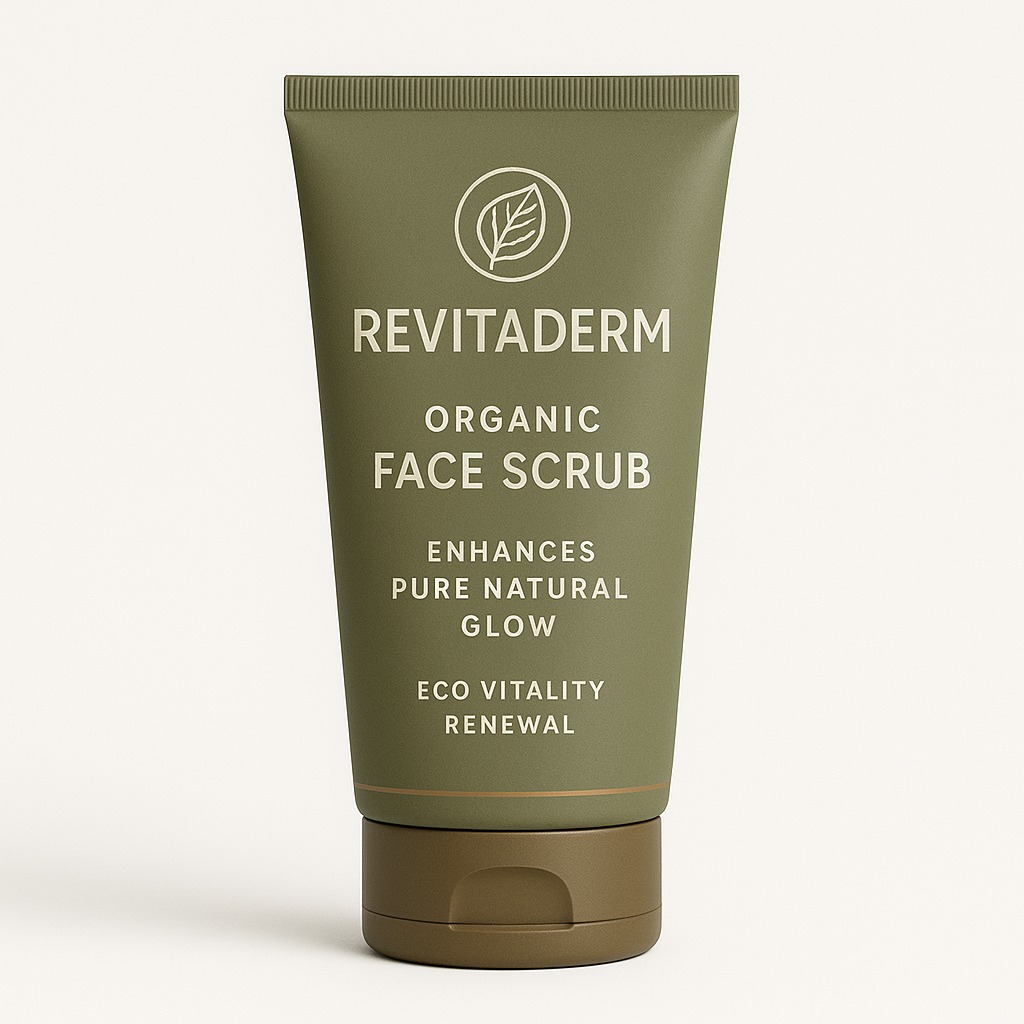}%
\includegraphics[width=0.2\textwidth]{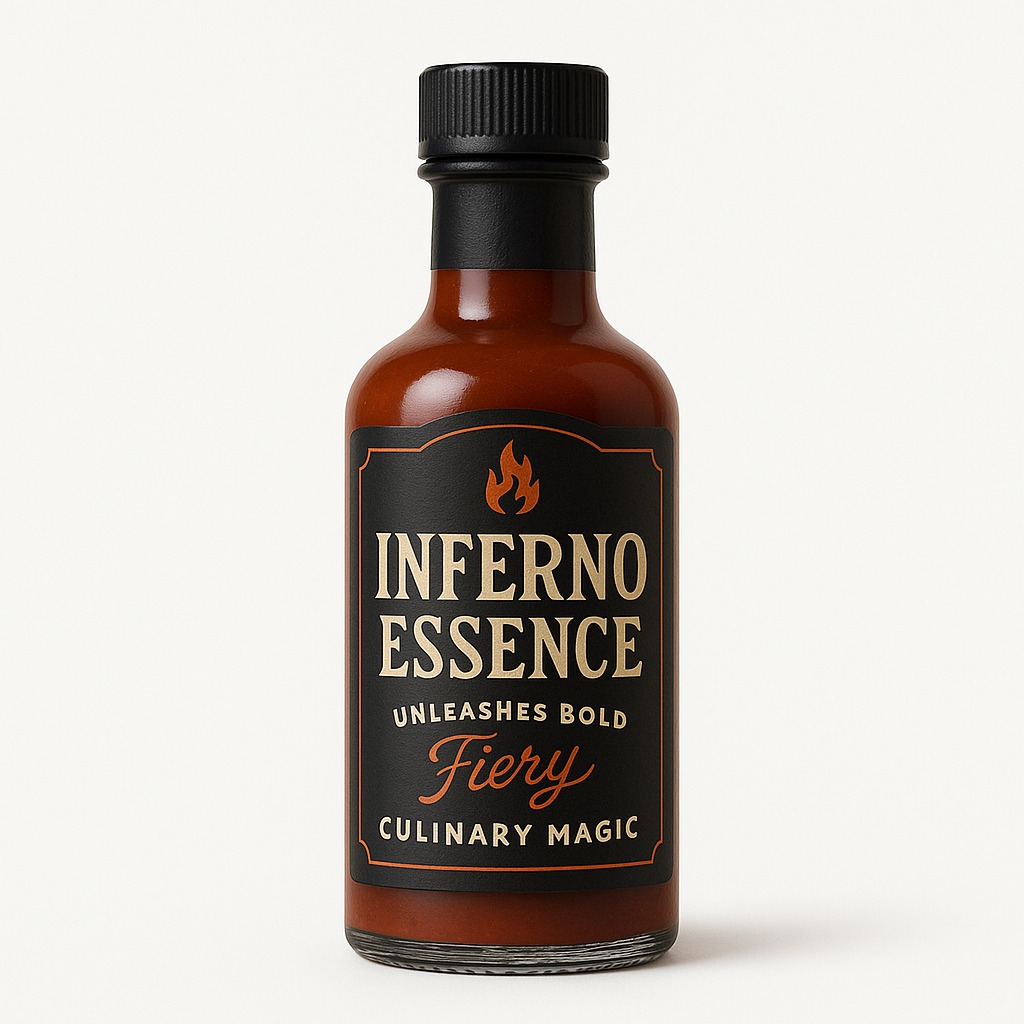}%
\includegraphics[width=0.2\textwidth]{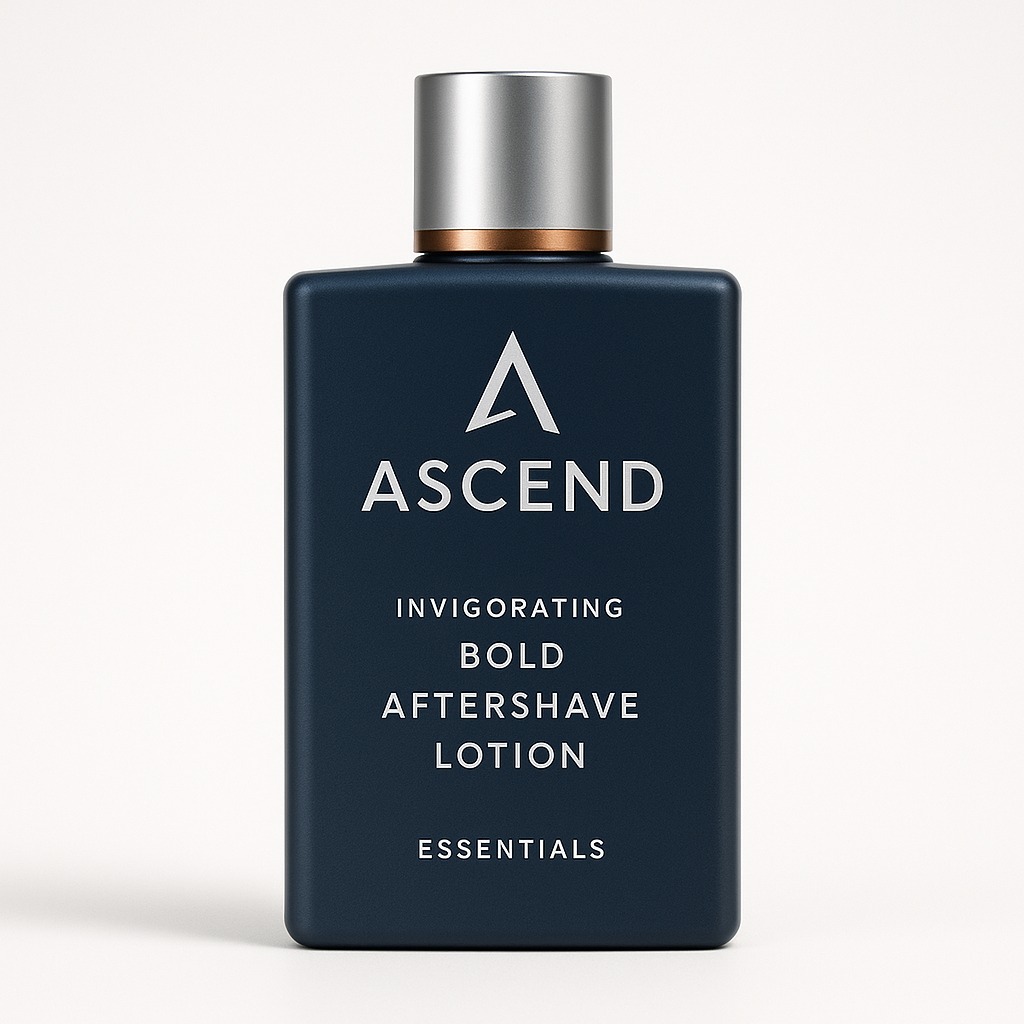}%
\includegraphics[width=0.2\textwidth]{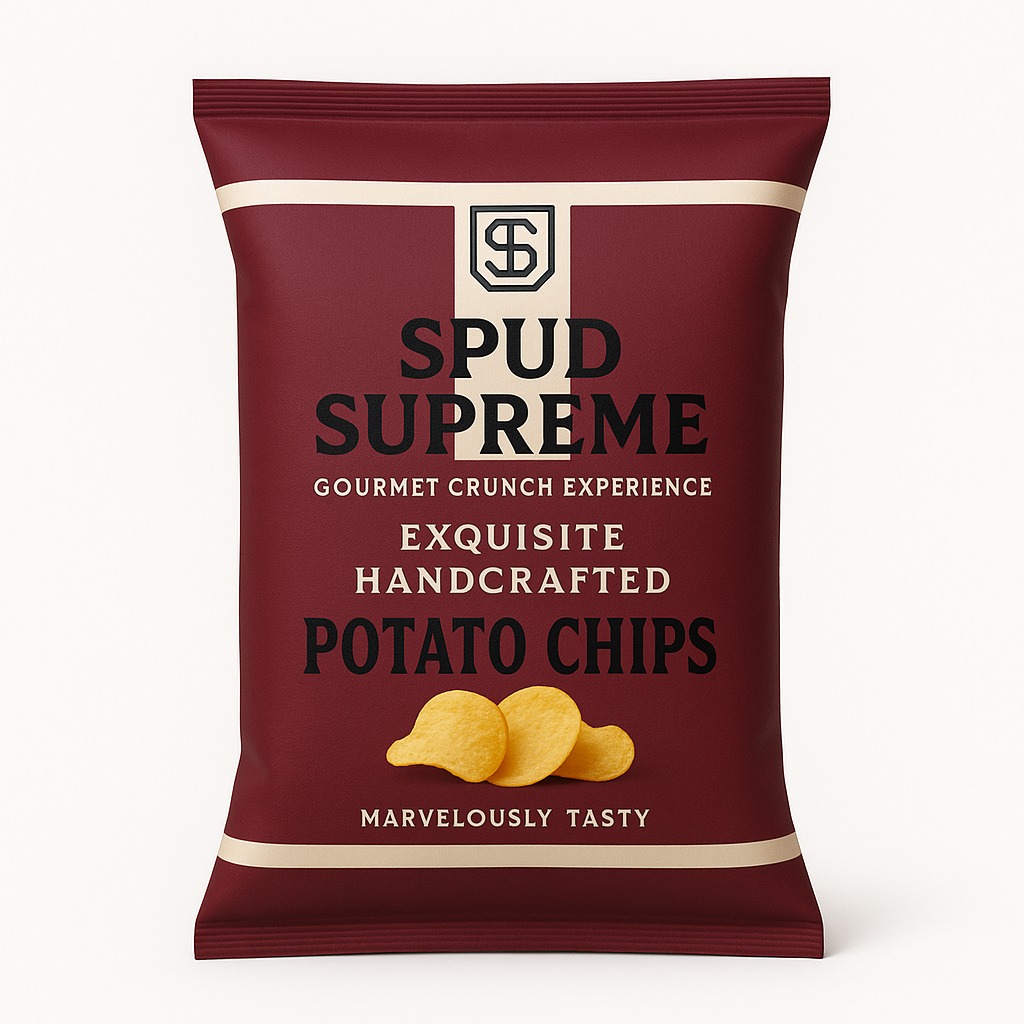}%
\includegraphics[width=0.2\textwidth]{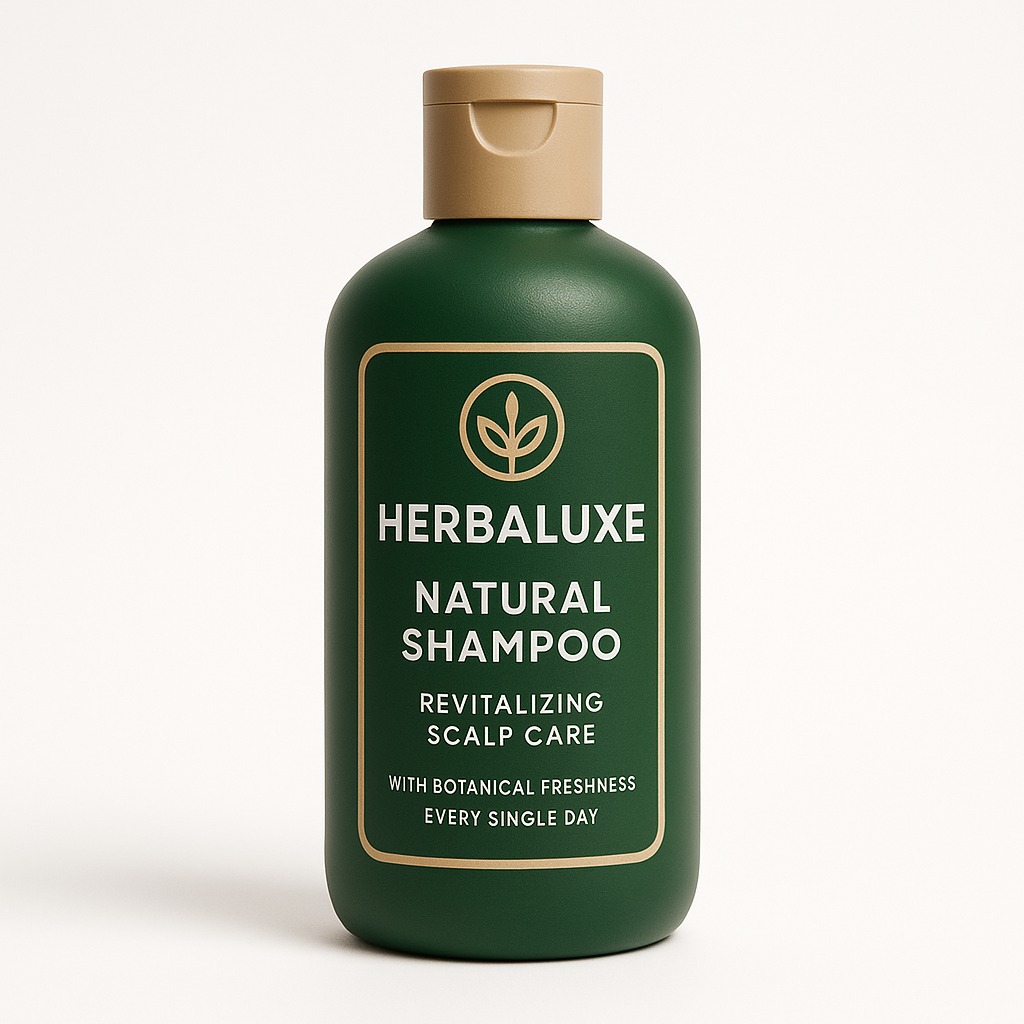}
\caption{
Examples of Product Images for the ProductConsistency-RL dataset.
The edit instructions for the images from left to right are:
(a) Place the product near a dumbbell rack with weights receding into the background, captured head-on with shallow depth of field.
(b) Position the product on a shelf with a subtle textured backing while keeping the shelf otherwise empty, photographed front-facing.
(c) Place the product on a wide urban plaza surface, captured front-facing during golden hour as warm sunlight washes over the city backdrop.
(d) Display the product next to a serene spa pool with still water and soft ambient light, captured head-on for a calm premium feel.
(e) Set the product on a weathered urban curb, framed closely from the front with street markings and asphalt texture visible.
}
    \label{fig:rl_dataset_examples}
\end{figure*}

\begin{figure*}[t]
\centering

\begin{minipage}[c]{0.48\textwidth}
\centering
\includegraphics[width=0.495\textwidth]{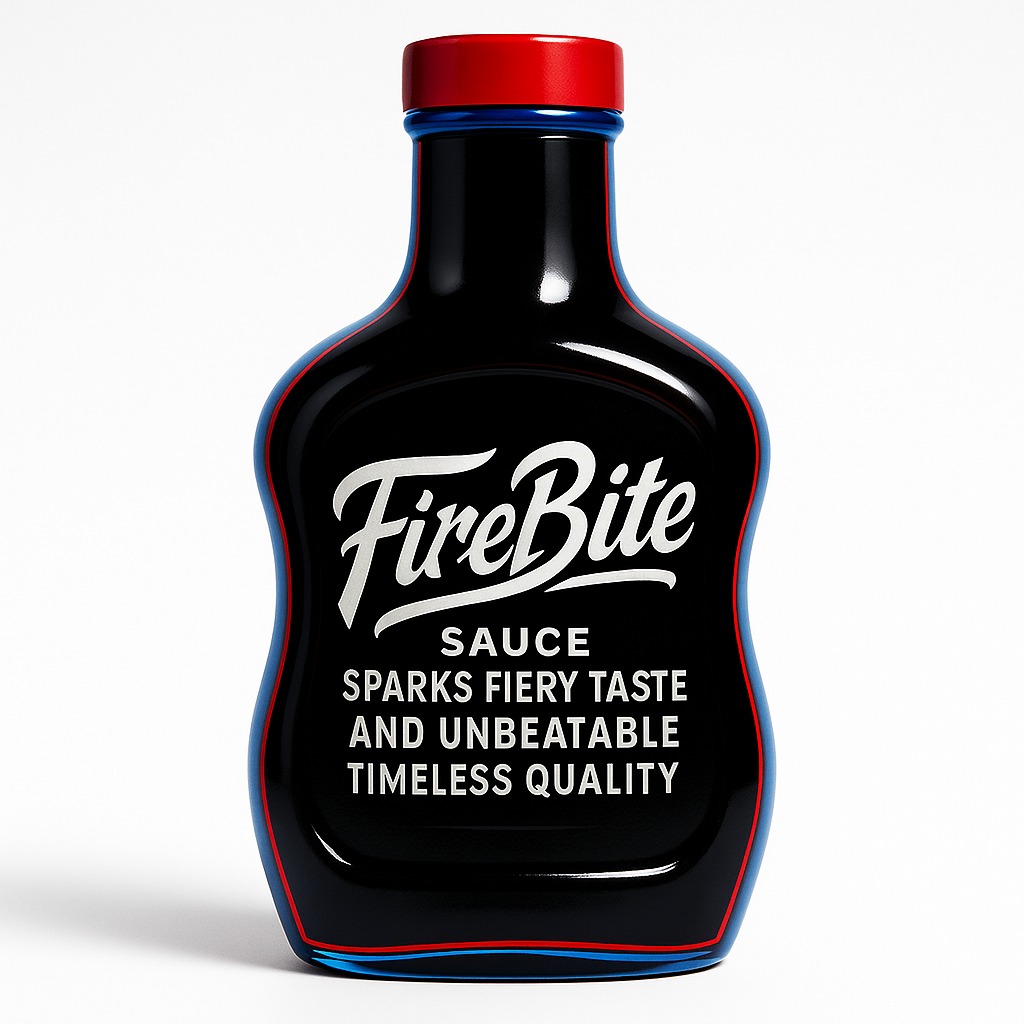}
\includegraphics[width=0.495\textwidth]{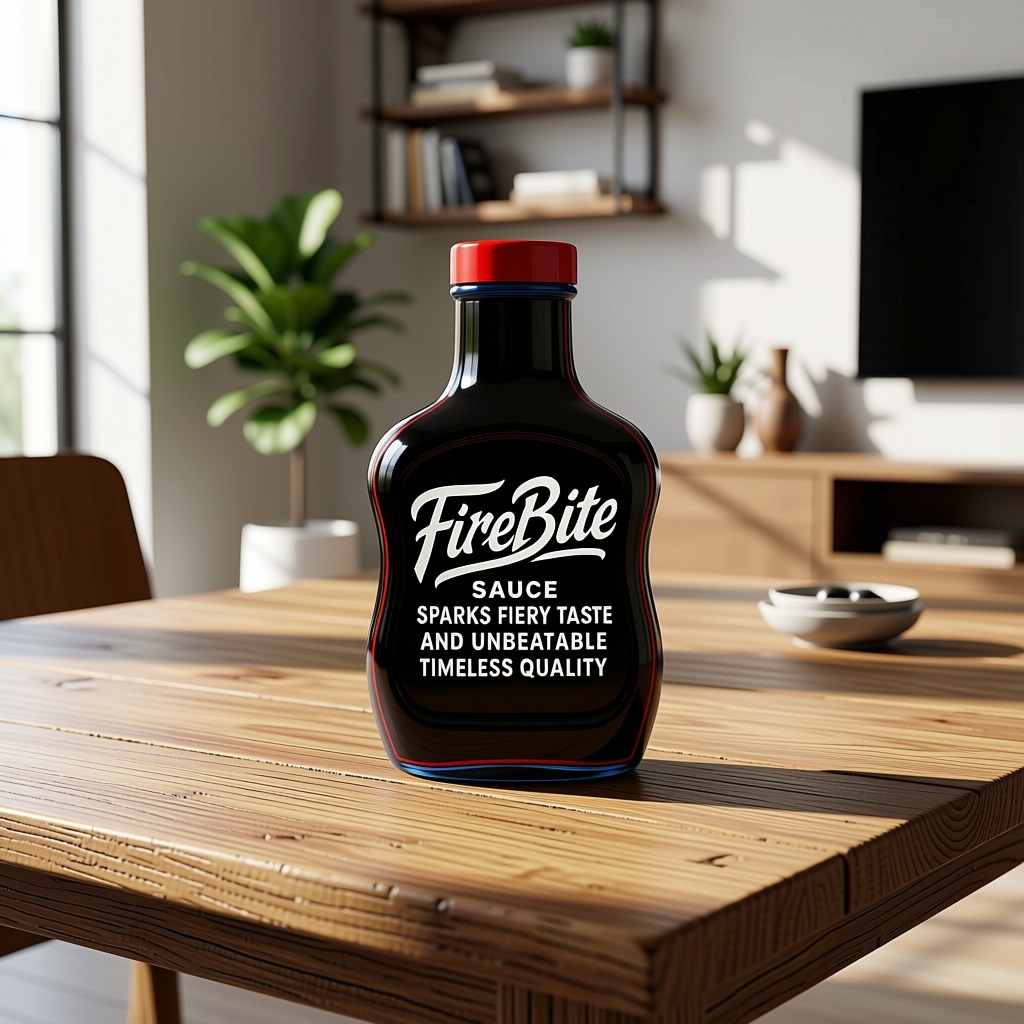}

{\footnotesize (a) Input $\rightarrow$ Output}
\end{minipage}\hspace{4pt}
\begin{minipage}[c]{0.48\textwidth}
\centering
\includegraphics[width=0.495\textwidth]{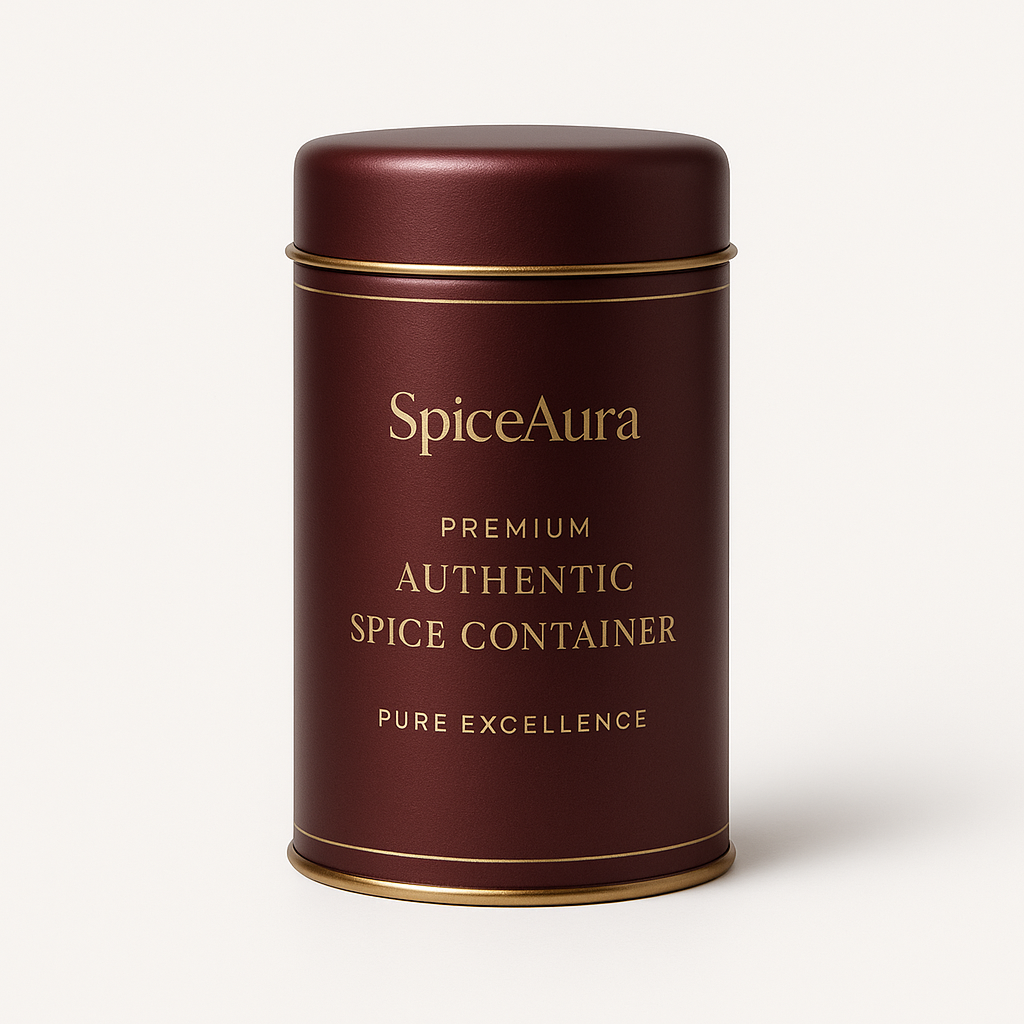}%
\includegraphics[width=0.495\textwidth]{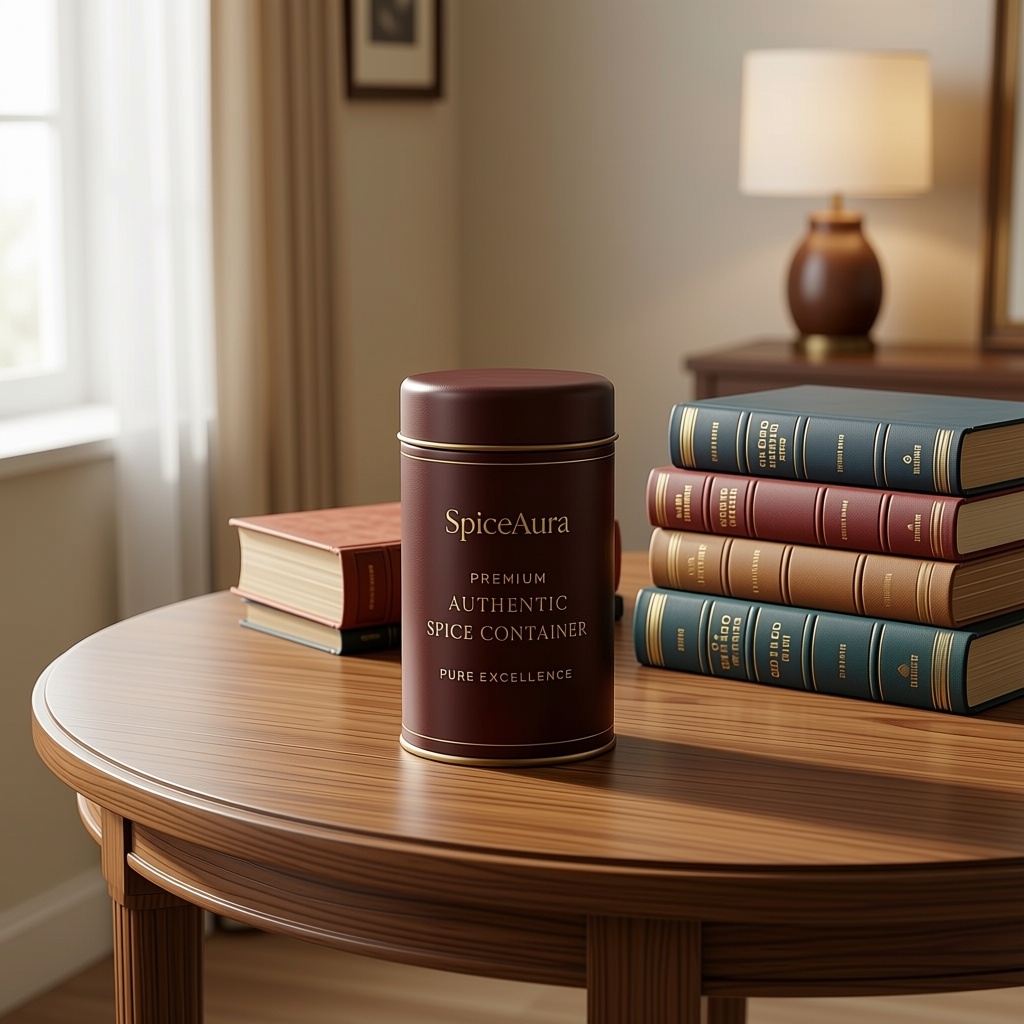}

{\footnotesize (b) Input $\rightarrow$ Output}
\end{minipage}

\vspace{4pt}

\begin{minipage}[c]{0.48\textwidth}
\centering
\includegraphics[width=0.495\textwidth]{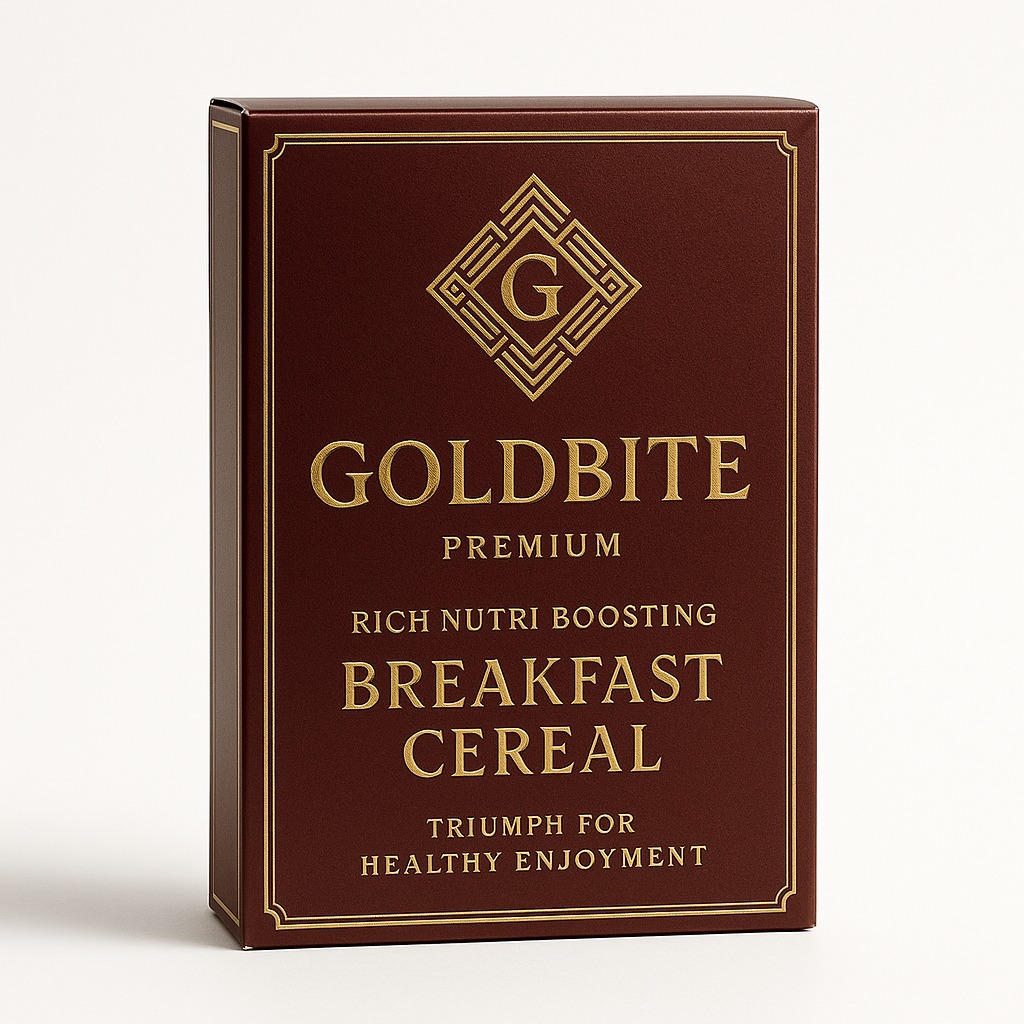}%
\includegraphics[width=0.495\textwidth]{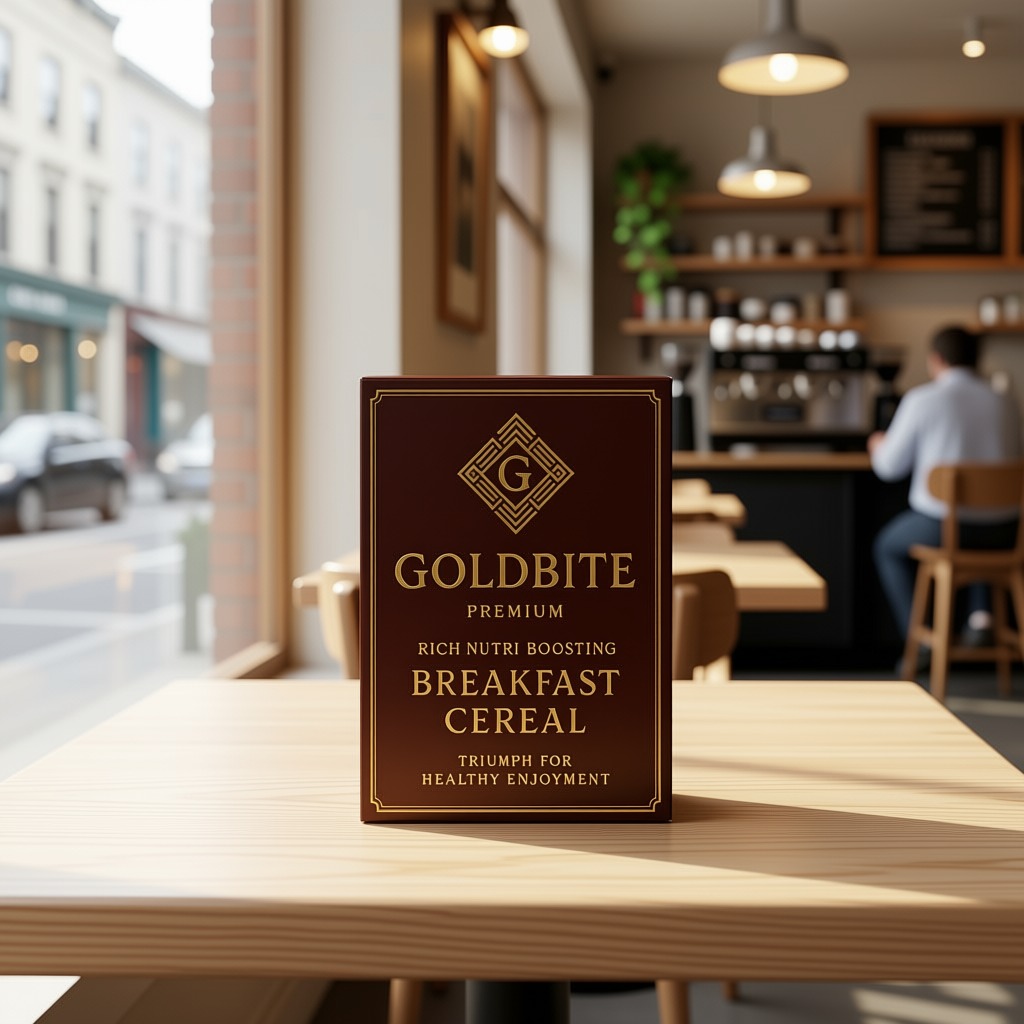}

{\footnotesize (c) Input $\rightarrow$ Output}
\end{minipage}\hspace{4pt}
\begin{minipage}[c]{0.48\textwidth}
\centering
\includegraphics[width=0.495\textwidth]{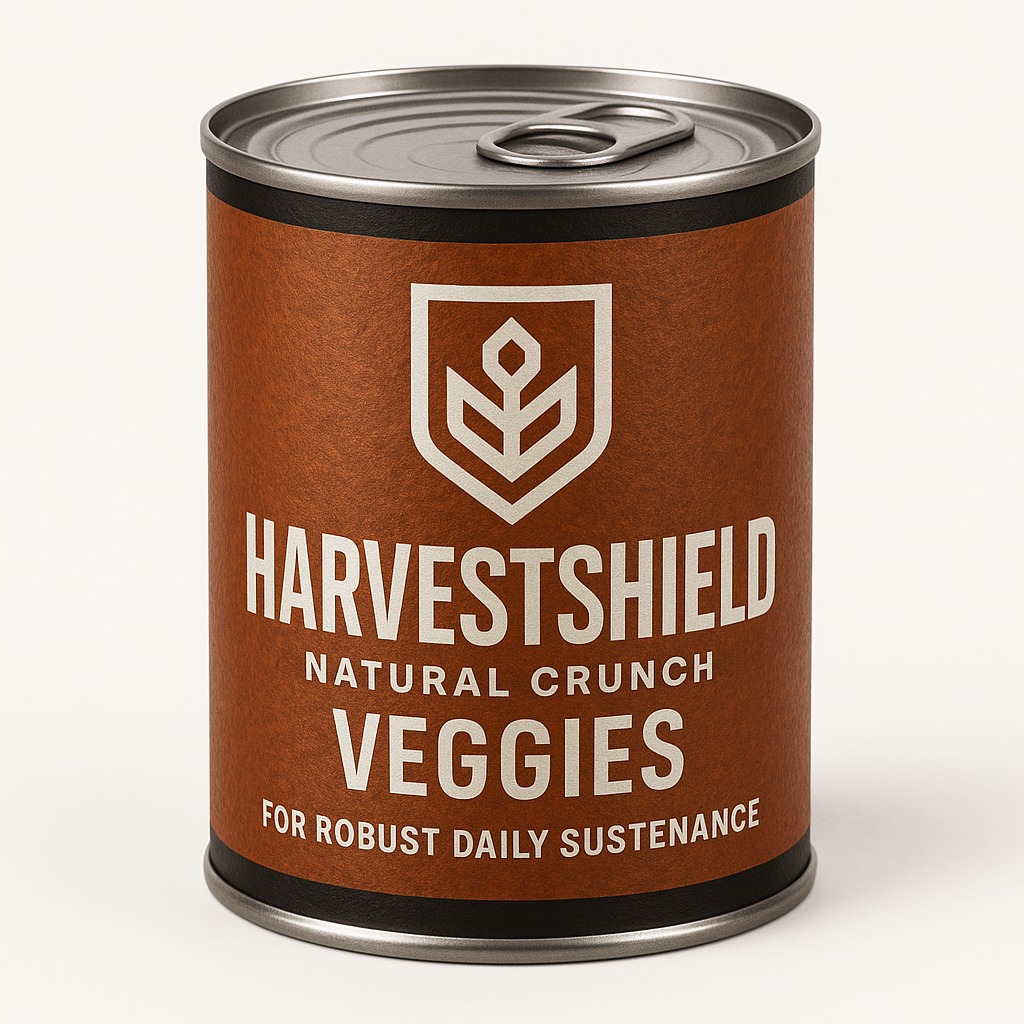}%
\includegraphics[width=0.495\textwidth]{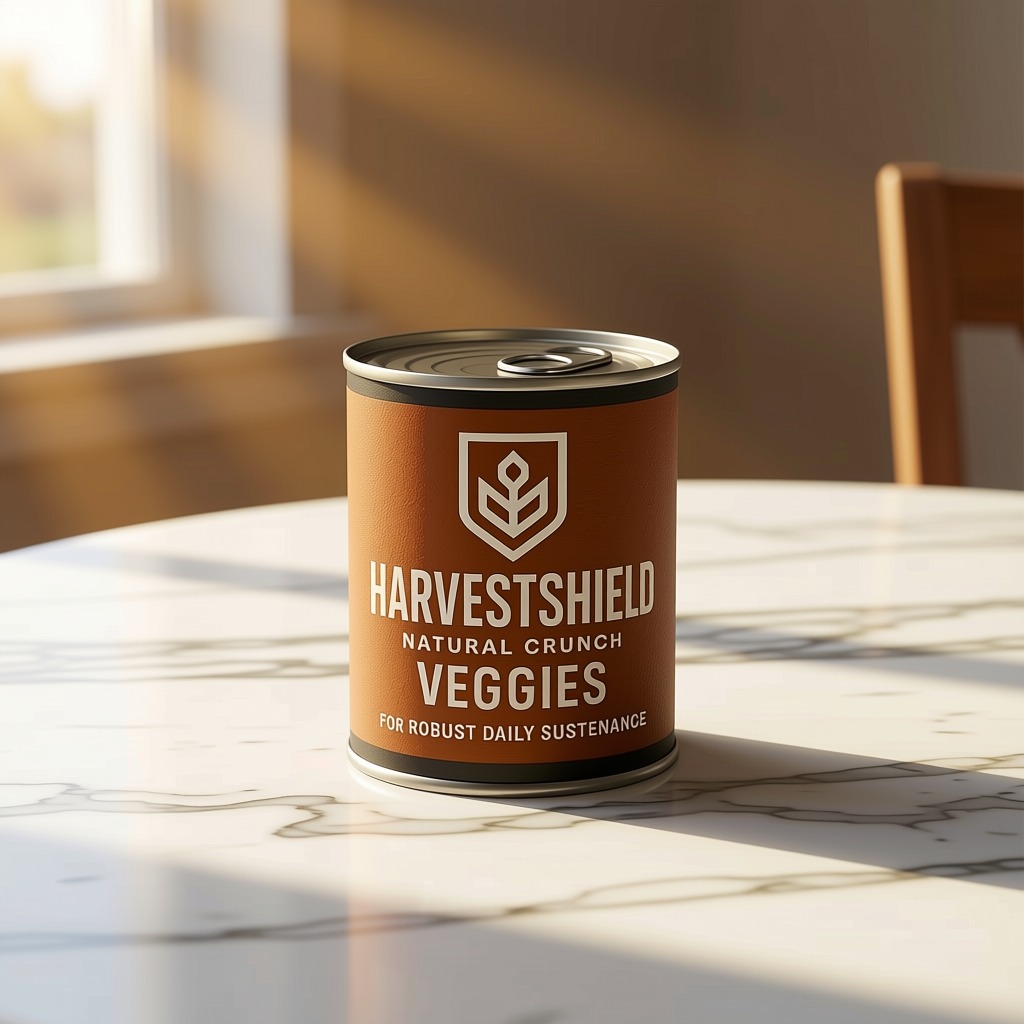}

{\footnotesize (d) Input $\rightarrow$ Output}
\end{minipage}

\vspace{4pt}

\begin{minipage}[c]{0.48\textwidth}
\centering
\includegraphics[width=0.495\textwidth]{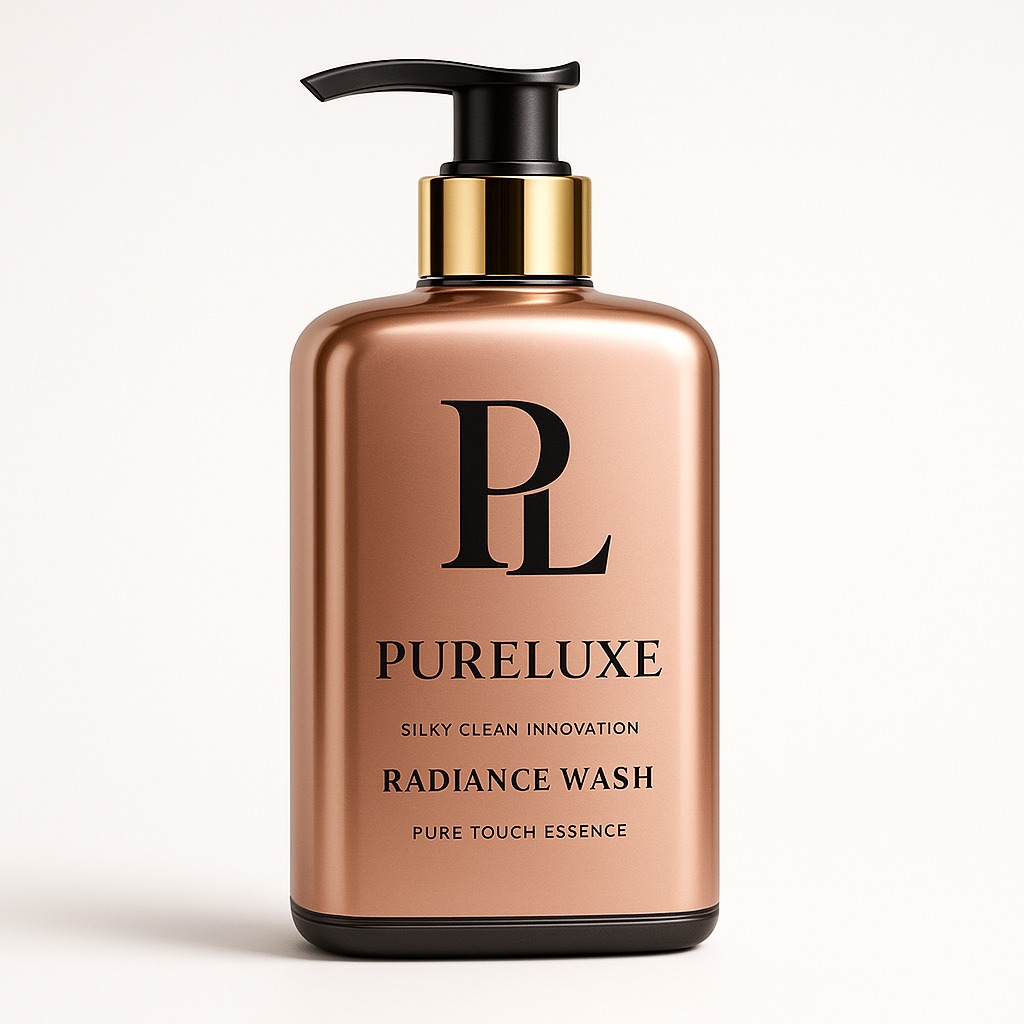}%
\includegraphics[width=0.495\textwidth]{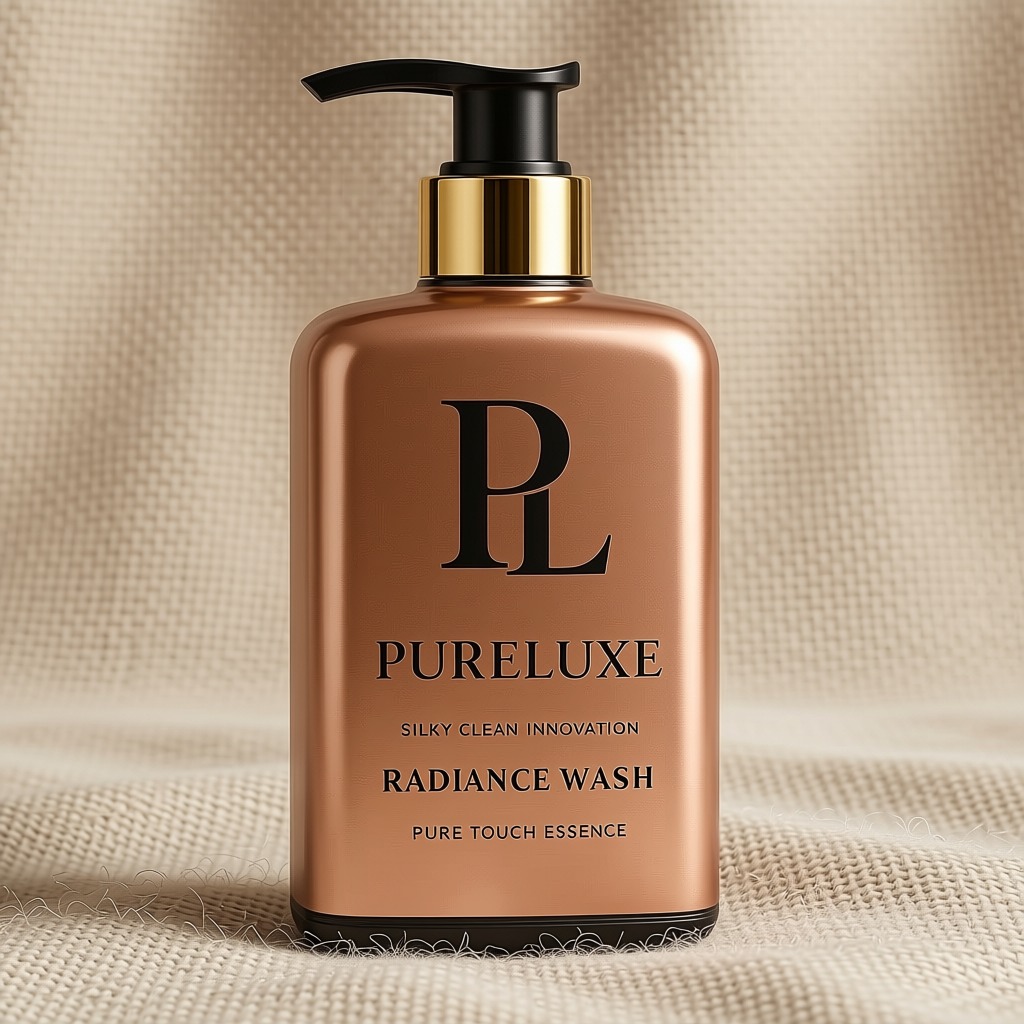}

{\footnotesize (e) Input $\rightarrow$ Output}
\end{minipage}\hspace{4pt}
\begin{minipage}[c]{0.48\textwidth}
\centering
\includegraphics[width=0.495\textwidth]{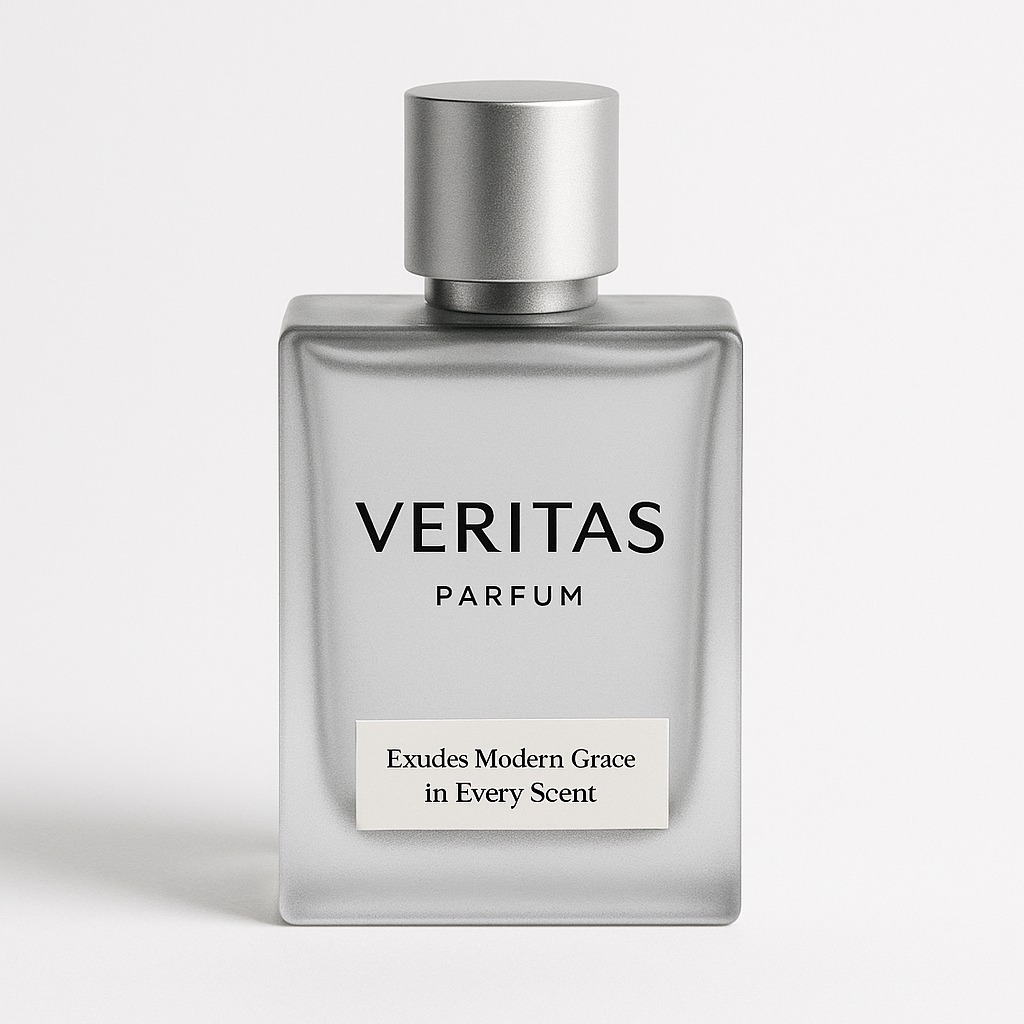}%
\includegraphics[width=0.495\textwidth]{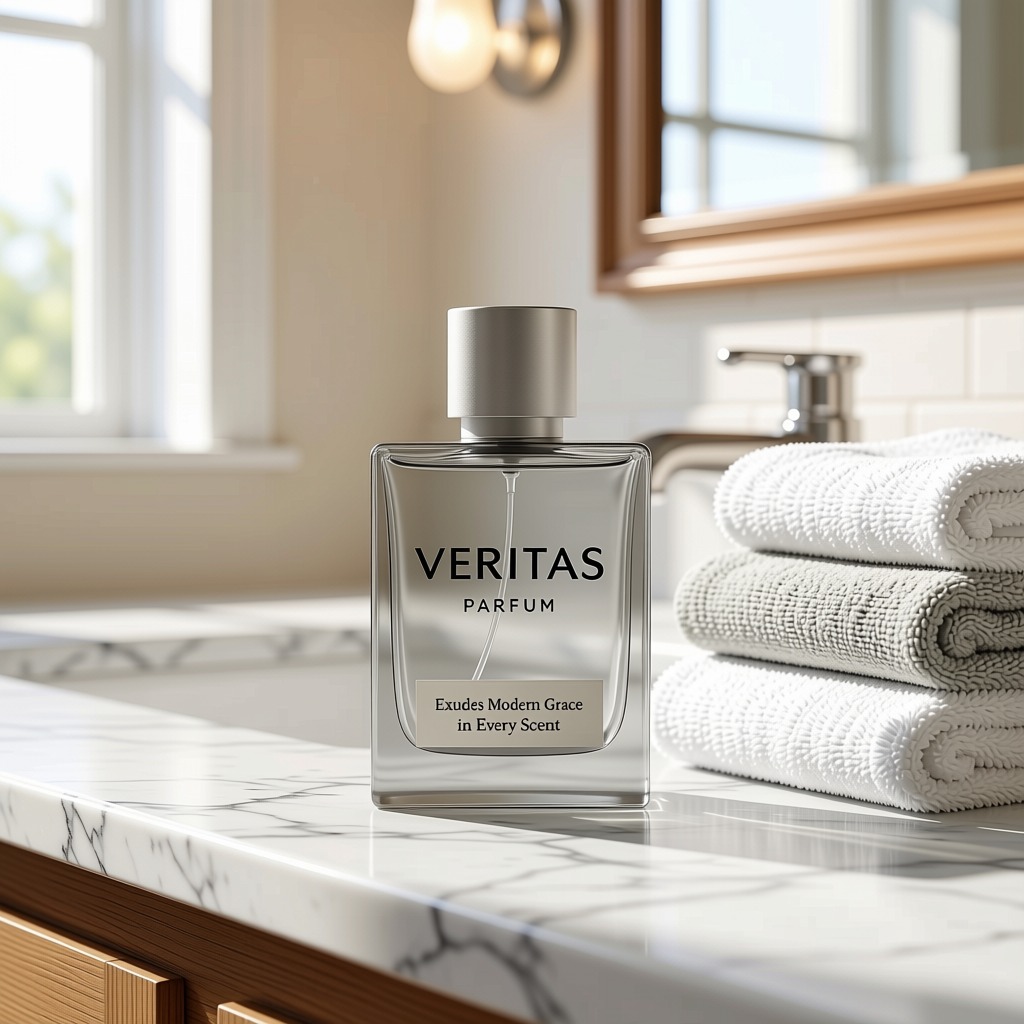}

{\footnotesize (f) Input $\rightarrow$ Output}
\end{minipage}

\caption{
Examples from the ProductConsistency-SFT dataset.
Each pair shows the input image (left) and the corresponding ground-truth edited output (right). 
The prompts are: 
(a) Place the product on a handcrafted wooden table in a contemporary home setting, shot front-facing with indirect sunlight illuminating the surface.
(b) Display the product on a refined lifestyle table with a stack of hardcover books nearby, photographed at eye level in calm, evenly balanced indoor lighting.
(c) Set the product on a light wood table inside a quiet neighborhood café, framed head-on with soft daylight filling the space.
(d) Set the product on a white marble table with subtle gray veining, photographed from the front as warm afternoon sunlight filters through a nearby window.
(e) Position the product on a neutral wool surface with subtle grain, photographed from the front for a timeless brand aesthetic.
(f) Position the product on a bathroom counter beside neatly folded towels, photographed front-facing in clean natural daylight.
}
\label{fig:sft_dataset_examples}
\end{figure*}

\textbf{SFT Target image construction:} To train our model on the SFT dataset, we need high quality target images. For each of the 113,400 source–instruction pairs in the SFT dataset, we generated multiple edited image candidates using the Qwen-Image-Lightning model with an 8-step inference process. We selected Qwen-Image-Lightning because it offers both faster generation and stronger text rendering compared to Qwen-Image-Edit-2511. The efficient 8-step inference allows us to generate candidates at scale, enabling a generate-and-filter strategy in which multiple outputs are produced and the best ones are retained. All generated images are again filtered using OCR-based text consistency checks. Only Images with correct text are selected as part of the final dataset. After filtering, the final ProductConsistency-SFT dataset is created that contains 87,242 high-quality image pairs. Images where no good outputs are generated are discarded. Examples of the RL dataset are presented in Figure \ref{fig:rl_dataset_examples}, and examples of the SFT dataset are presented in Figure \ref{fig:sft_dataset_examples}.

\subsection{Benchmark Test Set}

\begin{figure*}[t]
    \centering
    
    \begin{subfigure}[b]{0.48\textwidth}
        \centering
        \includegraphics[width=\textwidth]{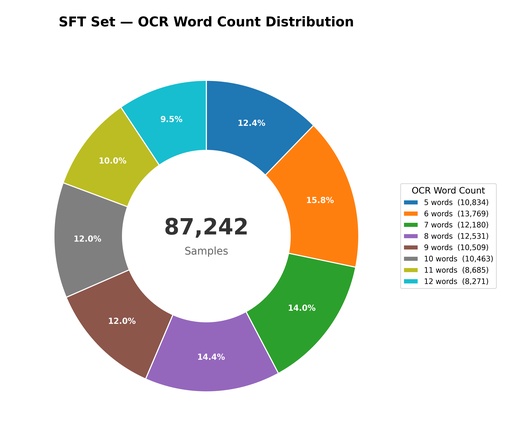}
        \caption{SFT Training Set}
        \label{fig:word_sft}
    \end{subfigure}
    \hfill
    \begin{subfigure}[b]{0.48\textwidth}
        \centering
        \includegraphics[width=\textwidth]{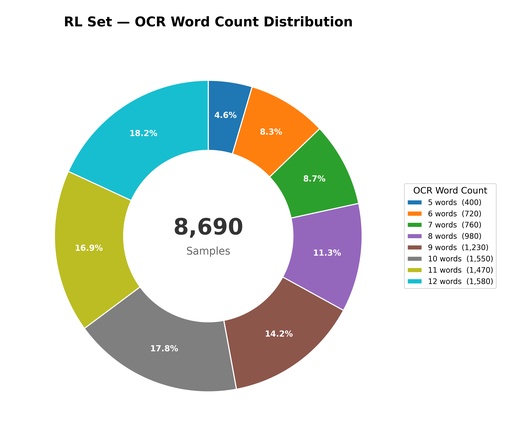}
        \caption{RL Training Set}
        \label{fig:word_rl}
    \end{subfigure}

    \vspace{0.5em}

    \begin{subfigure}[b]{0.6\textwidth}
        \centering
        \includegraphics[width=\textwidth]{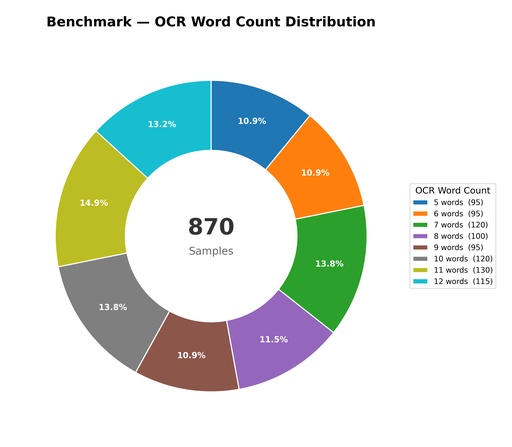}
        \caption{Benchmark Test Set}
        \label{fig:word_test}
    \end{subfigure}

    \caption{
        \textbf{OCR Word Count Distribution across datasets.}
        The word count ranges from 5 to 12, introducing natural variation in text complexity.
        Both training (SFT, RL) and benchmark sets exhibit an approximately uniform distribution.
    }
    \label{fig:data_distribution_word_count}
\end{figure*}

\begin{figure*}[t]
\centering

\includegraphics[width=0.2\textwidth]{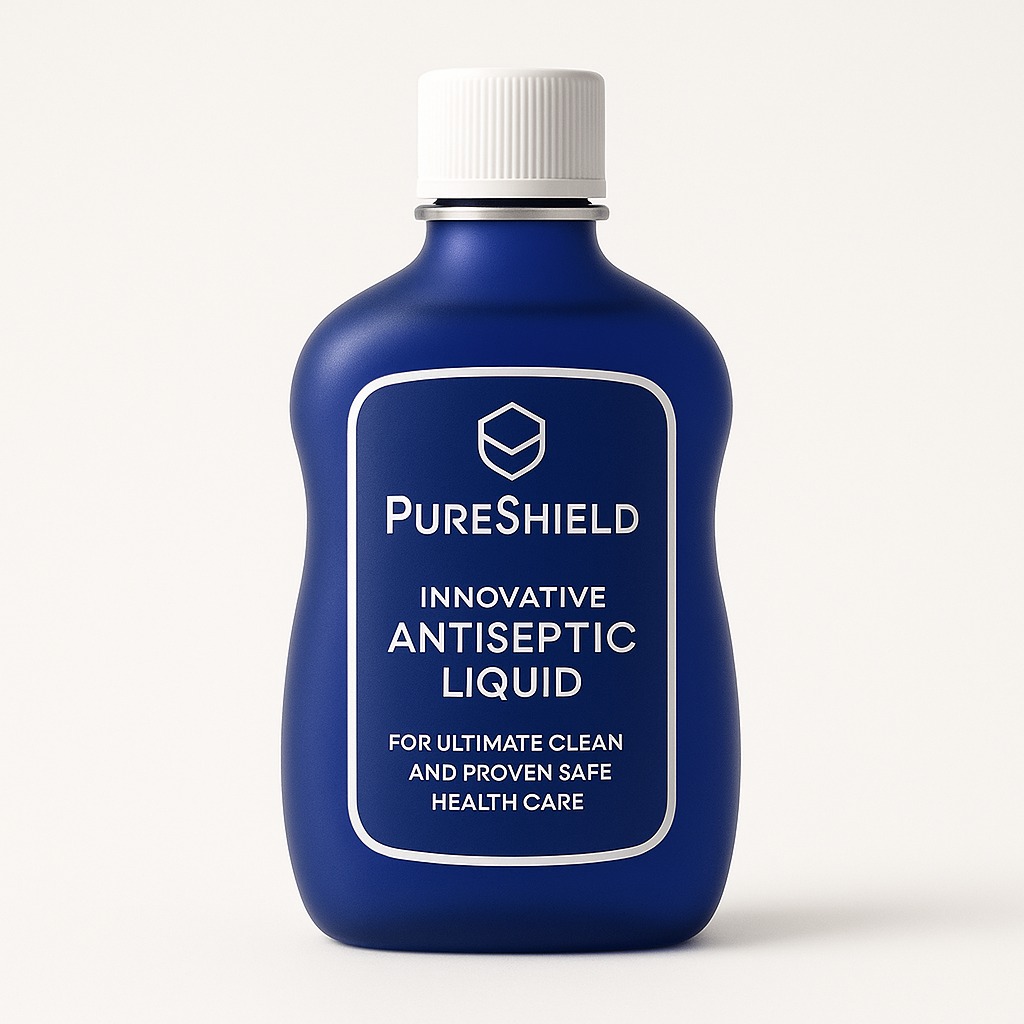}\hspace{5pt}%
\includegraphics[width=0.2\textwidth]{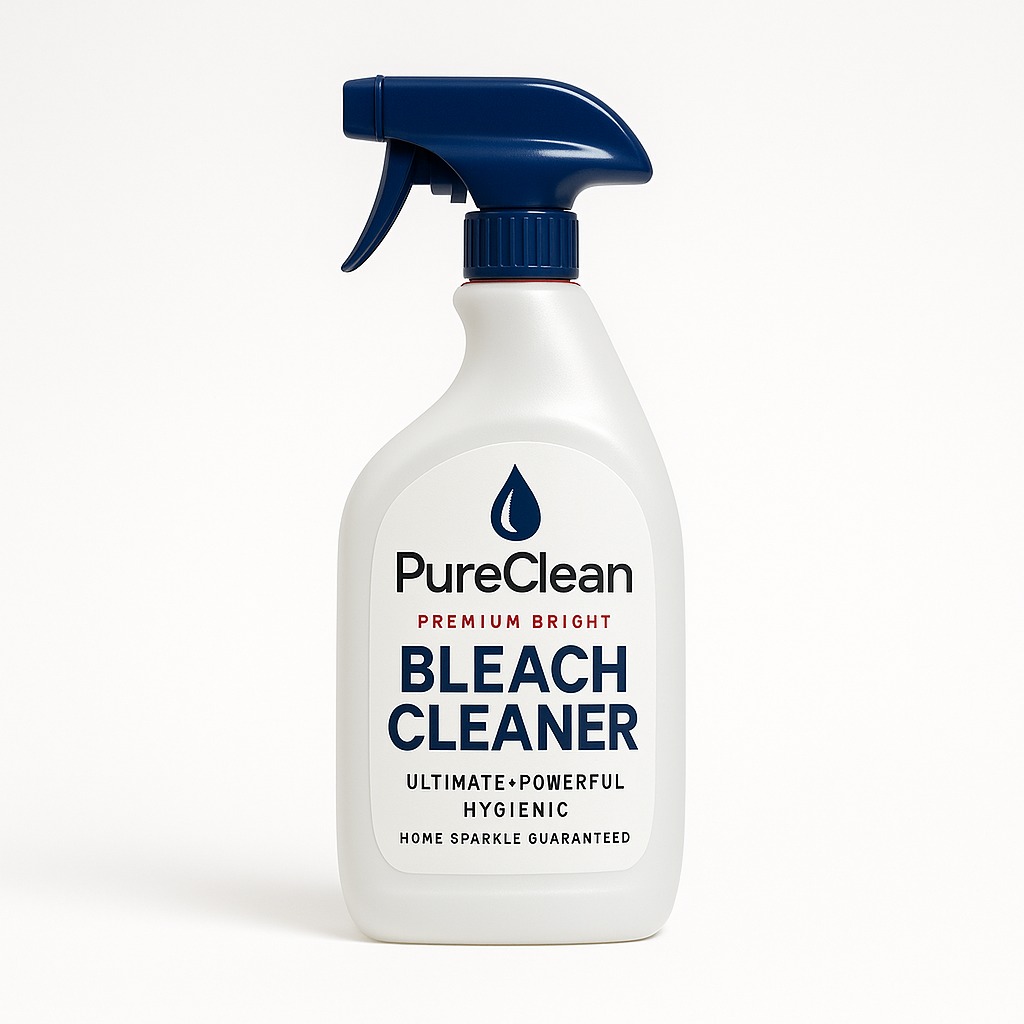}\hspace{5pt}%
\includegraphics[width=0.2\textwidth]{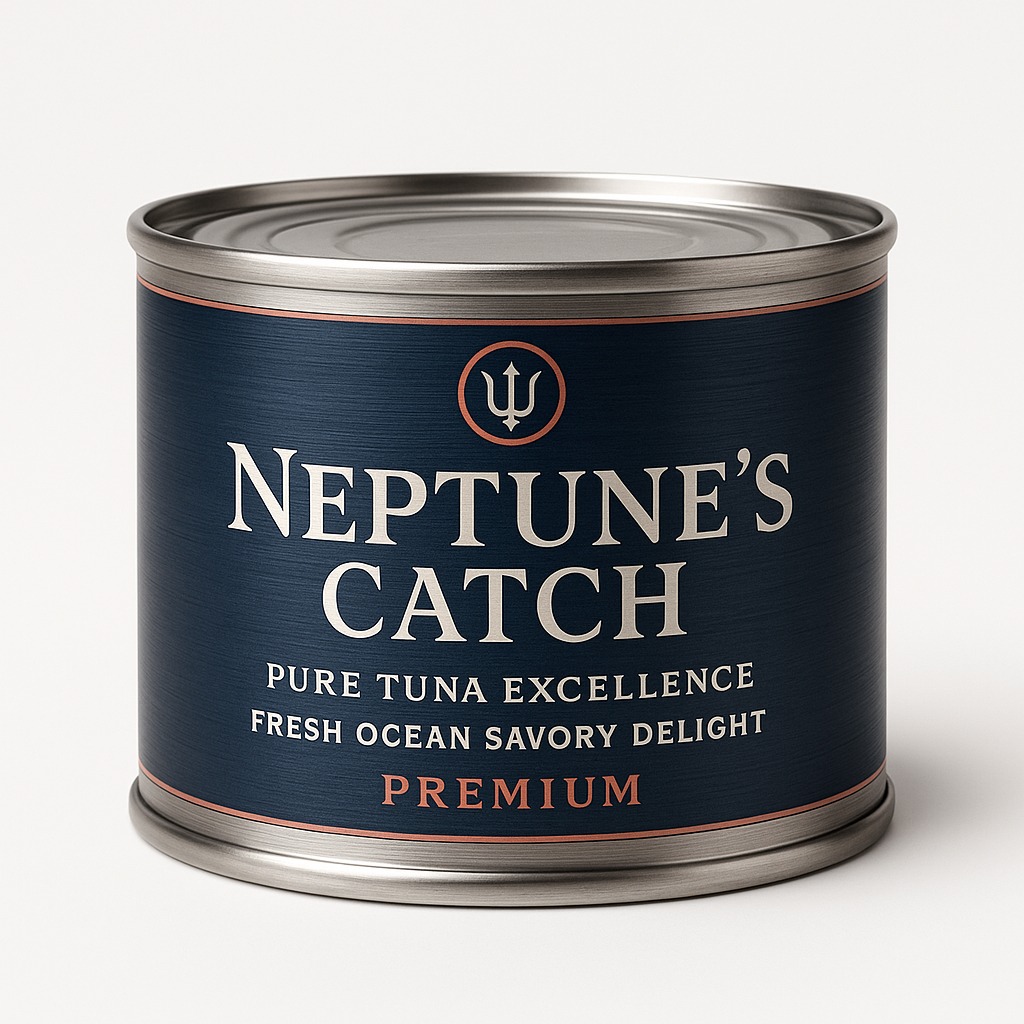}\hspace{5pt}%
\includegraphics[width=0.2\textwidth]{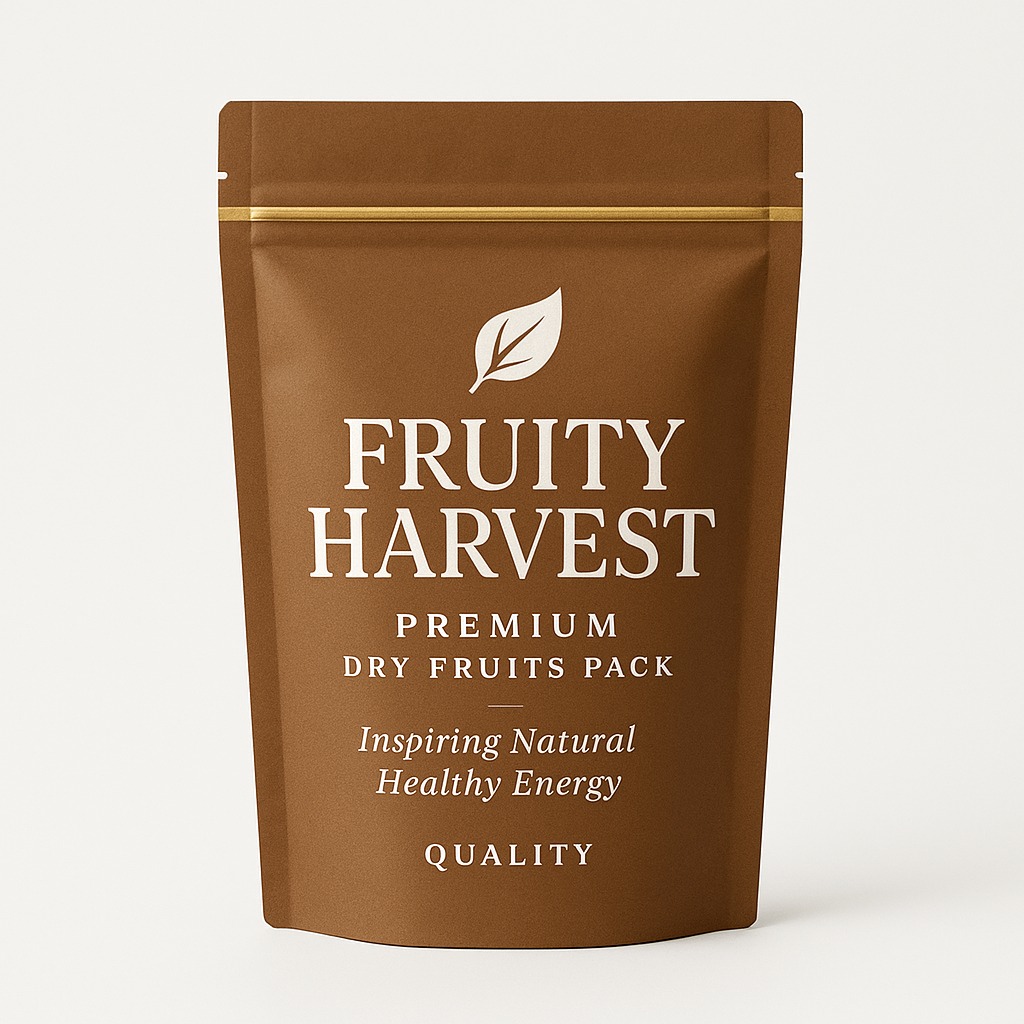}

\vspace{3pt}

\includegraphics[width=0.2\textwidth]{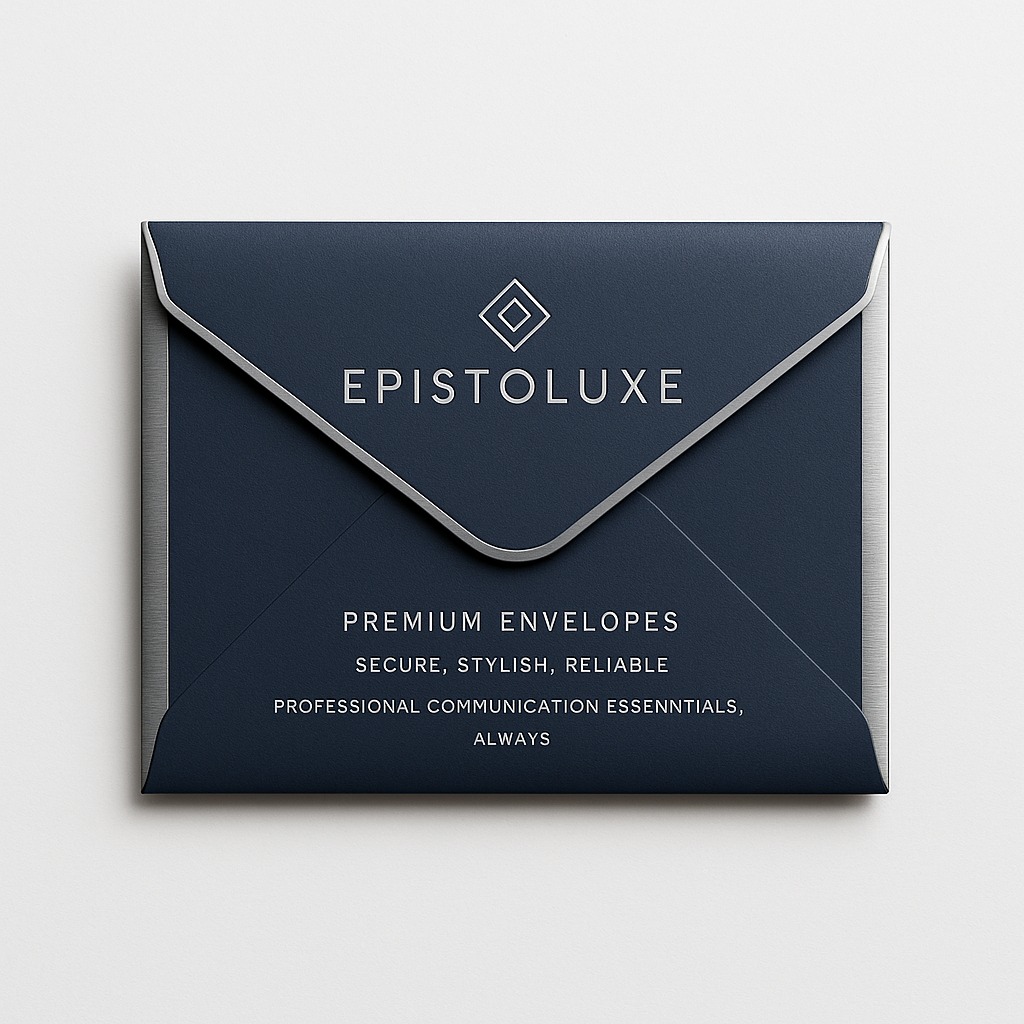}\hspace{5pt}%
\includegraphics[width=0.2\textwidth]{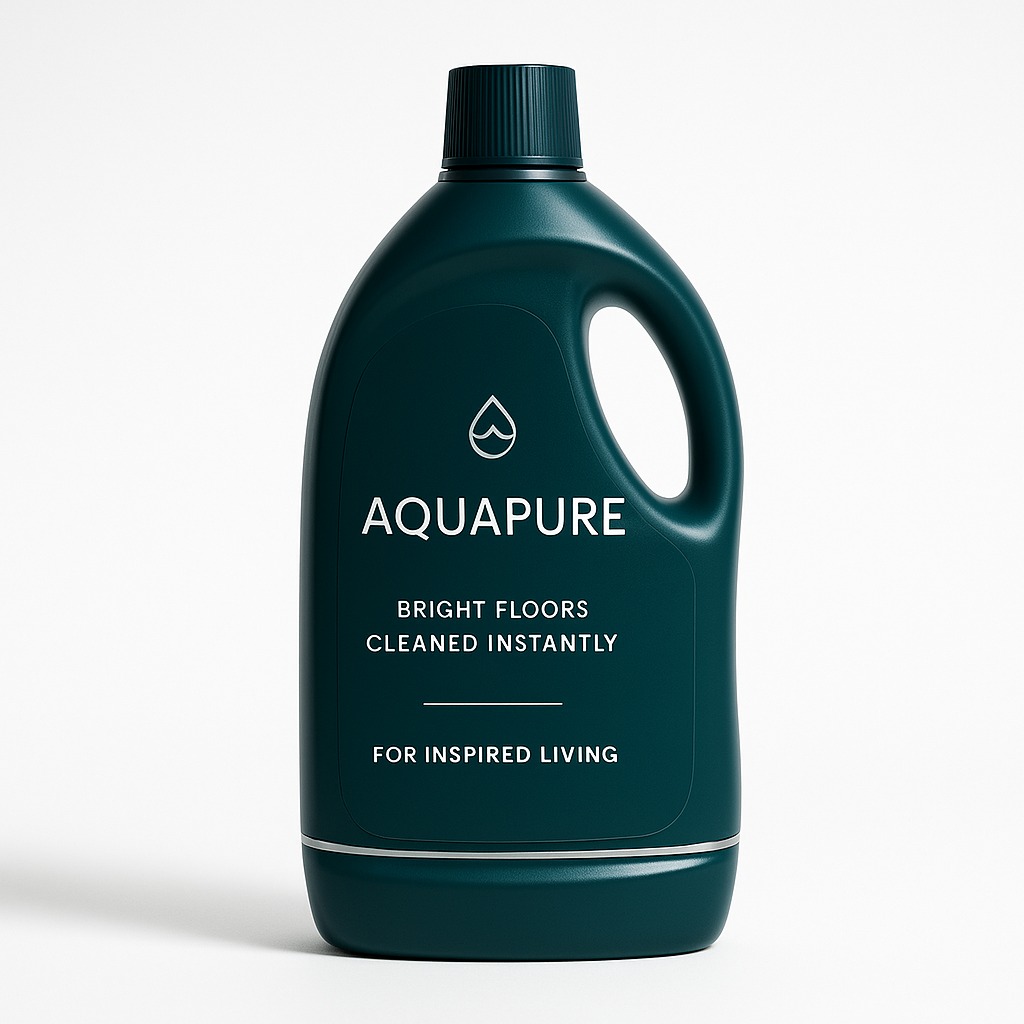}\hspace{5pt}%
\includegraphics[width=0.2\textwidth]{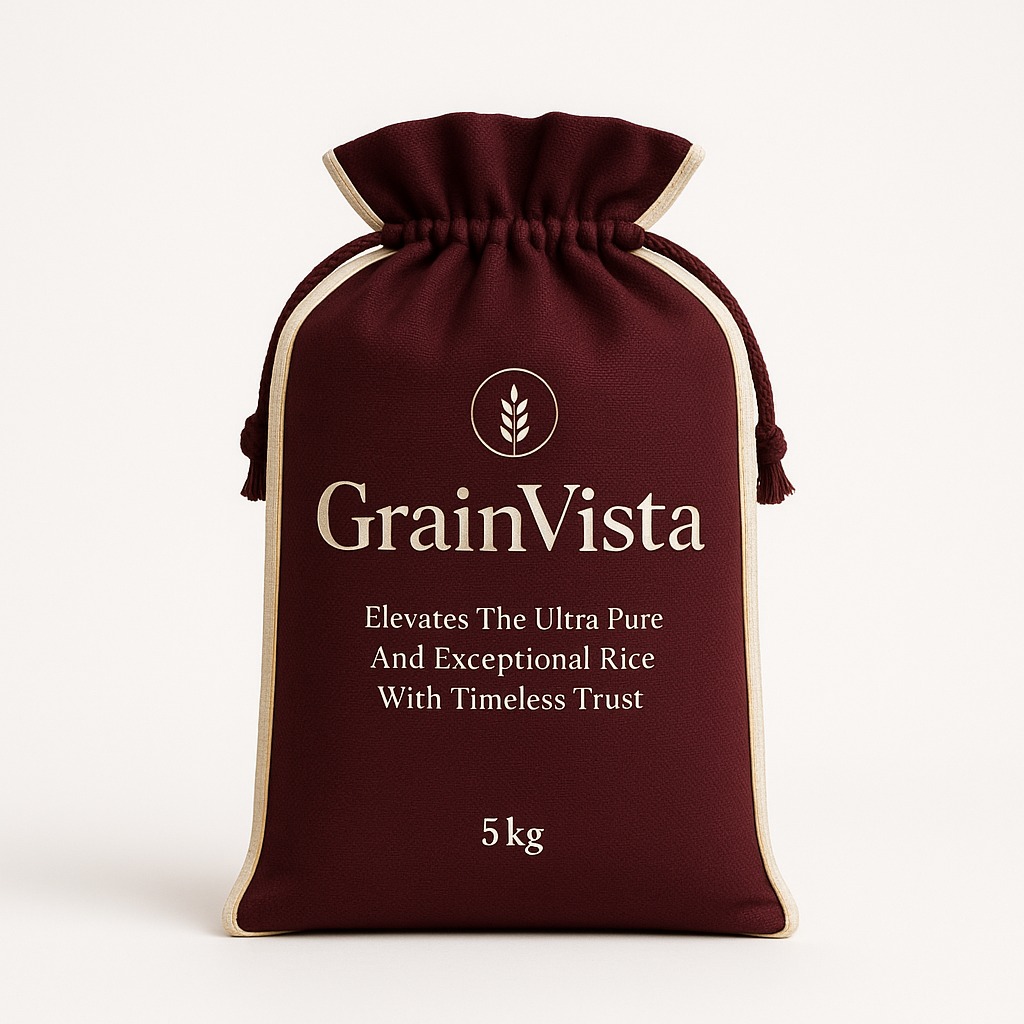}\hspace{5pt}%
\includegraphics[width=0.2\textwidth]{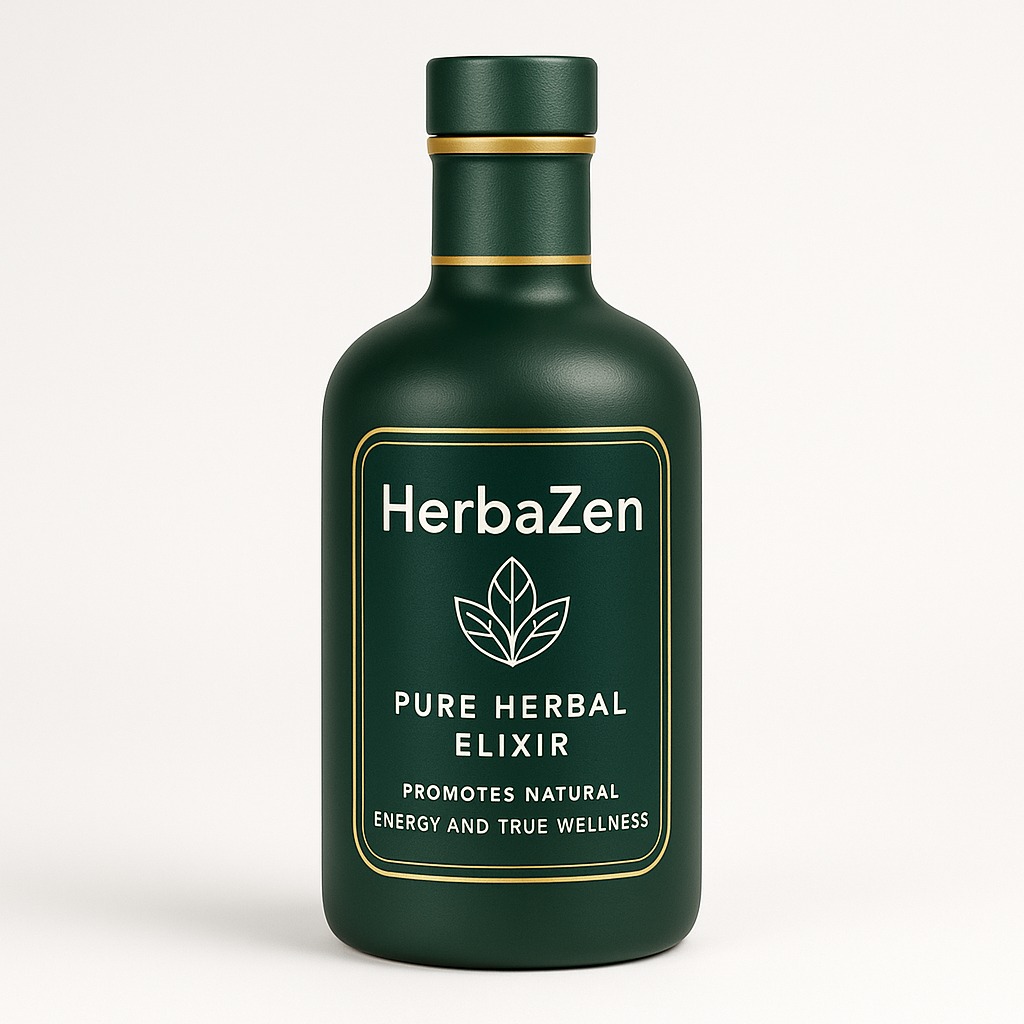}

\vspace{3pt}

\includegraphics[width=0.2\textwidth]{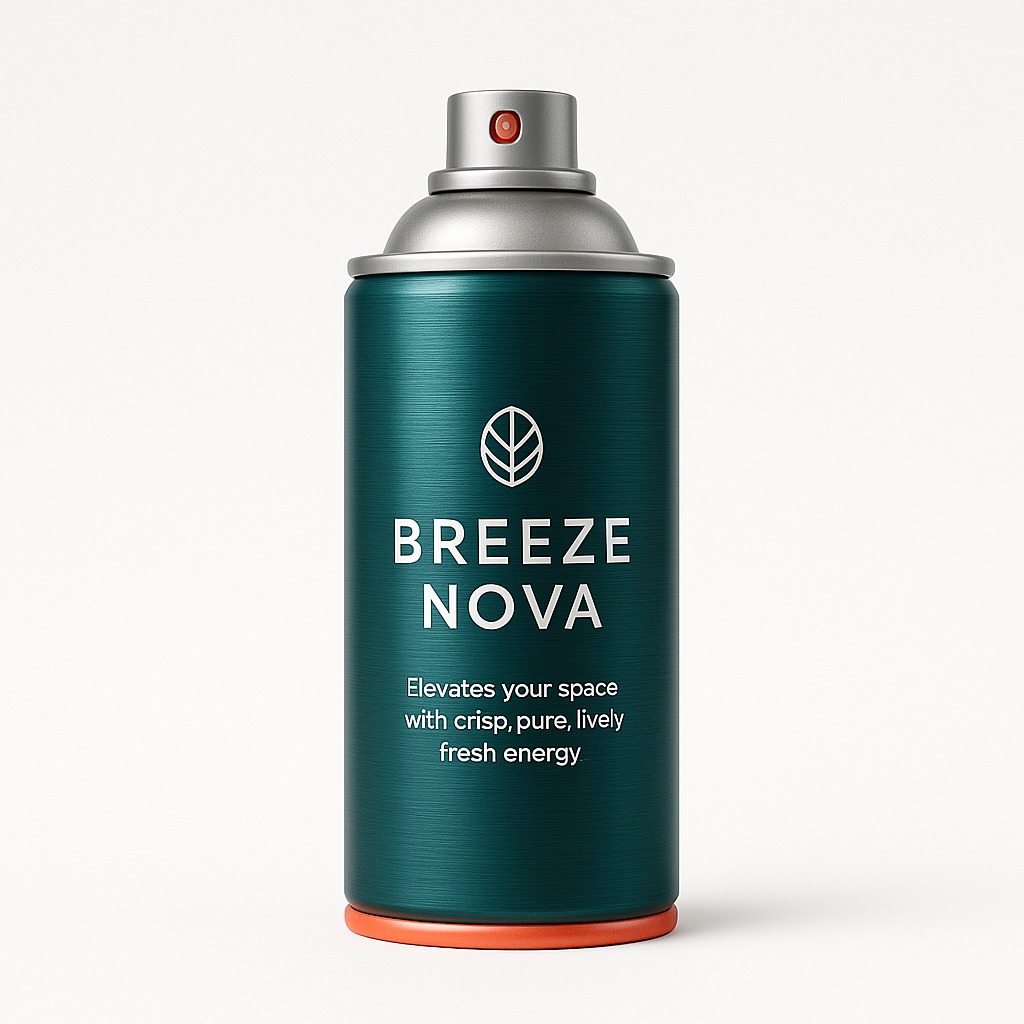}\hspace{5pt}%
\includegraphics[width=0.2\textwidth]{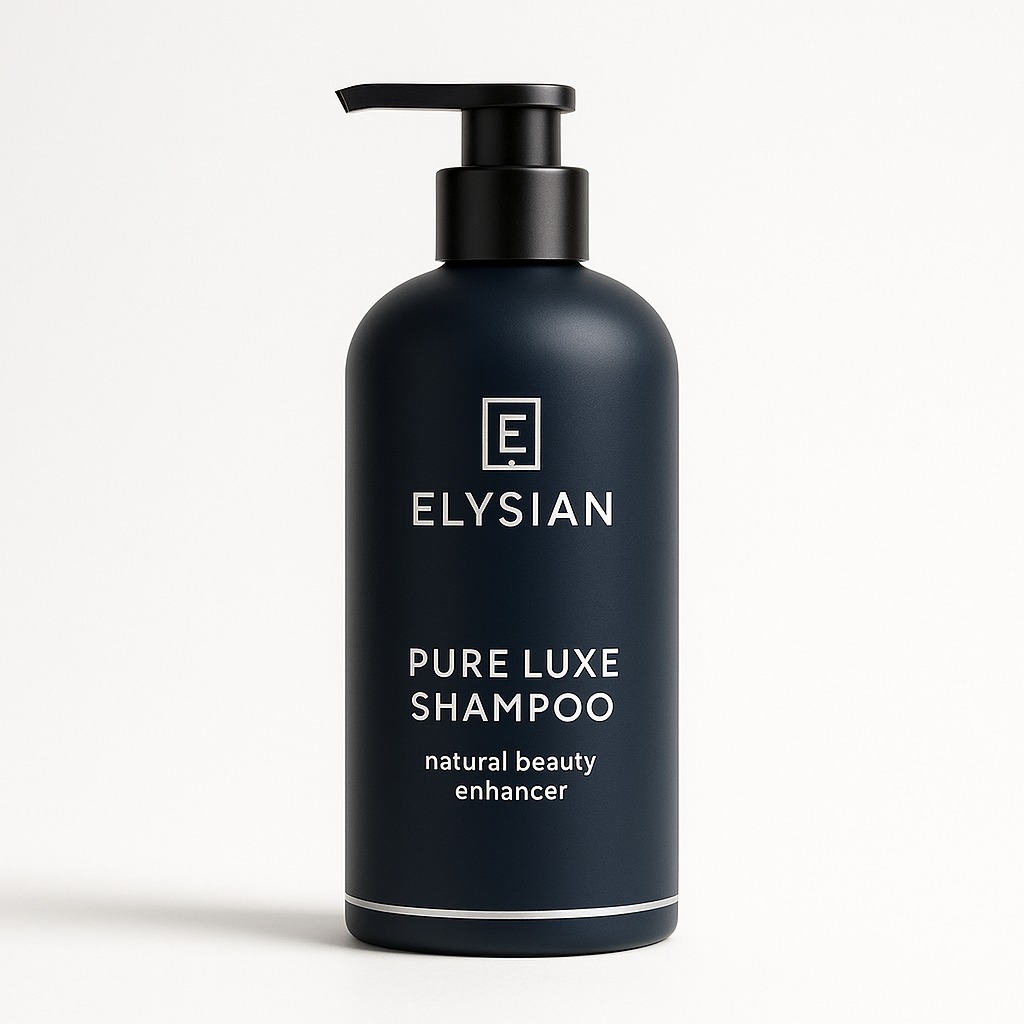}\hspace{5pt}%
\includegraphics[width=0.2\textwidth]{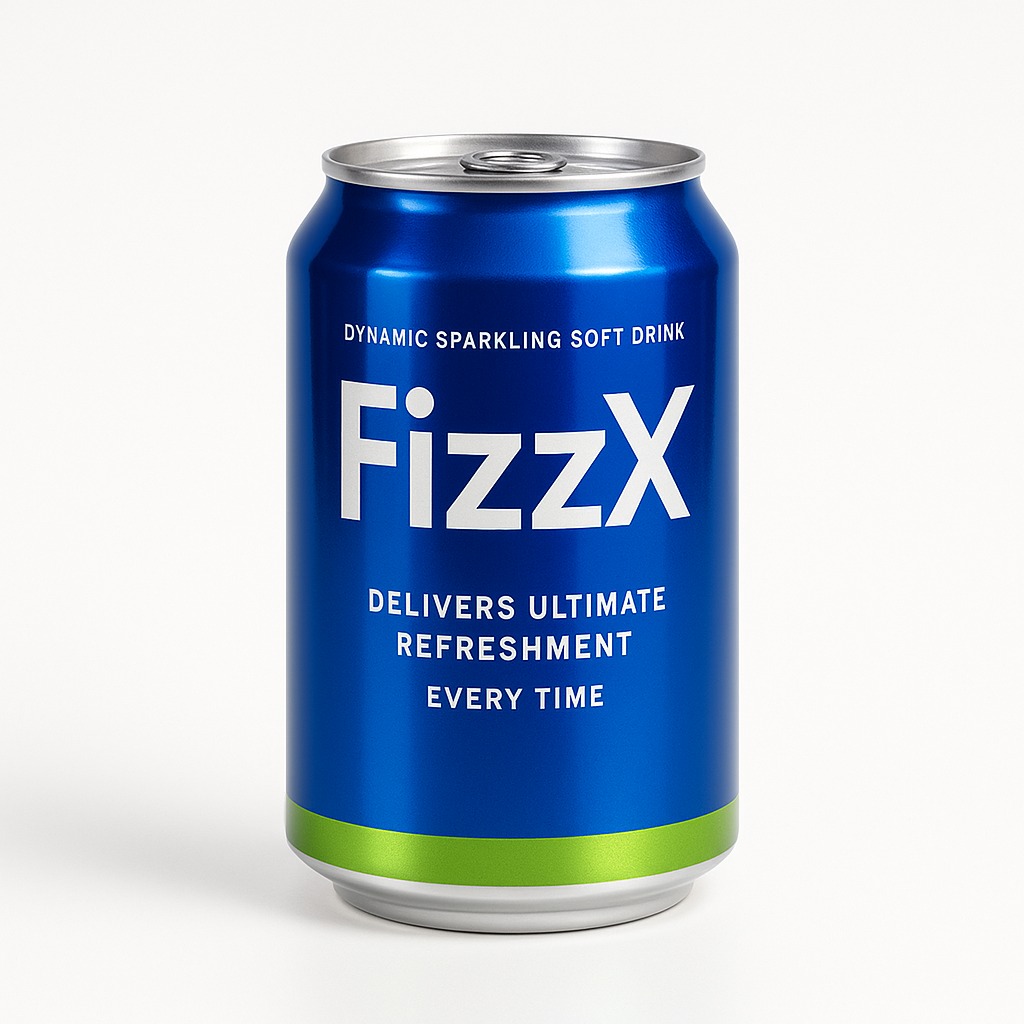}\hspace{5pt}%
\includegraphics[width=0.2\textwidth]{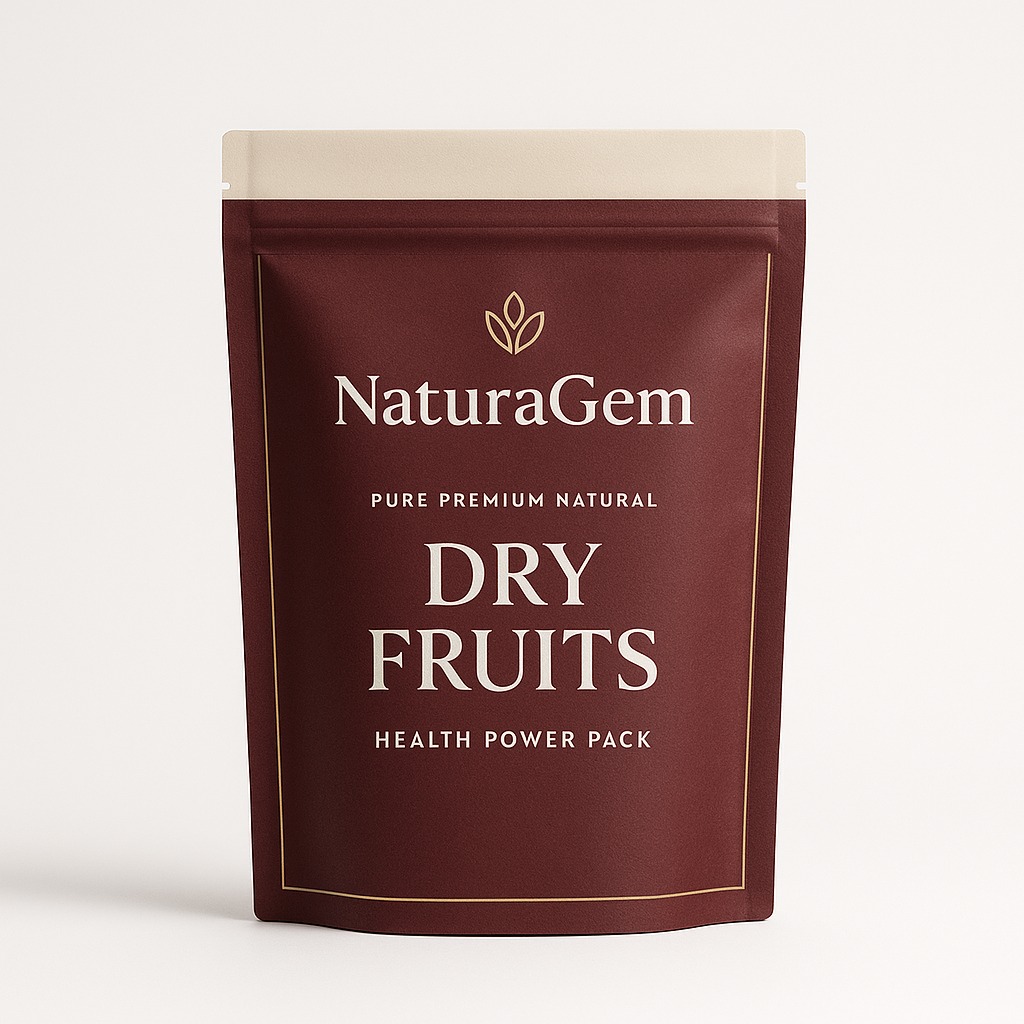}
\caption{Examples of product images from the ProductConsistency benchmark.  
Examples of all 5 Edit instructions for the first image:  
1) Place the bottle on a modern bathroom countertop with a large mirror reflecting soft morning light; include a neatly folded white towel and a small potted succulent as accents; warm ambient lighting to create a clean, inviting atmosphere; subtle reflections on the countertop to enhance the bottle's frosted finish; avoid clutter or personal items.  
2) Position the bottle outdoors on a wooden picnic table, surrounded by fresh herbs such as mint and basil; dappled sunlight filtering through tree leaves casts gentle shadows; a natural, health-focused context with soft, earthy tones; ensure the label remains clear and legible; no human presence or distracting elements.  
3) Set the bottle on a desk next to an open laptop and a steaming cup of herbal tea; create a calm, focused workspace environment with soft, indirect office lighting; background elements slightly blurred to emphasize the product; maintain a minimalist and uncluttered scene to highlight the product's sleek design; avoid cables and personal items.  
4) Display the bottle in a clean medical environment on a sterile metal tray with a few medical instruments in the periphery; bright, clinical overhead lighting; white and silver tones dominate to enhance the sense of sterility and safety; ensure the product remains central and clearly visible; avoid any clutter or brand logos.  
5) Show the bottle on a minimalist spa shelf with flickering candlelight providing a warm, soothing ambiance; include folded white linens and a small bowl of lavender buds as props; dim, calming lighting that highlights the bottle’s contours and enhances the elegant design; ensure the label remains readable and prominent; avoid any water or steam effects.}
\label{fig:benchmark_dataset_examples}
\end{figure*}

To enable a controlled and fair evaluation, we also construct the ProductConsistency benchmark by uniformly sampling product images across all categories and across the full range of rendered text lengths. We follow the same generation and OCR-filtering pipeline used for training data. We added a human verification step to ensure that the images have the correct text. For each product image, we prompt GPT to generate 5 different edit instructions unique to each product image. The resulting benchmark consists of 174 product images across 8 product categories with 5 edit instructions per image, resulting in a total of 870 samples. The category distribution for the train and test set can be found in Table \ref{tab:category_distribution} and the OCR word count distribution is present in Figure \ref{fig:data_distribution_word_count}. Some examples of the benchmark test set are shown in Figure \ref{fig:benchmark_dataset_examples}

\begin{table*}[t]
\centering
\caption{Quantitative comparison across OCR CER (Character error rate) and segmentation-based perceptual metrics along with GPT-based evaluation. Lower CER and higher Seg CLIP-I and Seg DINO-I are better. SFT and RL training with the ProductConsistency datasets show consistent improvement across both Qwen-Image-Edit-2511 and Flux.1-Kontext-dev models. The best scores in each metric are made bold. The overall score is computed by averaging the individual overall scores obtained for each run.}
\label{tab:main_results}

\resizebox{\textwidth}{!}{
\begin{tabular}{lccc|cccc}
\toprule
 & \multicolumn{3}{c|}{AI Eval} & \multicolumn{4}{c}{GPT Eval} \\
\cmidrule(r){2-4} \cmidrule(l){5-8}
\textbf{Model} & \textbf{CER $\downarrow$} & \textbf{Seg CLIP-I $\uparrow$} & \textbf{Seg DINO-I $\uparrow$} 
& \textbf{Product Consistency $\uparrow$} & \textbf{Aesthetics $\uparrow$} & \textbf{Text Fidelity $\uparrow$} & \textbf{Overall $\uparrow$} \\
\midrule

HiDream-E1-1 & 3.8774 & 0.8390 & 0.7240 & 6.5828 & 7.4134 & 3.4477 & 5.8146 \\
OmniGen2 & 1.7094 & 0.8858 & 0.7790 & 7.6739 & 7.8613 & 4.9908 & 6.8422 \\
BAGEL & 1.6810 & 0.8767 & 0.7515 & 7.7260 & 7.8088 & 6.2203 & 7.2520 \\
Step1x-edit-v1p2 & 1.1909 & 0.8626 & 0.7157 & 7.5812 & 7.4636 & 7.0414 & 7.3621 \\
RePlan-Flux & 0.2914 & 0.9174 & 0.8085 & 8.4118 & 6.8085 & 8.8727 & 8.0311 \\
RePlan-Qwen & 0.5164 & 0.9010 & 0.7419 & 7.6963 & 6.3391 & 7.6542 & 7.2298 \\
Nano Banana & 1.1868 & 0.8860 & 0.7020 & 8.8256 & 8.3552 & 8.0839 & 8.4167 \\
Qwen-Image-Lightning & 0.6073 & 0.8920 & 0.7680 & 8.5188 & 8.1941 & 8.1977 & 8.2738 \\
Edit-R1-Qwen & 0.4430 & 0.9046 & 0.7597 & 8.5834 & 8.3314 & 8.1542 & 8.3565 \\
Edit-R1-Flux & 0.1550 & 0.9195 & 0.7966 & 8.7015 & 8.0226 & 9.0142 & 8.5798 \\
GPT-Image-1 High & 0.3315 & 0.9080 & 0.7800 & \textbf{9.0134} & \textbf{8.5598} & 8.4077 & 8.6300 \\

\midrule

Qwen-Image-Edit-2511 & 1.0682 & 0.8728 & 0.7080 & 8.4578 & 8.2467 & 7.5958 & 8.1003 \\
+ SFT & 0.7803 & 0.8785 & 0.6950 & 8.2763 & 8.1571 & 7.5885 & 8.0062 \\
+ SFT + Cyclic Reward & 0.2080 & \textbf{0.9245} & 0.7990 & 8.8866 & 8.3373 & 8.8923 & \textbf{8.7055} \\

\midrule

Flux.1-Kontext-dev & 0.1490 & 0.9210 & 0.8110 & 8.7111 & 7.9506 & 8.9096 & 8.5240 \\
+ SFT & 0.1293 & 0.9216 & 0.7990 & 8.7283 & 7.9467 & 8.9797 & 8.5519 \\
+ SFT + Cyclic Reward & \textbf{0.1204} & 0.9224 & \textbf{0.8115} & 8.7996 & 7.9901 & \textbf{9.0740} & 8.6216 \\

\bottomrule
\end{tabular}
}
\end{table*}

\subsection{Model Training}

We fine-tune the Qwen-Image-Edit-2511 model and Flux.1-Kontext-dev model using Low-Rank Adaptation (LoRA) with rank $r=64$. Supervised fine-tuning is performed for one epoch with a total batch size of 96 and 4 gradient accumulation steps. A learning rate of $2 \times e^{-4}$ is used for Qwen and $1 \times e^{-5}$ for Flux. We use the AdamW optimizer in 8-bit mode and train at a resolution of $1024 \times 1024$.

For reinforcement learning, we continue training from the SFT checkpoints. We adopt the FlowGRPO\cite{liu2025flowgrpo} algorithm with mixed stochastic differential equation (SDE) and ordinary differential equation (ODE) sampling \cite{he2025tempflowgrpo,li2025mixgrpo}. Following prior observations that reward optimization does not require high-resolution images \cite{ping2025paco_rl}, training is conducted at a resolution of $512 \times 512$. We used a group size of 16, sampling 48 images per optimization step, and performing two gradient steps per epoch with a learning rate of $3 \times 10^{-4}$. We apply exponential moving average (EMA) regularization with a decay rate of $0.9$. RL checkpoints are trained until convergence based on the validation reward, and the best-performing checkpoint on the validation set is used for the evaluation. The train and validation sets are kept consistent during all training runs. All experiments were conducted on 8 Nvidia A100 80GB GPUs. 

\textbf{Reward Functions}: 
Since current reward models do not account for product consistency directly and, finetuning them would require datasets with annotated preferences. We design a proxy reward using caption similarity between the original and the generated image as a stand-in for product consistency. Specifically, we introduce a Cyclic Consistency reward. Let $c_{\text{gt}}$ denote the original product caption used to generate the input image, and let $c_{\text{gen}}$ denote the caption generated from the edited image using Qwen3-VL. We compute SigLIP-2 embeddings $\phi(\cdot)$ for both captions and define the reward as their cosine similarity:
    \[
    R_{\text{cycle}} = \left\langle \phi(c_{\text{gt}}), \phi(c_{\text{gen}}) \right\rangle.
    \]
This reward serves as a proxy for semantic product similarity. Qwen3-VL is prompted to caption the product in the center of the image. Although, this approach sometimes detects the text on the image; however, we combine this reward with an OCR based reward that separately uses Character Error Rate as defined above. We found that explicit OCR guidance leads to better performance. For reward aggregation, we use the GDPO \cite{liu2026gdpo} algorithm with equal weight to both rewards.

\section{Evaluation and Results}

We evaluated all models on the \textbf{ProductConsistency benchmark}. All images were generated using a fixed random seed to ensure reproducibility. Model specific hyper-parameters are present in Table \ref{tab:generation_settings}. We report performance across multiple metrics designed to capture product fidelity and text consistency. Text correctness is evaluated using the CER score as defined in the methodology section. Following \cite{ruiz2023dreambooth_seg_clip}, we also report Seg CLIP-I \cite{radford2021learning} and Seg DINO-I metrics to quantify the product fidelity. For these metrics, we first crop the product region from both the input image and the edited image using the product category as the tag with GroundingDino \cite{liu2024grounding} and the SAM-2 model \cite{ravi2024sam}, and then compute the localized cosine similarity between the corresponding CLIP and DINO embeddings \cite{malhi2025preserving_seg_clip,ruiz2023dreambooth_seg_clip,zhu2025multibooth_seg_clip}. 

The main results are presented in Table \ref{tab:main_results}. Earlier models such as Hidream-E1-1 and multi-modal image generators like BAGEL exhibit very high character error rates and low perceptual scores, indicating a strong inability to preserve product text and identity during editing. Step1x-edit-v1-p2 and OmniGen2 also struggle to preserve fine-grained text despite architectural improvements. Conversely, strong base editors such as Qwen-Image-Lightning and closed-source models GPT-Image-1 and Nano Banana exhibit relatively lower CER, and better perceptual similarity, but it is not sufficient for consistent product fidelity across edits. Models like RePlan that add reasoning capabilities but are not trained on product-aware data still show failures in textual fidelity and have lower aesthetics as they are not able to ground the edit instruction correctly.Edit-R1 models achieve substantial improvements in CER and perceptual metrics. However, they still fall short of models trained with our \textbf{ProductConsistency} dataset, highlighting the importance of product-aware supervision.

Fine-tuning on our ProductConsistency dataset substantially improves performance across both Qwen-Image-Edit-2511 and Flux.1-Kontext-dev models. For Qwen-Image-Edit-2511, SFT on the ProductConsistency SFT dataset reduces the CER from 1.0682 to 0.7803. Reinforcement learning using our cyclic consistency reward leads to significantly larger gains, reducing the CER to \textbf{0.2080} and increasing Seg CLIP-I to \textbf{0.9245} and Seg DINO-I to \textbf{0.7990}. This represents a nearly \textbf{5$\times$ reduction in text error} compared to the base model while simultaneously improving product fidelity. With Flux.1-Kontext-dev we observe a similar trend. Although absolute improvements are smaller due to the strong baseline performance, the reinforcement learning stage consistently improves text fidelity without degrading perceptual similarity. The Flux.1-Kontext-dev model finetuned on the ProductConsistency dataset with Cyclic consistency loss achieves the lowest CER of \textbf{0.1204} and the best DINO-I score of \textbf{0.8115}.

In addition to classical metrics, we employ an MLLM as a judge \cite{chen2024mllm, ku2024viescore} and use the GPT-5.1 model\cite{openai2025gpt51} as an evaluator to assess generated images along three qualitative axes: product consistency, Text fidelity, and visual aesthetics and instruction following. Text fidelity judges not only character correctness but also text color, font type, and text placement. The complete system prompt is presented in Figure \ref{fig:eval_system_prompt}. The scores are averaged over 3 runs with the temperature as 0 to ensure reproducibility. We prompt GPT to output the reasoning for its evaluation followed by the score. 
\begin{table*}[t]
\centering
\caption{Ablation of reward training strategies on Qwen-Image-Edit-2511. 
Segmented visual consistency achieves the best scores on automated metrics but was observed to overfit to these evaluation metrics due to reward hacking. GPT evaluation showed that the aesthetics, prompt following and composition are much worse than base model, thus making the model unusable for downstream product advertisement tasks. The overall score is computed by averaging the individual overall scores obtained for each run.}
\label{tab:reward_ablation}

\resizebox{0.90\textwidth}{!}{
\begin{tabular}{lccc|cccc}
\toprule
& \multicolumn{3}{c}{\textbf{AI Metrics Eval}} & \multicolumn{4}{c}{\textbf{GPT Eval}} \\
\cmidrule(r){2-4} \cmidrule(l){5-8}
\textbf{Model} & \textbf{CER $\downarrow$} & \textbf{Seg CLIP-I $\uparrow$} & \textbf{Seg DINO-I $\uparrow$} 
& \textbf{Product Consistency $\uparrow$} & \textbf{Aesthetics $\uparrow$} & \textbf{Text Fidelity $\uparrow$} & \textbf{Overall $\uparrow$} \\
\midrule

Qwen-Image-Edit-2511 & 1.0682 & 0.8728 & 0.7080
& 8.4578 & 8.2467 & 7.5958 & 8.1003 \\

\midrule

EditReward & 0.2765 & 0.9190 & 0.7805 
& 8.7586 & 8.3138 & 8.9533 & 8.6752 \\

Seg Visual Consistency & \textbf{0.2069} & \textbf{0.9540} & \textbf{0.8800} 
& \textbf{9.0916} & 5.3919 & \textbf{9.2115} & 7.8980 \\

Cyclic & 0.2080 & 0.9245 & 0.7990
& 8.8866 & \textbf{8.3373} & 8.8923 & \textbf{8.7055} \\

\bottomrule
\end{tabular}
}
\end{table*}

\begin{figure*}[t]
  \centering
  \setlength{\tabcolsep}{2pt}
  \renewcommand{\arraystretch}{1.04}

  \resizebox{\textwidth}{!}{%
  \begin{tabular}{@{}ccccccc@{}}
    \toprule
    \textbf{Input} & \textbf{Step1x-Edit} & \textbf{HiDream-E1-1} &
    \textbf{Qwen-Image-Lighting} & \textbf{BAGEL} &
    \textbf{Nano Banana} & \textbf{GPT-Image-1 High} \\
    \midrule

    \includegraphics[width=0.16\linewidth]{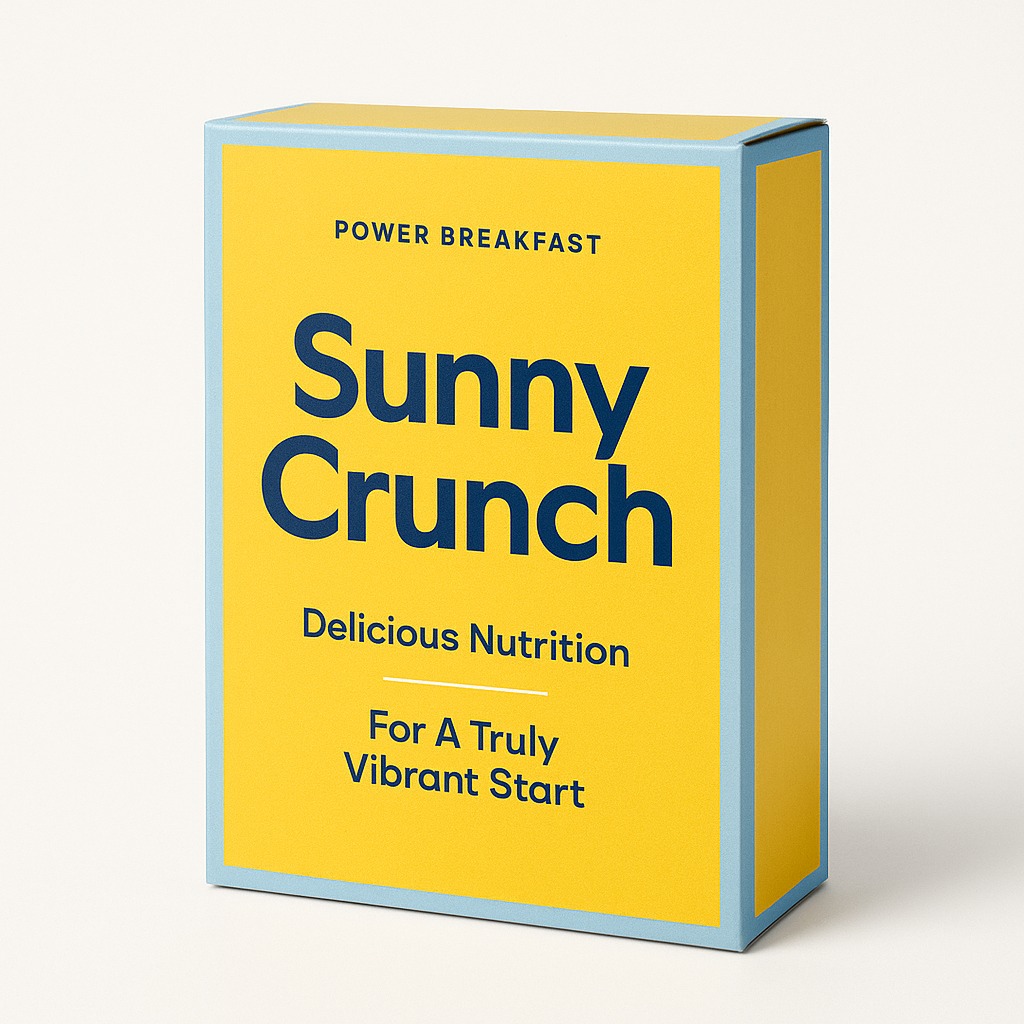} &
    \includegraphics[width=0.16\linewidth]{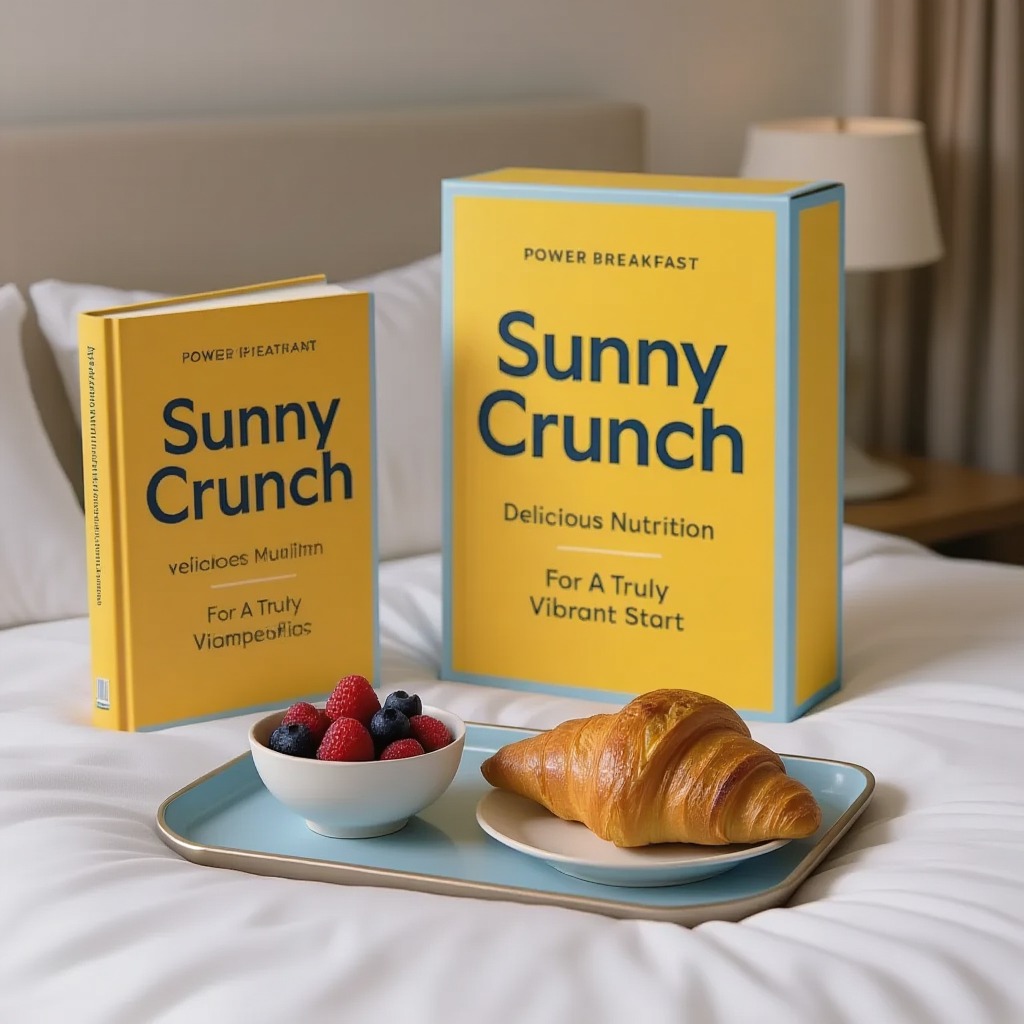} &
    \includegraphics[width=0.16\linewidth]{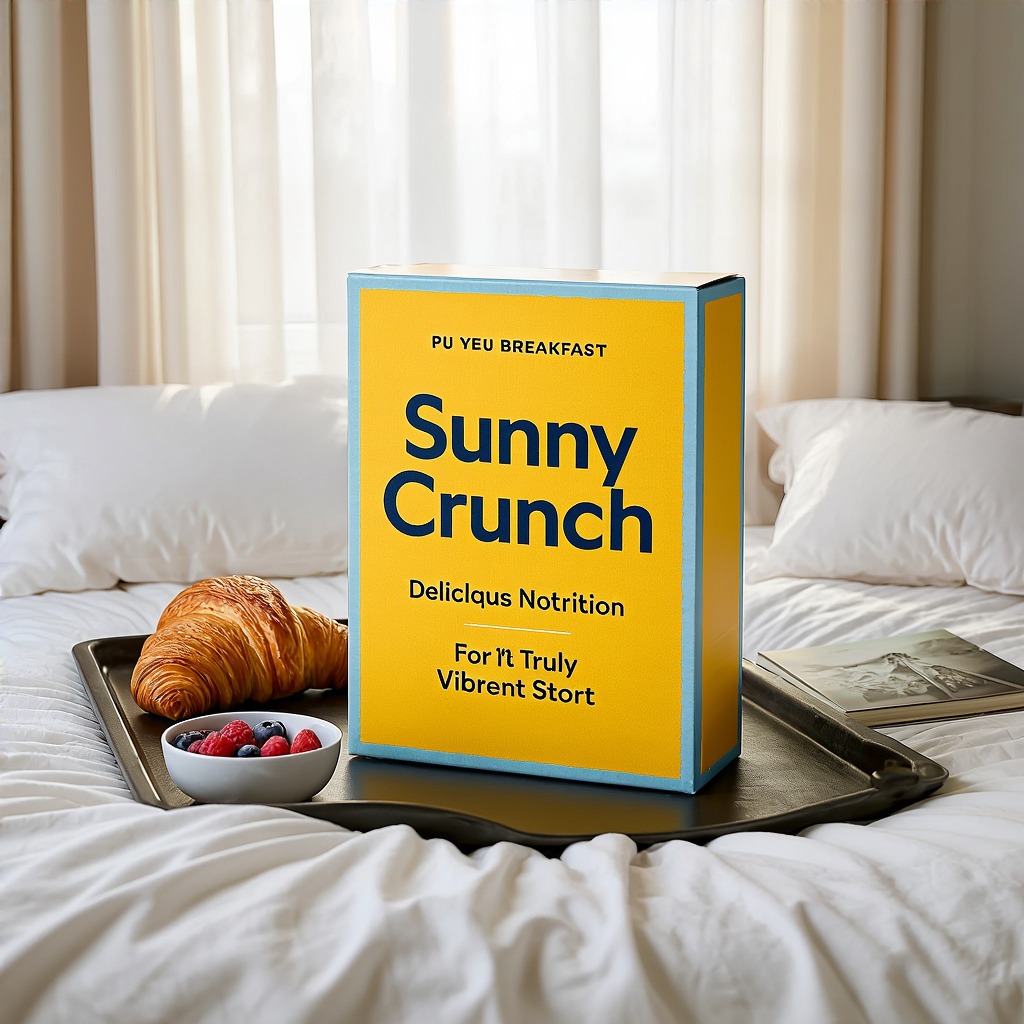} &
    \includegraphics[width=0.16\linewidth]{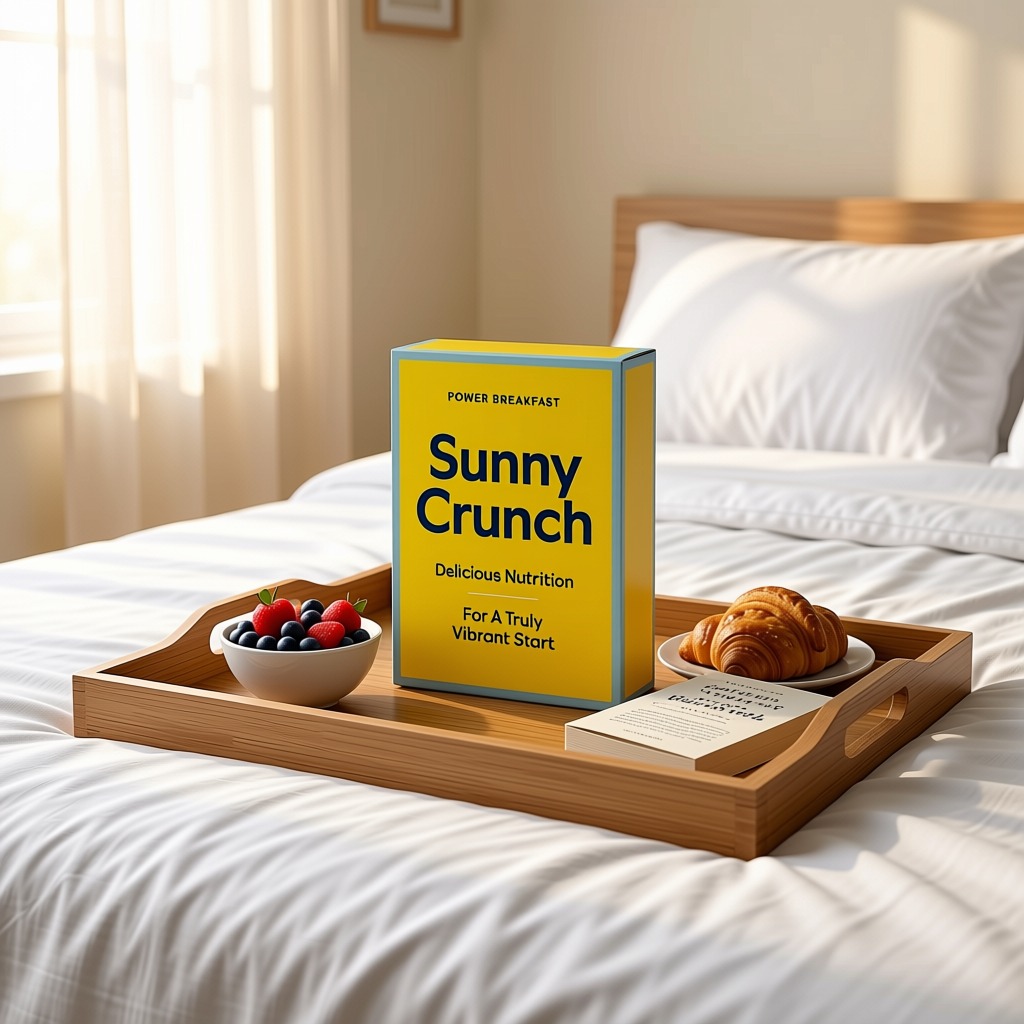} &
    \includegraphics[width=0.16\linewidth]{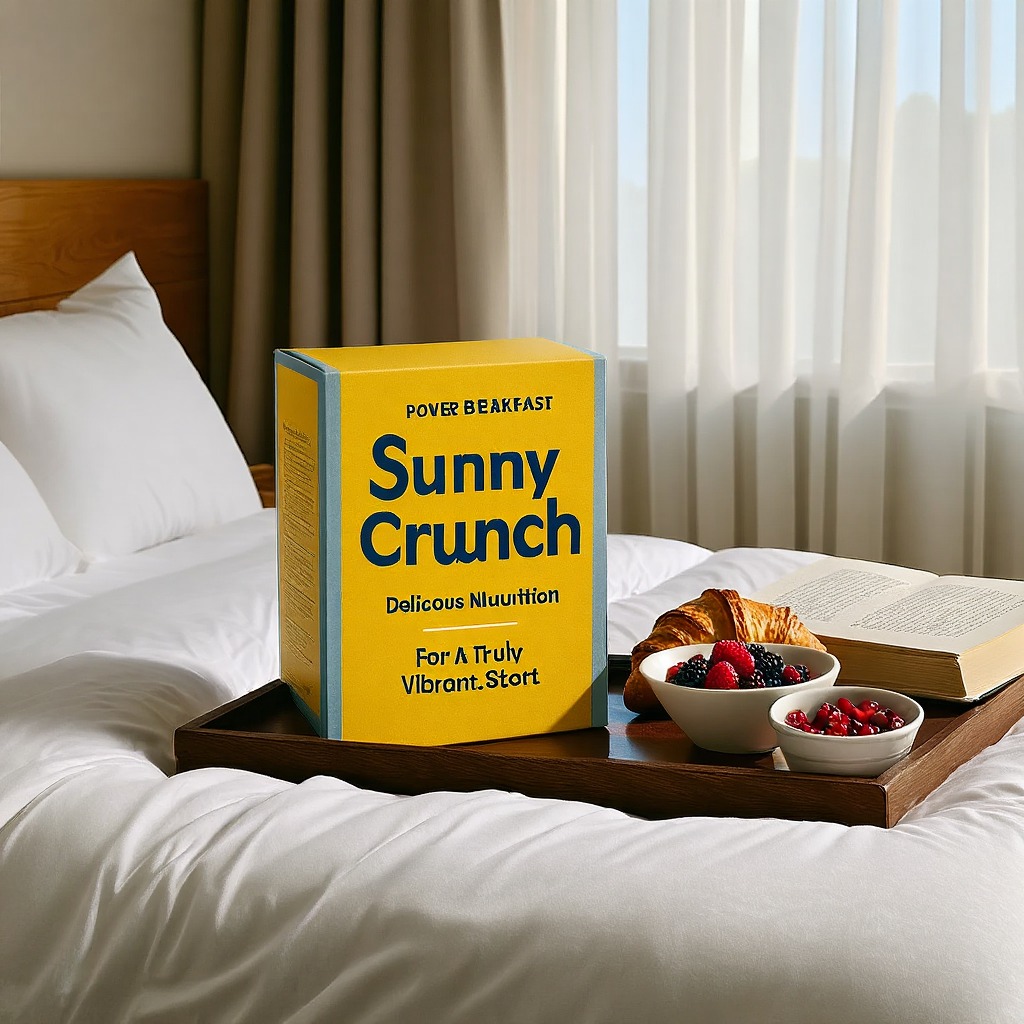} &
    \includegraphics[width=0.16\linewidth]{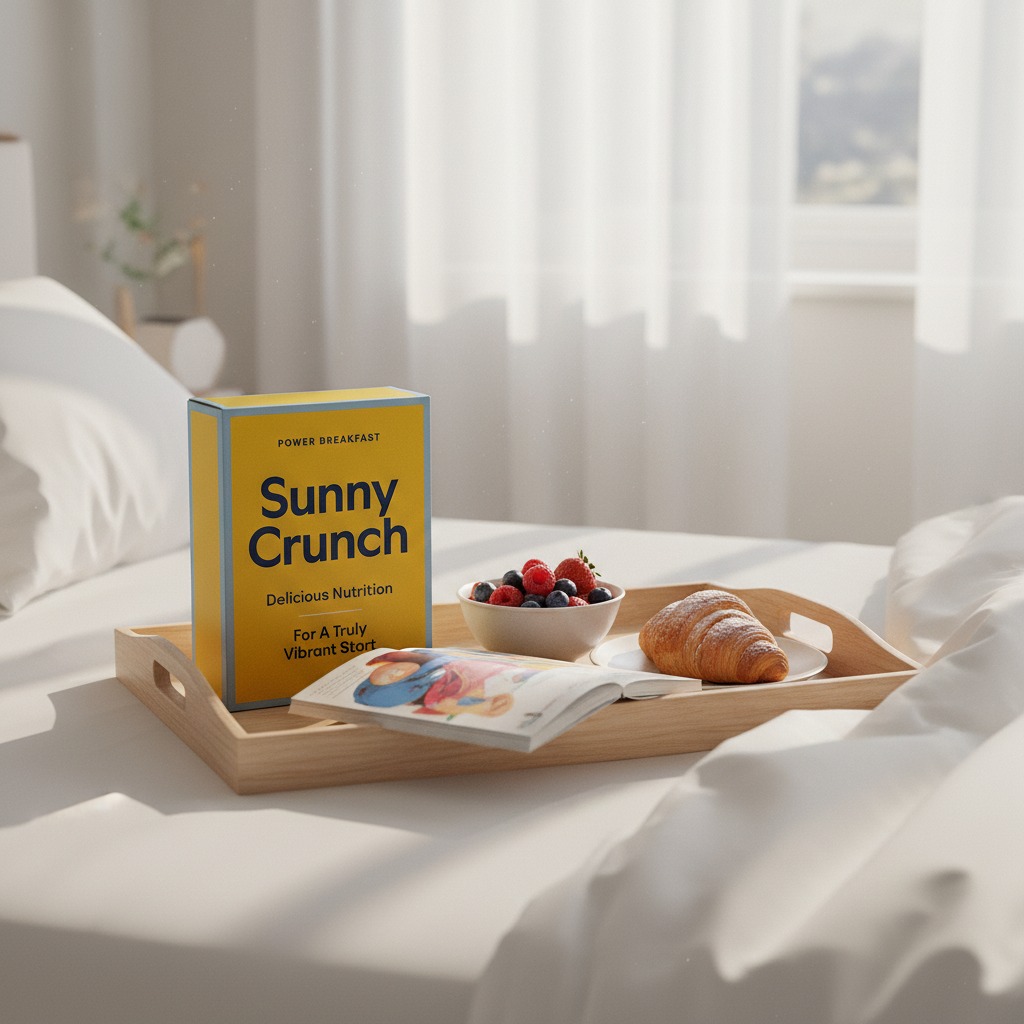} &
    \includegraphics[width=0.16\linewidth]{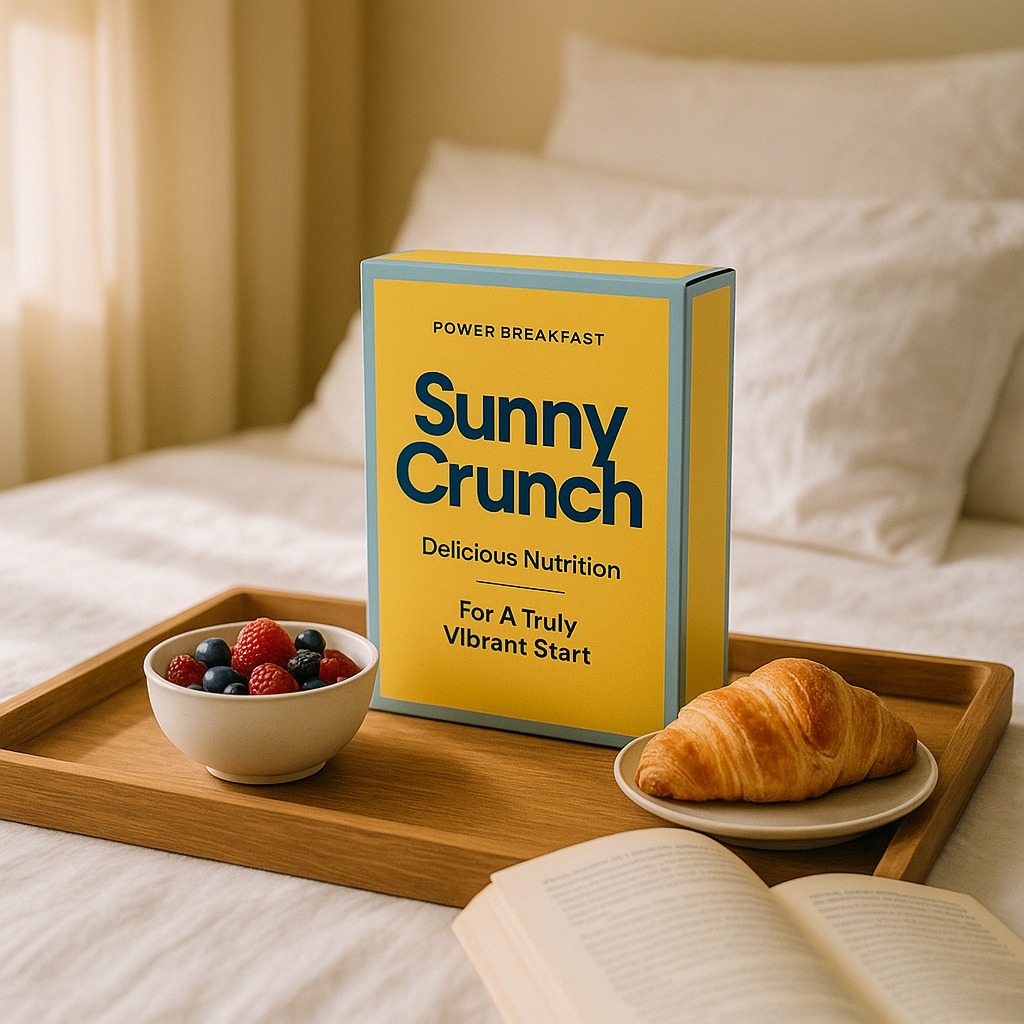} \\
    \multicolumn{7}{p{\textwidth}}{\scriptsize\emph{Feature the cereal box on a breakfast tray on a neatly made bed with soft white linens; include a small bowl of berries, a croissant, and a novel as supporting elements; gentle morning light filtering through sheer curtains for a cozy, indulgent mood; keep the composition balanced and the product sharply in focus.}} \\[2pt]

    \includegraphics[width=0.16\linewidth]{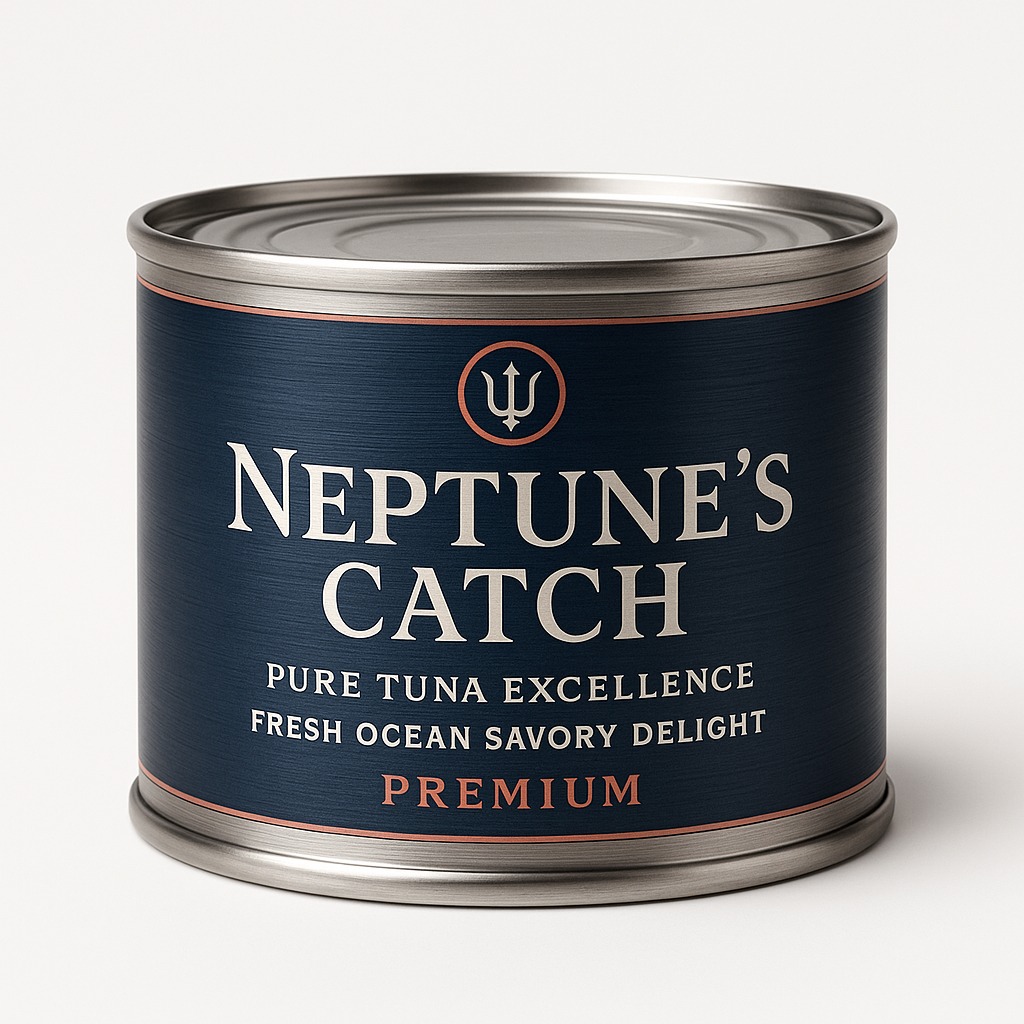} &
    \includegraphics[width=0.16\linewidth]{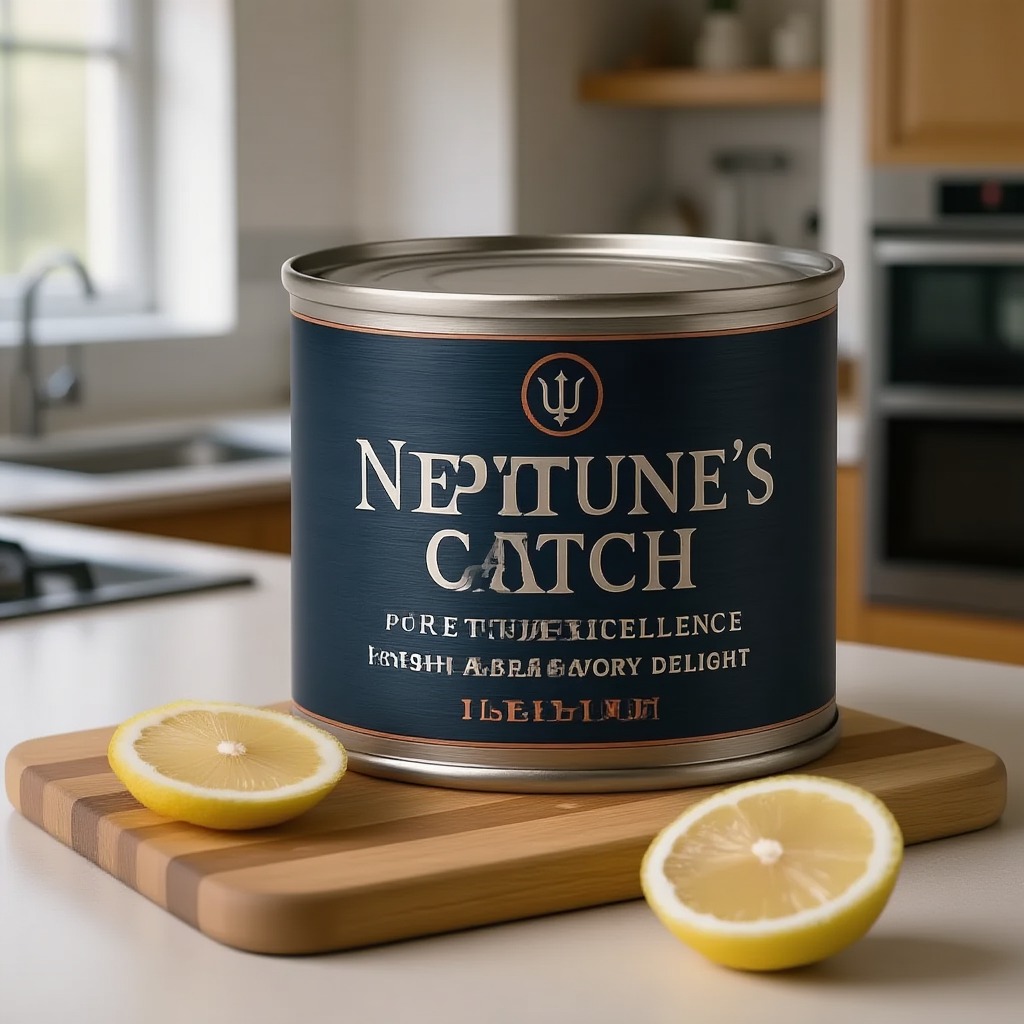} &
    \includegraphics[width=0.16\linewidth]{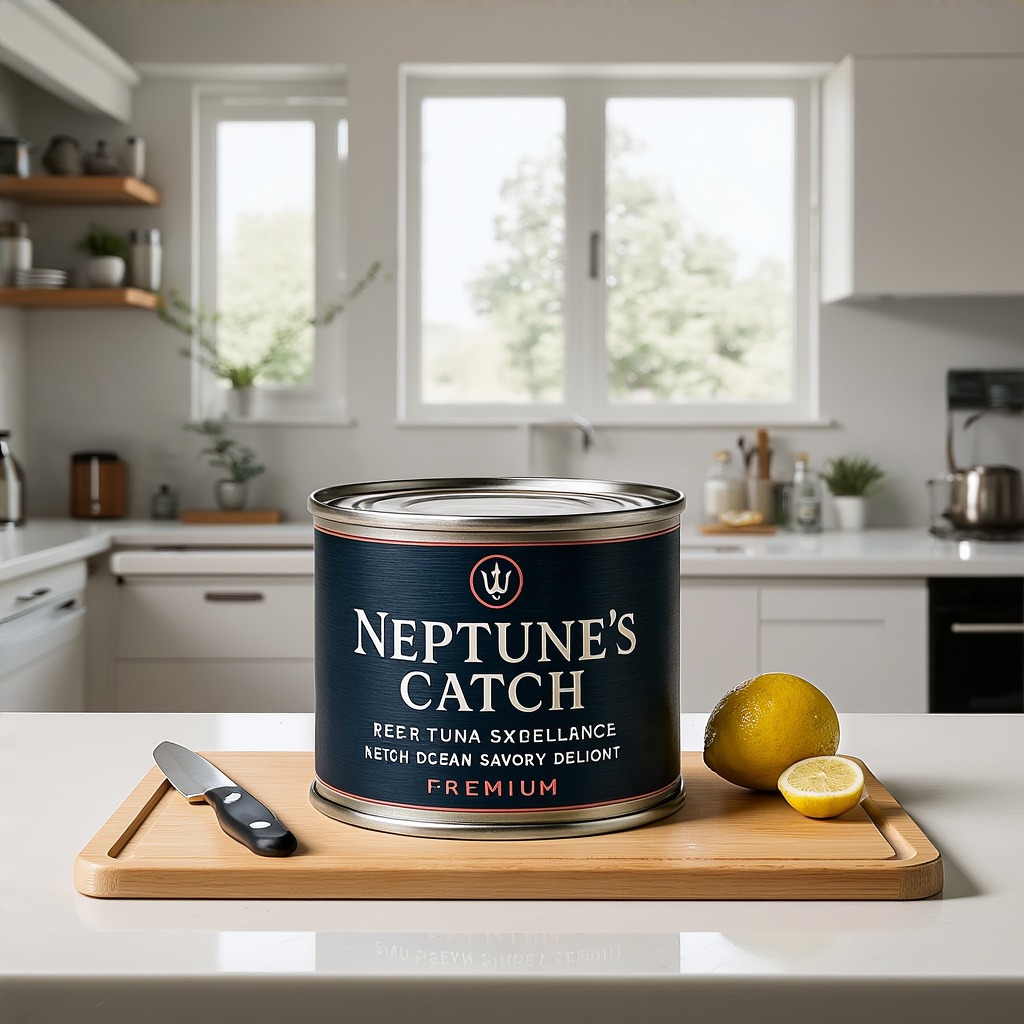} &
    \includegraphics[width=0.16\linewidth]{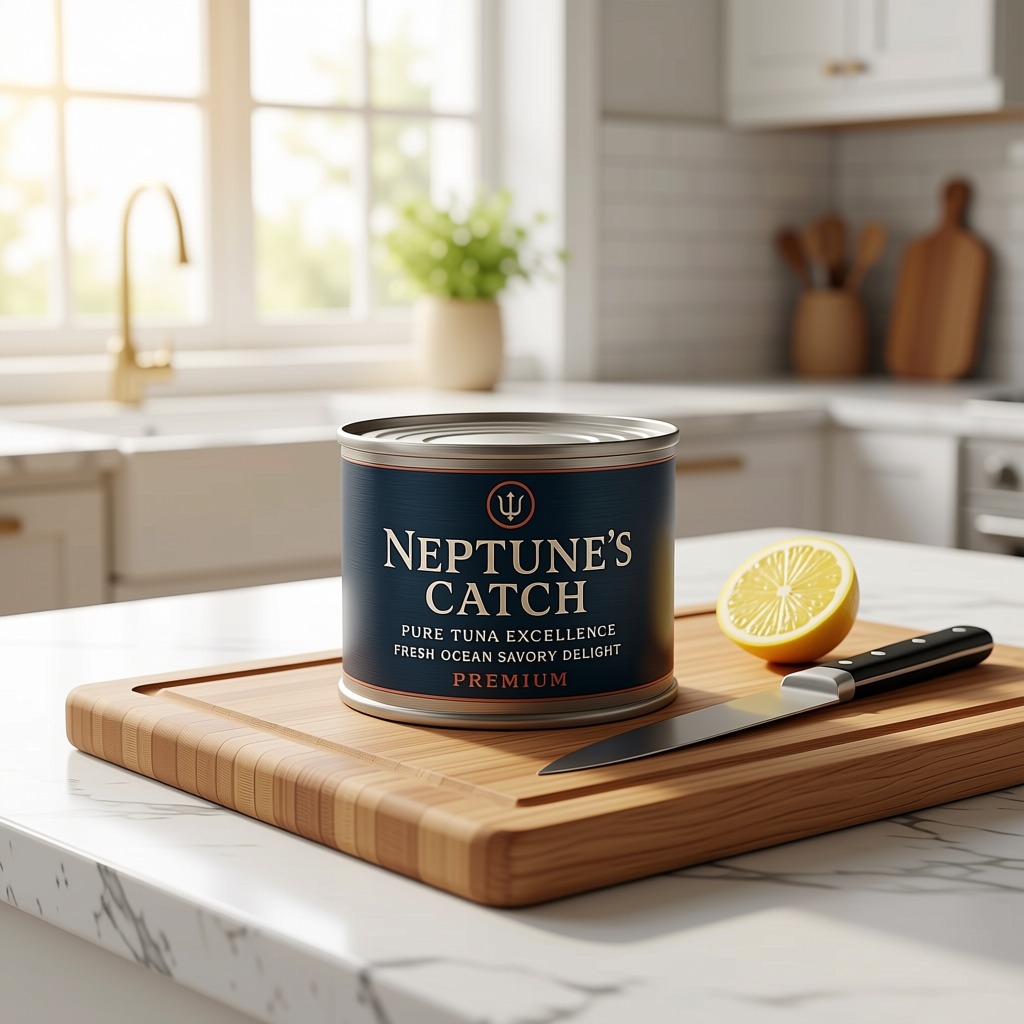} &
    \includegraphics[width=0.16\linewidth]{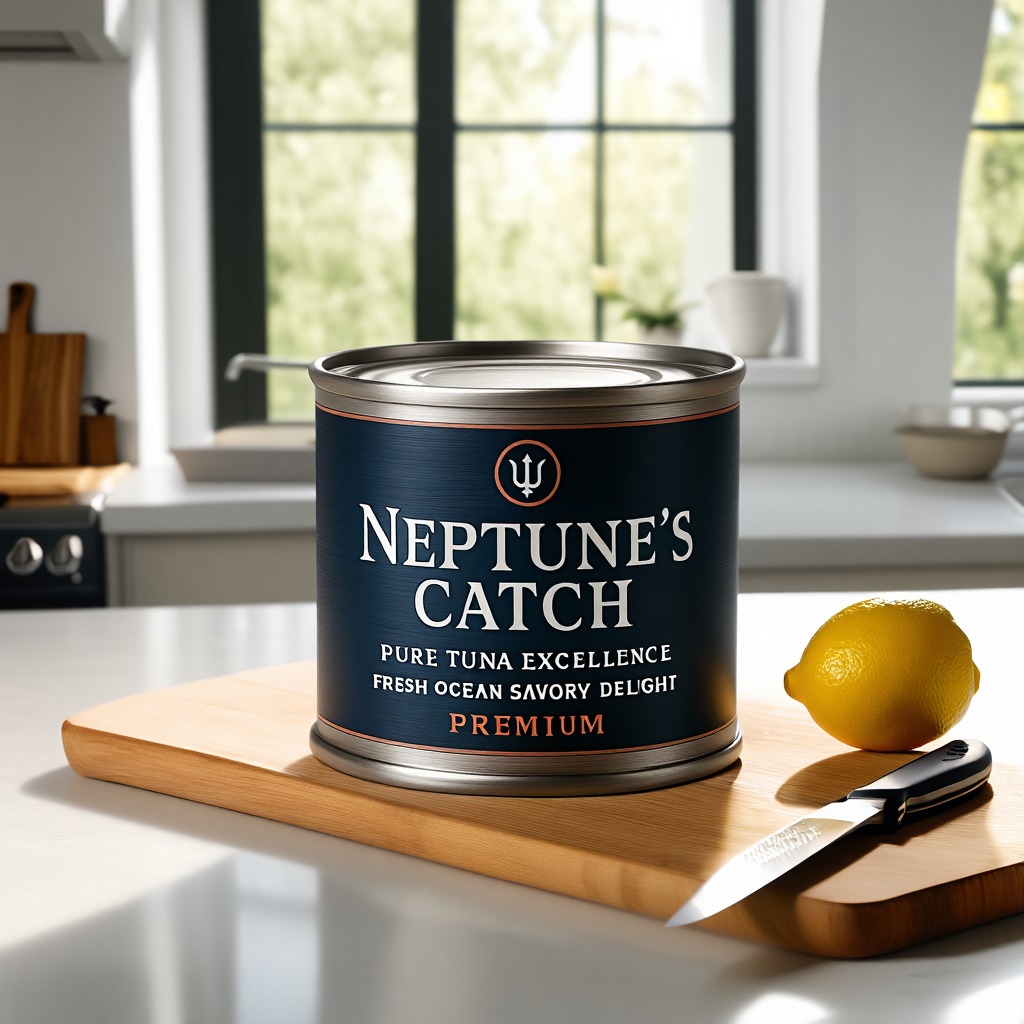} &
    \includegraphics[width=0.16\linewidth]{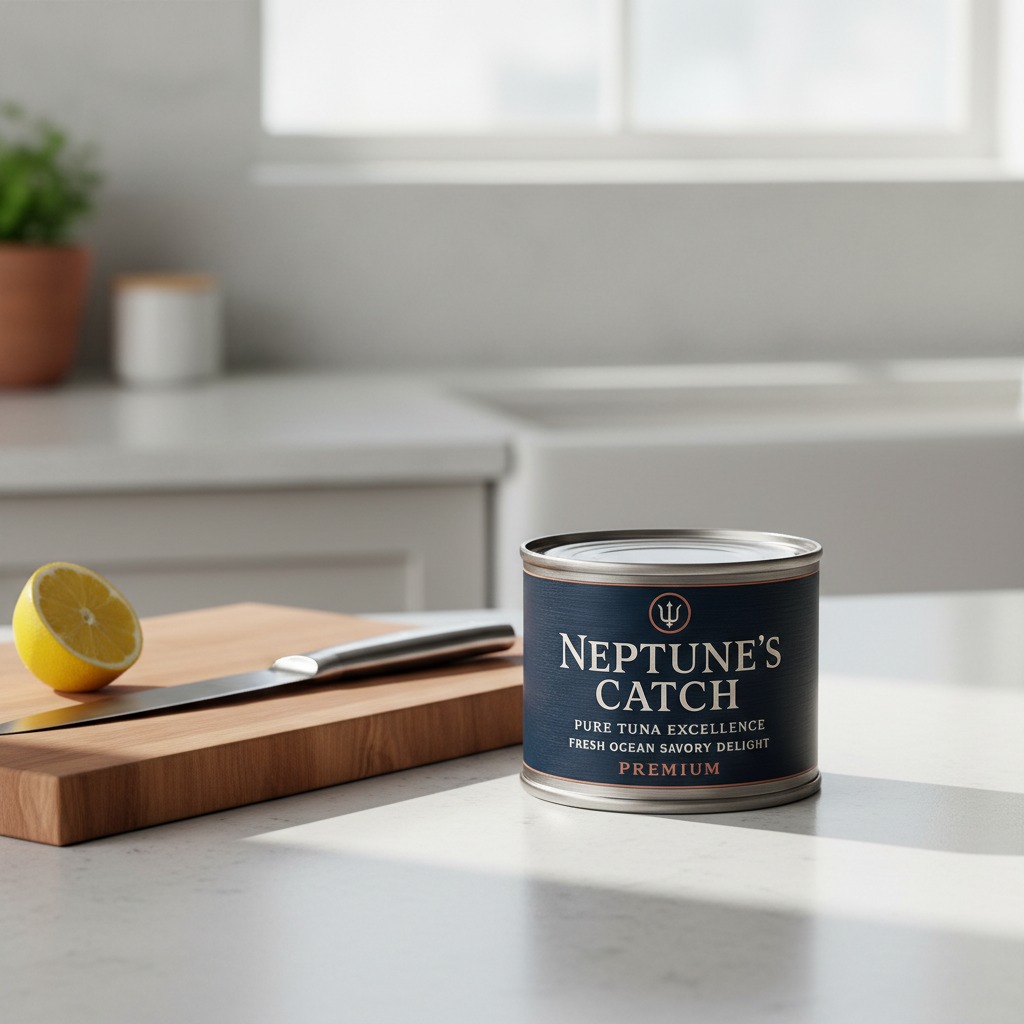} &
    \includegraphics[width=0.16\linewidth]{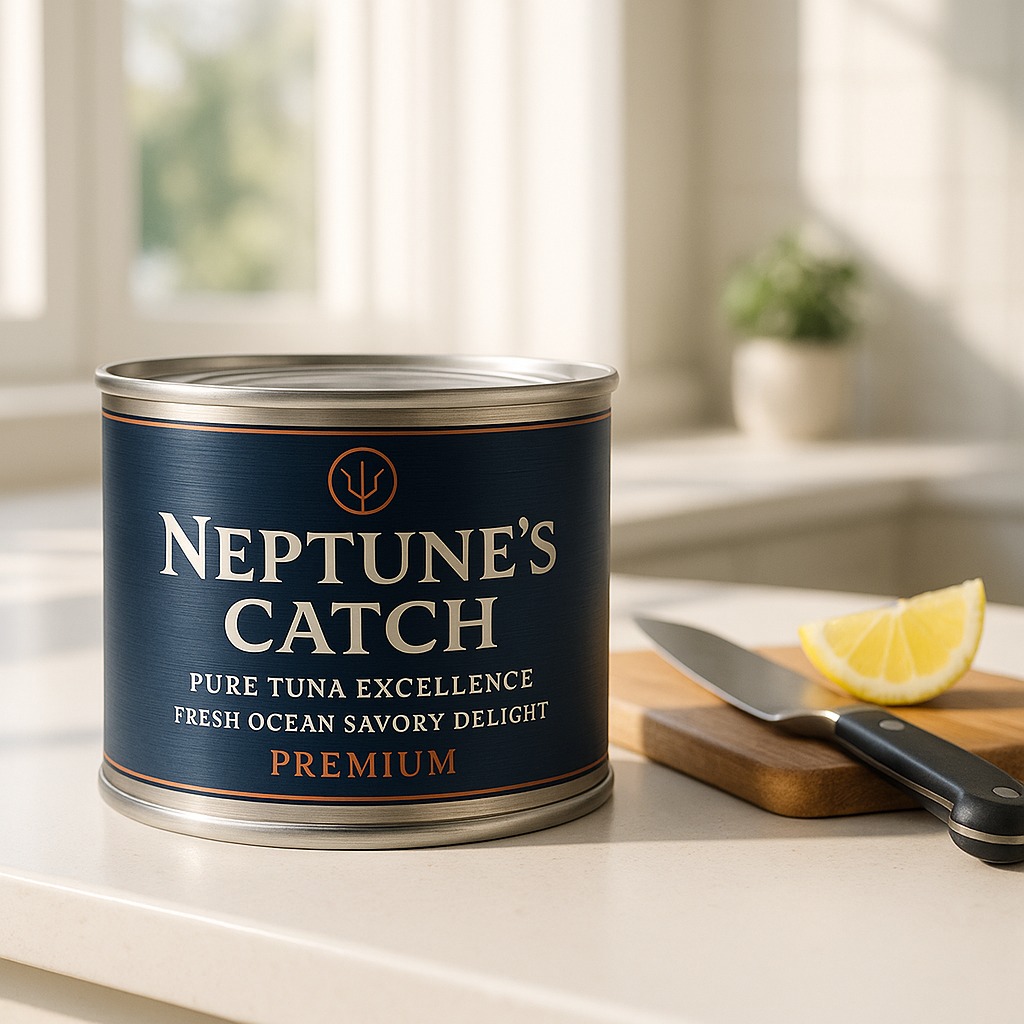} \\
    \multicolumn{7}{p{\textwidth}}{\scriptsize\emph{Position the can on a clean, minimalist kitchen countertop; include a high-quality wooden cutting board with a knife and a lemon slice as props; bathe the scene in soft, ambient daylight from a large kitchen window; ensure the product is hero-lit, with focus on the label and metallic finishes.}} \\[2pt]

    \includegraphics[width=0.16\linewidth]{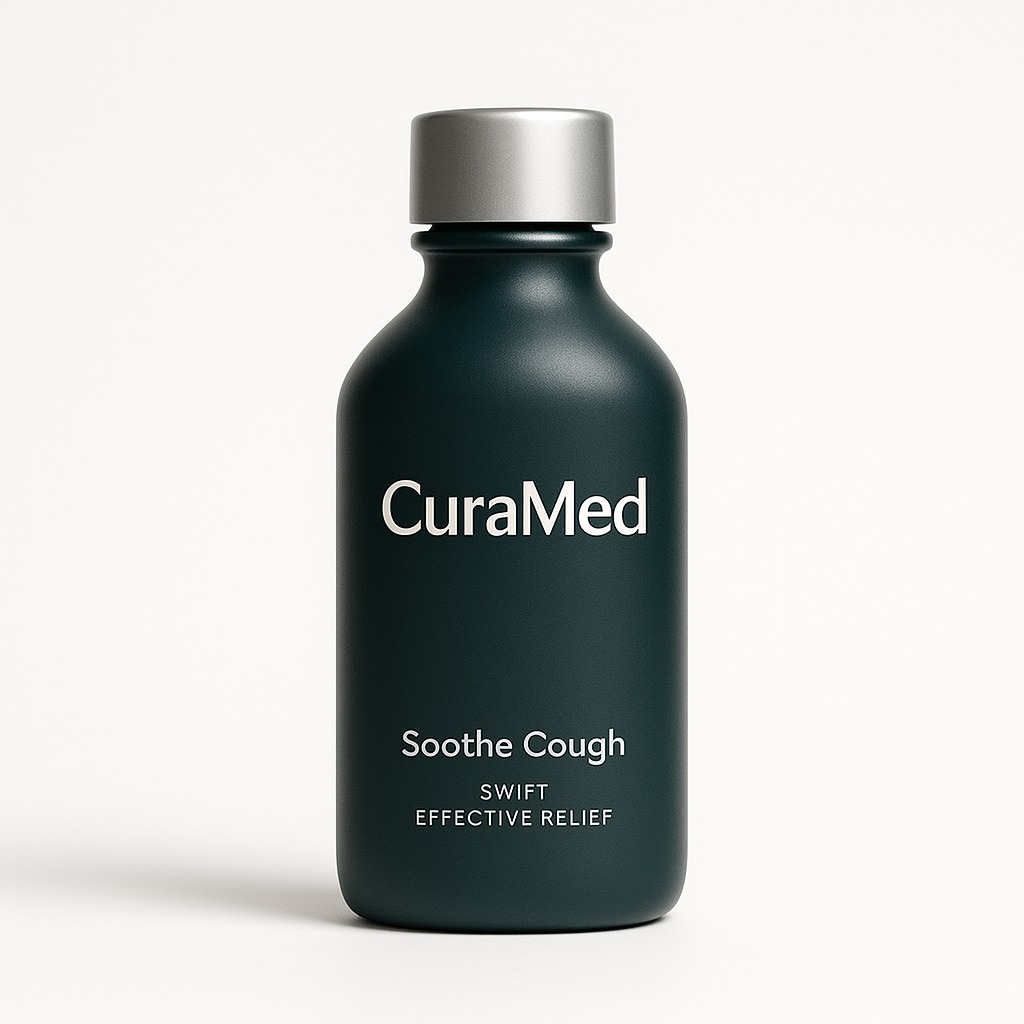} &
    \includegraphics[width=0.16\linewidth]{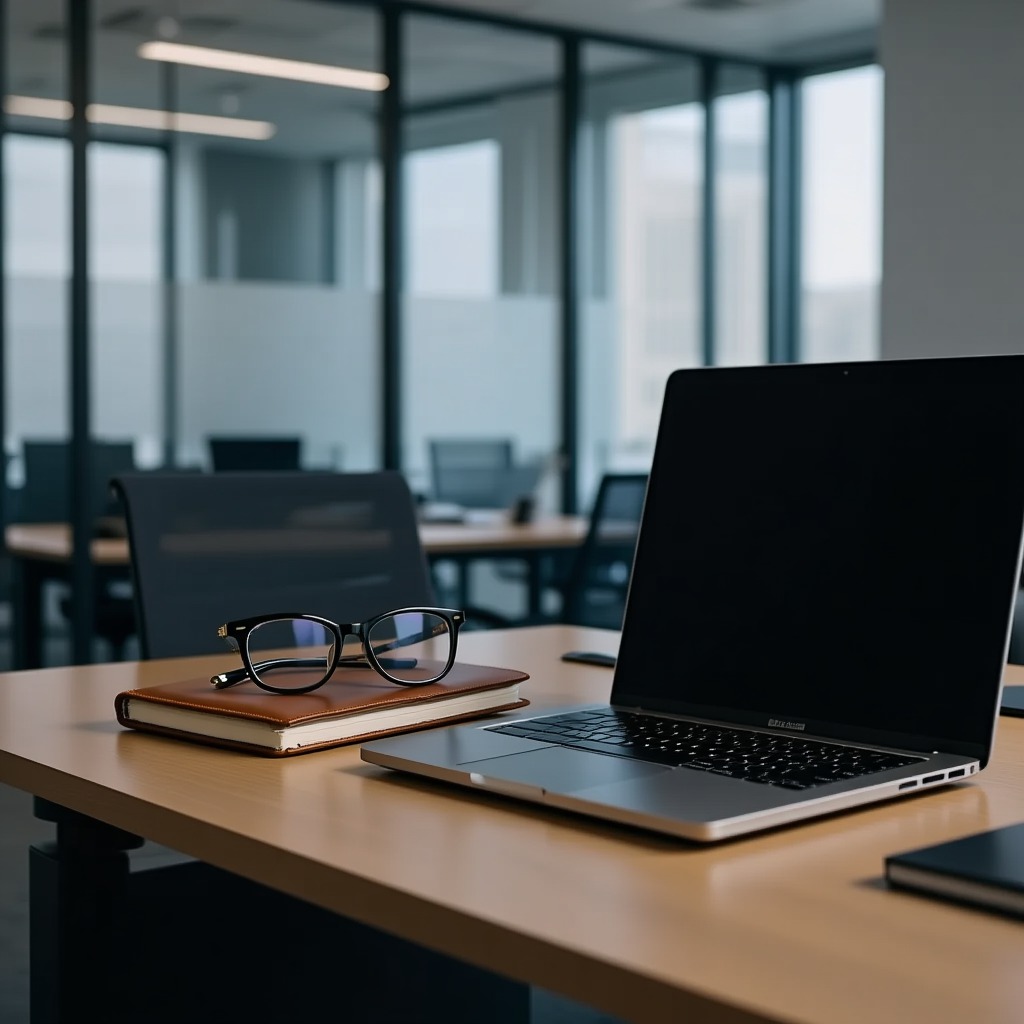} &
    \includegraphics[width=0.16\linewidth]{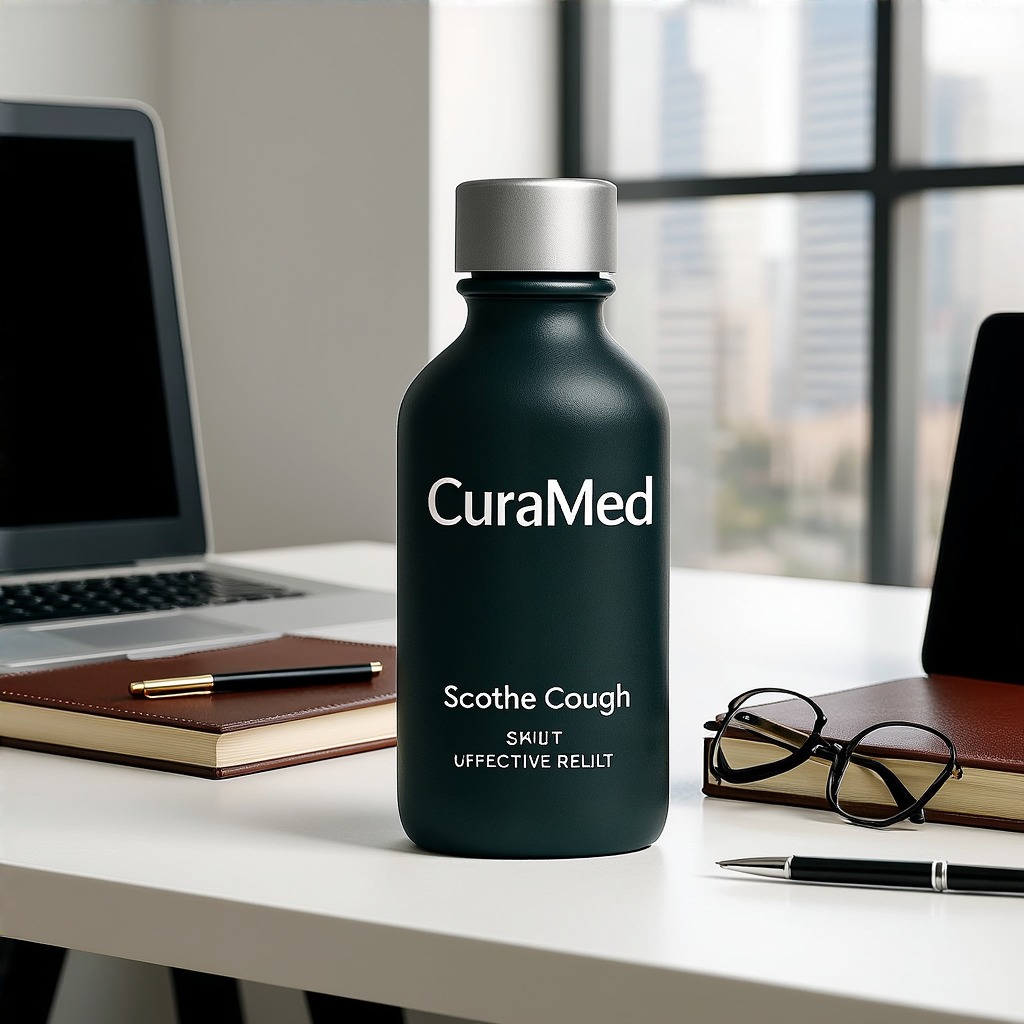} &
    \includegraphics[width=0.16\linewidth]{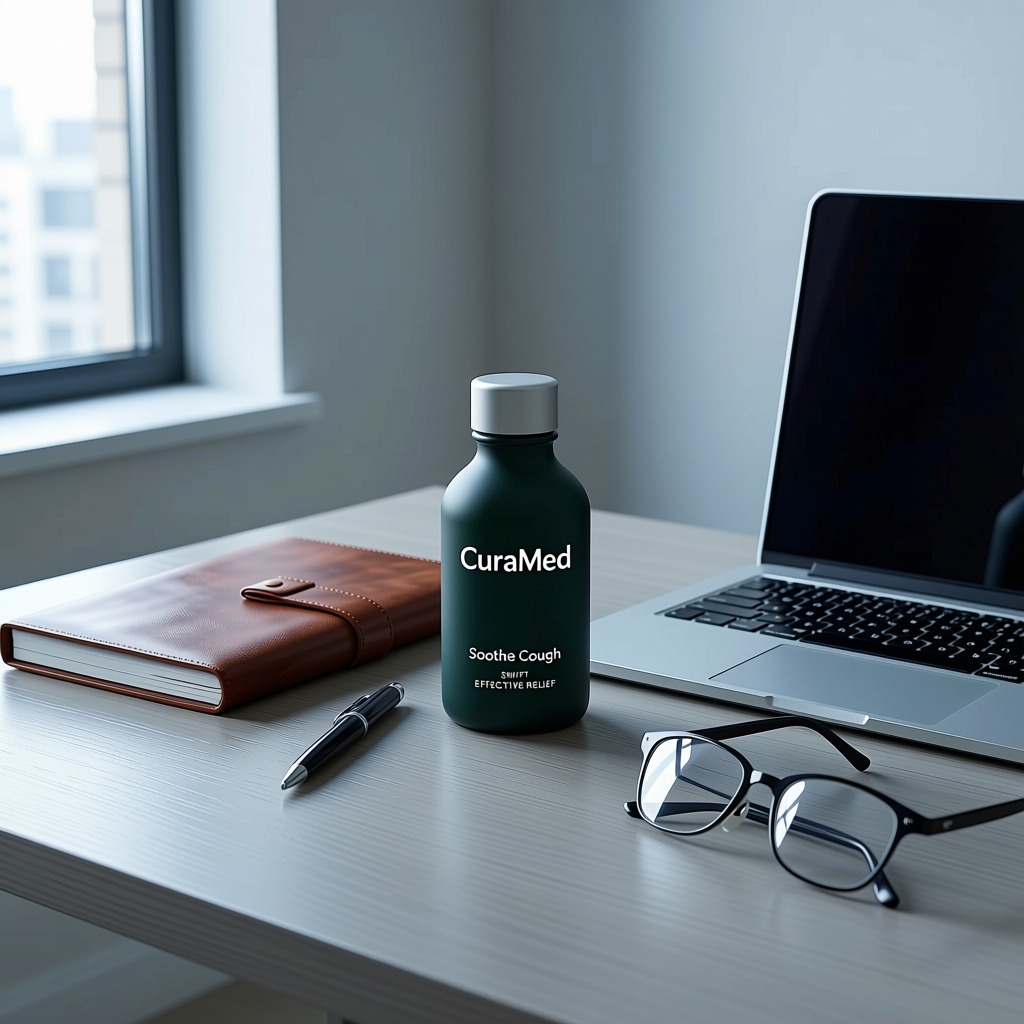} &
    \includegraphics[width=0.16\linewidth]{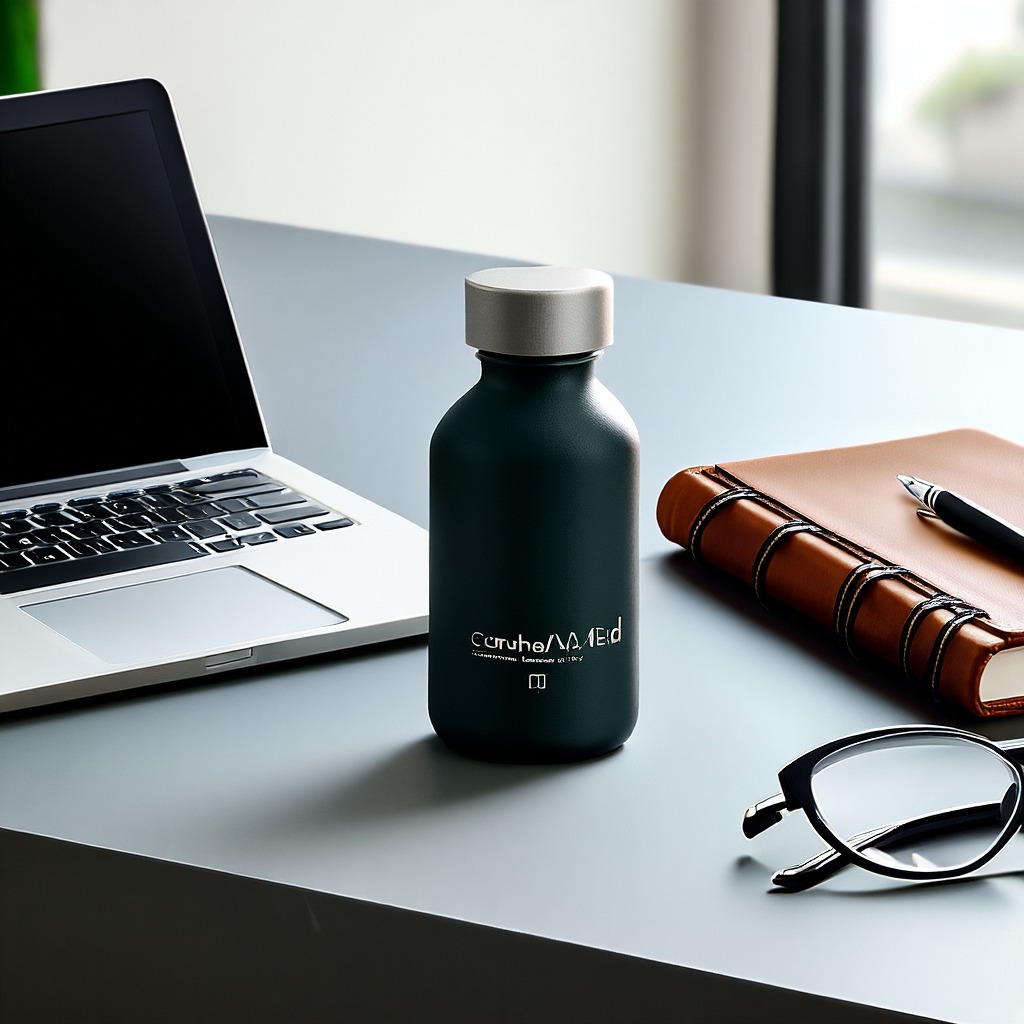} &
    \includegraphics[width=0.16\linewidth]{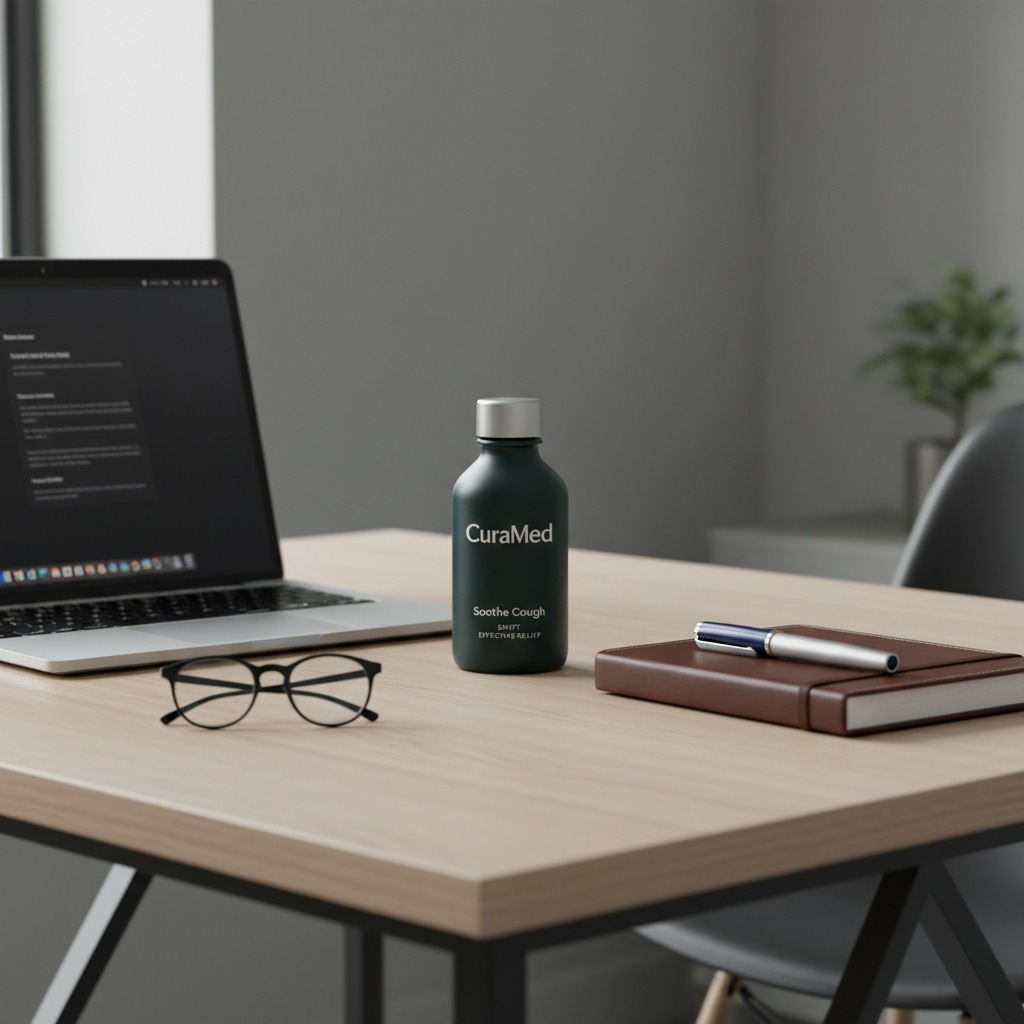} &
    \includegraphics[width=0.16\linewidth]{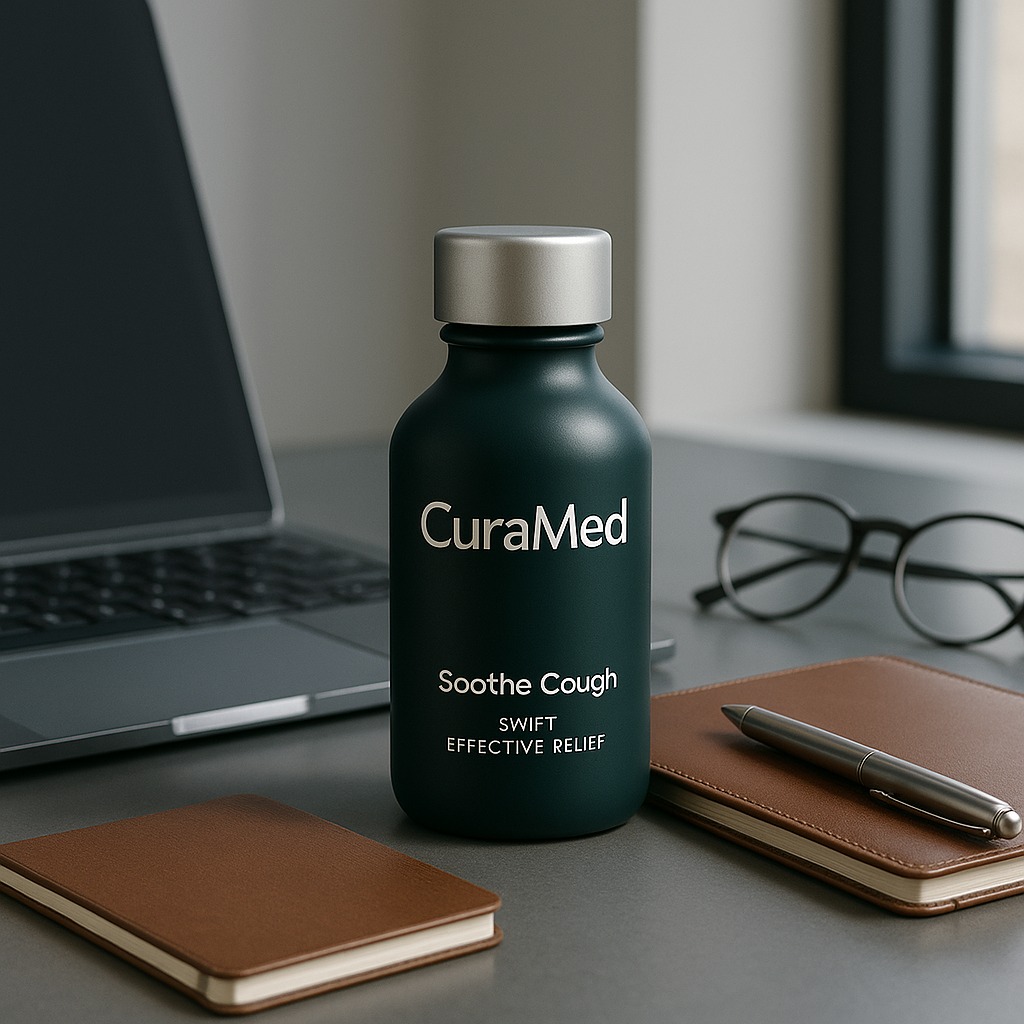} \\
    \multicolumn{7}{p{\textwidth}}{\scriptsize\emph{Place the bottle on a sleek, modern office desk next to a laptop and a stylish, leather-bound notebook; include a pen and a pair of reading glasses to suggest a productive work environment; cool, indirect daylight from a nearby window enhances the minimalist appeal.}} \\[3pt]

    \includegraphics[width=0.16\linewidth]{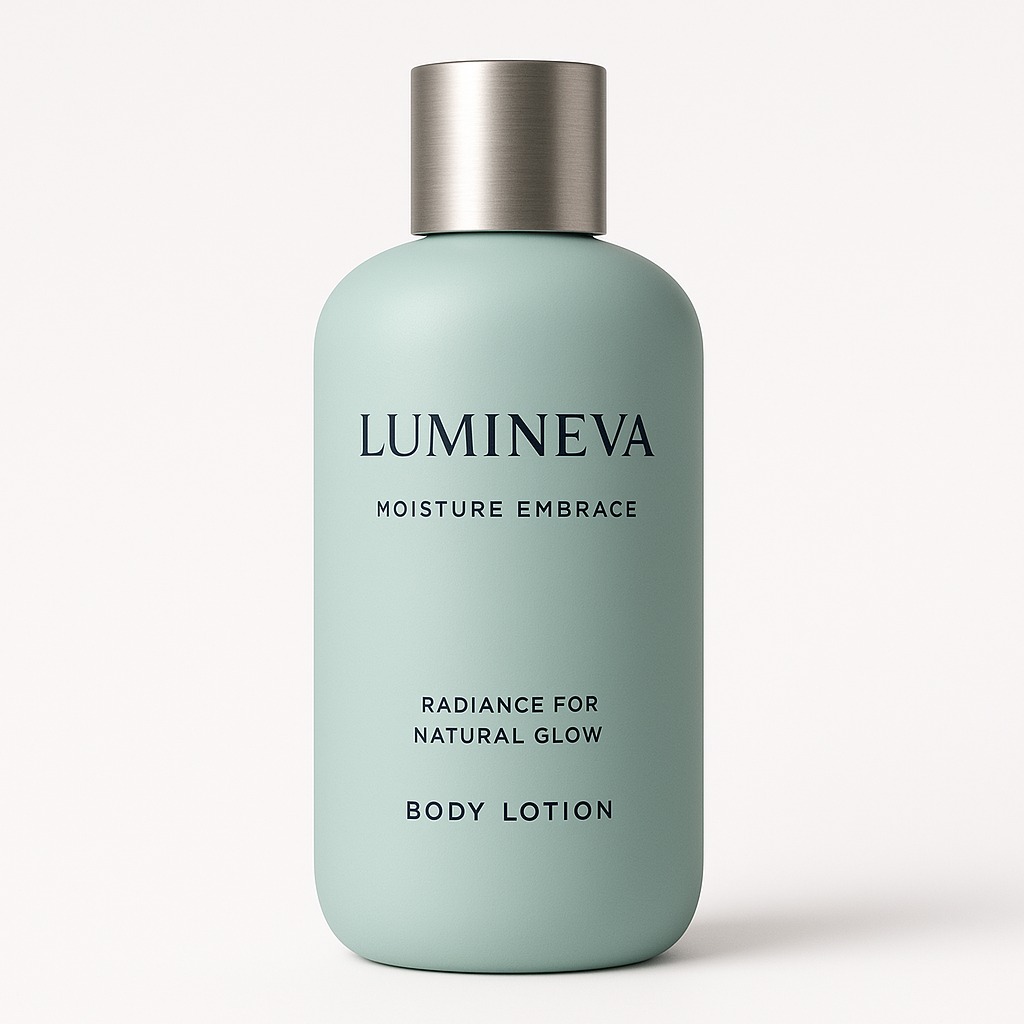} &
    \includegraphics[width=0.16\linewidth]{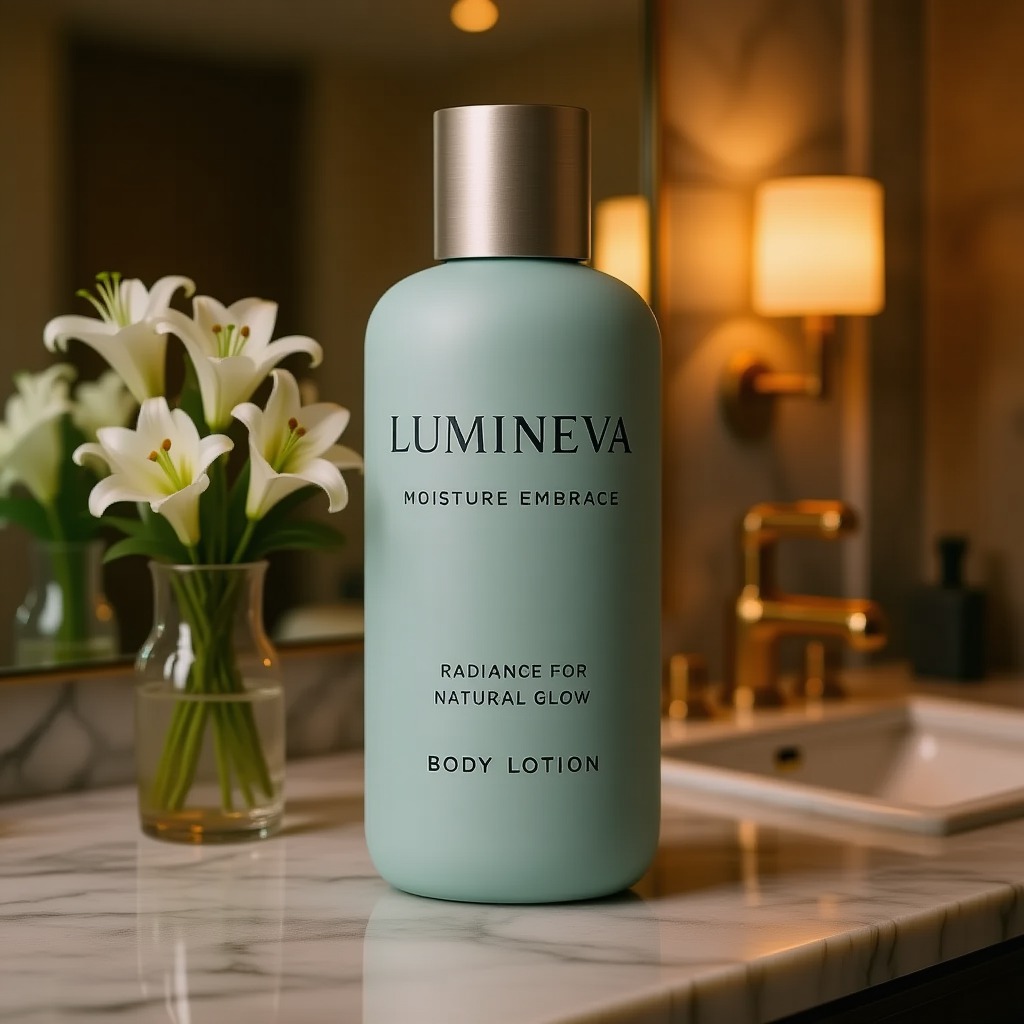} &
    \includegraphics[width=0.16\linewidth]{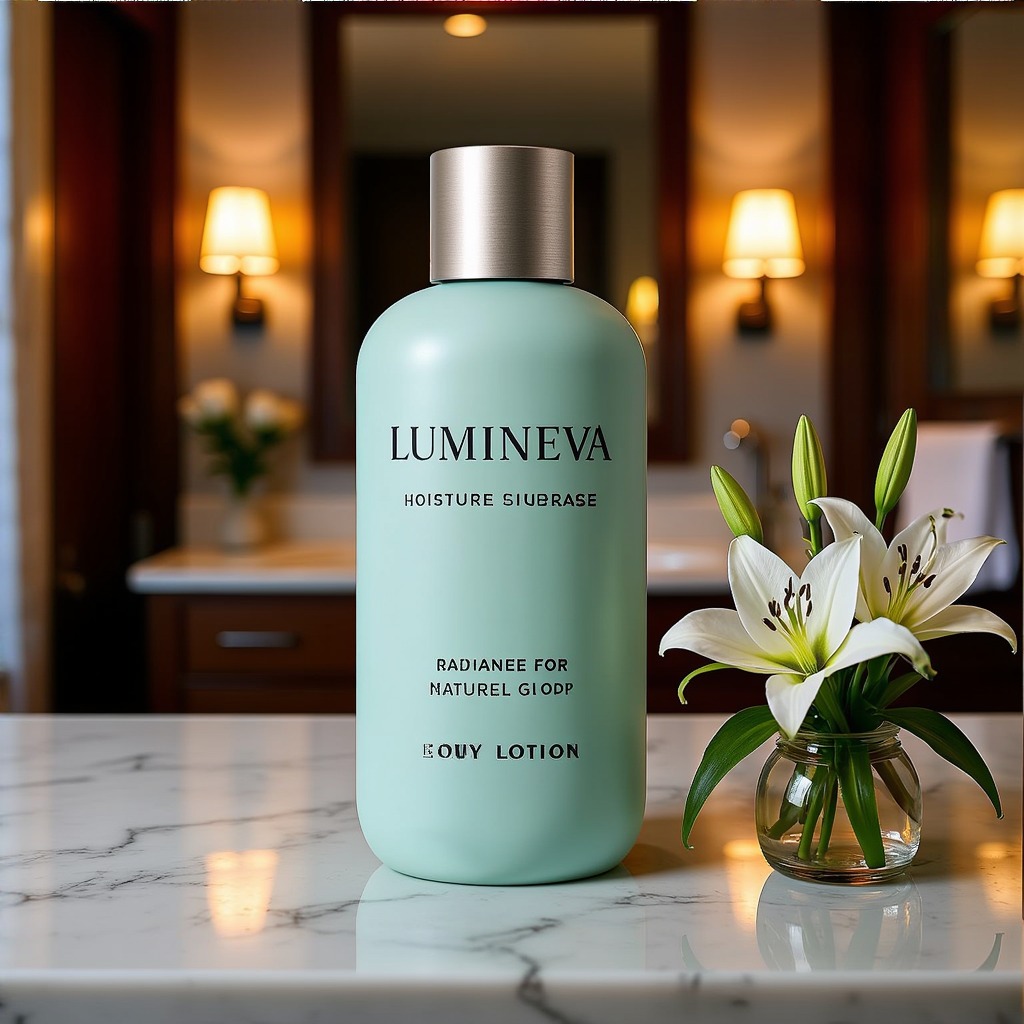} &
    \includegraphics[width=0.16\linewidth]{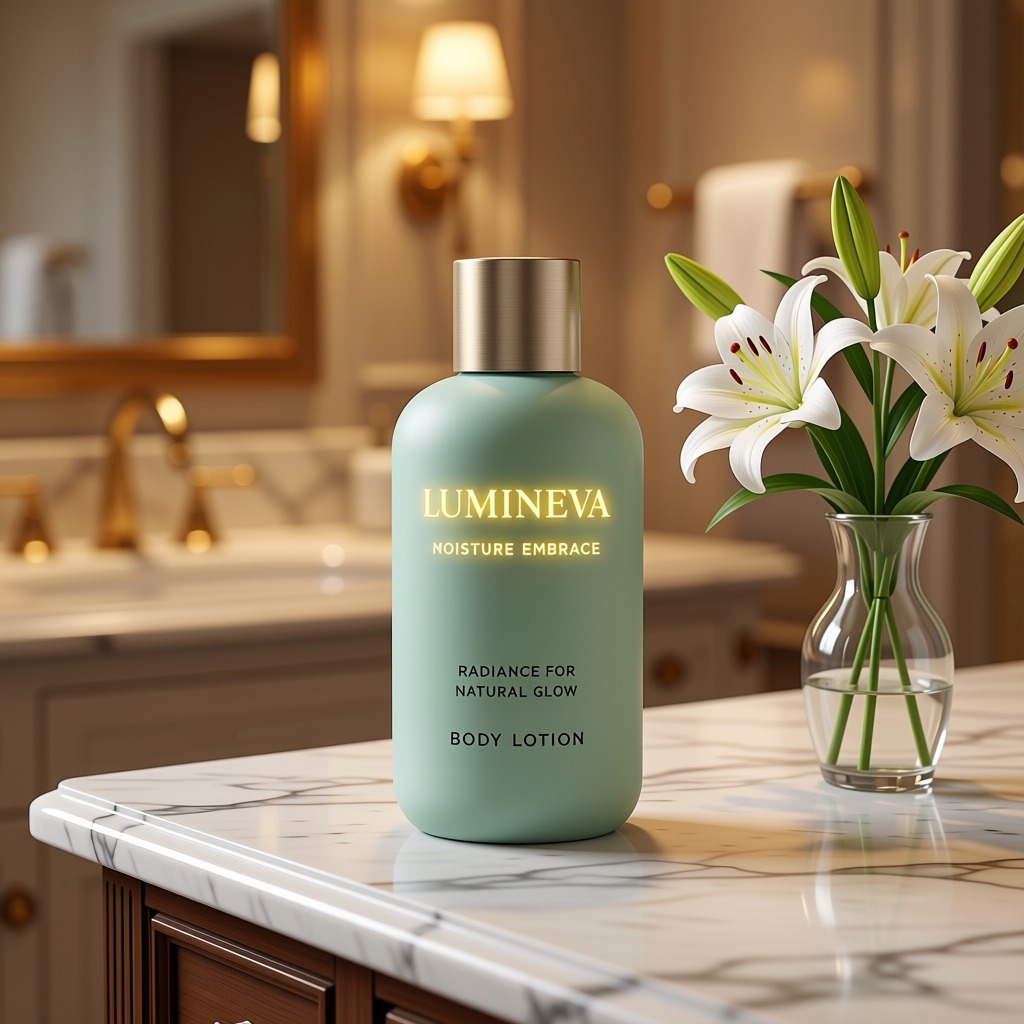} &
    \includegraphics[width=0.16\linewidth]{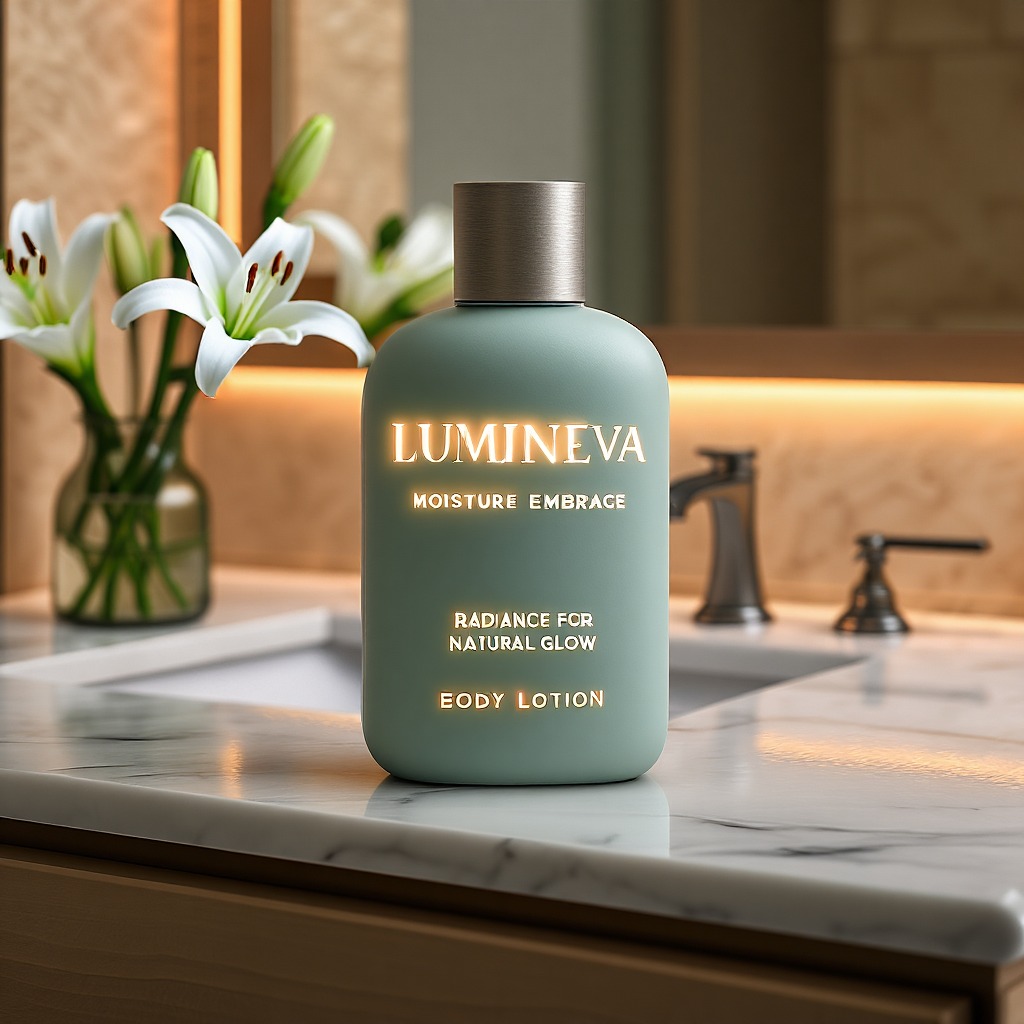} &
    \includegraphics[width=0.16\linewidth]{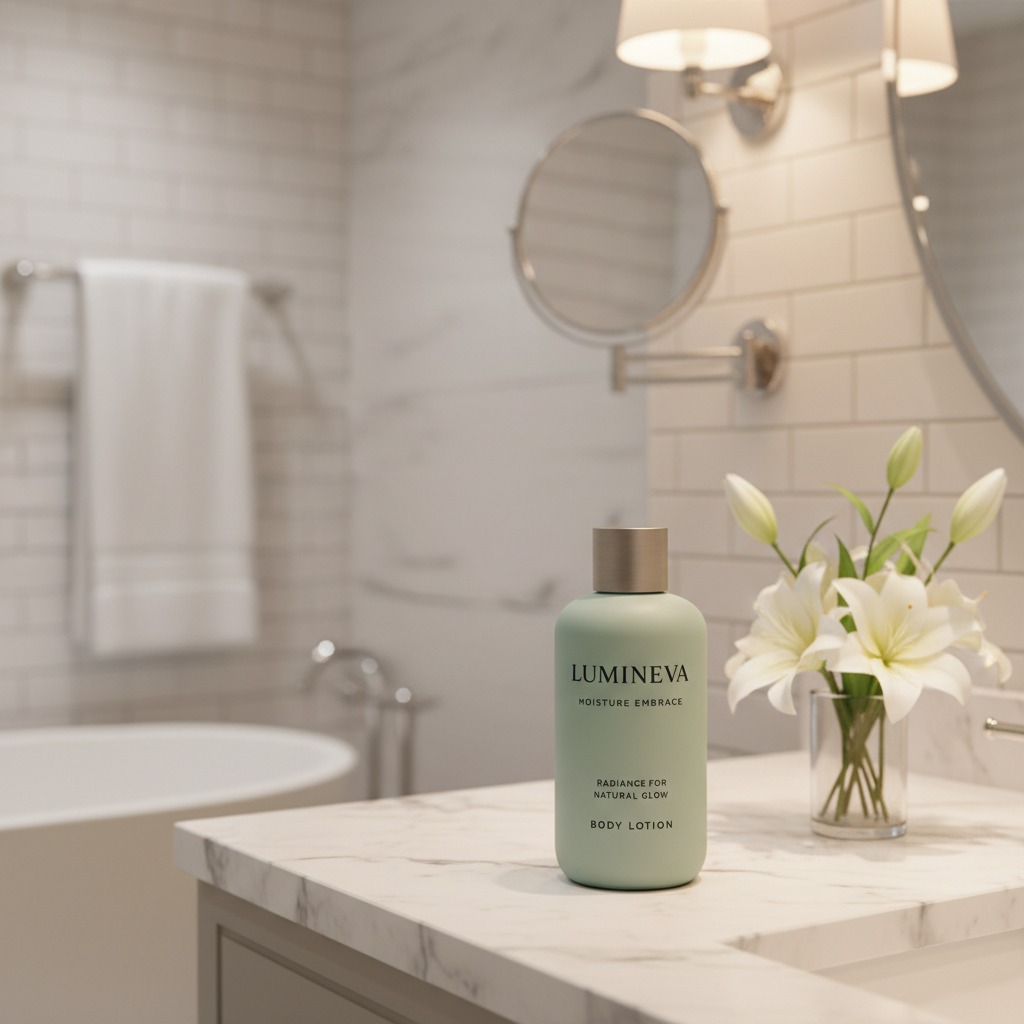} &
    \includegraphics[width=0.16\linewidth]{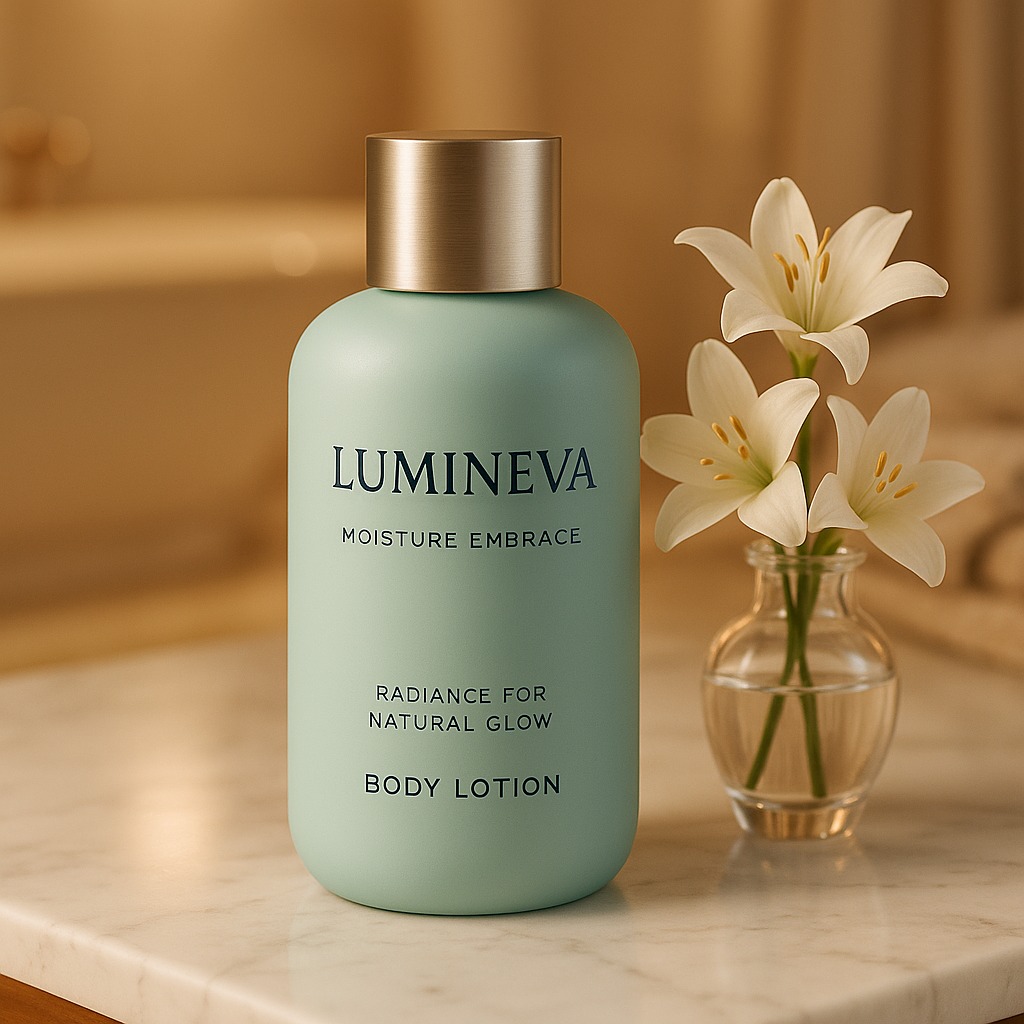} \\
    \multicolumn{7}{p{\textwidth}}{\scriptsize\emph{Place the body lotion bottle on a marble bathroom vanity with a blurred background of a luxurious bathroom; include a small vase with fresh white lilies nearby; warm, soft ambient lighting with a gentle glow to create an inviting atmosphere; ensure the logo is prominently lit with soft reflections on the bottle.}} \\[2pt]

    \includegraphics[width=0.16\linewidth]{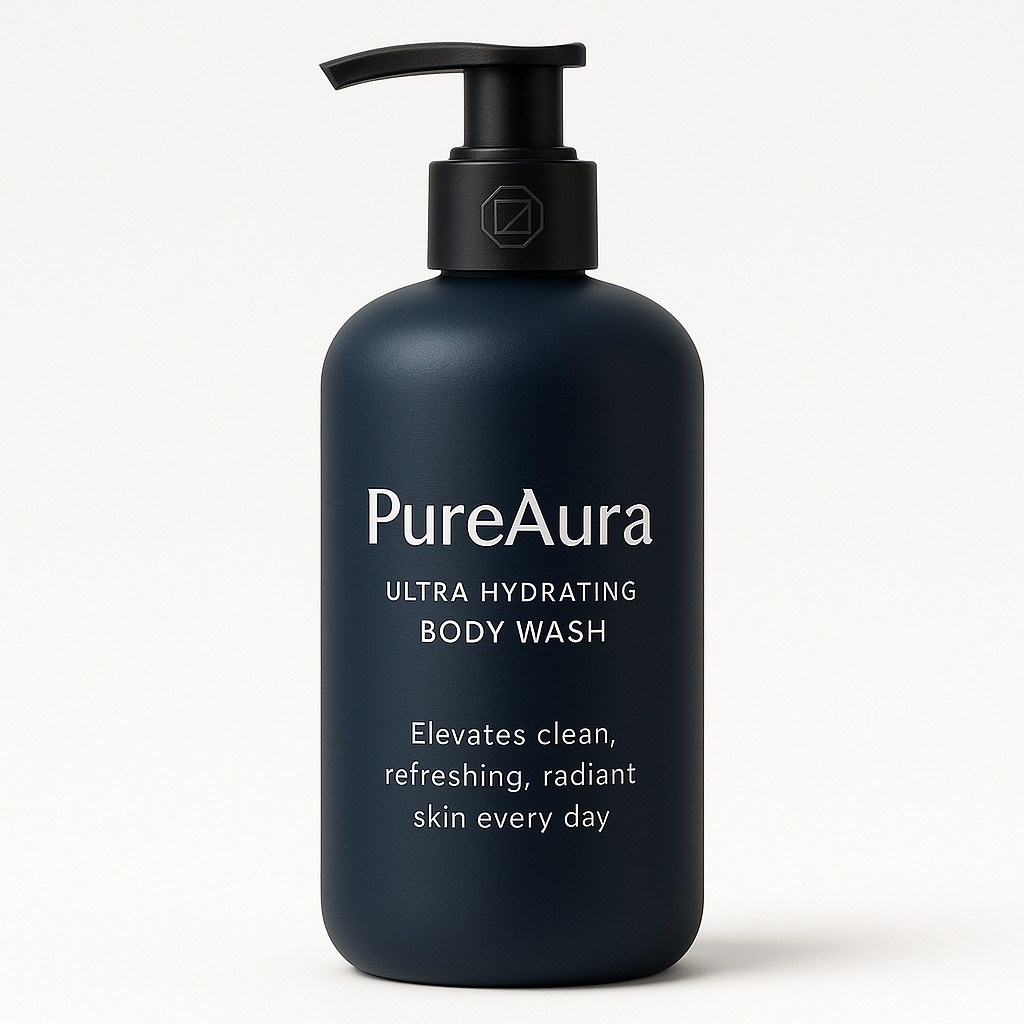} &
    \includegraphics[width=0.16\linewidth]{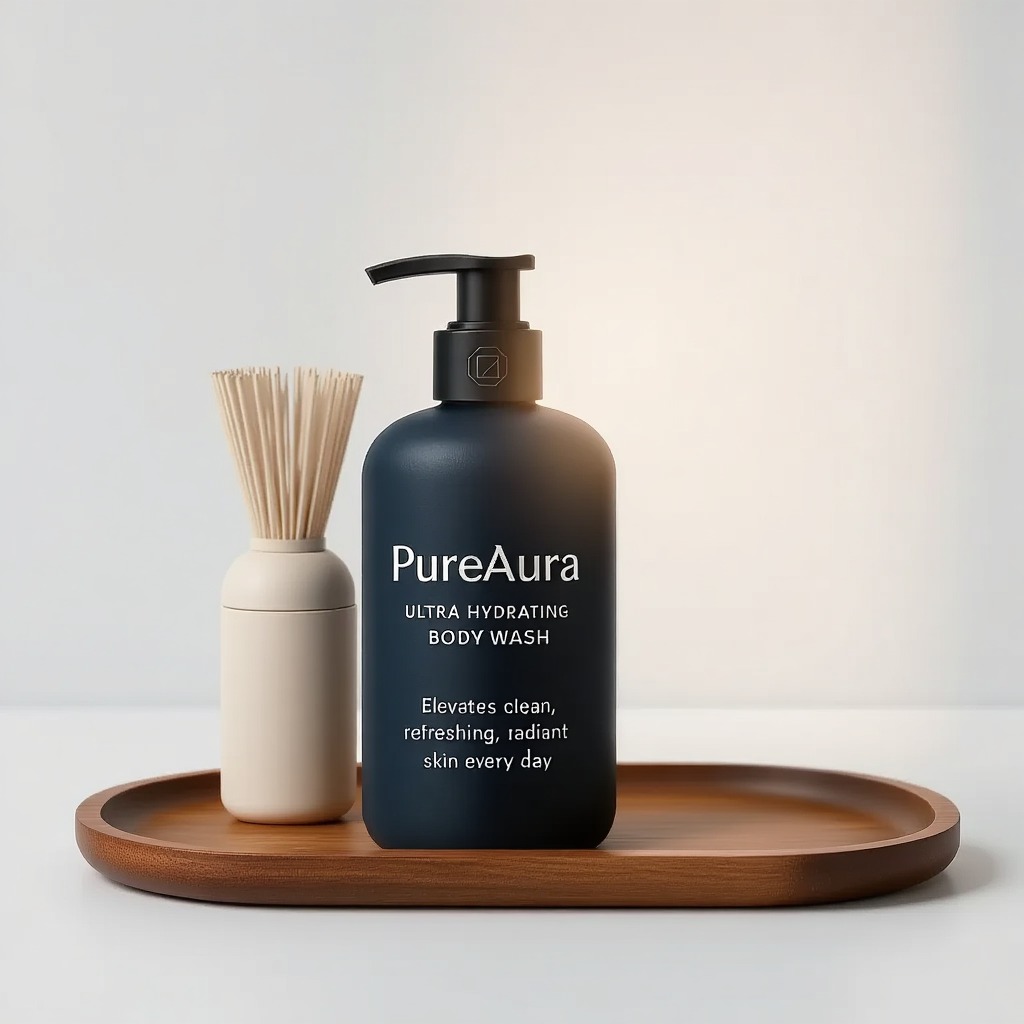} &
    \includegraphics[width=0.16\linewidth]{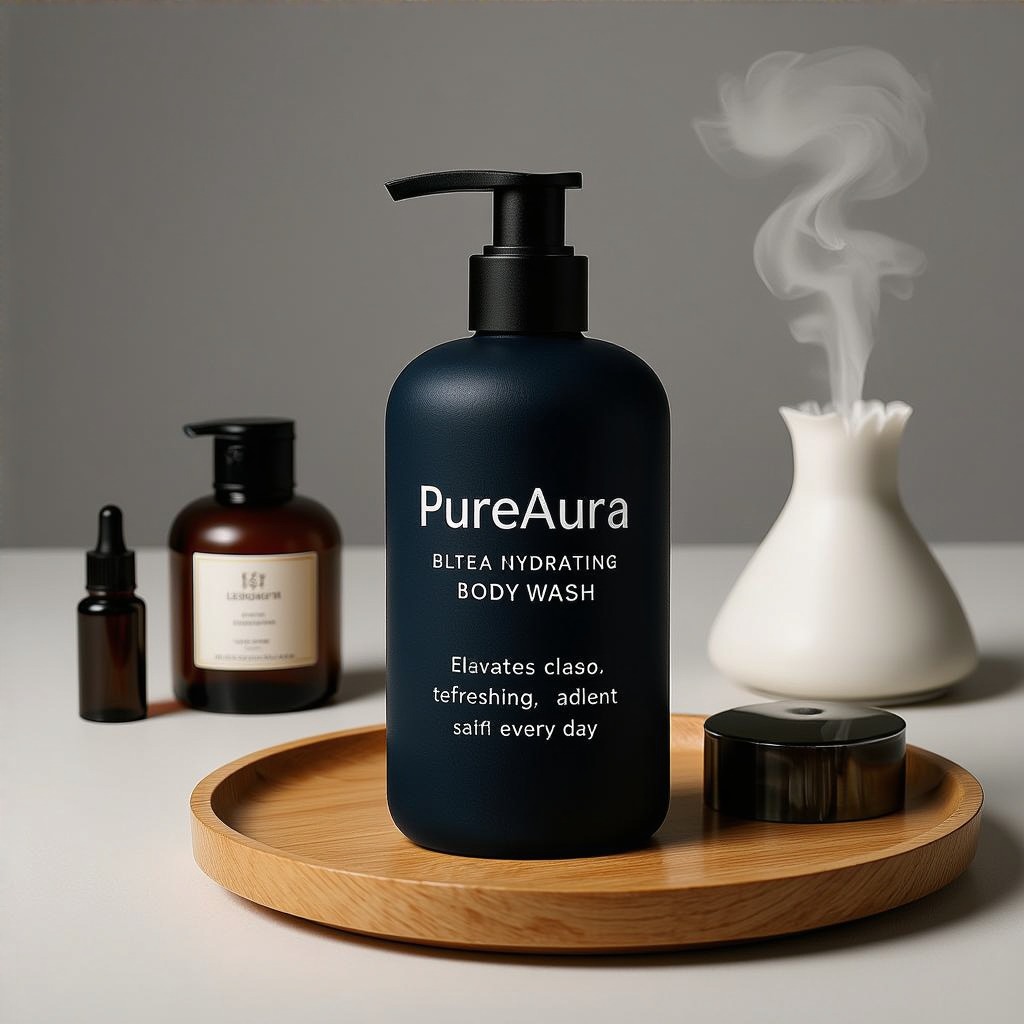} &
    \includegraphics[width=0.16\linewidth]{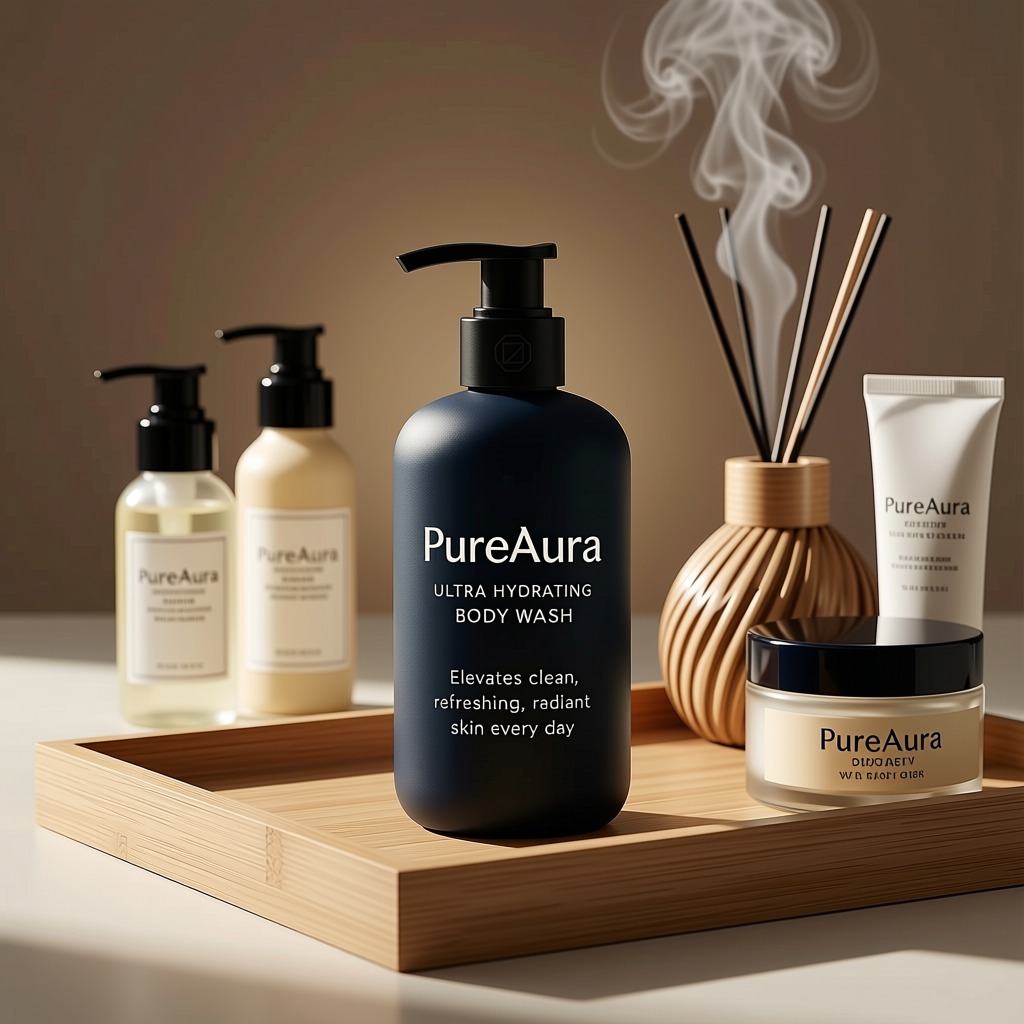} &
    \includegraphics[width=0.16\linewidth]{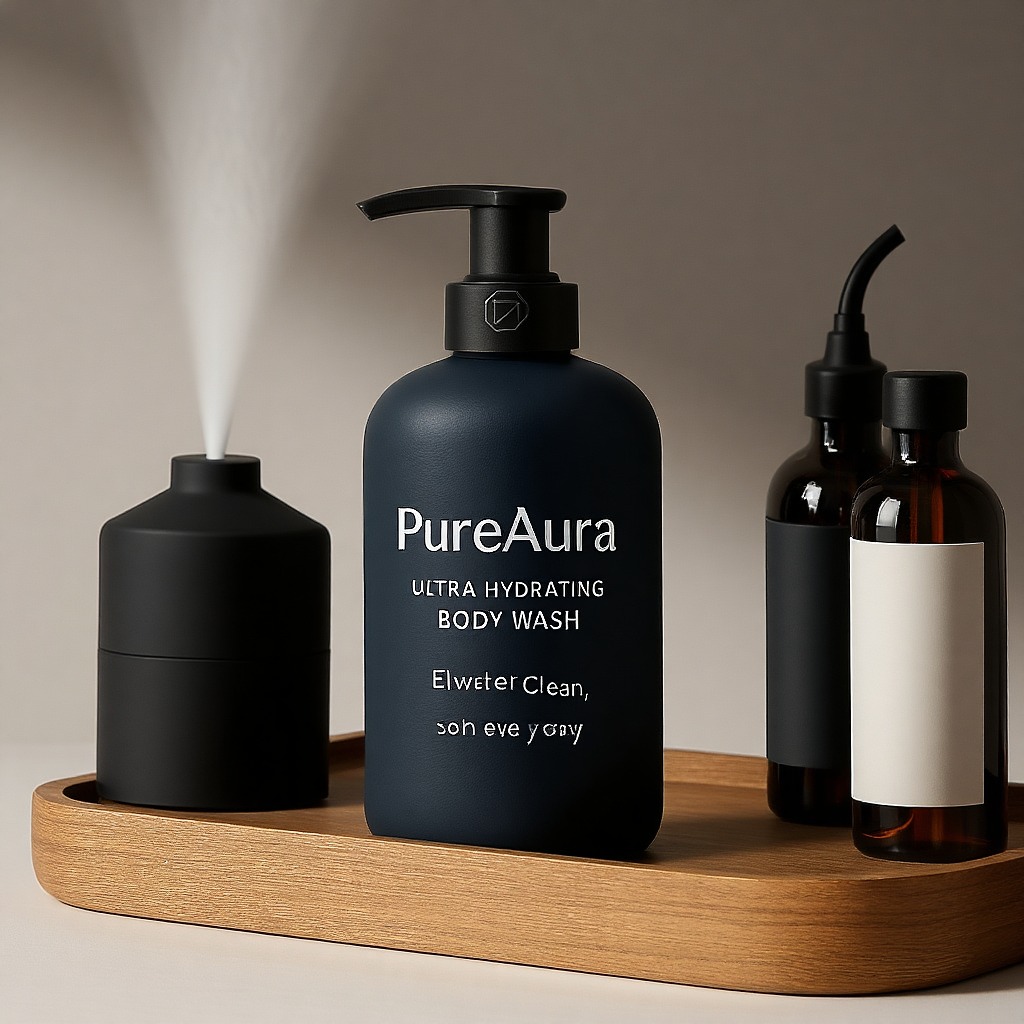} &
    \includegraphics[width=0.16\linewidth]{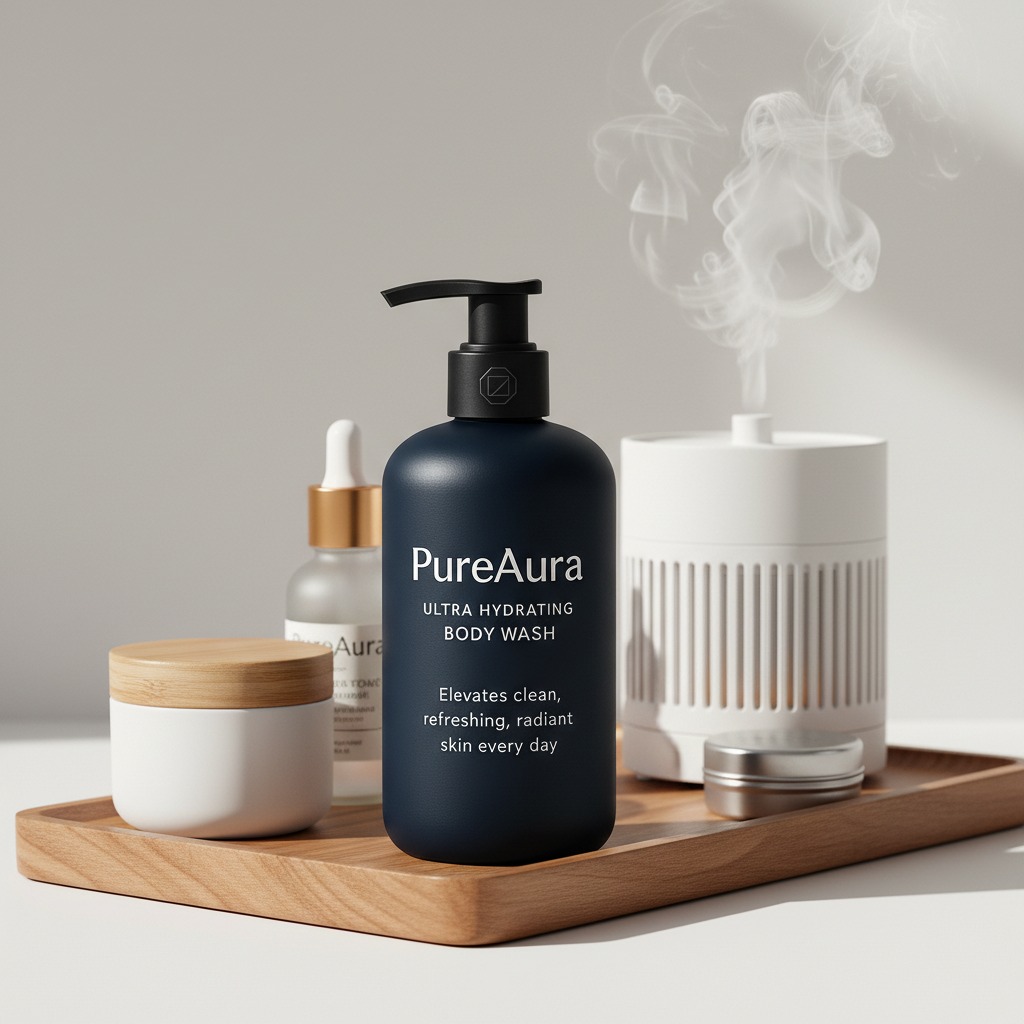} &
    \includegraphics[width=0.16\linewidth]{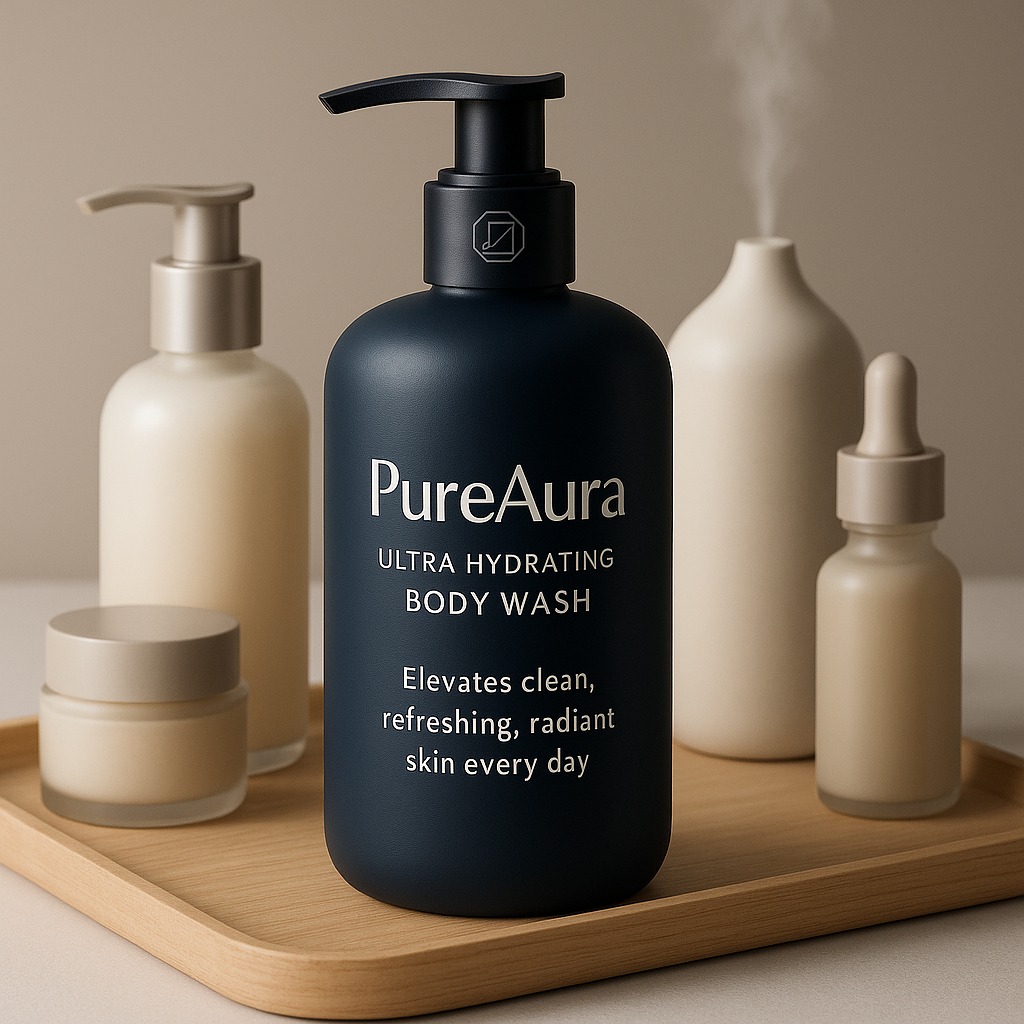} \\
    \multicolumn{7}{p{\textwidth}}{\scriptsize\emph{Place the bottle on a minimalist wooden tray amidst a selection of high-end skincare products; soft, directional lighting highlighting the bottle's silhouette; include a small, stylish diffuser emitting a gentle mist in the background for a calming and rejuvenating environment; maintain a sense of elegance and harmony.}} \\[2pt]

    \includegraphics[width=0.16\linewidth]{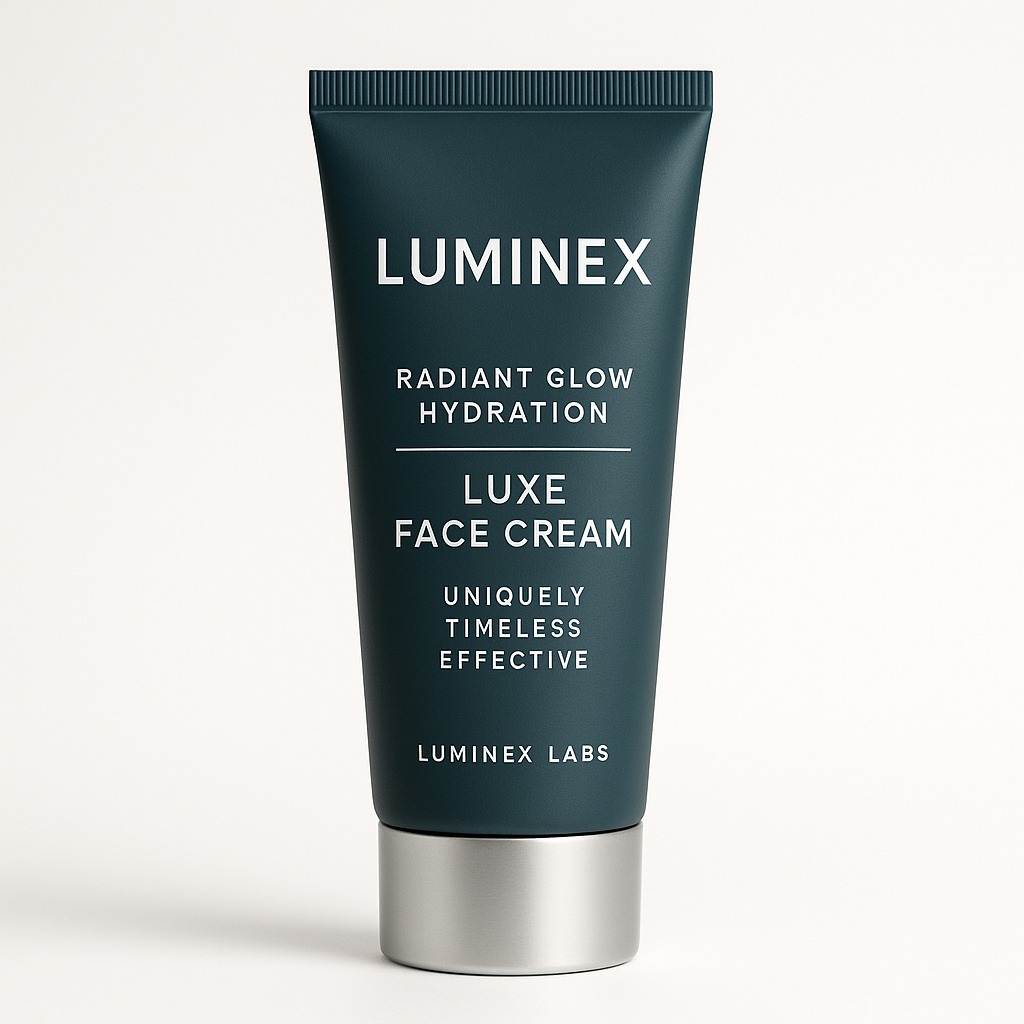} &
    \includegraphics[width=0.16\linewidth]{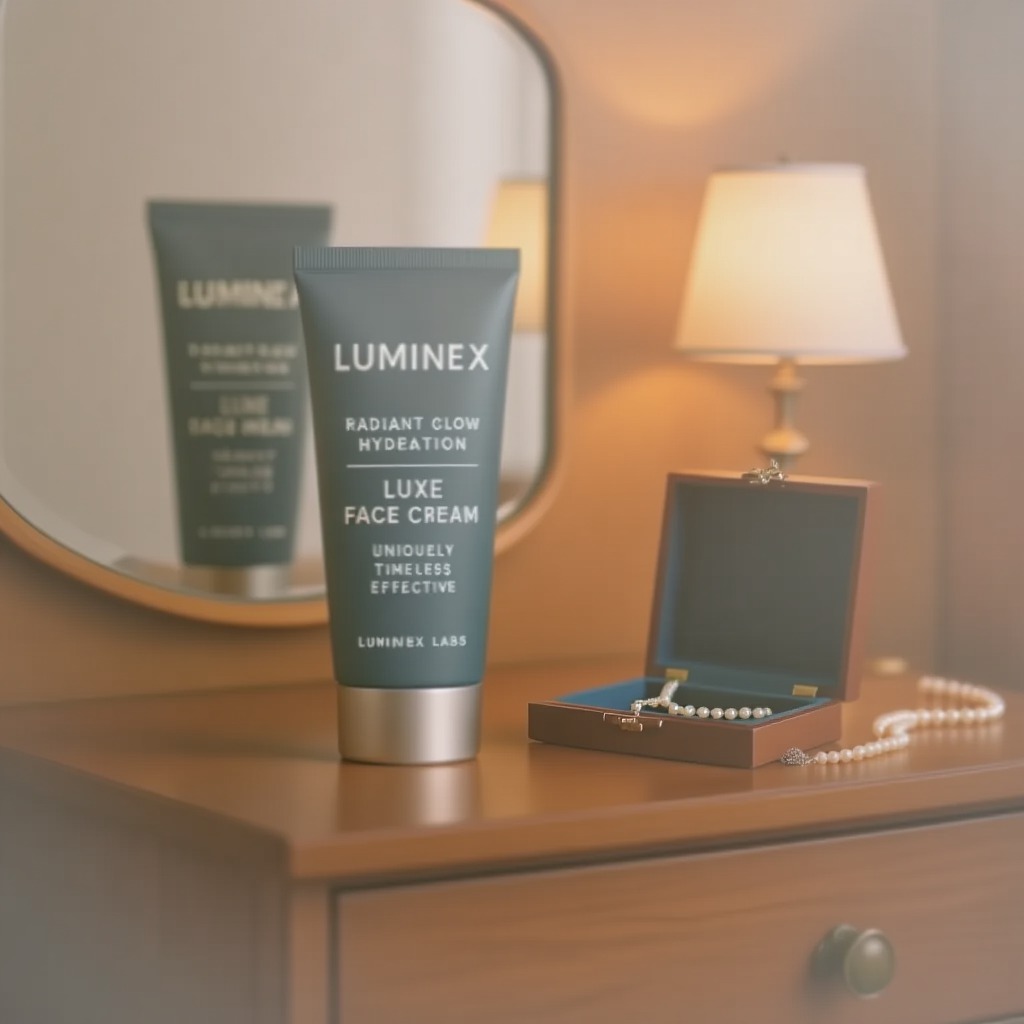} &
    \includegraphics[width=0.16\linewidth]{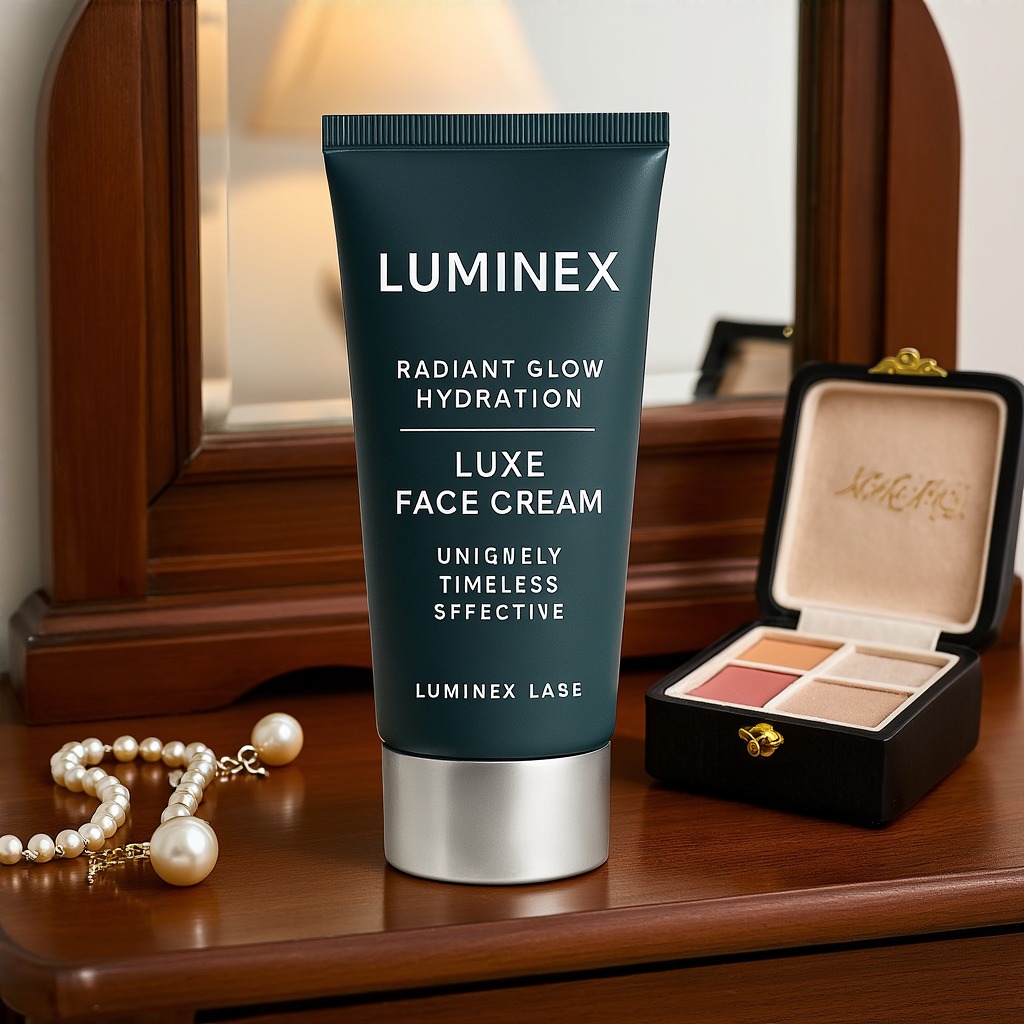} &
    \includegraphics[width=0.16\linewidth]{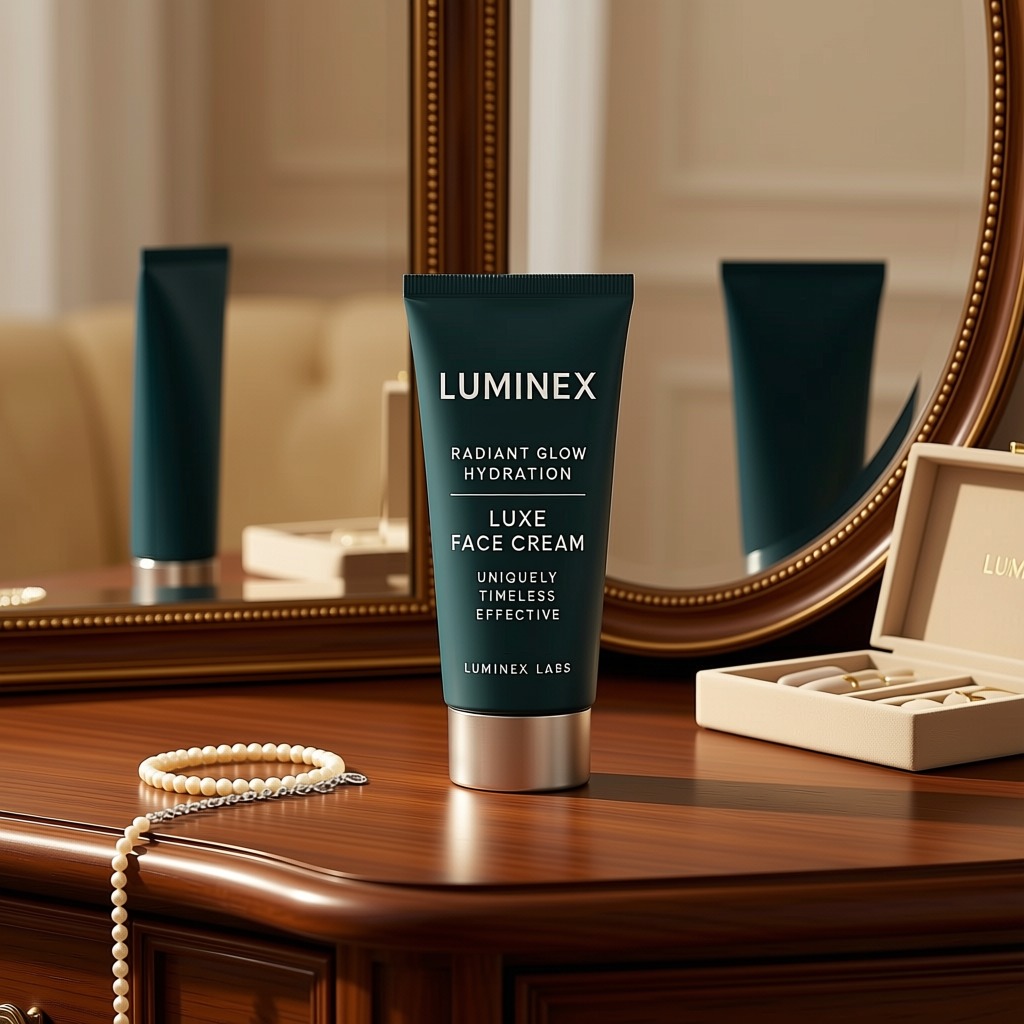} &
    \includegraphics[width=0.16\linewidth]{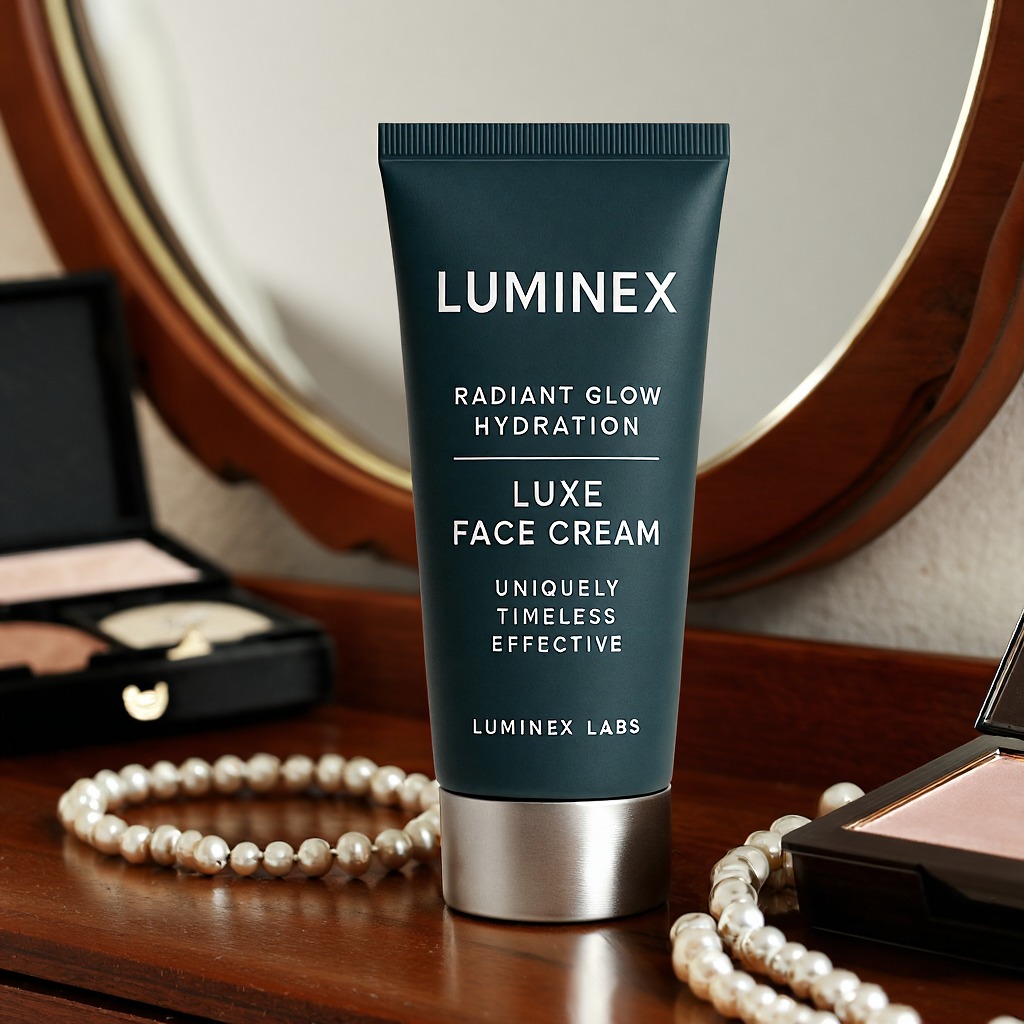} &
    \includegraphics[width=0.16\linewidth]{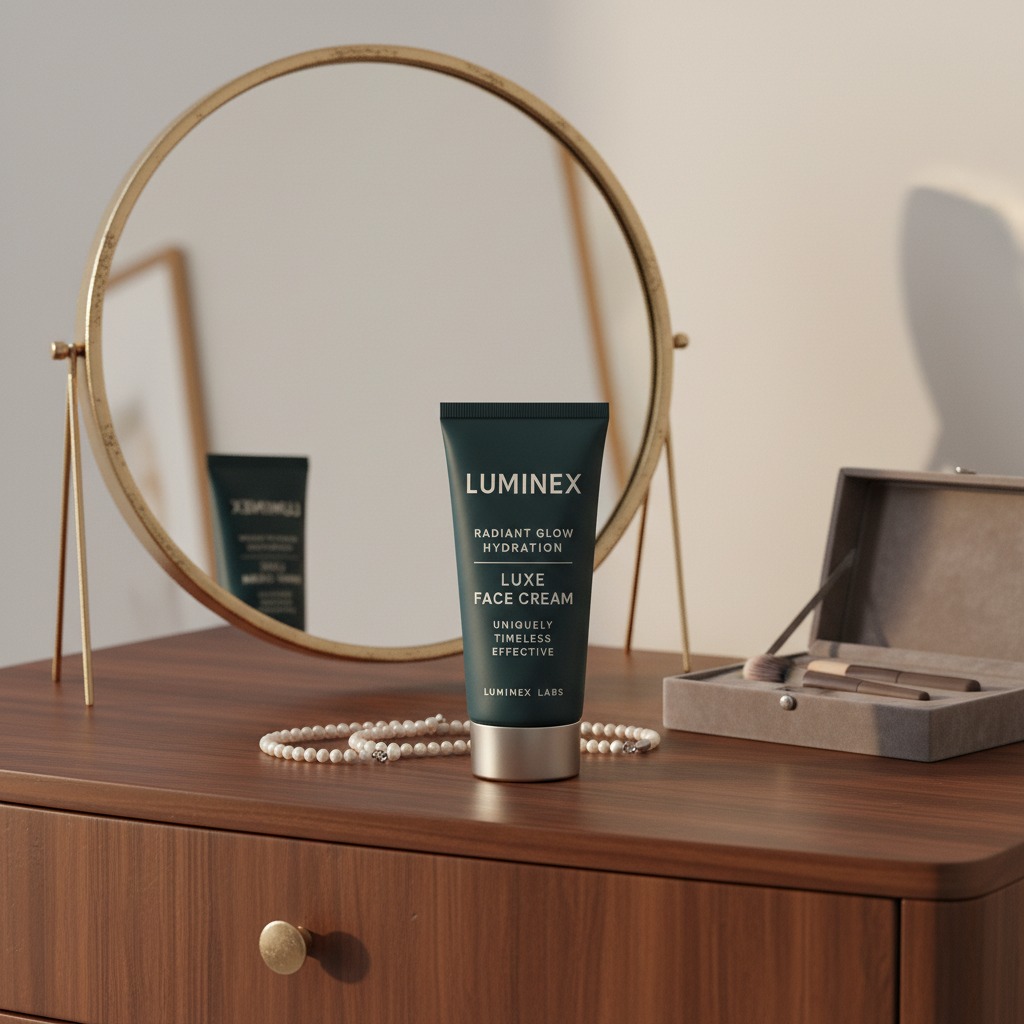} &
    \includegraphics[width=0.16\linewidth]{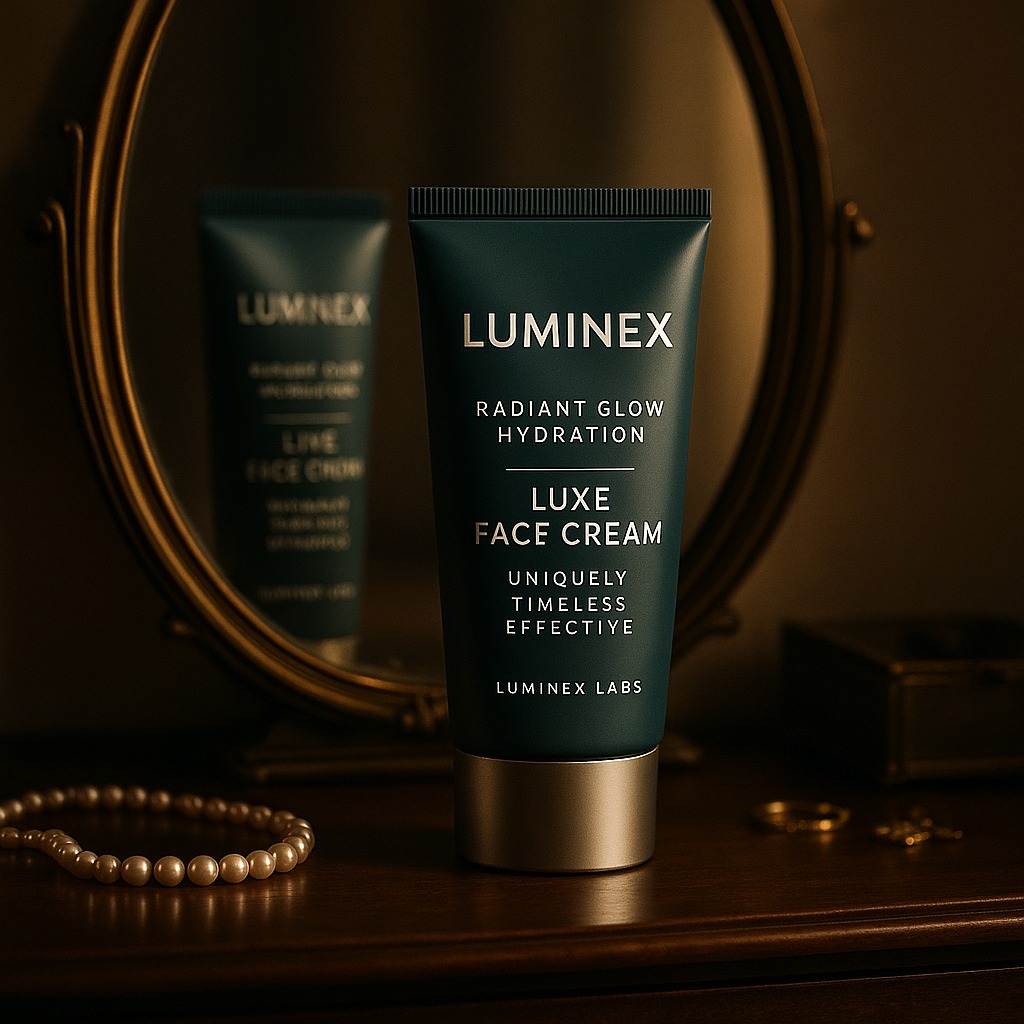} \\
    \multicolumn{7}{p{\textwidth}}{\scriptsize\emph{Position the face cream tube on a sleek wooden dresser with a vintage mirror reflecting its image; use soft, ambient lighting to create warm highlights on the silver trim; include delicate jewelry like a pearl necklace and an open cosmetics box in soft focus around it, emphasizing elegance and luxury.}} \\
    \bottomrule
  \end{tabular}
  }

  \caption{Comparison of various models across six edit instructions. Columns: Input, Step1x‑Edit, HiDream‑E1‑1, Qwen-Image-Lightning, BAGEL, Nano Banana, GPT-Image-1 High. The edit instruction for each input image is present below it.}
  \label{fig:other_models_grid_6x7_fit}
\end{figure*}

\begin{figure*}[t]
  \centering
  \setlength{\tabcolsep}{2pt}
  \renewcommand{\arraystretch}{1.04}

  \resizebox{0.95\textwidth}{!}{%
  \begin{tabular}{@{}cccccc@{}}
    \toprule
    \textbf{Input} & \textbf{Omnigen2} & \textbf{Edit R1 Qwen} & \textbf{Edit R1 Flux} & \textbf{Replan Qwen} & \textbf{Replan Flux} \\
    \midrule

    \includegraphics[width=0.16\linewidth]{sec/selected/other_models/input/BreakfastCerealBox_11words_1__edit5.jpg} &
    \includegraphics[width=0.16\linewidth]{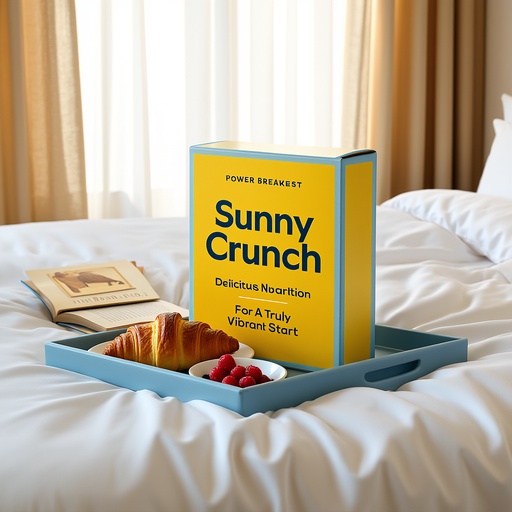} &
    \includegraphics[width=0.16\linewidth]{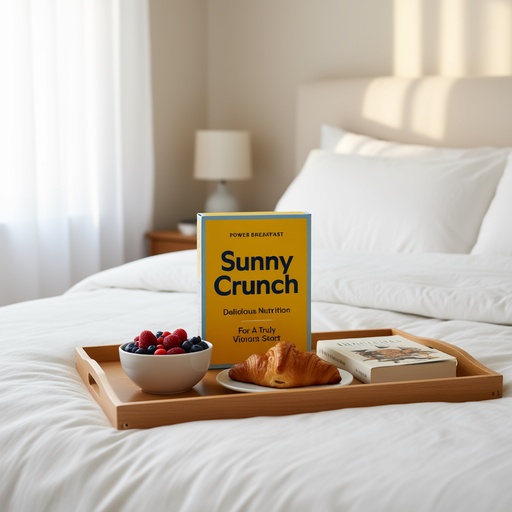} &
    \includegraphics[width=0.16\linewidth]{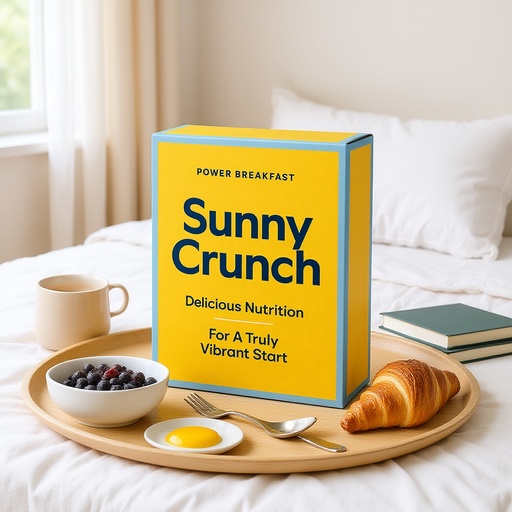} &
    \includegraphics[width=0.16\linewidth]{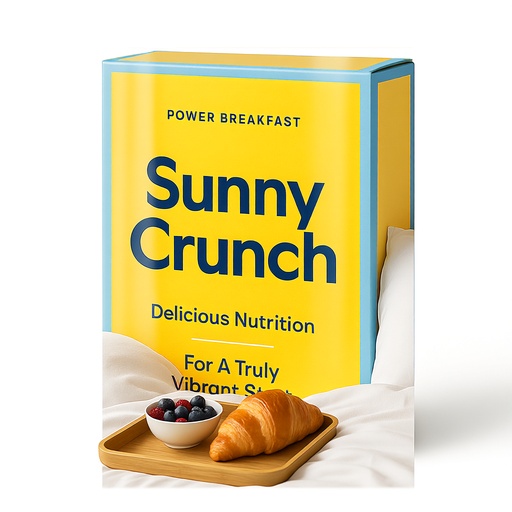} &
    \includegraphics[width=0.16\linewidth]{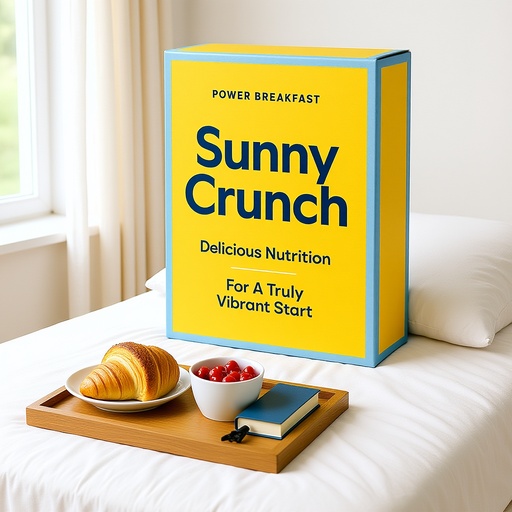} \\
    \multicolumn{6}{p{\textwidth}}{\scriptsize\emph{Feature the cereal box on a breakfast tray on a neatly made bed with soft white linens; include a small bowl of berries, a croissant, and a novel as supporting elements; gentle morning light filtering through sheer curtains for a cozy, indulgent mood; keep the composition balanced and the product sharply in focus.}} \\[2pt]

    \includegraphics[width=0.16\linewidth]{sec/selected/other_models/input/CannedTunaCan_10words_1__edit2.jpg} &
    \includegraphics[width=0.16\linewidth]{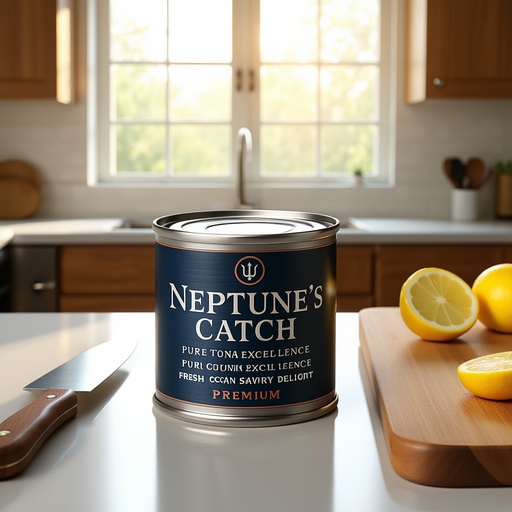} &
    \includegraphics[width=0.16\linewidth]{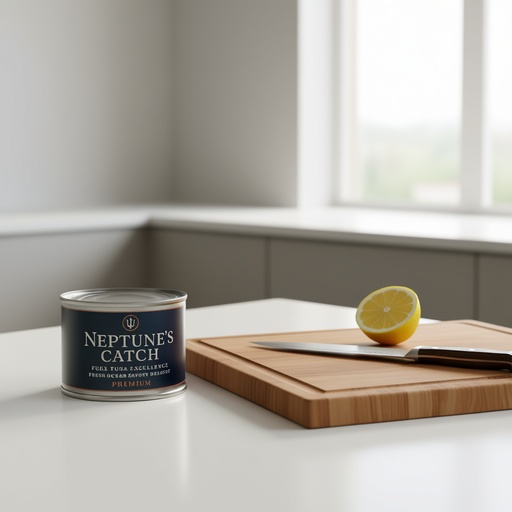} &
    \includegraphics[width=0.16\linewidth]{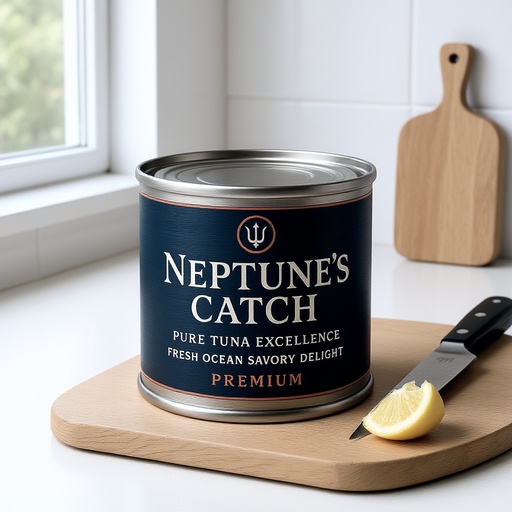} &
    \includegraphics[width=0.16\linewidth]{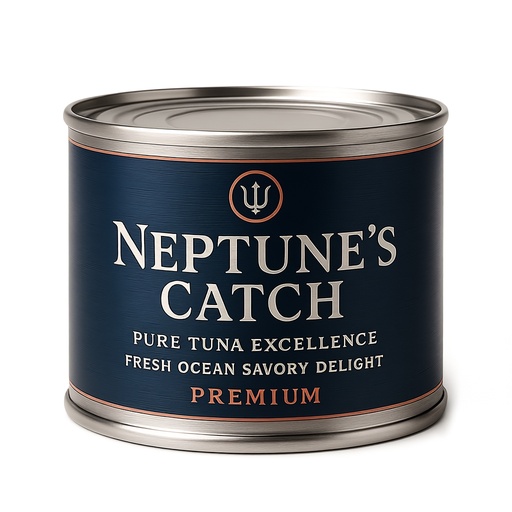} &
    \includegraphics[width=0.16\linewidth]{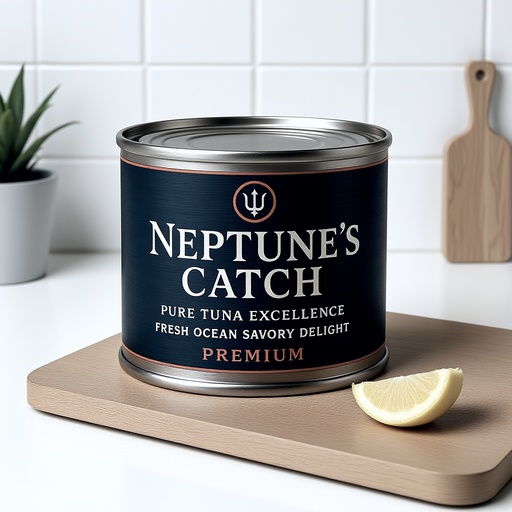} \\
    \multicolumn{6}{p{\textwidth}}{\scriptsize\emph{Position the can on a clean, minimalist kitchen countertop; include a high-quality wooden cutting board with a knife and a lemon slice as props; bathe the scene in soft, ambient daylight from a large kitchen window; ensure the product is hero-lit, with focus on the label and metallic finishes.}} \\[2pt]

    \includegraphics[width=0.16\linewidth]{sec/selected/other_models/input/CoughSyrupBottle_6words_1__edit4.jpg} &
    \includegraphics[width=0.16\linewidth]{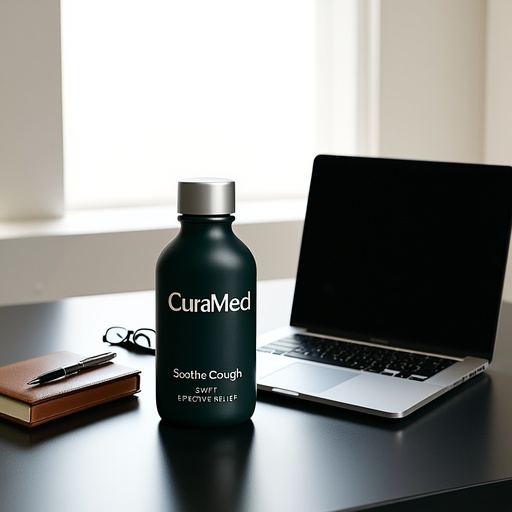} &
    \includegraphics[width=0.16\linewidth]{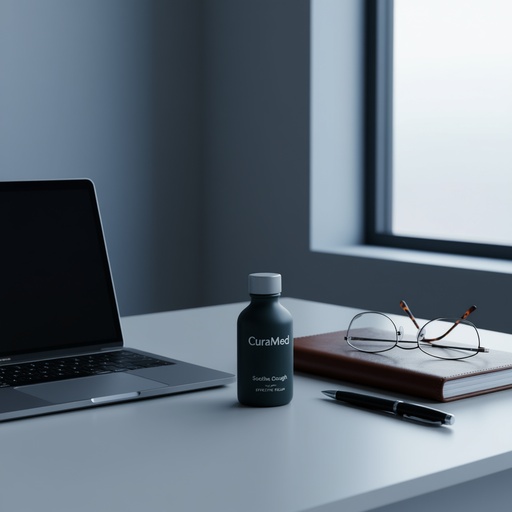} &
    \includegraphics[width=0.16\linewidth]{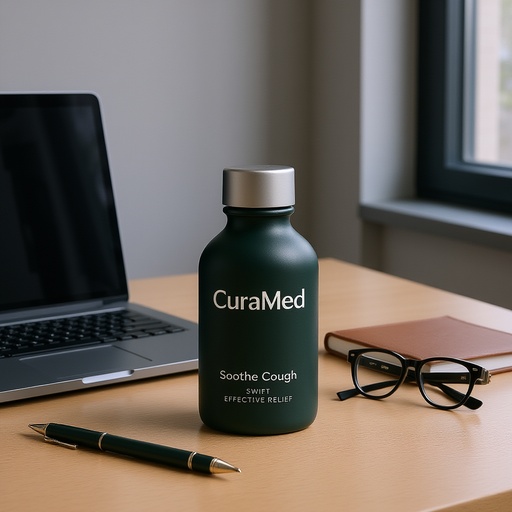} &
    \includegraphics[width=0.16\linewidth]{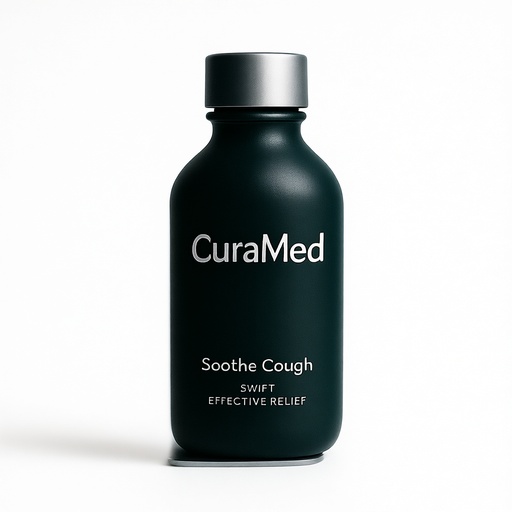} &
    \includegraphics[width=0.16\linewidth]{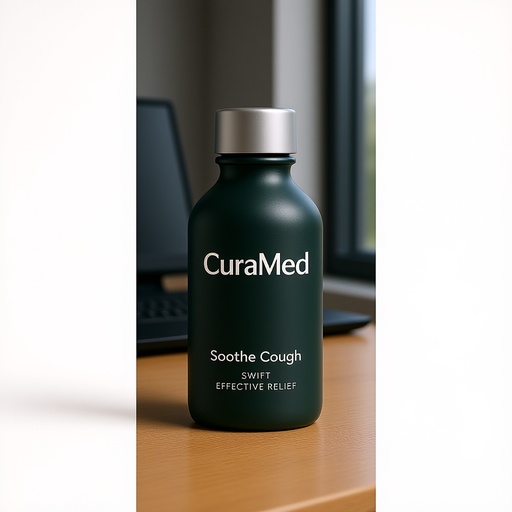} \\
    \multicolumn{6}{p{\textwidth}}{\scriptsize\emph{Place the bottle on a sleek, modern office desk next to a laptop and a stylish, leather-bound notebook; include a pen and a pair of reading glasses to suggest a productive work environment; cool, indirect daylight from a nearby window enhances the minimalist appeal.}} \\[3pt]

    \includegraphics[width=0.16\linewidth]{sec/selected/other_models/input/BodyLotionBottle_9words_1__edit1.jpg} &
    \includegraphics[width=0.16\linewidth]{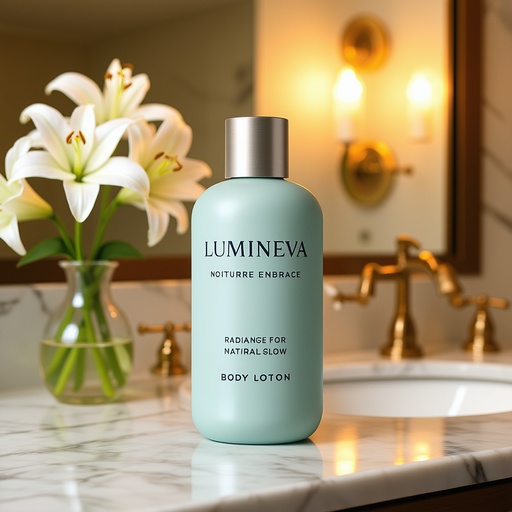} &
    \includegraphics[width=0.16\linewidth]{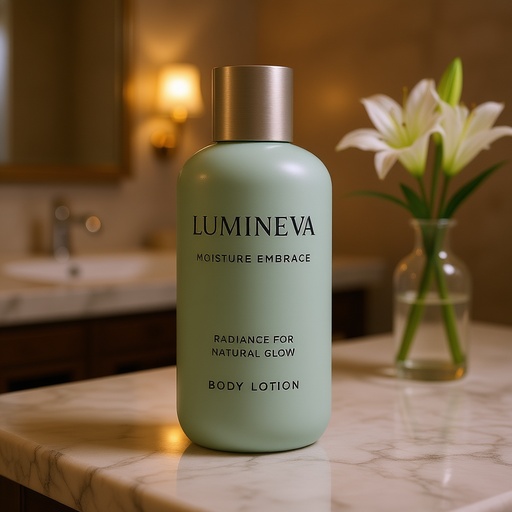} &
    \includegraphics[width=0.16\linewidth]{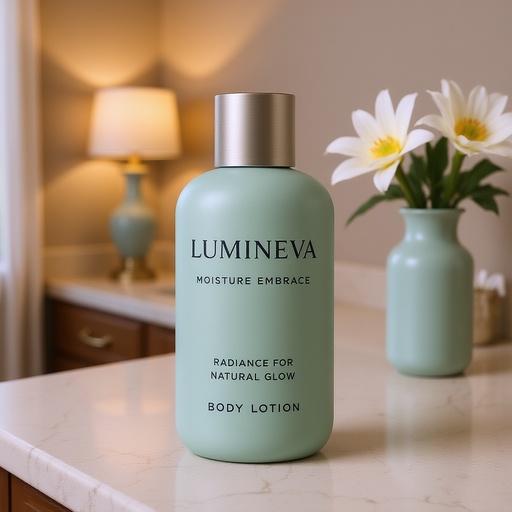} &
    \includegraphics[width=0.16\linewidth]{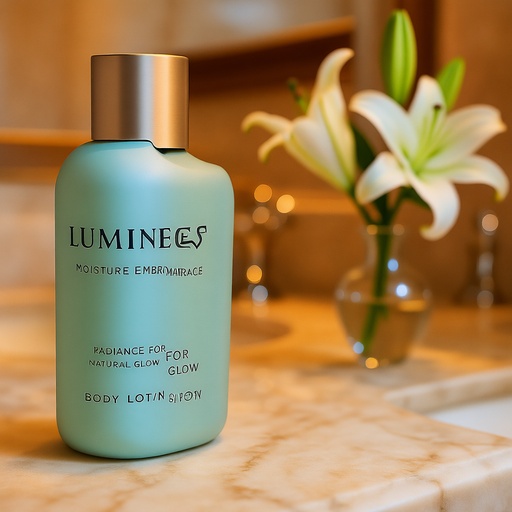} &
    \includegraphics[width=0.16\linewidth]{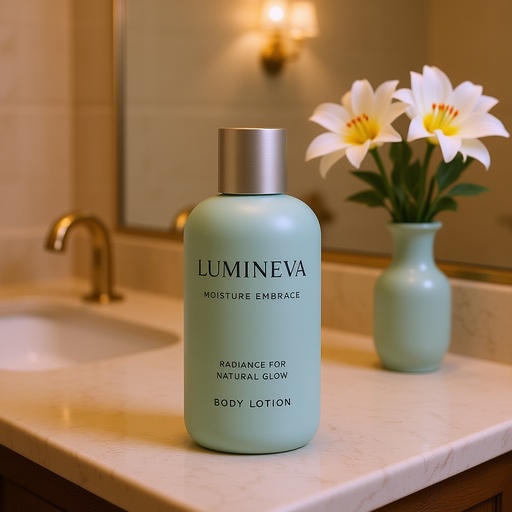} \\
    \multicolumn{6}{p{\textwidth}}{\scriptsize\emph{Place the body lotion bottle on a marble bathroom vanity with a blurred background of a luxurious bathroom; include a small vase with fresh white lilies nearby; warm, soft ambient lighting with a gentle glow to create an inviting atmosphere; ensure the logo is prominently lit with soft reflections on the bottle.}} \\[2pt]

    \includegraphics[width=0.16\linewidth]{sec/selected/other_models/input/BodyWashBottle_12words_1__edit5.jpg} &
    \includegraphics[width=0.16\linewidth]{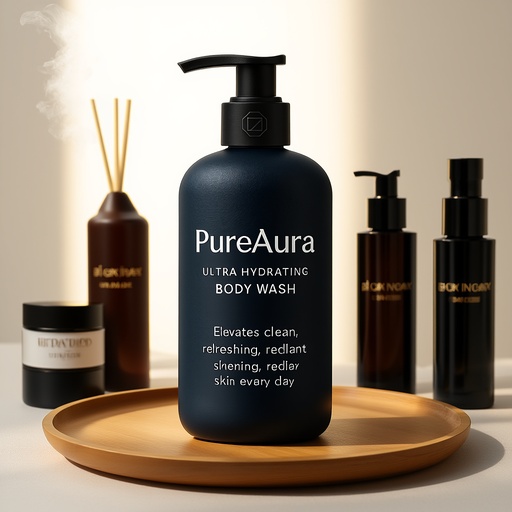} &
    \includegraphics[width=0.16\linewidth]{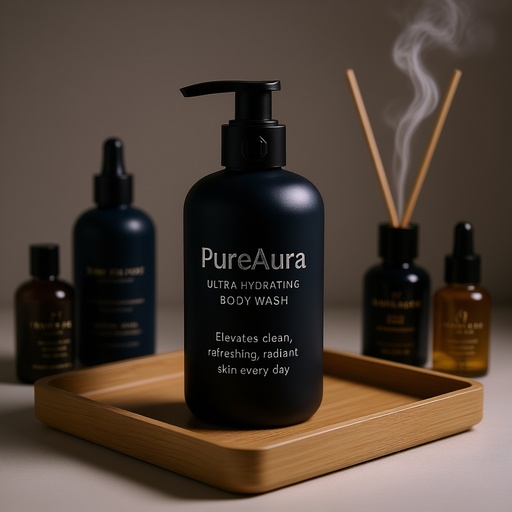} &
    \includegraphics[width=0.16\linewidth]{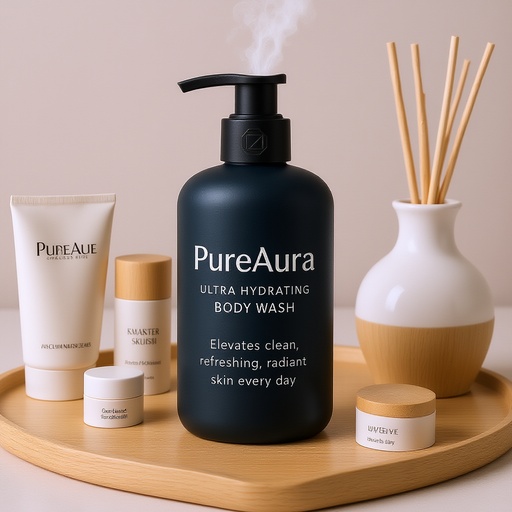} &
    \includegraphics[width=0.16\linewidth]{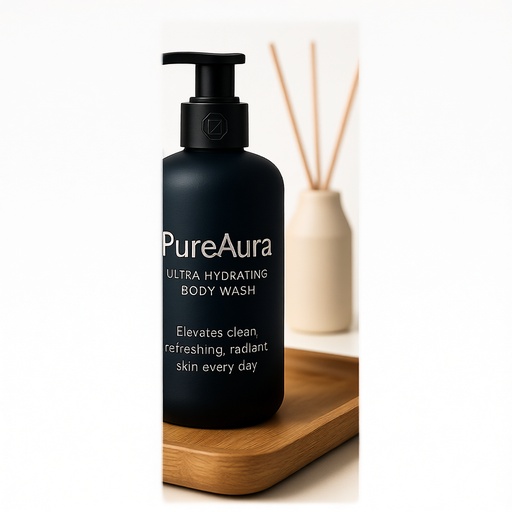} &
    \includegraphics[width=0.16\linewidth]{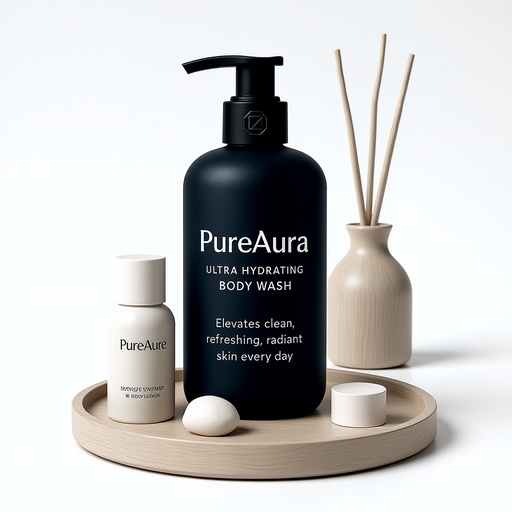} \\
    \multicolumn{6}{p{\textwidth}}{\scriptsize\emph{Place the bottle on a minimalist wooden tray amidst a selection of high-end skincare products; soft, directional lighting highlighting the bottle's silhouette; include a small, stylish diffuser emitting a gentle mist in the background for a calming and rejuvenating environment; maintain a sense of elegance and harmony.}} \\[2pt]

    \includegraphics[width=0.16\linewidth]{sec/selected/other_models/input/FaceCreamTube_10words_1__edit3.jpg} &
    \includegraphics[width=0.16\linewidth]{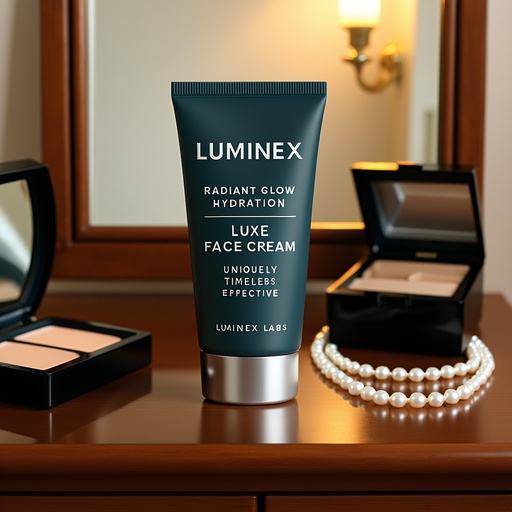} &
    \includegraphics[width=0.16\linewidth]{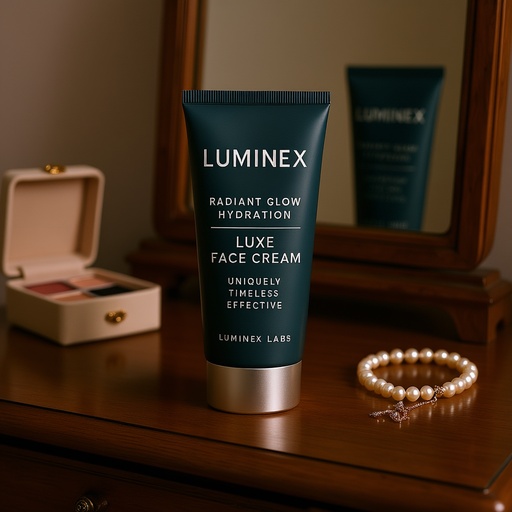} &
    \includegraphics[width=0.16\linewidth]{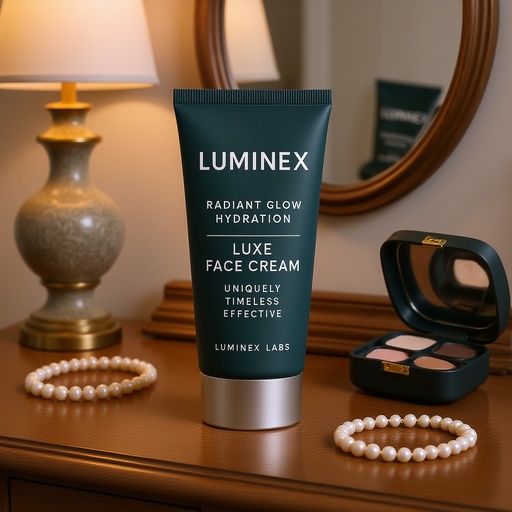} &
    \includegraphics[width=0.16\linewidth]{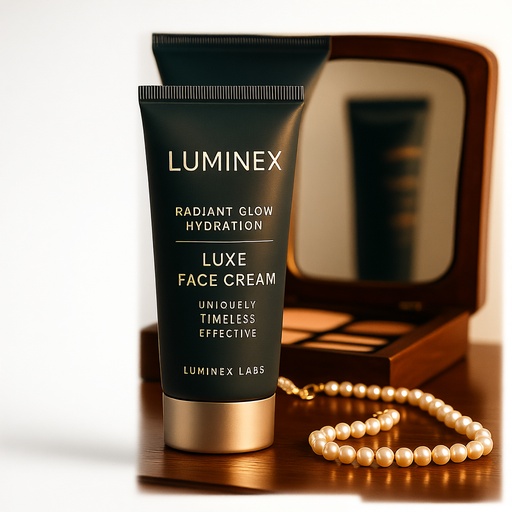} &
    \includegraphics[width=0.16\linewidth]{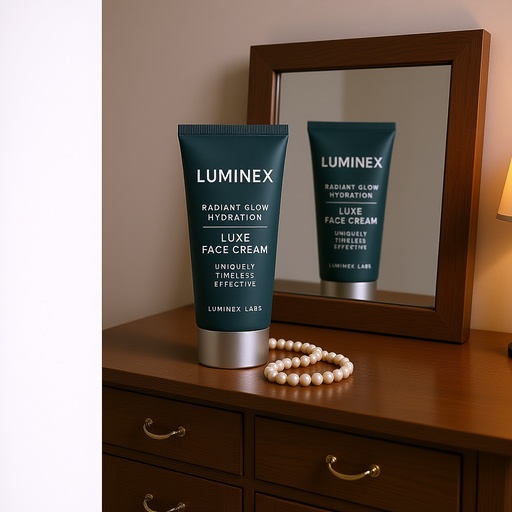} \\
    \multicolumn{6}{p{\textwidth}}{\scriptsize\emph{Position the face cream tube on a sleek wooden dresser with a vintage mirror reflecting its image; use soft, ambient lighting to create warm highlights on the silver trim; include delicate jewelry like a pearl necklace and an open cosmetics box in soft focus around it, emphasizing elegance and luxury.}} \\
    \bottomrule
  \end{tabular}
  }

    \caption{Continuing comparison across six edit instructions. Columns: Input, Omnigen2, Edit-R1-Qwen, Edit-R1-Flux, Replan-Qwen, Replan-Flux. The edit instruction for each input image is present below it.}
  \label{fig:other_models_grid_6x6_fit}
\end{figure*}

As shown in Table \ref{tab:main_results}, the GPT-based evaluation follows trends consistent with our other metrics. Earlier models such as BAGEL\cite{deng2025emerging_bagel}, Hidream-E1-1, Step1x-edit-v1-p2, and OmniGen2 perform noticeably worse across all dimensions, particularly in text rendering, where Hidream-E1-1 receives a score of only 3.44. In contrast, models trained with our approach show clear improvements. For Qwen-Image-Edit-2511, the cyclic reward model increases the text rendering score from 7.59 in the base model to \textbf{8.89}, while also improving overall performance from 8.10 to \textbf{8.70}. Flux1.Kontext already performs strongly but still benefits from cyclic reward training, achieving the highest text rendering score of \textbf{9.074}. These results demonstrate that the improvements observed in automated metrics translate into perceptible gains in MLLM-as-a-judge evaluation as well. Qualitative examples for the Qwen and Flux baseline and finetuned models are present in Figures \ref{fig:qwen_grid} and \ref{fig:flux_grid}, respectively and for other models in Figure \ref{fig:other_models_grid_6x7_fit} and Figure \ref{fig:other_models_grid_6x6_fit}.

\section{Ablation Study}

To understand the contribution of the cyclic consistency reward, we performed an ablation study comparing different reward configurations for GRPO training. In all experiments, RL training is initialized from the Qwen checkpoint fine-tuned on the ProductConsistency-SFT dataset. We replace the Cyclic Consistency reward with several auxiliary reward models designed to encourage product consistency during editing. All training and evaluation settings are kept the same

First, we evaluated the EditReward model, which measures instruction adherence and overall perceptual alignment between the input and edited images. This reward model is trained to capture general instruction following and perceptual quality and is not explicitly optimized to preserve the identity of the product. Second, we evaluated a Segmented Visual Consistency reward, which measures cosine similarity between SigLip-2 \cite{tschannen2025siglip} embeddings extracted from segmented product regions (via GroundingDino and SAM-2) in the input and edited images. By restricting the comparison to the segmented product area, this reward encourages the preservation of localized visual features while remaining invariant to background changes.

\begin{figure*}[t]
\centering

\begin{minipage}[c]{0.4\textwidth}
\centering
\includegraphics[width=0.495\textwidth]{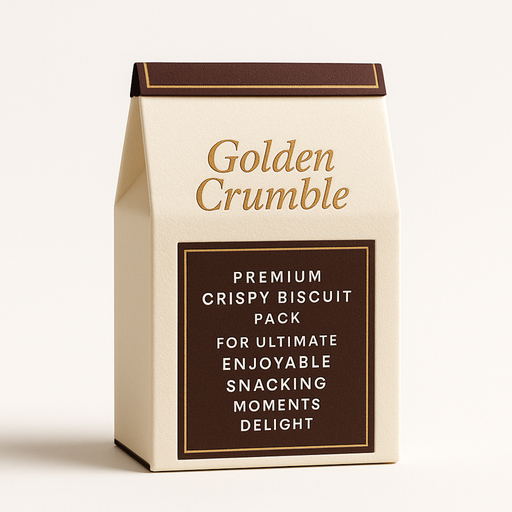}%
\includegraphics[width=0.495\textwidth]{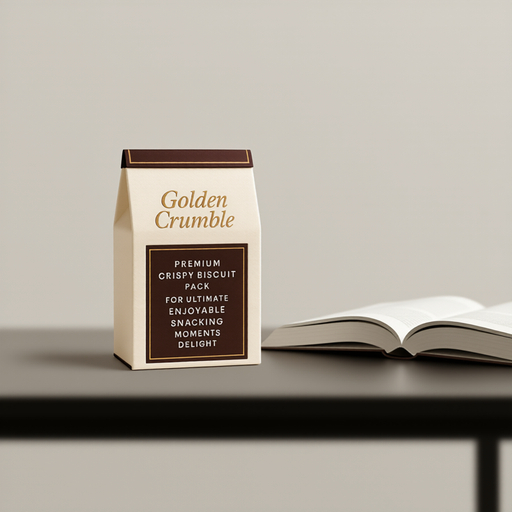}

{\footnotesize (a) Input $\rightarrow$ Output}
\end{minipage}\hspace{4pt}
\begin{minipage}[c]{0.4\textwidth}
\centering
\includegraphics[width=0.495\textwidth]{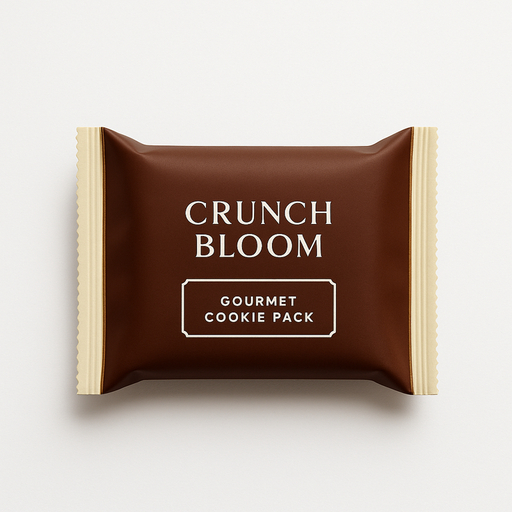}%
\includegraphics[width=0.495\textwidth]{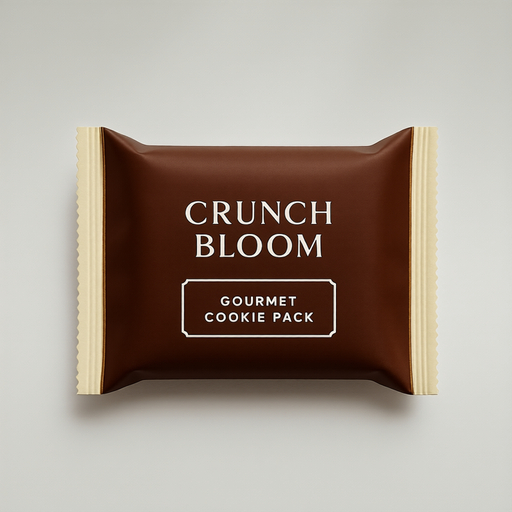}

{\footnotesize (b) Input $\rightarrow$ Output}
\end{minipage}

\vspace{4pt}

\begin{minipage}[c]{0.48\textwidth}
\centering
\includegraphics[width=0.495\textwidth]{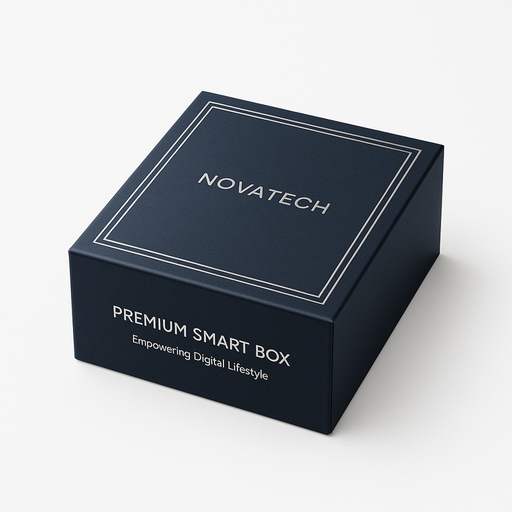}%
\includegraphics[width=0.495\textwidth]{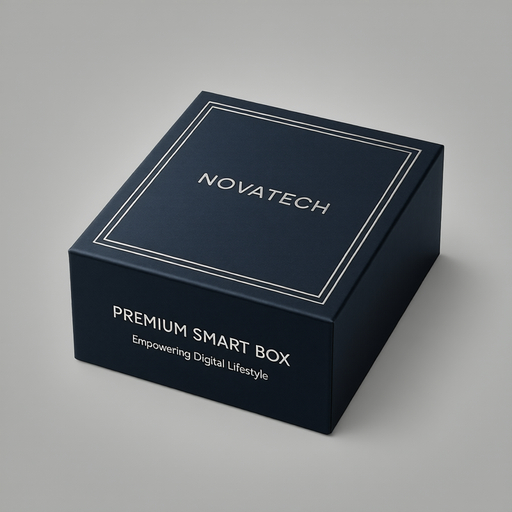}

{\footnotesize (c) Input $\rightarrow$ Output}
\end{minipage}\hspace{4pt}
\begin{minipage}[c]{0.48\textwidth}
\centering
\includegraphics[width=0.495\textwidth]{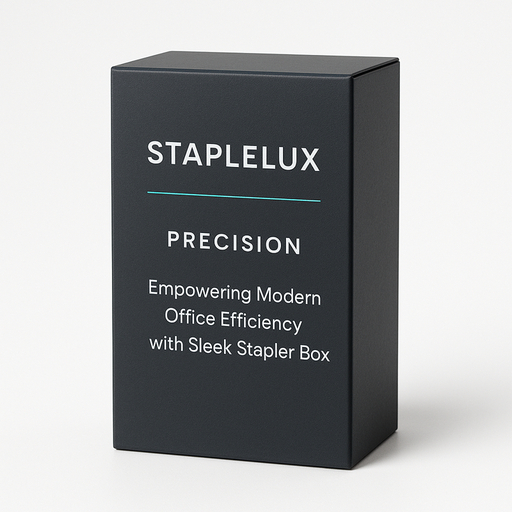}%
\includegraphics[width=0.495\textwidth]{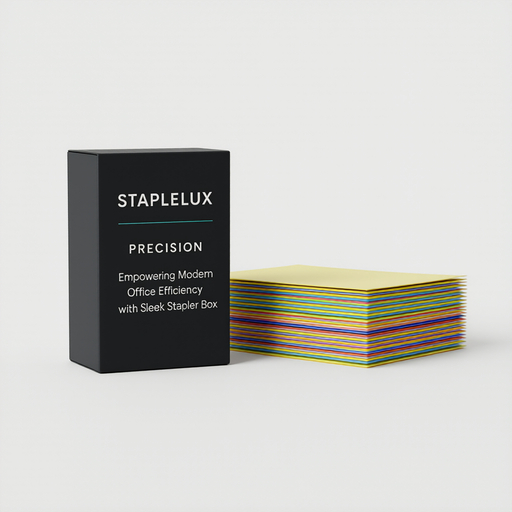}

{\footnotesize (d) Input $\rightarrow$ Output}
\end{minipage}

\caption{\textbf{Example outputs from Segmented Visual consistency reward demonstrating overfitting.}
Edit instructions —
(a) Display the biscuit pack on a dark wooden coffee table alongside an open book and a cozy throw blanket in a softly lit living room; flickering fireplace in the blurred background; intimate and comforting mood; warm tones and soft focus emphasize relaxation.
(b) Place the cookie pack atop an elegant dessert table at a chic outdoor garden party, accompanied by a small arrangement of fresh flowers and a vintage silver tray; dappled sunlight through leaves adds a natural, upscale ambiance; soft focus on surrounding elements keeps the pack as the centerpiece.
(c) Set the box against a luxurious black velvet backdrop with subtle low-key lighting; include a soft-focus silver ribbon partially unwrapped beside it; focus on reflective silver accents with a spotlight creating a vignette effect.
(d) Place the stapler box next to a neatly arranged stack of colorful stapled documents on a vibrant modern coworking table; include an upscale coffee cup and digital tablet; bright lighting conveys productivity.
These examples illustrate the failure mode discussed in the ablation: the model stops following the edit instruction and instead collapses to copying the original product image with minimal changes.}
\label{fig:ablation}
\end{figure*}

Quantitative results are reported in Table~\ref{tab:reward_ablation}. Although the segmented visual consistency reward achieves the highest scores on automated perceptual metrics, we find that it is prone to reward hacking. The model learns to maximize the embedding similarity by reproducing the input image rather than performing the intended edit. This behavior inflates similarity-based metrics, but results in visually degraded outputs that fail to follow the editing instruction, often producing images with poor composition and aesthetic quality. In contrast, the cyclic consistency reward provides a more robust training signal outperforming the EditReward model with a lower CER and better CLIP-I and DINO-I scores and also comes out ahead in GPT-evaluation as well. Qualitative examples that illustrate the reward hacking failure cases of Segmented Visual Consistency are shown in Figure~\ref{fig:ablation}.

\section{Conclusion}
We propose the \textbf{ProductConsistency} dataset designed to improve product-centric instruction-based image editing. Our approach addresses a key limitation of existing editing models by explicitly introducing training data and objectives that enforce product and text consistency, which is largely absent from current datasets.  Our framework includes a SFT dataset for product editing, a RL dataset for reward-driven optimization, and a new evaluation suite, the \textbf{ProductConsistency Benchmark}, for rigorous assessment of product-centric editing capabilities. To guide training, we introduce a \textbf{cyclic consistency reward} that aligns captions generated from edited images with the original product description, while incorporating OCR-based rewards to ensure accurate text rendering. Extensive experiments demonstrate that models fine-tuned with our dataset significantly outperform strong baselines across multiple automated metrics and MLLM-based evaluations. These results highlight the effectiveness of explicitly modeling product consistency for instruction-based image editing.



\newpage
{
    \small
    \bibliographystyle{ieeenat_fullname}
    \bibliography{main}

\begin{thebibliography}{57}
\providecommand{\natexlab}[1]{#1}
\providecommand{\url}[1]{\texttt{#1}}
\expandafter\ifx\csname urlstyle\endcsname\relax
  \providecommand{\doi}[1]{doi: #1}\else
  \providecommand{\doi}{doi: \begingroup \urlstyle{rm}\Url}\fi

\bibitem[Bai et~al.(2020)Bai, Chen, Yu, Wang, and Zhang]{bai2020products}
Yalong Bai, Yuxiang Chen, Wei Yu, Linfang Wang, and Wei Zhang.
\newblock Products-10k: A large-scale product recognition dataset.
\newblock \emph{arXiv preprint arXiv:2008.10545}, 2020.

\bibitem[{Black Forest Labs}(2024)]{blackforestlabs_flux1_fill_dev_2024_blx}
{Black Forest Labs}.
\newblock Flux.1 fill [dev], 2024.
\newblock Model repository on Hugging Face.

\bibitem[Brooks et~al.(2023)Brooks, Holynski, and Efros]{brooks2023instructpix2pix}
Tim Brooks, Aleksander Holynski, and Alexei~A Efros.
\newblock Instructpix2pix: Learning to follow image editing instructions.
\newblock In \emph{Proceedings of the IEEE/CVF conference on computer vision and pattern recognition}, pages 18392--18402, 2023.

\bibitem[Cai et~al.(2025)Cai, Chen, Chen, Li, Long, Pan, Qiu, Zhang, Gao, Xu, et~al.]{cai2025hidream}
Qi Cai, Jingwen Chen, Yang Chen, Yehao Li, Fuchen Long, Yingwei Pan, Zhaofan Qiu, Yiheng Zhang, Fengbin Gao, Peihan Xu, et~al.
\newblock Hidream-i1: A high-efficient image generative foundation model with sparse diffusion transformer.
\newblock \emph{arXiv preprint arXiv:2505.22705}, 2025.

\bibitem[Chen et~al.(2024)Chen, Chen, Zhang, Wang, Liu, Zhou, Zhang, Wan, Zhou, and Sun]{chen2024mllm}
Dongping Chen, Ruoxi Chen, Shilin Zhang, Yaochen Wang, Yinuo Liu, Huichi Zhou, Qihui Zhang, Yao Wan, Pan Zhou, and Lichao Sun.
\newblock Mllm-as-a-judge: Assessing multimodal llm-as-a-judge with vision-language benchmark.
\newblock In \emph{Forty-first International Conference on Machine Learning}, 2024.

\bibitem[Cho et~al.(2022)Cho, Yoon, Kale, Dernoncourt, Bui, and Bansal]{cho2022fineclipmodel}
Jaemin Cho, Seunghyun Yoon, Ajinkya Kale, Franck Dernoncourt, Trung Bui, and Mohit Bansal.
\newblock Fine-grained image captioning with clip reward.
\newblock In \emph{Findings of the Association for Computational Linguistics: NAACL 2022}, pages 517--527, 2022.

\bibitem[Collins et~al.(2022)Collins, Goel, Deng, Luthra, Xu, Gundogdu, Zhang, Vicente, Dideriksen, Arora, et~al.]{collins2022abo}
Jasmine Collins, Shubham Goel, Kenan Deng, Achleshwar Luthra, Leon Xu, Erhan Gundogdu, Xi Zhang, Tomas F~Yago Vicente, Thomas Dideriksen, Himanshu Arora, et~al.
\newblock Abo: Dataset and benchmarks for real-world 3d object understanding.
\newblock In \emph{Proceedings of the IEEE/CVF conference on computer vision and pattern recognition}, pages 21126--21136, 2022.

\bibitem[Deng et~al.(2025)Deng, Zhu, Li, Gou, Li, Wang, Zhong, Yu, Nie, Song, et~al.]{deng2025emerging_bagel}
Chaorui Deng, Deyao Zhu, Kunchang Li, Chenhui Gou, Feng Li, Zeyu Wang, Shu Zhong, Weihao Yu, Xiaonan Nie, Ziang Song, et~al.
\newblock Emerging properties in unified multimodal pretraining.
\newblock \emph{arXiv preprint arXiv:2505.14683}, 2025.

\bibitem[Dong et~al.(2023)Dong, Xue, Duan, and Han]{dong2023prompt_inversion}
Wenkai Dong, Song Xue, Xiaoyue Duan, and Shumin Han.
\newblock Prompt tuning inversion for text-driven image editing using diffusion models.
\newblock In \emph{Proceedings of the IEEE/CVF international conference on computer vision}, pages 7430--7440, 2023.

\bibitem[Guo et~al.(2025)Guo, Yang, Zhang, Song, Wang, Zhu, Xu, Zhang, Ma, Bi, et~al.]{guo2025deepseekr1}
Daya Guo, Dejian Yang, Haowei Zhang, Junxiao Song, Peiyi Wang, Qihao Zhu, Runxin Xu, Ruoyu Zhang, Shirong Ma, Xiao Bi, et~al.
\newblock Deepseek-r1: Incentivizing reasoning capability in llms via reinforcement learning.
\newblock \emph{arXiv preprint arXiv:2501.12948}, 2025.

\bibitem[He et~al.(2025)He, Fu, Zhao, Li, Yang, Yin, Rao, and Zhang]{he2025tempflowgrpo}
Xiaoxuan He, Siming Fu, Yuke Zhao, Wanli Li, Jian Yang, Dacheng Yin, Fengyun Rao, and Bo Zhang.
\newblock Tempflow-grpo: When timing matters for grpo in flow models.
\newblock \emph{arXiv preprint arXiv:2508.04324}, 2025.

\bibitem[Ju et~al.(2024)Ju, Liu, Wang, Bian, Shan, and Xu]{ju2024brushnet}
Xuan Ju, Xian Liu, Xintao Wang, Yuxuan Bian, Ying Shan, and Qiang Xu.
\newblock Brushnet: A plug-and-play image inpainting model with decomposed dual-branch diffusion.
\newblock In \emph{European Conference on Computer Vision}, pages 150--168. Springer, 2024.

\bibitem[Ku et~al.(2024)Ku, Jiang, Wei, Yue, and Chen]{ku2024viescore}
Max Ku, Dongfu Jiang, Cong Wei, Xiang Yue, and Wenhu Chen.
\newblock Viescore: Towards explainable metrics for conditional image synthesis evaluation.
\newblock In \emph{Proceedings of the 62nd Annual Meeting of the Association for Computational Linguistics (Volume 1: Long Papers)}, pages 12268--12290, 2024.

\bibitem[Labs et~al.(2025)Labs, Batifol, Blattmann, Boesel, Consul, Diagne, Dockhorn, English, English, Esser, et~al.]{labs2025flux}
Black~Forest Labs, Stephen Batifol, Andreas Blattmann, Frederic Boesel, Saksham Consul, Cyril Diagne, Tim Dockhorn, Jack English, Zion English, Patrick Esser, et~al.
\newblock Flux. 1 kontext: Flow matching for in-context image generation and editing in latent space.
\newblock \emph{arXiv preprint arXiv:2506.15742}, 2025.

\bibitem[Li et~al.(2025{\natexlab{a}})Li, Zhang, Zheng, Guo, Jia, Feng, Yu, Liu, Feng, Pei, et~al.]{li2025editthinker}
Hongyu Li, Manyuan Zhang, Dian Zheng, Ziyu Guo, Yimeng Jia, Kaituo Feng, Hao Yu, Yexin Liu, Yan Feng, Peng Pei, et~al.
\newblock Editthinker: Unlocking iterative reasoning for any image editor.
\newblock \emph{arXiv preprint arXiv:2512.05965}, 2025{\natexlab{a}}.

\bibitem[Li et~al.(2025{\natexlab{b}})Li, Cui, Huang, Ma, Fan, Yang, and Zhong]{li2025mixgrpo}
Junzhe Li, Yutao Cui, Tao Huang, Yinping Ma, Chun Fan, Miles Yang, and Zhao Zhong.
\newblock Mixgrpo: Unlocking flow-based grpo efficiency with mixed ode-sde.
\newblock \emph{arXiv preprint arXiv:2507.21802}, 2025{\natexlab{b}}.

\bibitem[Li et~al.(2024{\natexlab{a}})Li, Yang, Kuang, Wu, Wang, Xiao, and Chen]{li2024controlnet++}
Ming Li, Taojiannan Yang, Huafeng Kuang, Jie Wu, Zhaoning Wang, Xuefeng Xiao, and Chen Chen.
\newblock Controlnet++: Improving conditional controls with efficient consistency feedback: Project page: liming-ai. github. io/controlnet\_plus\_plus.
\newblock In \emph{European Conference on Computer Vision}, pages 129--147. Springer, 2024{\natexlab{a}}.

\bibitem[Li et~al.(2023)Li, Van De~Weijer, Hu, Khan, Hou, Wang, Yang, and Cheng]{li2023stylediffusion_inversion}
Senmao Li, Joost Van De~Weijer, Taihang Hu, Fahad~Shahbaz Khan, Qibin Hou, Yaxing Wang, Jian Yang, and Ming-Ming Cheng.
\newblock Stylediffusion: Prompt-embedding inversion for text-based editing.
\newblock \emph{arXiv preprint arXiv:2303.15649}, 2023.

\bibitem[Li et~al.(2025{\natexlab{c}})Li, Kallidromitis, Gokul, Koneru, Kato, Kozuka, and Grover]{li2025reflect}
Shufan Li, Konstantinos Kallidromitis, Akash Gokul, Arsh Koneru, Yusuke Kato, Kazuki Kozuka, and Aditya Grover.
\newblock Reflect-dit: Inference-time scaling for text-to-image diffusion transformers via in-context reflection.
\newblock In \emph{Proceedings of the IEEE/CVF International Conference on Computer Vision}, pages 15657--15668, 2025{\natexlab{c}}.

\bibitem[Li et~al.(2024{\natexlab{b}})Li, Liu, Chen, and Liu]{li2024instructrl4pix}
Tiancheng Li, Jinxiu Liu, Huajun Chen, and Qi Liu.
\newblock Instructrl4pix: Training diffusion for image editing by reinforcement learning.
\newblock \emph{arXiv preprint arXiv:2406.09973}, 2024{\natexlab{b}}.

\bibitem[Li et~al.(2024{\natexlab{c}})Li, Bian, Ju, Zhang, Zhuang, Shan, Zou, and Xu]{li2024brushedit}
Yaowei Li, Yuxuan Bian, Xuan Ju, Zhaoyang Zhang, Junhao Zhuang, Ying Shan, Yuexian Zou, and Qiang Xu.
\newblock Brushedit: All-in-one image inpainting and editing.
\newblock \emph{arXiv preprint arXiv:2412.10316}, 2024{\natexlab{c}}.

\bibitem[Li et~al.(2025{\natexlab{d}})Li, Liu, Zhang, Lin, Wu, Yuan, Yan, Ye, Yu, Niu, et~al.]{li2025uniworld}
Zongjian Li, Zheyuan Liu, Qihui Zhang, Bin Lin, Feize Wu, Shenghai Yuan, Zhiyuan Yan, Yang Ye, Wangbo Yu, Yuwei Niu, et~al.
\newblock Uniworld-v2: Reinforce image editing with diffusion negative-aware finetuning and mllm implicit feedback.
\newblock \emph{arXiv preprint arXiv:2510.16888}, 2025{\natexlab{d}}.

\bibitem[Liang et~al.(2025)Liang, Luo, Guo, and Bi]{liang2025evaluation_hfpc}
Yuqi Liang, Jun Luo, Xiaoxi Guo, and Jianqi Bi.
\newblock An evaluation framework for product images background inpainting based on human feedback and product consistency.
\newblock In \emph{Proceedings of the AAAI Conference on Artificial Intelligence}, pages 478--486, 2025.

\bibitem[Liu et~al.(2025{\natexlab{a}})Liu, Liu, Liang, Li, Liu, Wang, Wan, Zhang, and Ouyang]{liu2025flowgrpo}
Jie Liu, Gongye Liu, Jiajun Liang, Yangguang Li, Jiaheng Liu, Xintao Wang, Pengfei Wan, Di Zhang, and Wanli Ouyang.
\newblock Flow-grpo: Training flow matching models via online rl.
\newblock \emph{arXiv preprint arXiv:2505.05470}, 2025{\natexlab{a}}.

\bibitem[Liu et~al.(2024)Liu, Zeng, Ren, Li, Zhang, Yang, Jiang, Li, Yang, Su, et~al.]{liu2024grounding}
Shilong Liu, Zhaoyang Zeng, Tianhe Ren, Feng Li, Hao Zhang, Jie Yang, Qing Jiang, Chunyuan Li, Jianwei Yang, Hang Su, et~al.
\newblock Grounding dino: Marrying dino with grounded pre-training for open-set object detection.
\newblock In \emph{European conference on computer vision}, pages 38--55. Springer, 2024.

\bibitem[Liu et~al.(2025{\natexlab{b}})Liu, Han, Xing, Yin, Wang, Cheng, Liao, Wang, Fu, Han, et~al.]{liu2025step1x}
Shiyu Liu, Yucheng Han, Peng Xing, Fukun Yin, Rui Wang, Wei Cheng, Jiaqi Liao, Yingming Wang, Honghao Fu, Chunrui Han, et~al.
\newblock Step1x-edit: A practical framework for general image editing.
\newblock \emph{arXiv preprint arXiv:2504.17761}, 2025{\natexlab{b}}.

\bibitem[Liu et~al.(2026)Liu, Dong, Lu, Diao, Belcak, Liu, Chen, Yin, Wang, Cheng, et~al.]{liu2026gdpo}
Shih-Yang Liu, Xin Dong, Ximing Lu, Shizhe Diao, Peter Belcak, Mingjie Liu, Min-Hung Chen, Hongxu Yin, Yu-Chiang~Frank Wang, Kwang-Ting Cheng, et~al.
\newblock Gdpo: Group reward-decoupled normalization policy optimization for multi-reward rl optimization.
\newblock \emph{arXiv preprint arXiv:2601.05242}, 2026.

\bibitem[Luo et~al.(2025)Luo, Wang, Wu, Xiao, Jiang, Lian, Zhang, Liu, et~al.]{luo2025editscore}
Xin Luo, Jiahao Wang, Chenyuan Wu, Shitao Xiao, Xiyan Jiang, Defu Lian, Jiajun Zhang, Dong Liu, et~al.
\newblock Editscore: Unlocking online rl for image editing via high-fidelity reward modeling.
\newblock \emph{arXiv preprint arXiv:2509.23909}, 2025.

\bibitem[Ma et~al.(2025)Ma, Wu, Sun, and Li]{ma2025hpsv3}
Yuhang Ma, Xiaoshi Wu, Keqiang Sun, and Hongsheng Li.
\newblock Hpsv3: Towards wide-spectrum human preference score.
\newblock In \emph{Proceedings of the IEEE/CVF International Conference on Computer Vision}, pages 15086--15095, 2025.

\bibitem[Malhi et~al.(2025)Malhi, Dutta, Talius, Ma, Driscoll, Holden, Pruthi, and Narayanaswamy]{malhi2025preserving_seg_clip}
Ishaan Malhi, Praneet Dutta, Ellie Talius, Sally Ma, Brendan Driscoll, Krista Holden, Garima Pruthi, and Arunachalam Narayanaswamy.
\newblock Preserving product fidelity in large scale image recontextualization with diffusion models.
\newblock \emph{arXiv preprint arXiv:2503.08729}, 2025.

\bibitem[Mokady et~al.(2023)Mokady, Hertz, Aberman, Pritch, and Cohen-Or]{mokady2023null_inversion}
Ron Mokady, Amir Hertz, Kfir Aberman, Yael Pritch, and Daniel Cohen-Or.
\newblock Null-text inversion for editing real images using guided diffusion models.
\newblock In \emph{Proceedings of the IEEE/CVF conference on computer vision and pattern recognition}, pages 6038--6047, 2023.

\bibitem[Mou et~al.(2024)Mou, Wang, Xie, Wu, Zhang, Qi, and Shan]{mou2024t2i_adapter}
Chong Mou, Xintao Wang, Liangbin Xie, Yanze Wu, Jian Zhang, Zhongang Qi, and Ying Shan.
\newblock T2i-adapter: Learning adapters to dig out more controllable ability for text-to-image diffusion models.
\newblock In \emph{Proceedings of the AAAI conference on artificial intelligence}, pages 4296--4304, 2024.

\bibitem[OpenAI(2025{\natexlab{a}})]{openai2025gpt51}
OpenAI.
\newblock Gpt-5.1: A smarter, more conversational chatgpt, 2025{\natexlab{a}}.
\newblock OpenAI Product Release.

\bibitem[OpenAI(2025{\natexlab{b}})]{openai2025gptimage1}
OpenAI.
\newblock Introducing gpt image 1 (gpt-4o image generation), 2025{\natexlab{b}}.
\newblock Initial GPT Image 1 release (March 25, 2025).

\bibitem[Ping et~al.(2025)Ping, Jia, Luo, Xia, Shen, Dang, and Qian]{ping2025paco_rl}
Bowen Ping, Chengyou Jia, Minnan Luo, Changliang Xia, Xin Shen, Zhuohang Dang, and Hangwei Qian.
\newblock Paco-rl: Advancing reinforcement learning for consistent image generation with pairwise reward modeling.
\newblock \emph{arXiv preprint arXiv:2512.04784}, 2025.

\bibitem[Podell et~al.(2023)Podell, English, Lacey, Blattmann, Dockhorn, M{\"u}ller, Penna, and Rombach]{podell2023sdxl}
Dustin Podell, Zion English, Kyle Lacey, Andreas Blattmann, Tim Dockhorn, Jonas M{\"u}ller, Joe Penna, and Robin Rombach.
\newblock Sdxl: Improving latent diffusion models for high-resolution image synthesis.
\newblock \emph{arXiv preprint arXiv:2307.01952}, 2023.

\bibitem[Qin et~al.(2023)Qin, Zhang, Yu, Feng, Yang, Zhou, Wang, Niebles, Xiong, Savarese, et~al.]{qin2023unicontrol}
Can Qin, Shu Zhang, Ning Yu, Yihao Feng, Xinyi Yang, Yingbo Zhou, Huan Wang, Juan~Carlos Niebles, Caiming Xiong, Silvio Savarese, et~al.
\newblock Unicontrol: A unified diffusion model for controllable visual generation in the wild.
\newblock \emph{arXiv preprint arXiv:2305.11147}, 2023.

\bibitem[Qu et~al.(2025)Qu, Ke, Zhan, Tang, Liu, Peng, Yu, Yu, and Jia]{qu2025replan}
Tianyuan Qu, Lei Ke, Xiaohang Zhan, Longxiang Tang, Yuqi Liu, Bohao Peng, Bei Yu, Dong Yu, and Jiaya Jia.
\newblock Replan: Reasoning-guided region planning for complex instruction-based image editing.
\newblock \emph{arXiv preprint arXiv:2512.16864}, 2025.

\bibitem[Radford et~al.(2021)Radford, Kim, Hallacy, Ramesh, Goh, Agarwal, Sastry, Askell, Mishkin, Clark, et~al.]{radford2021learning}
Alec Radford, Jong~Wook Kim, Chris Hallacy, Aditya Ramesh, Gabriel Goh, Sandhini Agarwal, Girish Sastry, Amanda Askell, Pamela Mishkin, Jack Clark, et~al.
\newblock Learning transferable visual models from natural language supervision.
\newblock In \emph{International conference on machine learning}, pages 8748--8763. PmLR, 2021.

\bibitem[Ravi et~al.(2024)Ravi, Gabeur, Hu, Hu, Ryali, Ma, Khedr, R{\"a}dle, Rolland, Gustafson, et~al.]{ravi2024sam}
Nikhila Ravi, Valentin Gabeur, Yuan-Ting Hu, Ronghang Hu, Chaitanya Ryali, Tengyu Ma, Haitham Khedr, Roman R{\"a}dle, Chloe Rolland, Laura Gustafson, et~al.
\newblock Sam 2: Segment anything in images and videos.
\newblock \emph{arXiv preprint arXiv:2408.00714}, 2024.

\bibitem[Rombach et~al.(2022)Rombach, Blattmann, Lorenz, Esser, and Ommer]{1rombach2022highdiffusionmodel}
Robin Rombach, Andreas Blattmann, Dominik Lorenz, Patrick Esser, and Bj{\"o}rn Ommer.
\newblock High-resolution image synthesis with latent diffusion models.
\newblock In \emph{Proceedings of the IEEE/CVF conference on computer vision and pattern recognition}, pages 10684--10695, 2022.

\bibitem[Ruiz et~al.(2023)Ruiz, Li, Jampani, Pritch, Rubinstein, and Aberman]{ruiz2023dreambooth_seg_clip}
Nataniel Ruiz, Yuanzhen Li, Varun Jampani, Yael Pritch, Michael Rubinstein, and Kfir Aberman.
\newblock Dreambooth: Fine tuning text-to-image diffusion models for subject-driven generation.
\newblock In \emph{Proceedings of the IEEE/CVF conference on computer vision and pattern recognition}, pages 22500--22510, 2023.

\bibitem[Thon and Wilde(2025)]{thon2025laionaesthetics}
Jan-No{\"e}l Thon and Lukas~RA Wilde.
\newblock Introduction: Ai aesthetics.
\newblock In \emph{AI Aesthetics}, pages 1--21. Routledge, 2025.

\bibitem[Tschannen et~al.(2025)Tschannen, Gritsenko, Wang, Naeem, Alabdulmohsin, Parthasarathy, Evans, Beyer, Xia, Mustafa, et~al.]{tschannen2025siglip}
Michael Tschannen, Alexey Gritsenko, Xiao Wang, Muhammad~Ferjad Naeem, Ibrahim Alabdulmohsin, Nikhil Parthasarathy, Talfan Evans, Lucas Beyer, Ye Xia, Basil Mustafa, et~al.
\newblock Siglip 2: Multilingual vision-language encoders with improved semantic understanding, localization, and dense features.
\newblock \emph{arXiv preprint arXiv:2502.14786}, 2025.

\bibitem[Wu et~al.(2025{\natexlab{a}})Wu, Li, Zhou, Lin, Gao, Yan, Yin, Bai, Xu, Chen, et~al.]{wu2025qwen}
Chenfei Wu, Jiahao Li, Jingren Zhou, Junyang Lin, Kaiyuan Gao, Kun Yan, Sheng-ming Yin, Shuai Bai, Xiao Xu, Yilei Chen, et~al.
\newblock Qwen-image technical report.
\newblock \emph{arXiv preprint arXiv:2508.02324}, 2025{\natexlab{a}}.

\bibitem[Wu et~al.(2025{\natexlab{b}})Wu, Zheng, Yan, Xiao, Luo, Wang, Li, Jiang, Liu, Zhou, et~al.]{wu2025omnigen2}
Chenyuan Wu, Pengfei Zheng, Ruiran Yan, Shitao Xiao, Xin Luo, Yueze Wang, Wanli Li, Xiyan Jiang, Yexin Liu, Junjie Zhou, et~al.
\newblock Omnigen2: Exploration to advanced multimodal generation.
\newblock \emph{arXiv preprint arXiv:2506.18871}, 2025{\natexlab{b}}.

\bibitem[Wu et~al.(2025{\natexlab{c}})Wu, Jiang, Ku, Nie, Liu, and Chen]{wu2025editreward}
Keming Wu, Sicong Jiang, Max Ku, Ping Nie, Minghao Liu, and Wenhu Chen.
\newblock Editreward: A human-aligned reward model for instruction-guided image editing.
\newblock \emph{arXiv preprint arXiv:2509.26346}, 2025{\natexlab{c}}.

\bibitem[Wu et~al.(2023)Wu, Hao, Sun, Chen, Zhu, Zhao, and Li]{wu2023human_hps2}
Xiaoshi Wu, Yiming Hao, Keqiang Sun, Yixiong Chen, Feng Zhu, Rui Zhao, and Hongsheng Li.
\newblock Human preference score v2: A solid benchmark for evaluating human preferences of text-to-image synthesis.
\newblock \emph{arXiv preprint arXiv:2306.09341}, 2023.

\bibitem[Xiao et~al.(2025)Xiao, Wang, Zhou, Yuan, Xing, Yan, Li, Wang, Huang, and Liu]{xiao2025omnigen}
Shitao Xiao, Yueze Wang, Junjie Zhou, Huaying Yuan, Xingrun Xing, Ruiran Yan, Chaofan Li, Shuting Wang, Tiejun Huang, and Zheng Liu.
\newblock Omnigen: Unified image generation.
\newblock In \emph{Proceedings of the IEEE/CVF Conference on Computer Vision and Pattern Recognition}, pages 13294--13304, 2025.

\bibitem[Xu et~al.(2023)Xu, Liu, Wu, Tong, Li, Ding, Tang, and Dong]{xu2023imagereward}
Jiazheng Xu, Xiao Liu, Yuchen Wu, Yuxuan Tong, Qinkai Li, Ming Ding, Jie Tang, and Yuxiao Dong.
\newblock Imagereward: Learning and evaluating human preferences for text-to-image generation.
\newblock \emph{Advances in Neural Information Processing Systems}, 36:\penalty0 15903--15935, 2023.

\bibitem[Ye et~al.(2023)Ye, Zhang, Liu, Han, and Yang]{ye2023ip}
Hu Ye, Jun Zhang, Sibo Liu, Xiao Han, and Wei Yang.
\newblock Ip-adapter: Text compatible image prompt adapter for text-to-image diffusion models.
\newblock \emph{arXiv preprint arXiv:2308.06721}, 2023.

\bibitem[Yin et~al.(2025)Yin, Liu, Han, Wang, Xing, Wang, Cheng, Wang, Li, Yin, et~al.]{yin2025reasonedit}
Fukun Yin, Shiyu Liu, Yucheng Han, Zhibo Wang, Peng Xing, Rui Wang, Wei Cheng, Yingming Wang, Aojie Li, Zixin Yin, et~al.
\newblock Reasonedit: Towards reasoning-enhanced image editing models.
\newblock \emph{arXiv preprint arXiv:2511.22625}, 2025.

\bibitem[Zhang et~al.(2025{\natexlab{a}})Zhang, He, Yan, Shen, and Tang]{zhang2025r_genie}
Dong Zhang, Lingfeng He, Rui Yan, Fei Shen, and Jinhui Tang.
\newblock R-genie: Reasoning-guided generative image editing.
\newblock \emph{arXiv preprint arXiv:2505.17768}, 2025{\natexlab{a}}.

\bibitem[Zhang et~al.(2023)Zhang, Rao, and Agrawala]{zhang2023adding_controlnet}
Lvmin Zhang, Anyi Rao, and Maneesh Agrawala.
\newblock Adding conditional control to text-to-image diffusion models.
\newblock In \emph{Proceedings of the IEEE/CVF international conference on computer vision}, pages 3836--3847, 2023.

\bibitem[Zhang et~al.(2025{\natexlab{b}})Zhang, Yuan, Song, Wang, and Liu]{zhang2025easycontrol}
Yuxuan Zhang, Yirui Yuan, Yiren Song, Haofan Wang, and Jiaming Liu.
\newblock Easycontrol: Adding efficient and flexible control for diffusion transformer.
\newblock In \emph{Proceedings of the IEEE/CVF International Conference on Computer Vision}, pages 19513--19524, 2025{\natexlab{b}}.

\bibitem[Zhu et~al.(2025)Zhu, Li, Ma, He, and Li]{zhu2025multibooth_seg_clip}
Chenyang Zhu, Kai Li, Yue Ma, Chunming He, and Xiu Li.
\newblock Multibooth: Towards generating all your concepts in an image from text.
\newblock In \emph{Proceedings of the AAAI Conference on Artificial Intelligence}, pages 10923--10931, 2025.

\bibitem[Zou et~al.(2025)Zou, Yue, Du, Bao, Li, Xie, Xu, Zhou, Wang, Hu, et~al.]{zou2025beyond_MURE}
Zhentao Zou, Zhengrong Yue, Kunpeng Du, Binlei Bao, Hanting Li, Haizhen Xie, Guozheng Xu, Yue Zhou, Yali Wang, Jie Hu, et~al.
\newblock Beyond textual cot: Interleaved text-image chains with deep confidence reasoning for image editing.
\newblock \emph{arXiv preprint arXiv:2510.08157}, 2025.

\end{thebibliography}
}
\clearpage
\setcounter{page}{1}
\maketitlesupplementary

\section{Qualitative Evaluation}
\begin{figure*}[t]
  \centering
  \setlength{\tabcolsep}{3pt}
  \renewcommand{\arraystretch}{1.0}

  \begin{tabular}{@{}cccc@{}}
    \toprule
    \textbf{Input Image} & \textbf{Baseline} & \textbf{SFT} & \textbf{SFT+GRPO} \\
    \midrule

    \includegraphics[width=0.235\linewidth]{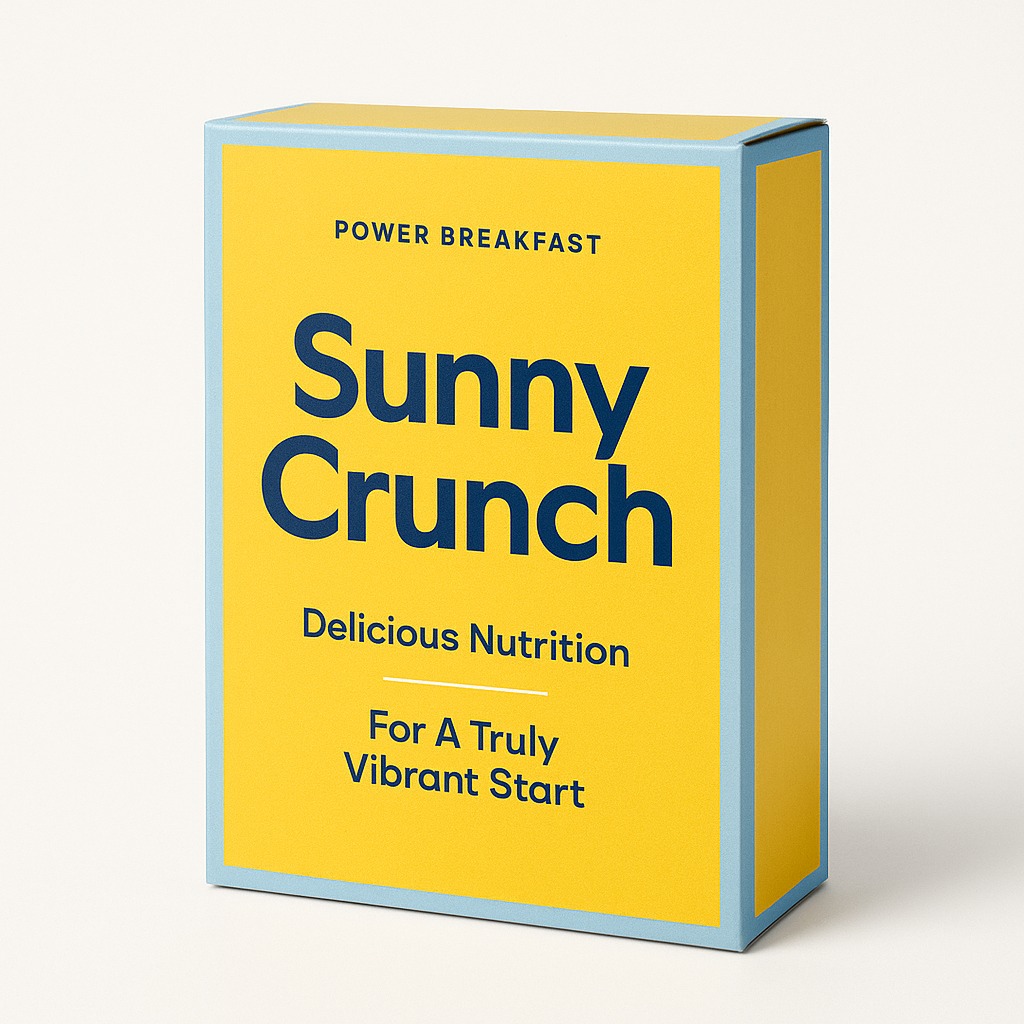} &
    \includegraphics[width=0.235\linewidth]{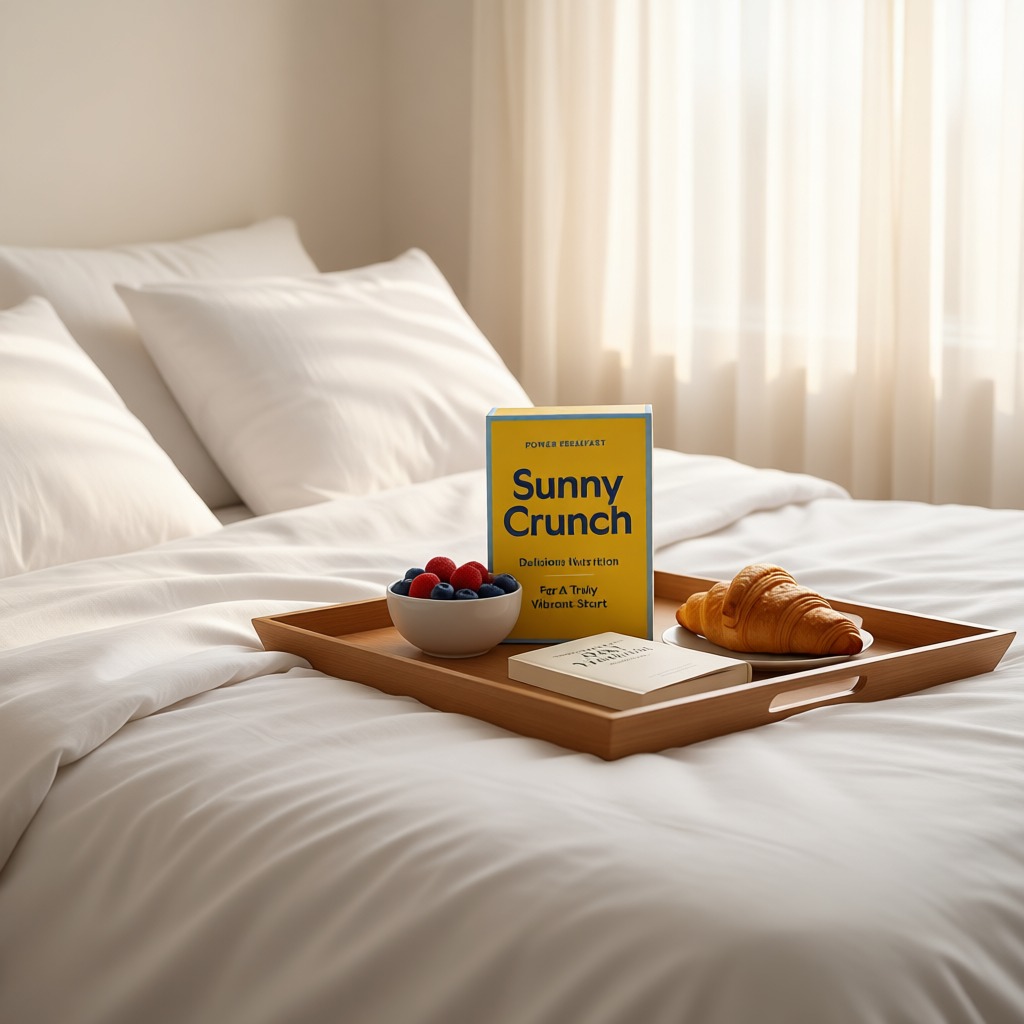} &
    \includegraphics[width=0.235\linewidth]{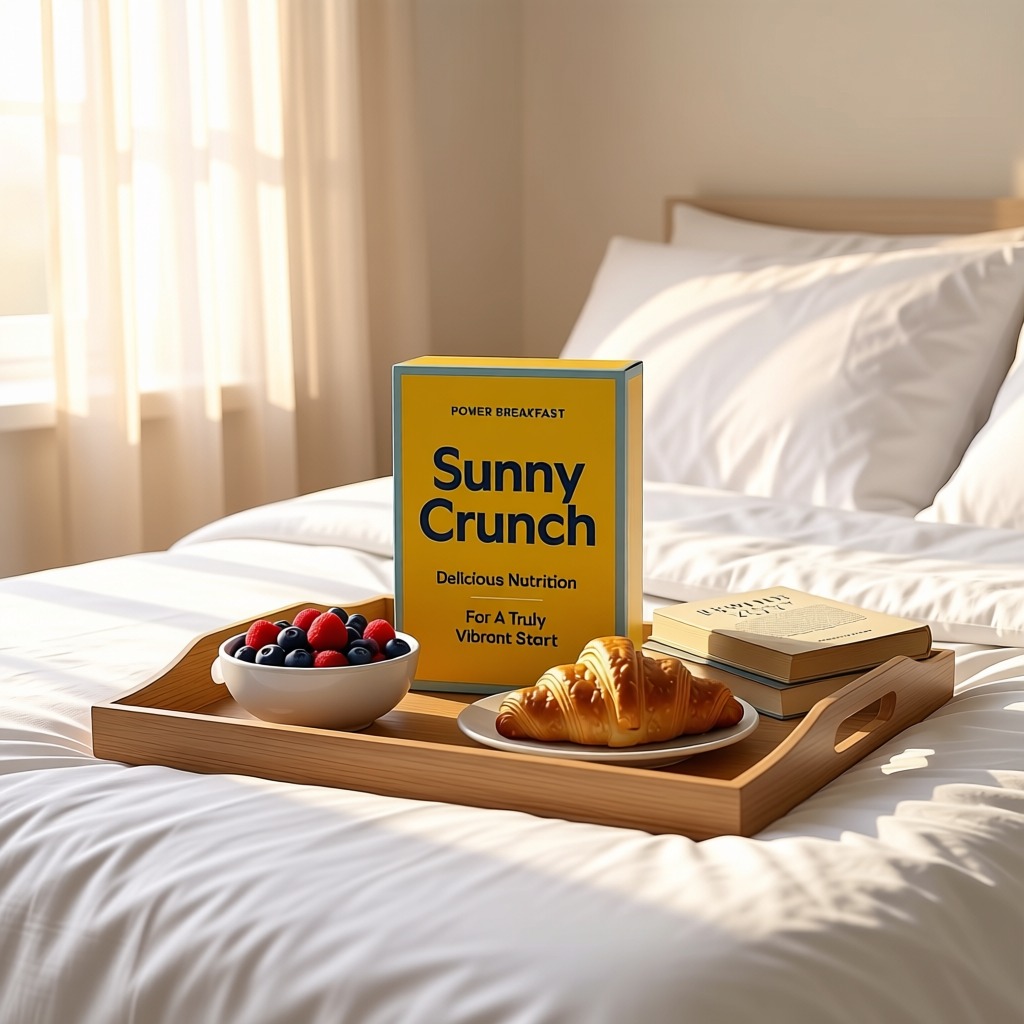} &
    \includegraphics[width=0.235\linewidth]{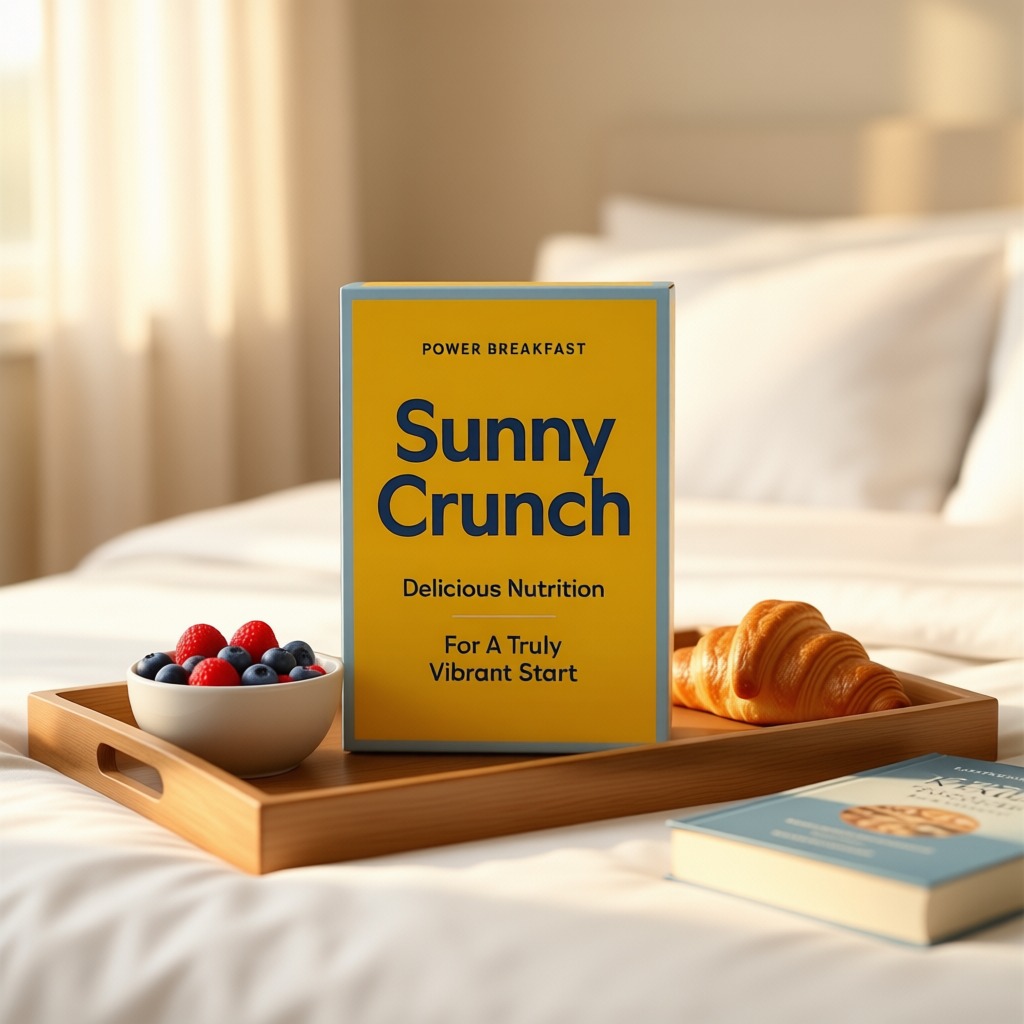} \\
    \multicolumn{4}{p{\dimexpr\linewidth-2\tabcolsep\relax}}{%
      \raggedright\scriptsize\emph{
        Feature the cereal box on a breakfast tray on a neatly made bed 
        with soft white linens; include a small bowl of berries, a croissant, 
        and a novel as supporting elements; gentle morning light filtering through 
        sheer curtains for a cozy, indulgent mood; keep the composition balanced 
        and the product sharply in focus.
      }
    }\\[4pt]

    \includegraphics[width=0.235\linewidth]{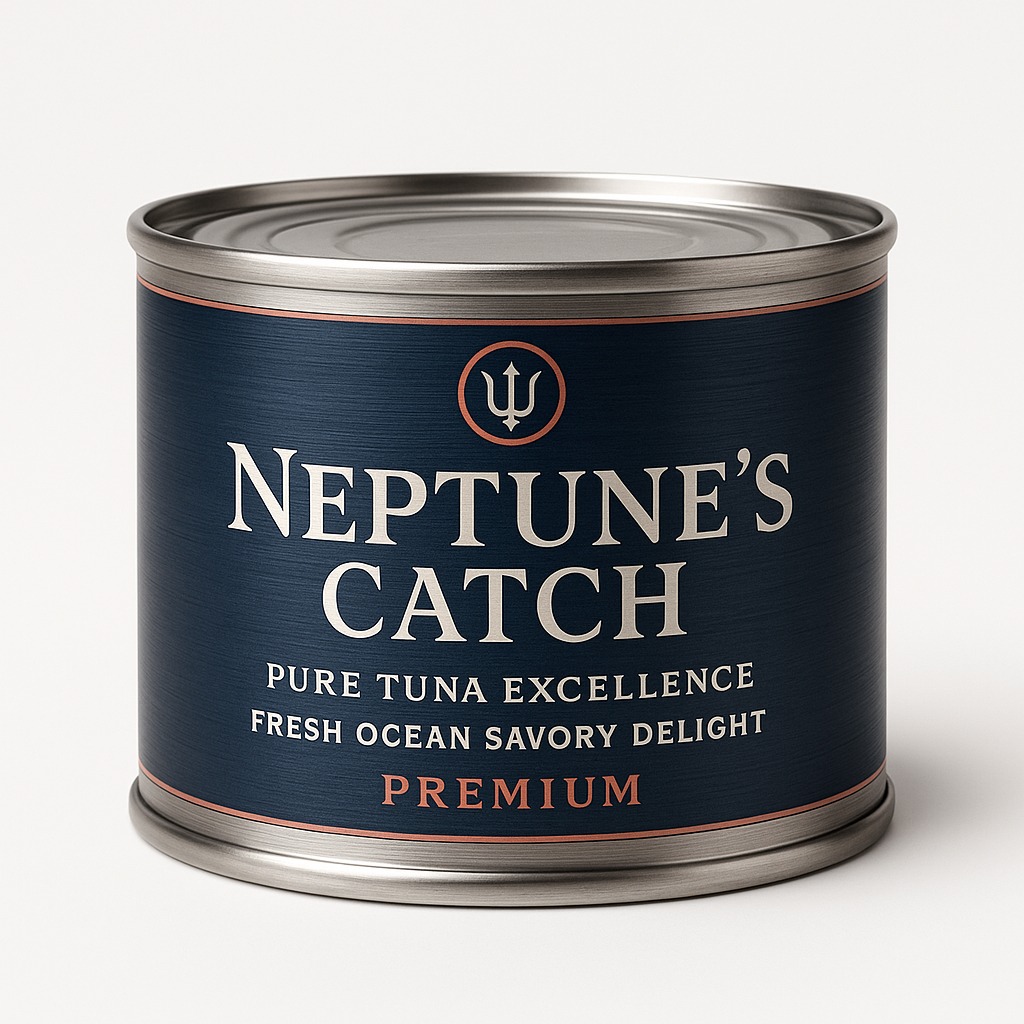} &
    \includegraphics[width=0.235\linewidth]{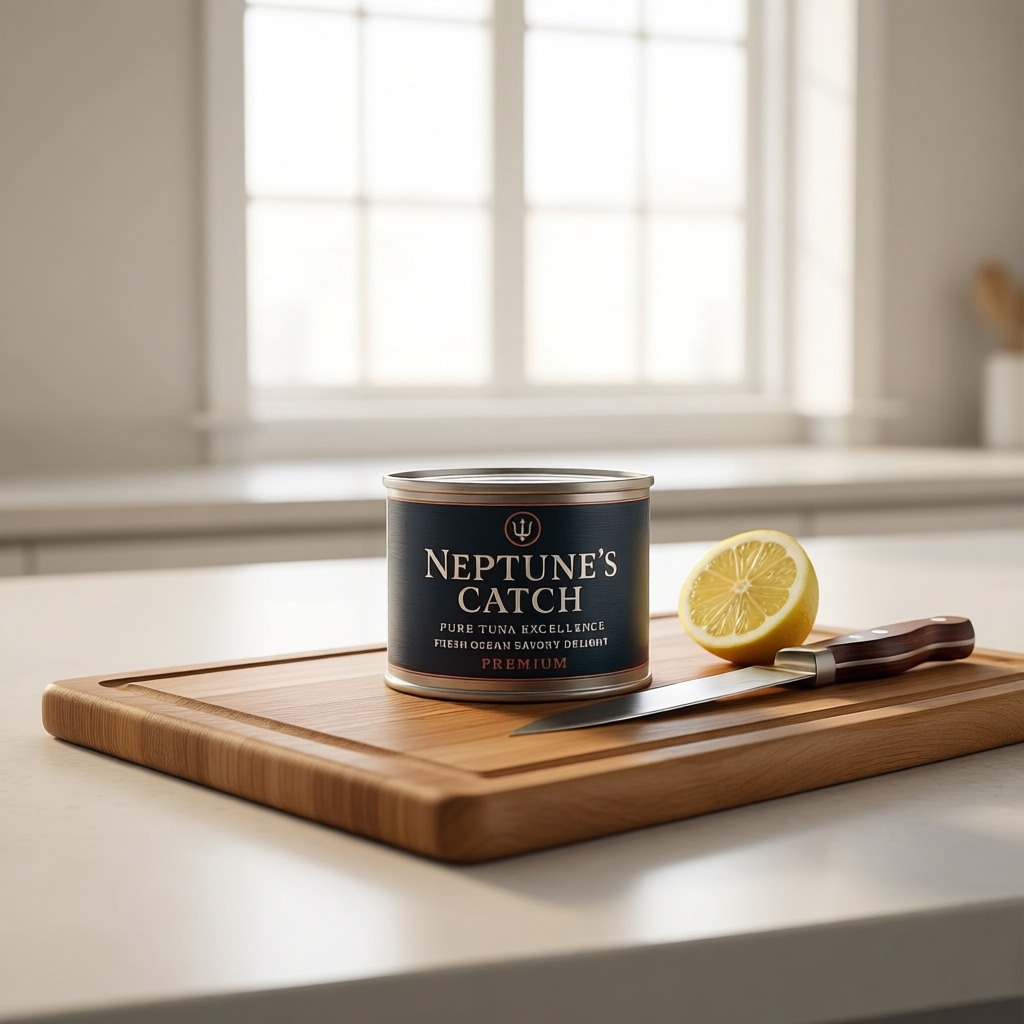} &
    \includegraphics[width=0.235\linewidth]{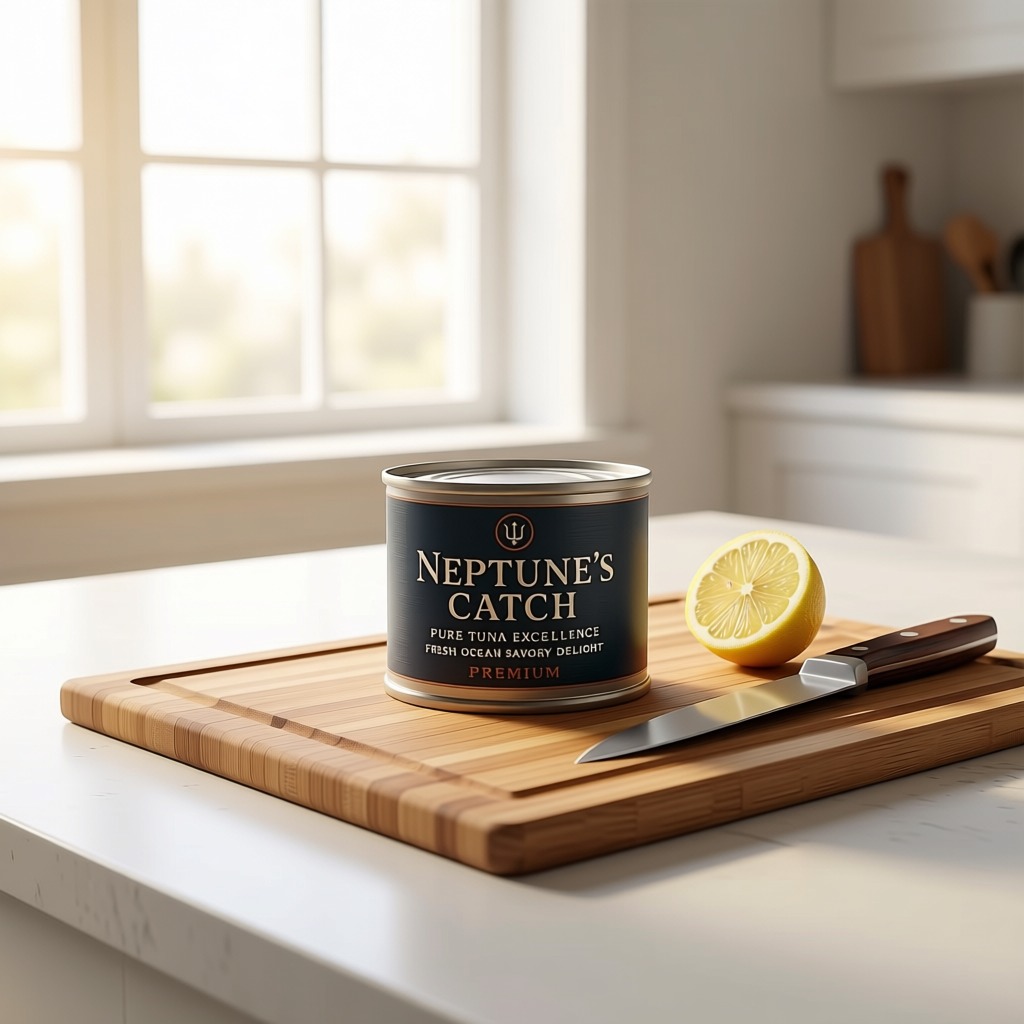} &
    \includegraphics[width=0.235\linewidth]{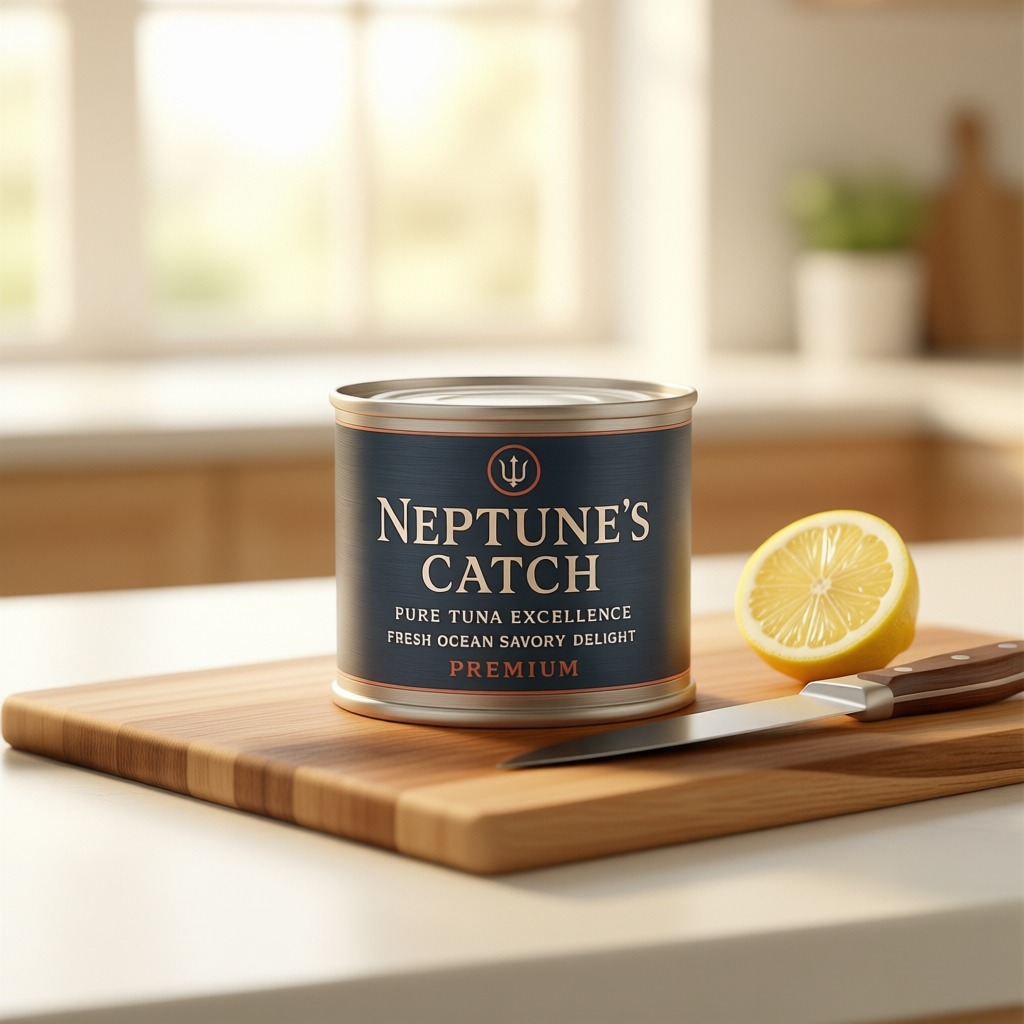} \\
    \multicolumn{4}{p{\dimexpr\linewidth-2\tabcolsep\relax}}{%
      \raggedright\scriptsize\emph{
        Position the can on a clean, minimalist kitchen countertop; 
        include a high-quality wooden cutting board with a knife and a lemon slice; 
        bathe the scene in soft, ambient daylight from a large kitchen window; 
        ensure the product is hero-lit, with focus on the label and metallic finishes.
      }
    }\\[4pt]

    \includegraphics[width=0.235\linewidth]{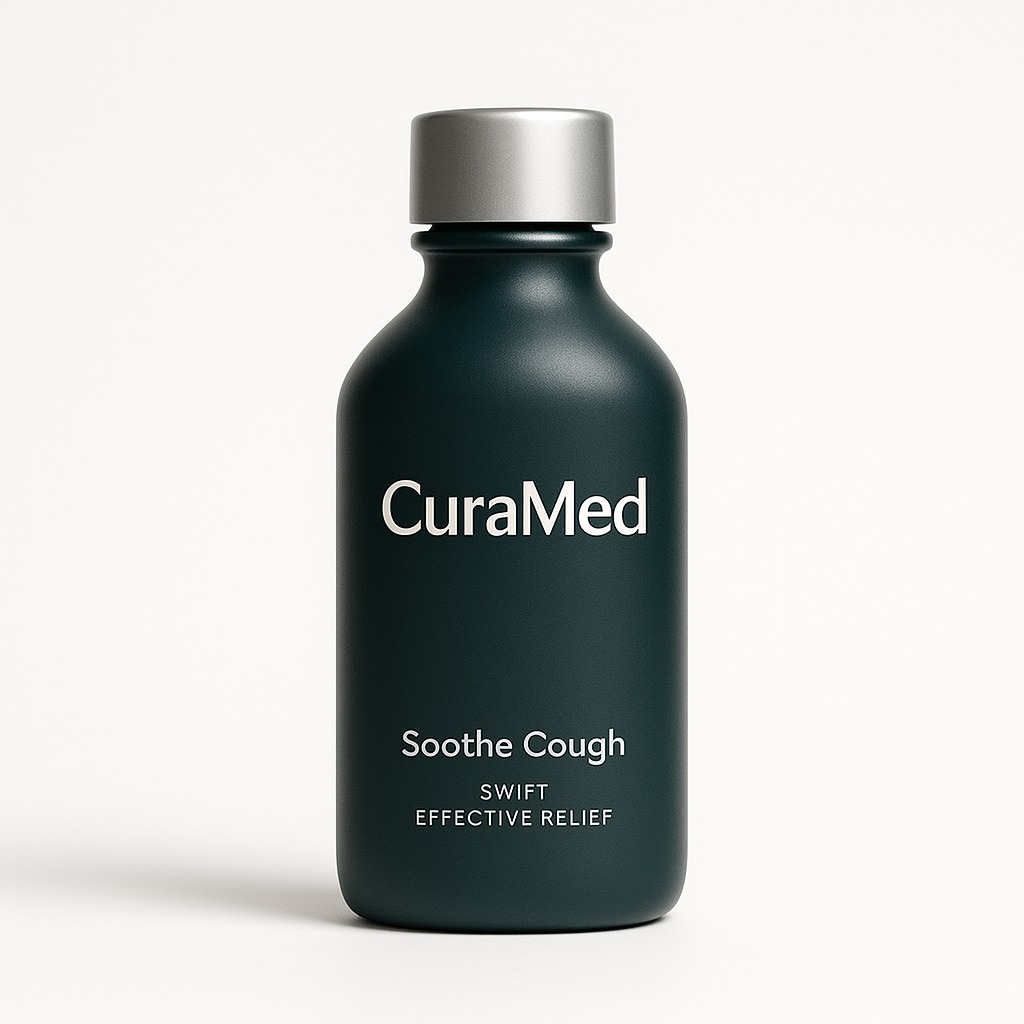} &
    \includegraphics[width=0.235\linewidth]{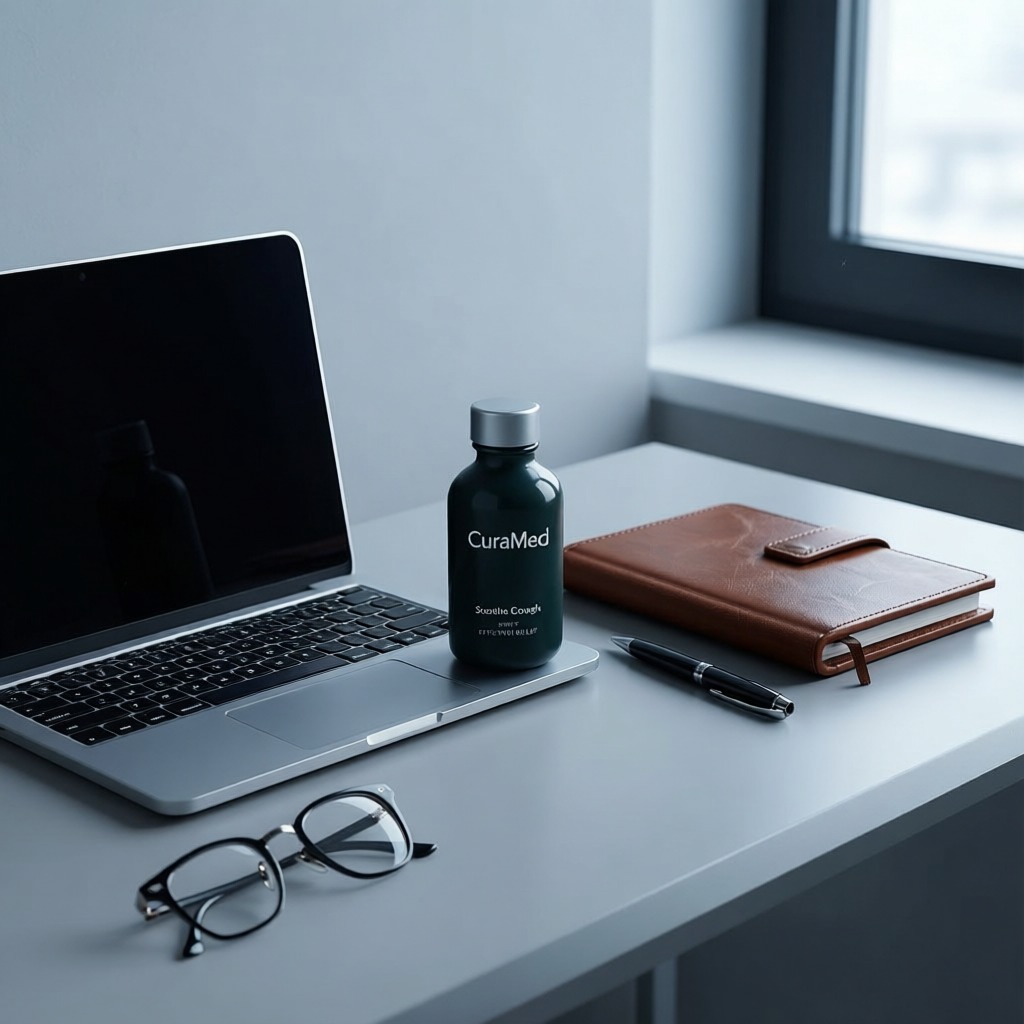} &
    \includegraphics[width=0.235\linewidth]{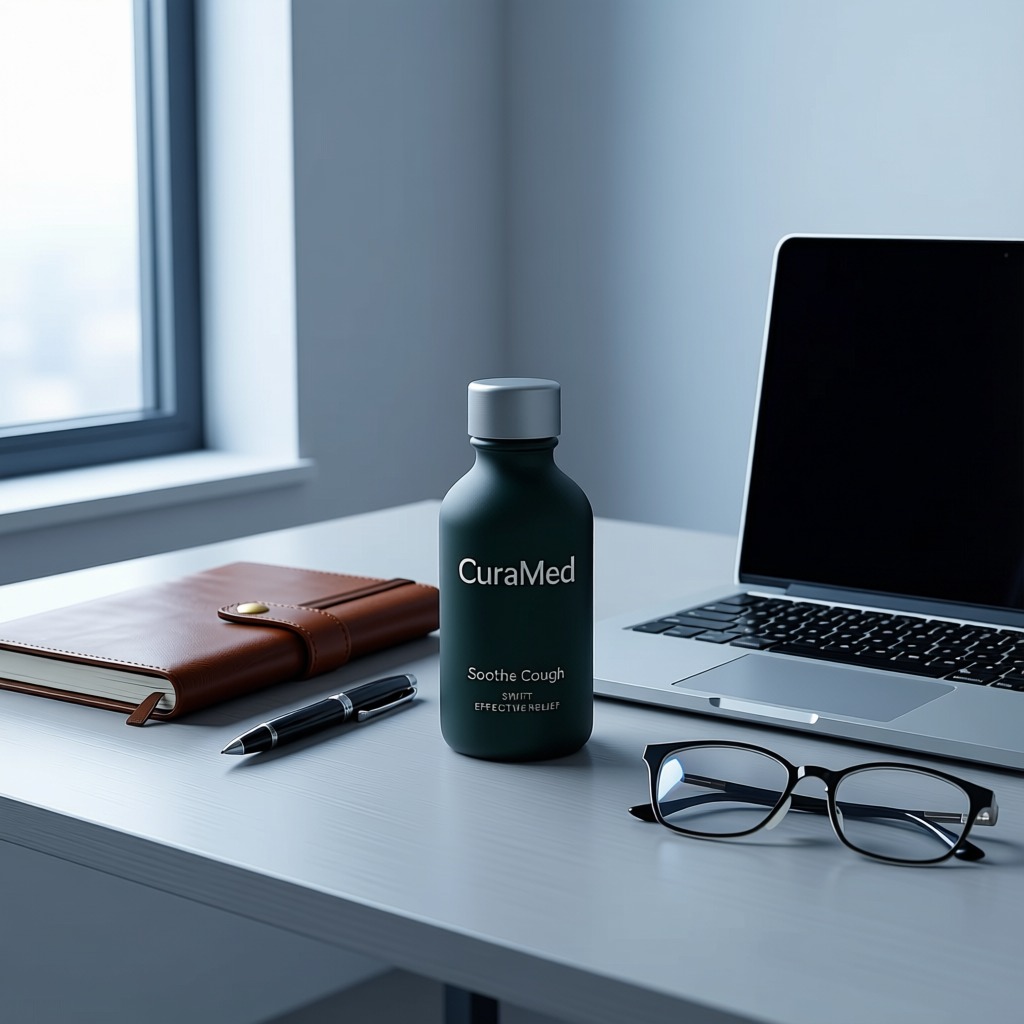} &
    \includegraphics[width=0.235\linewidth]{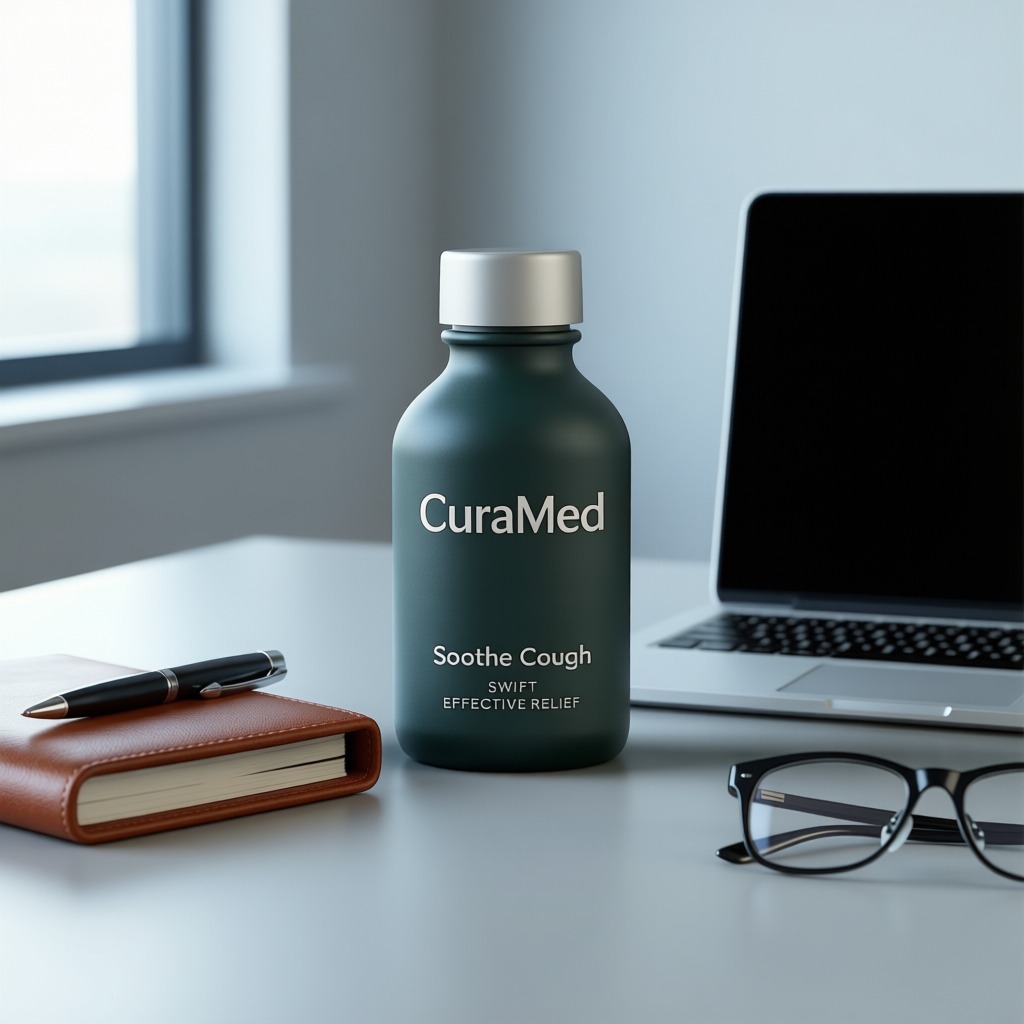} \\
    \multicolumn{4}{p{\dimexpr\linewidth-2\tabcolsep\relax}}{%
      \raggedright\scriptsize\emph{
        Place the bottle on a sleek, modern office desk next to a laptop 
        and a stylish leather-bound notebook; include a pen and a pair of reading glasses 
        to suggest a productive work environment; cool, indirect daylight from a nearby 
        window enhances the minimalist appeal.
      }
    }\\[4pt]

    \includegraphics[width=0.235\linewidth]{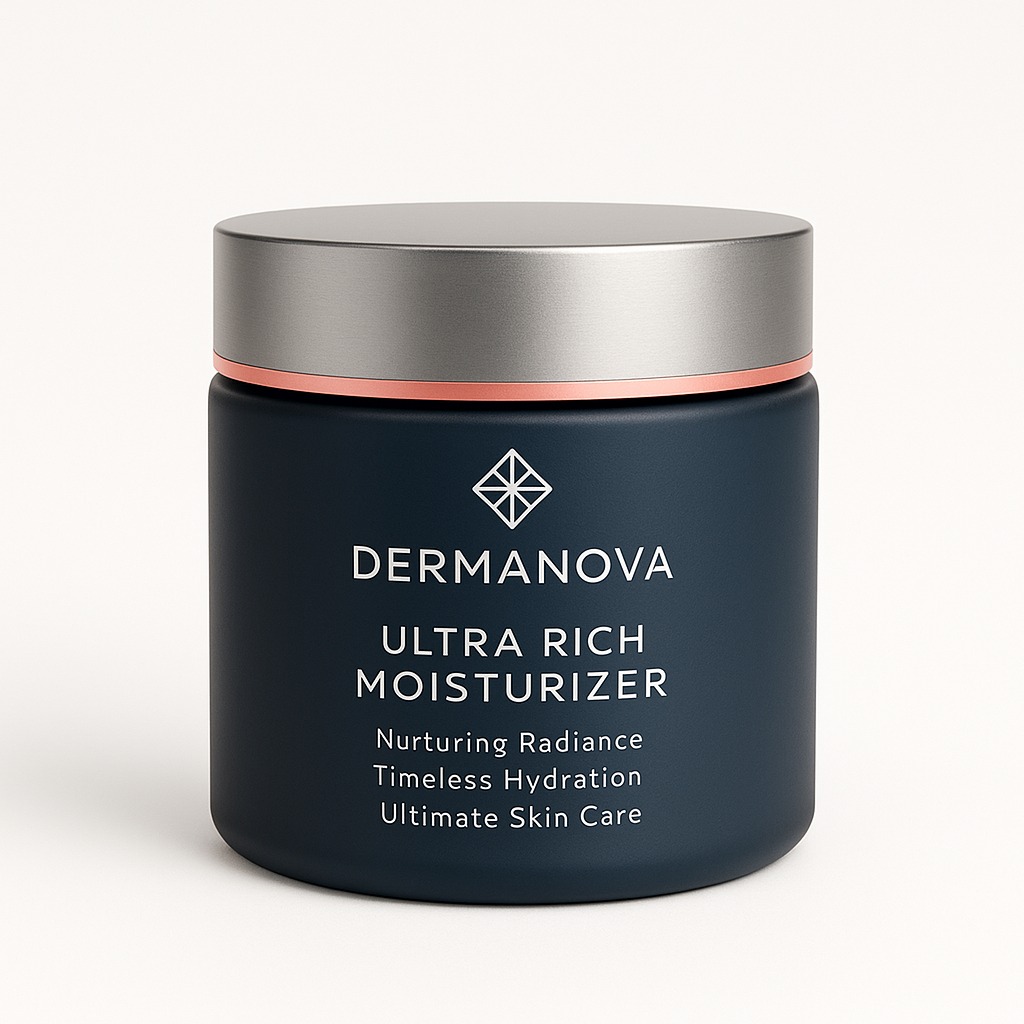} &
    \includegraphics[width=0.235\linewidth]{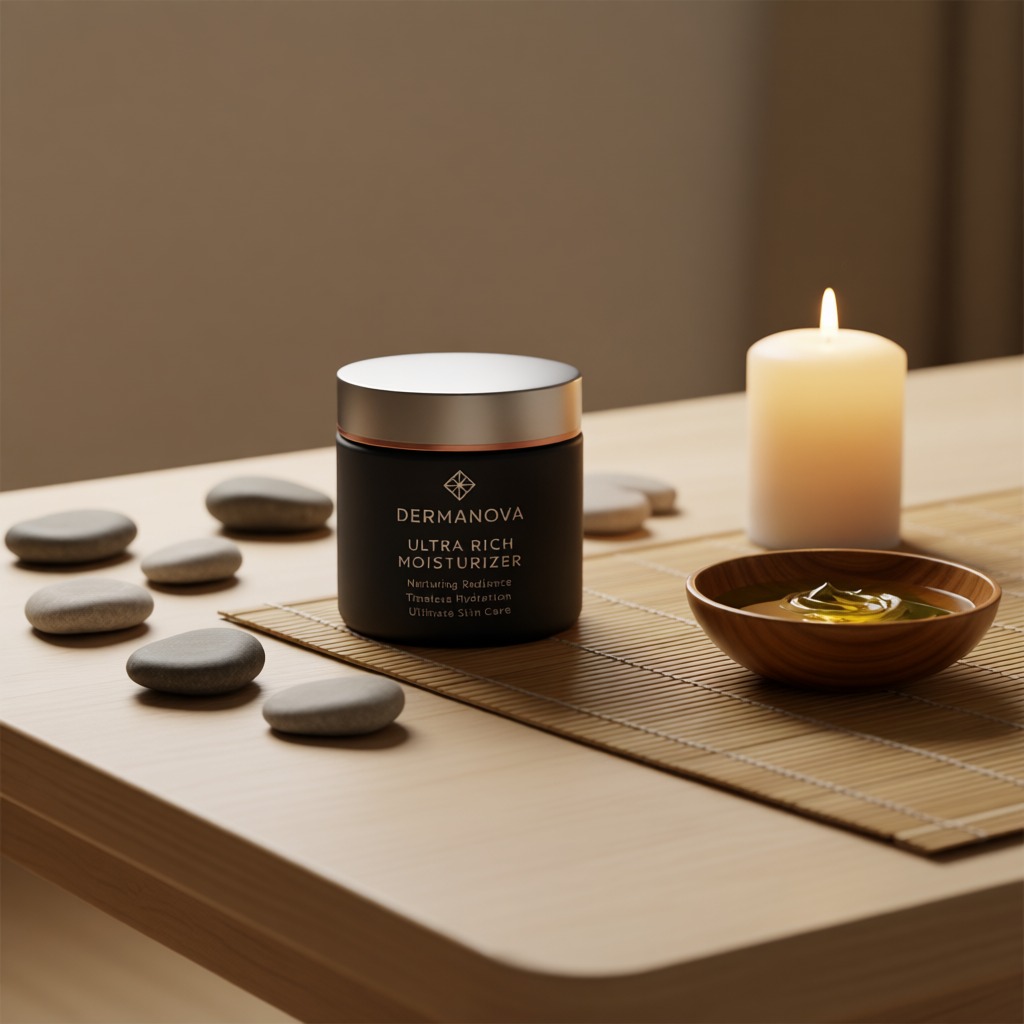} &
    \includegraphics[width=0.235\linewidth]{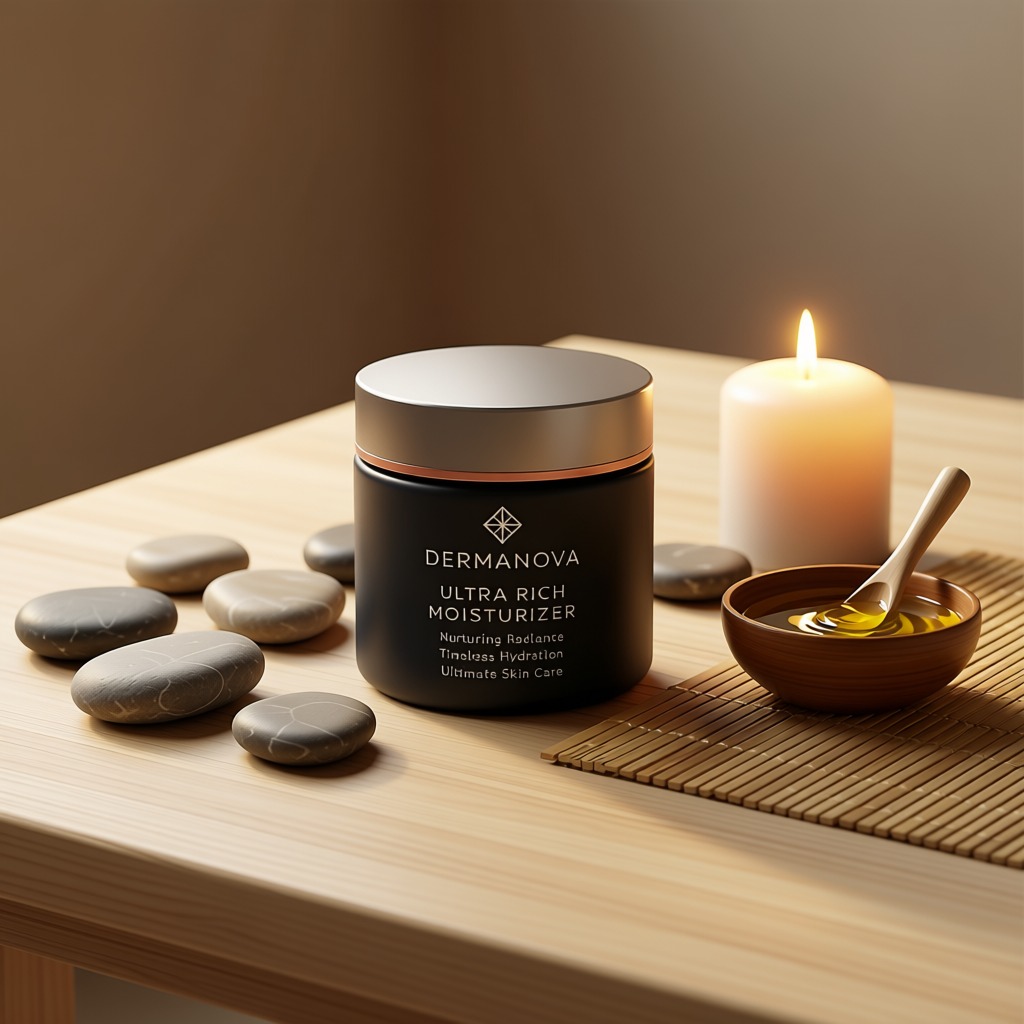} &
    \includegraphics[width=0.235\linewidth]{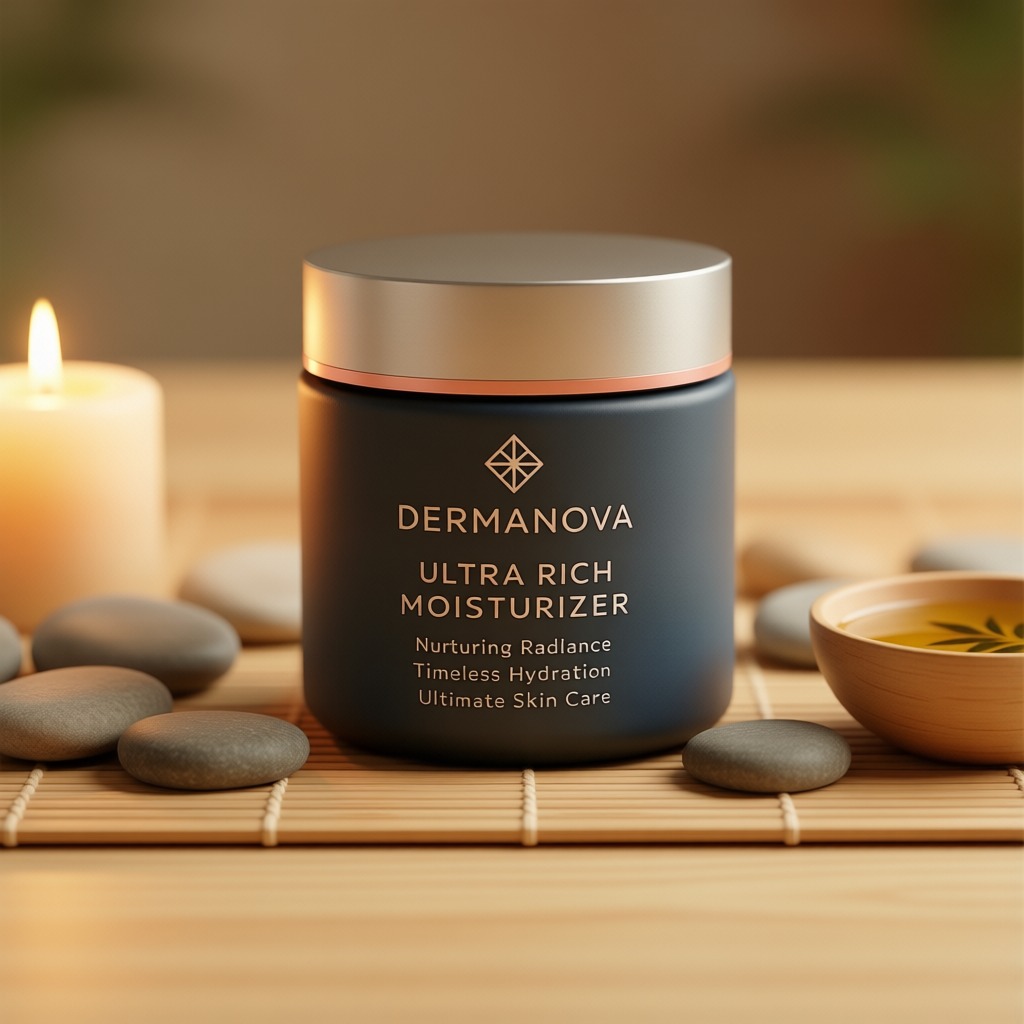} \\
    \multicolumn{4}{p{\dimexpr\linewidth-2\tabcolsep\relax}}{%
      \raggedright\scriptsize\emph{
        Set the jar on a light wooden spa table surrounded by smooth river stones 
        and a softly lit candle; diffused, warm spa lighting; incorporate a bamboo mat 
        and a small bowl of essential oils; soft shadows and a calming atmosphere; 
        ensure the jar remains the focal point.
      }
    }\\

    \bottomrule
  \end{tabular}

  \caption{
    Qualitative comparison on Qwen-Image-Edit-2511 across four inputs for the base model, 
    SFT trained checkpoint, and the final SFT + GRPO checkpoint trained with 
    Cyclic Consistency reward.
  }
  \label{fig:qwen_grid}
\end{figure*}

\begin{figure*}[t]
  \centering
  \setlength{\tabcolsep}{3pt} 
  \renewcommand{\arraystretch}{1.0}

  \begin{tabular}{@{}cccc@{}}
    \toprule
    \textbf{Input} & \textbf{Baseline} & \textbf{SFT} & \textbf{SFT+GRPO} \\
    \midrule

    \includegraphics[width=0.235\linewidth]{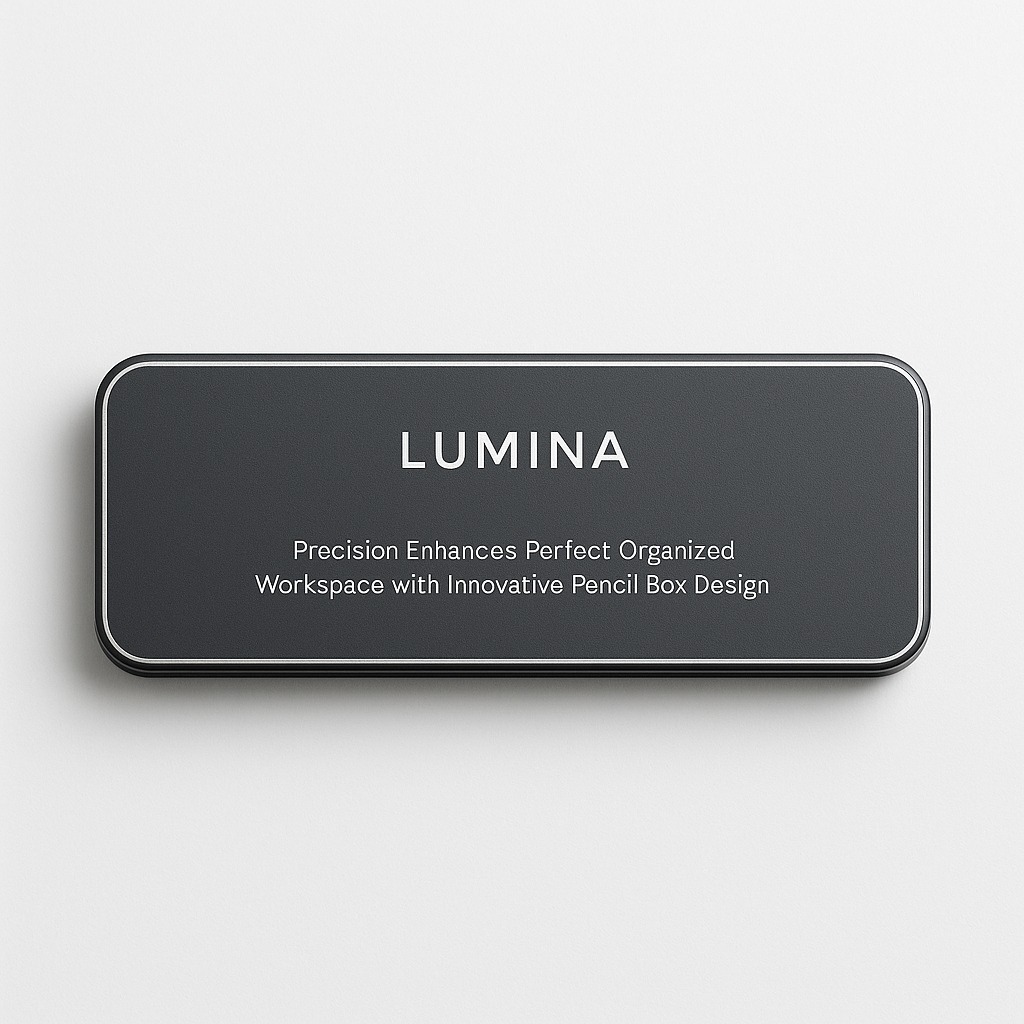} &
    \includegraphics[width=0.235\linewidth]{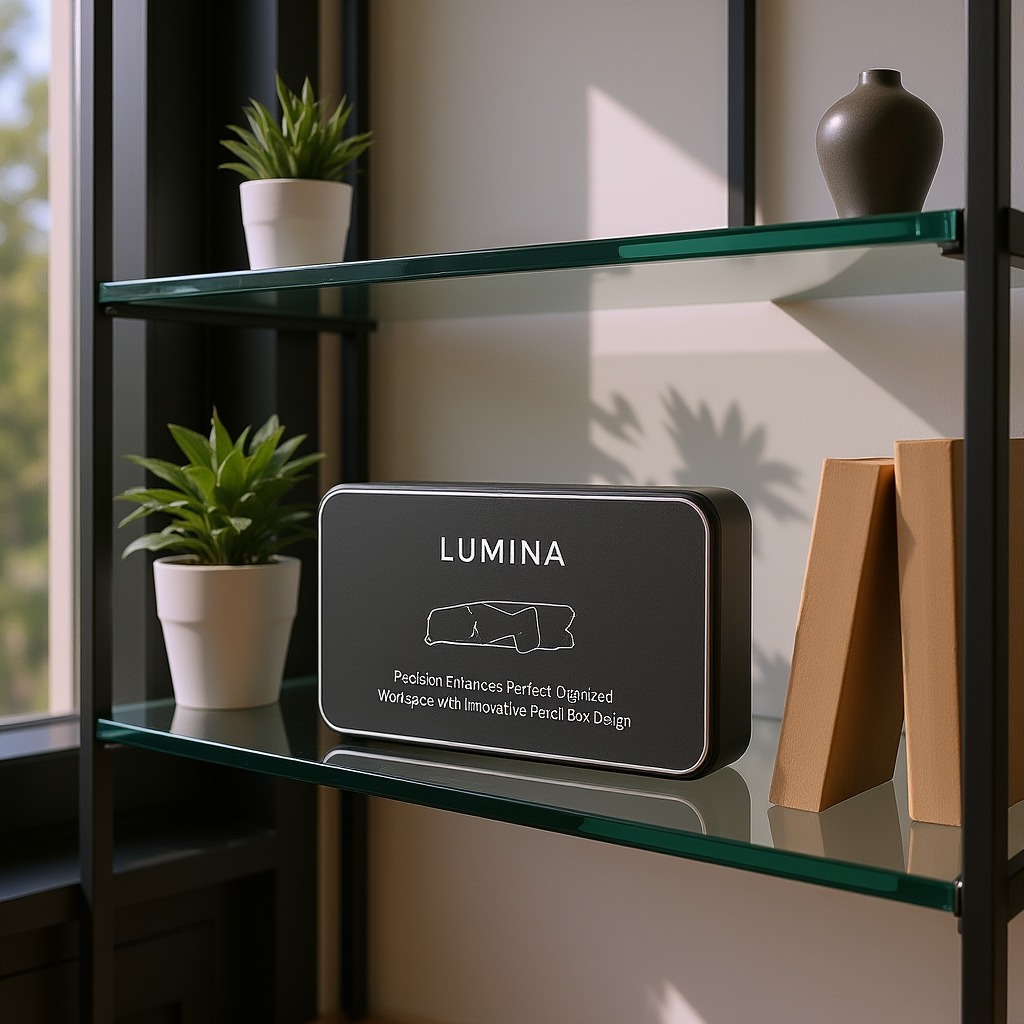} &
    \includegraphics[width=0.235\linewidth]{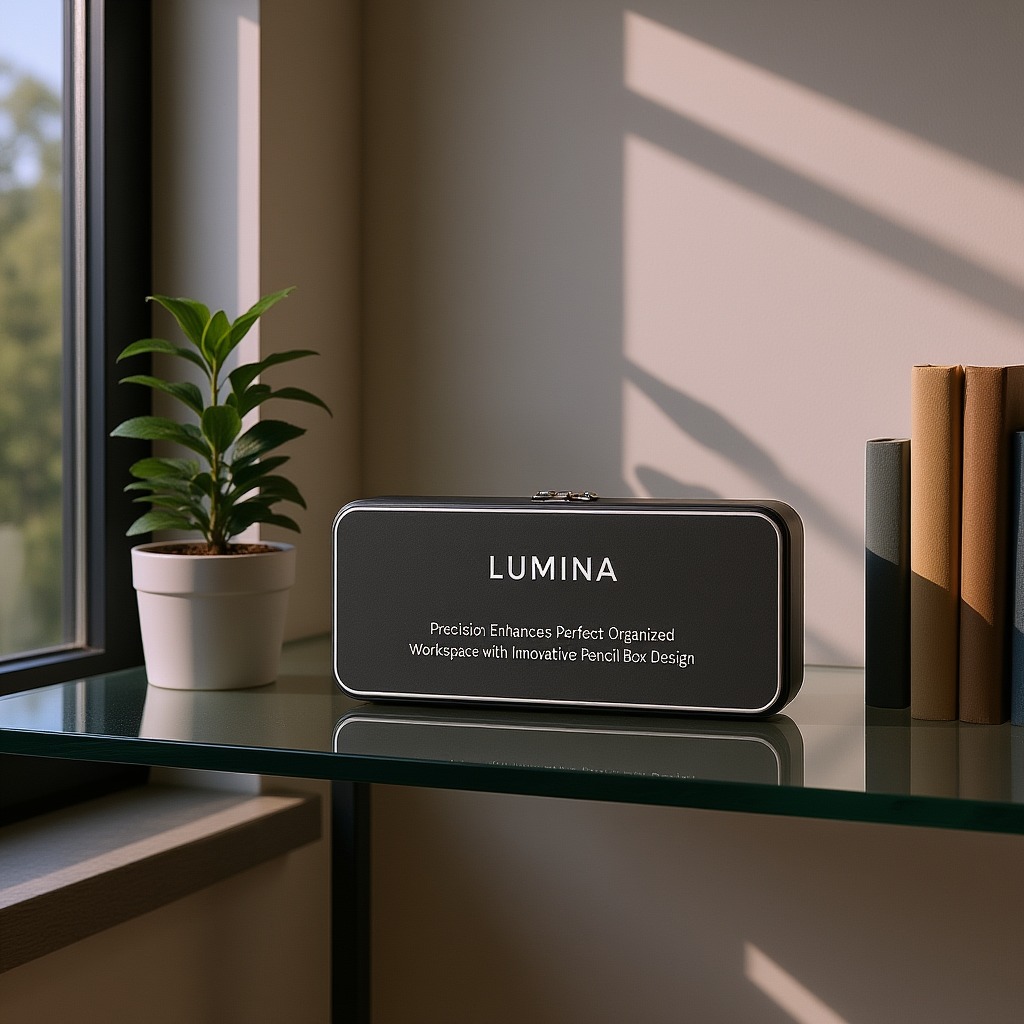} &
    \includegraphics[width=0.235\linewidth]{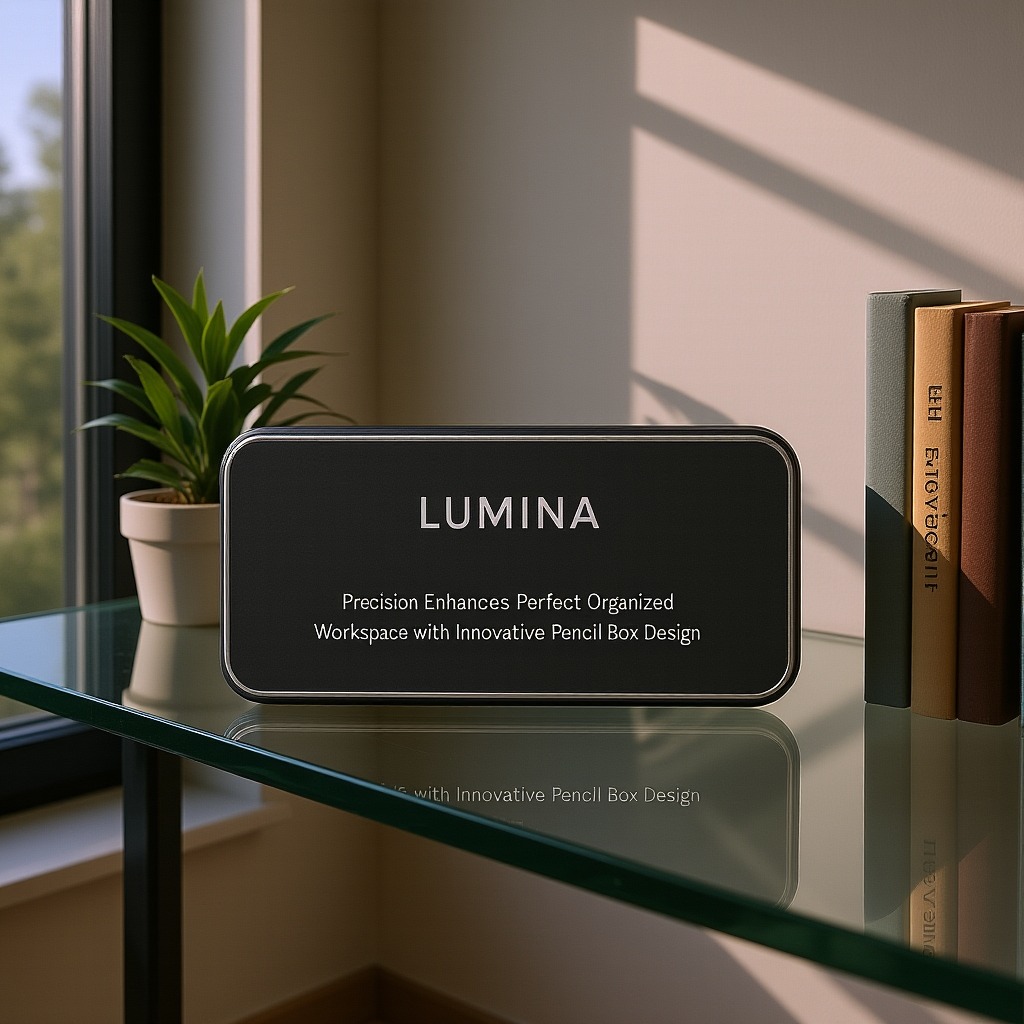} \\
    \multicolumn{4}{p{\dimexpr\linewidth-2\tabcolsep\relax}}{%
      \raggedright\scriptsize\emph{Place the product on a contemporary glass shelf within a chic home office environment; ambient natural light filters through a nearby window, casting gentle shadows; add a small potted plant and artistic bookends as decor accents; ensure the pencil box is centered and crisply lit, highlighting its minimalist design.}
    }\\[4pt]

    \includegraphics[width=0.235\linewidth]{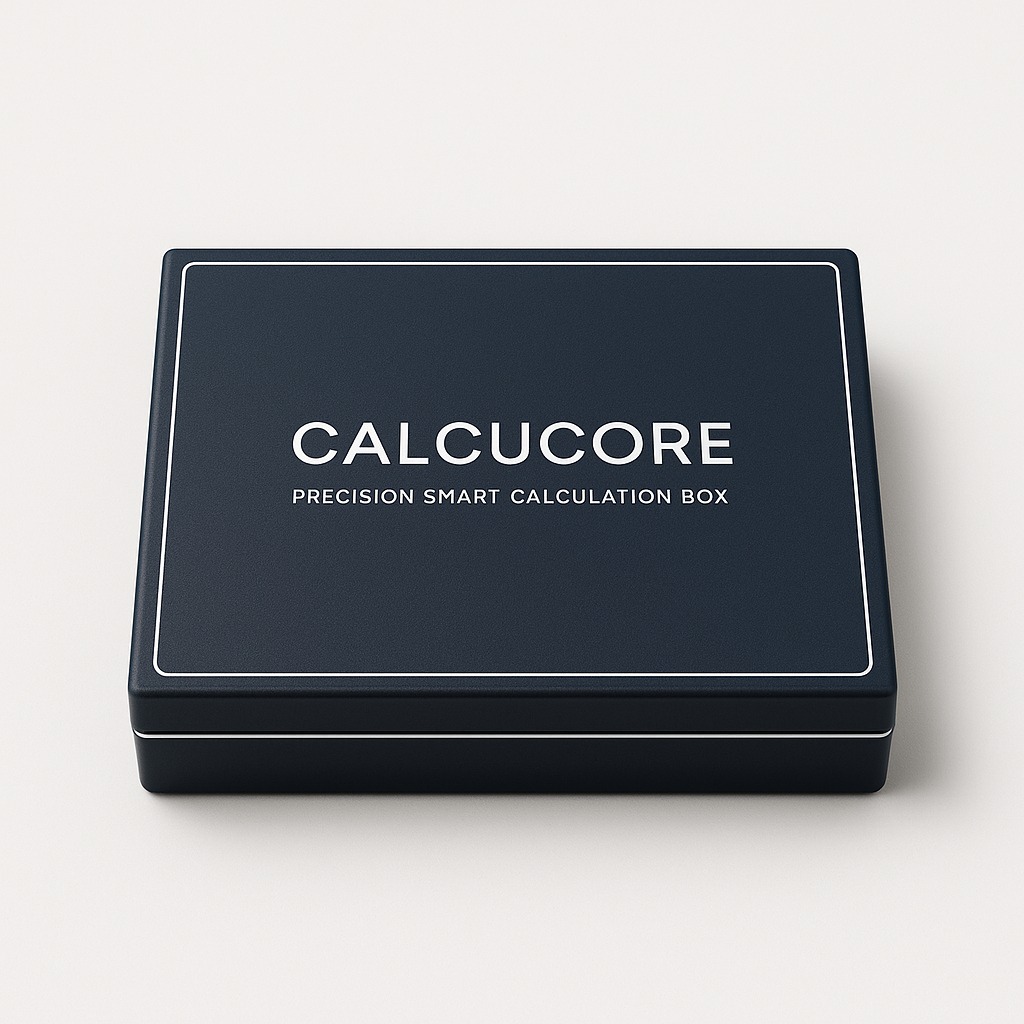} &
    \includegraphics[width=0.235\linewidth]{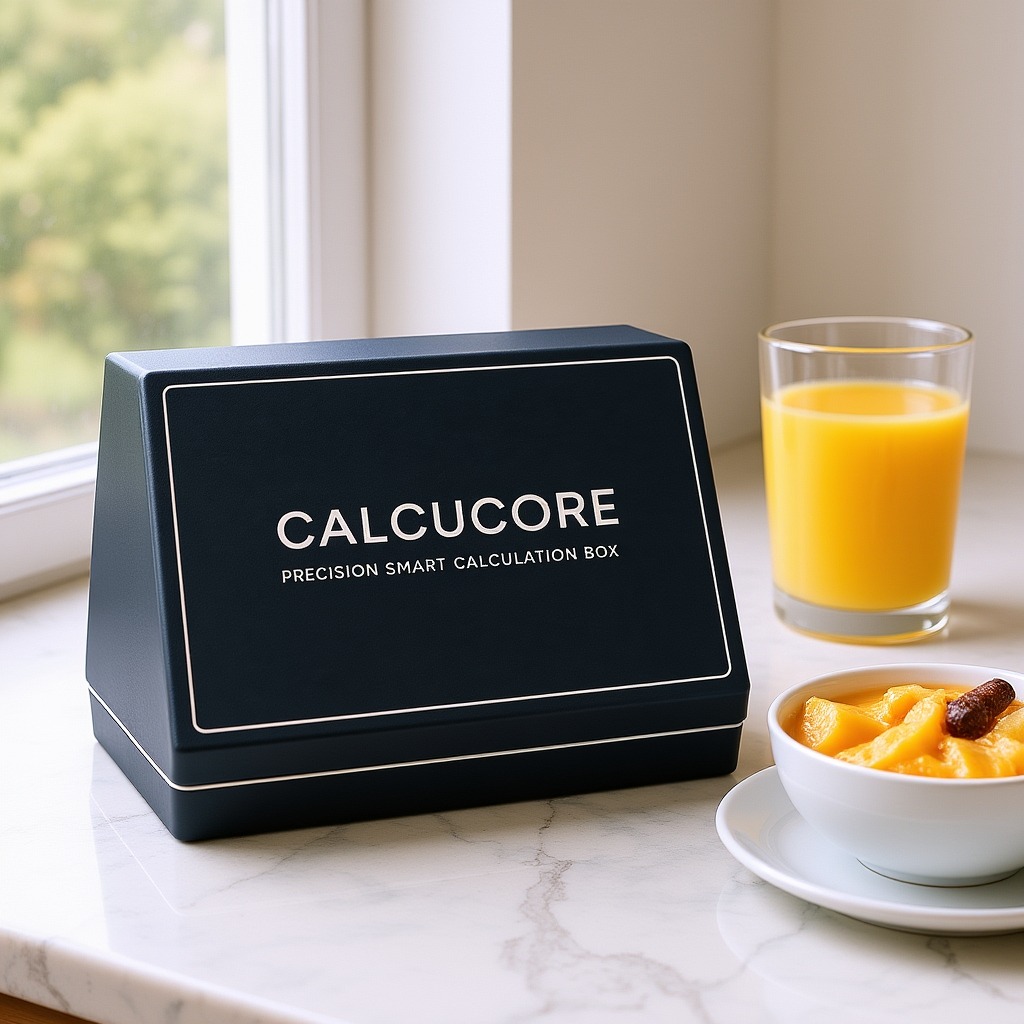} &
    \includegraphics[width=0.235\linewidth]{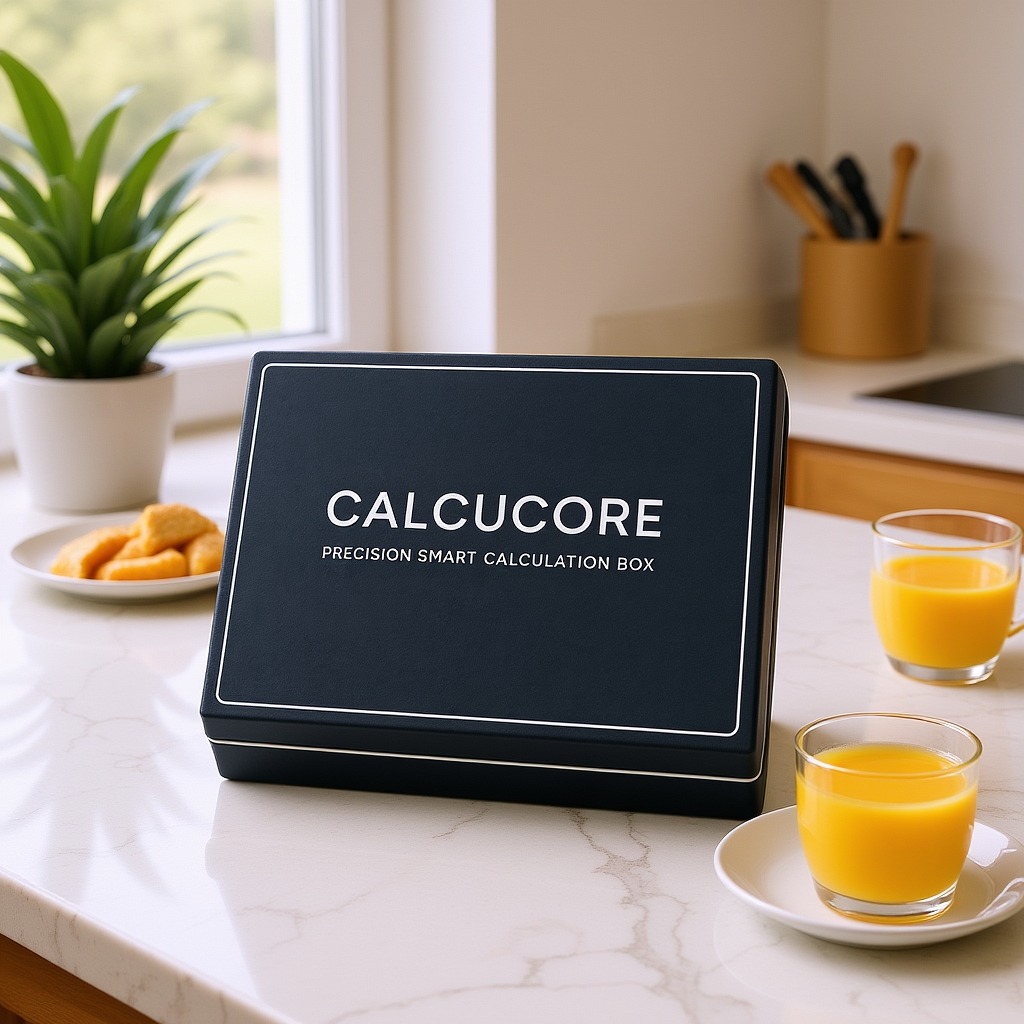} &
    \includegraphics[width=0.235\linewidth]{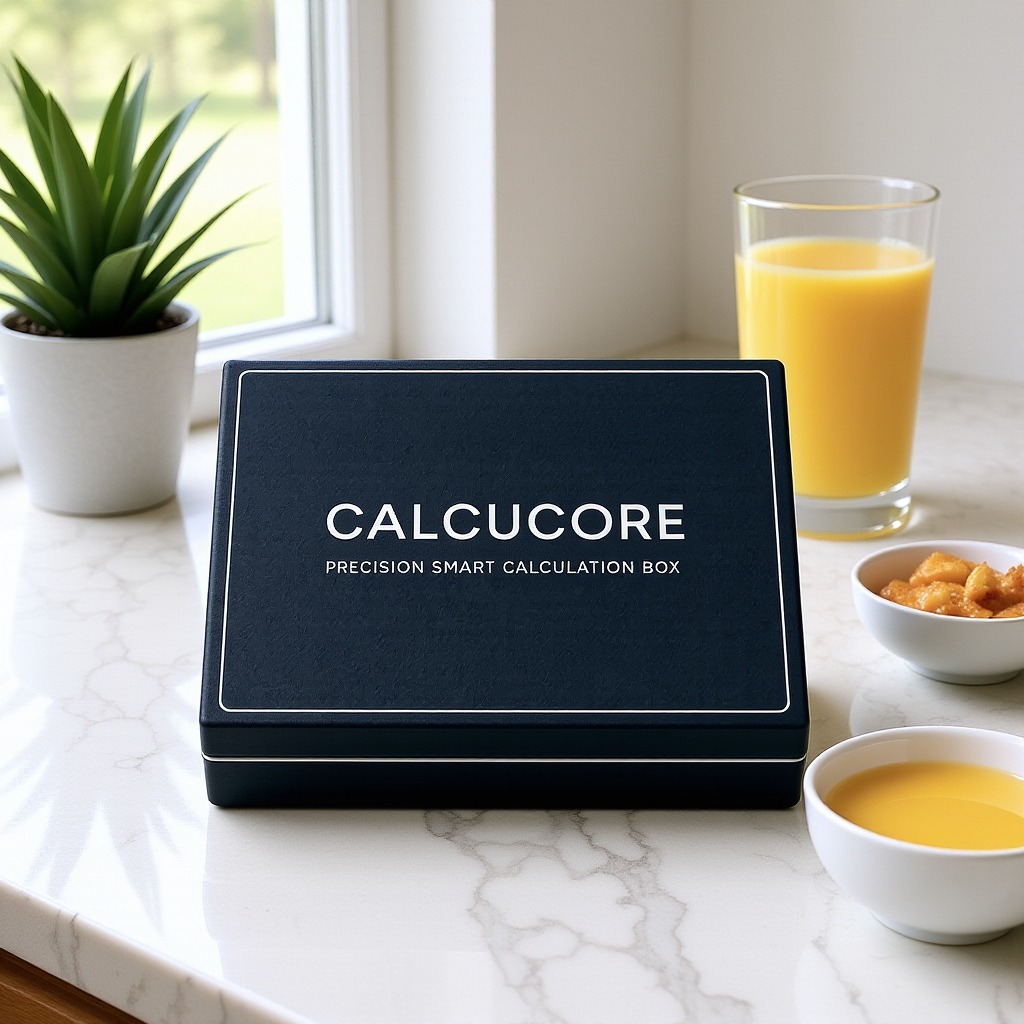} \\
    \multicolumn{4}{p{\dimexpr\linewidth-2\tabcolsep\relax}}{%
      \raggedright\scriptsize\emph{Position the box on a marble kitchen counter with a clean, luxurious breakfast setup featuring a glass of orange juice and a small fruit bowl; morning sunlight streaming through a window casting natural reflections; ensure the product is hero-lit with crisp branding visibility; avoid clutter and unnecessary props.}
    }\\[4pt]

    \includegraphics[width=0.235\linewidth]{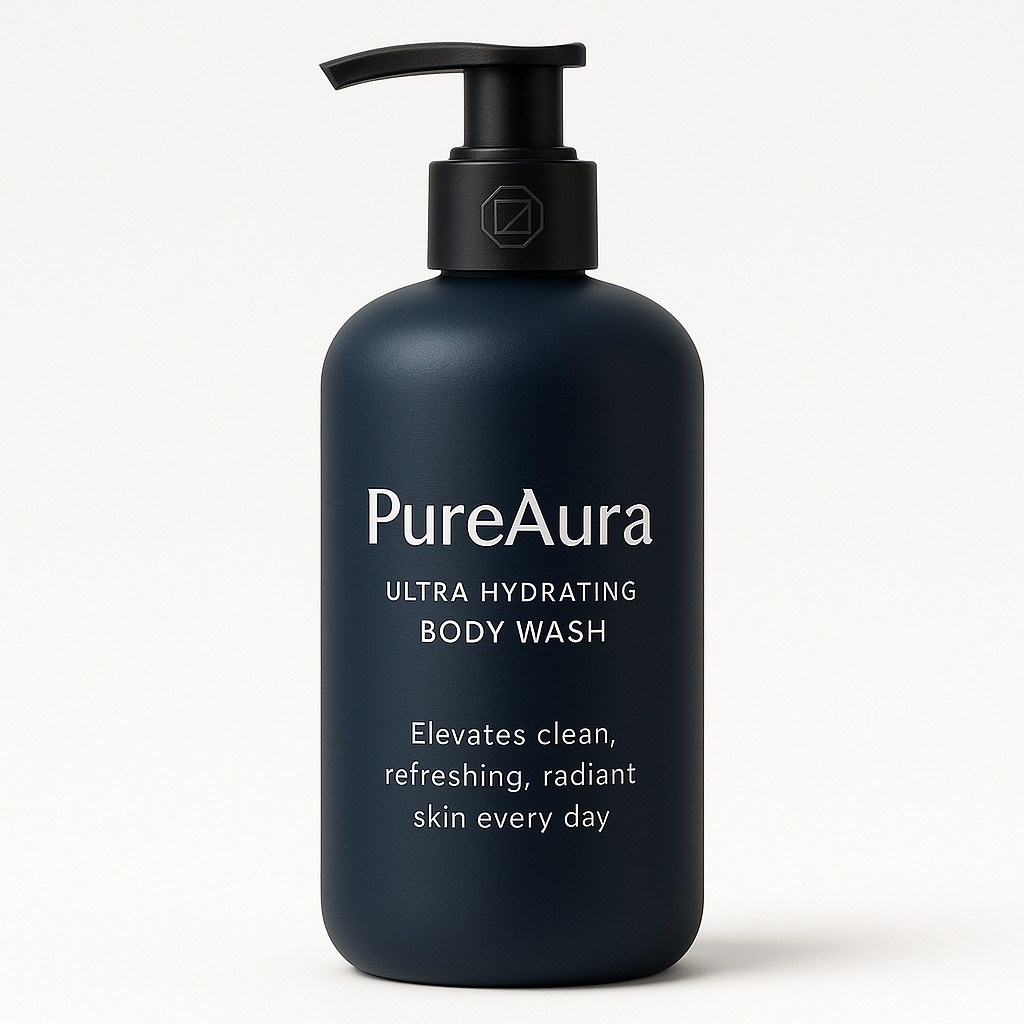} &
    \includegraphics[width=0.235\linewidth]{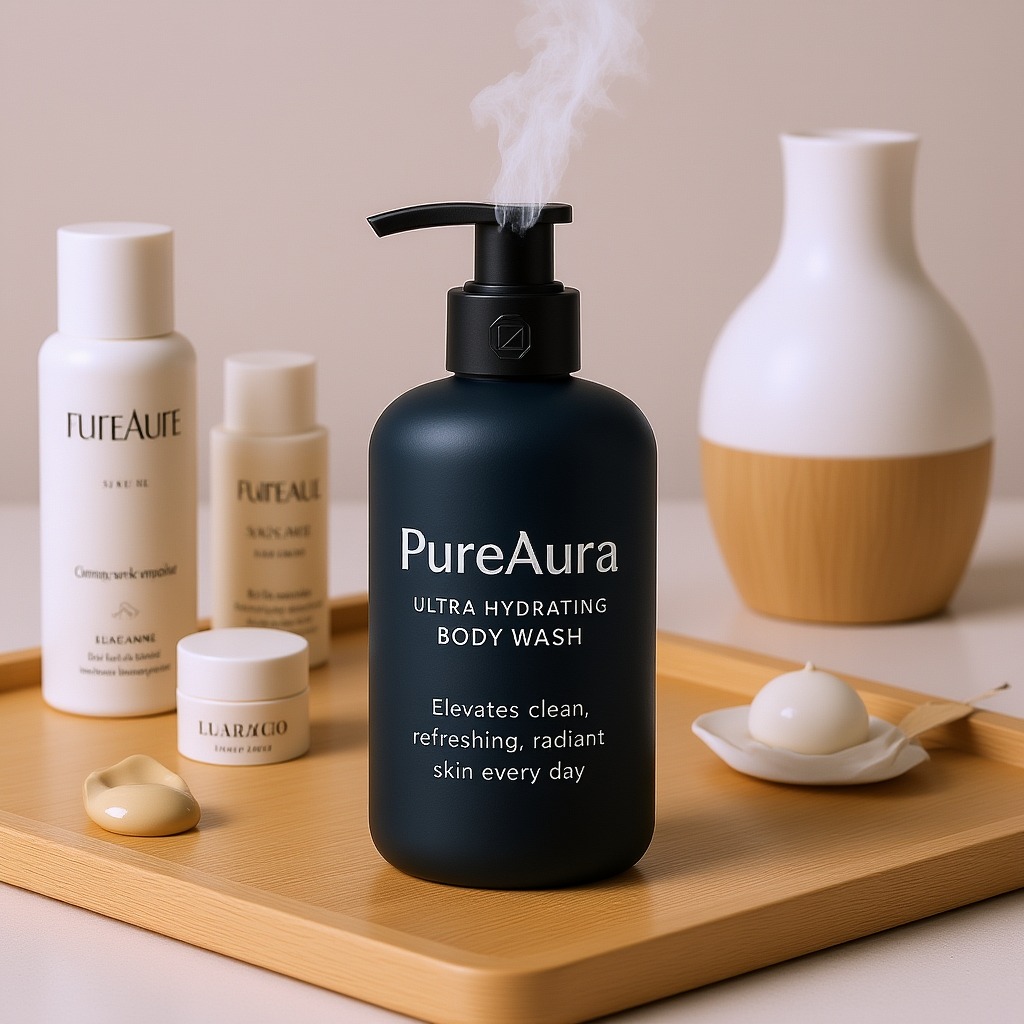} &
    \includegraphics[width=0.235\linewidth]{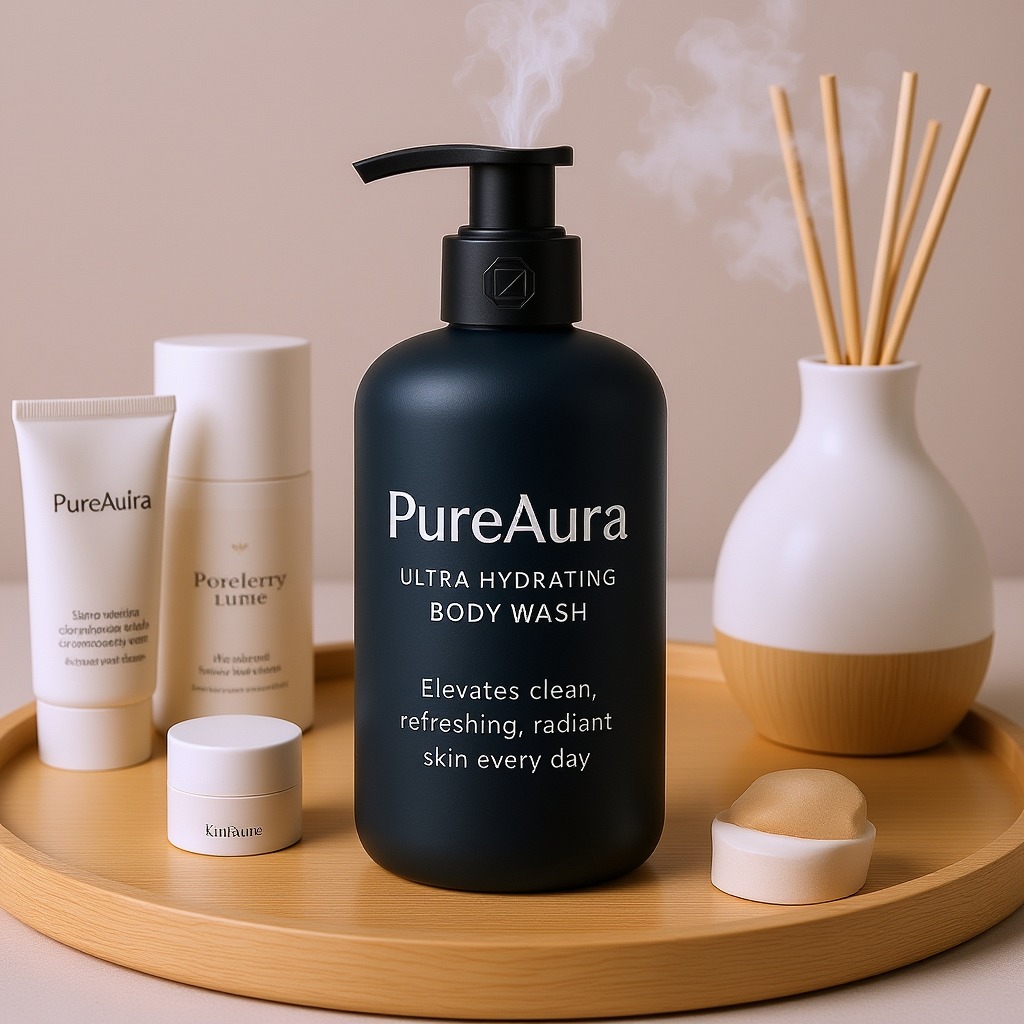} &
    \includegraphics[width=0.235\linewidth]{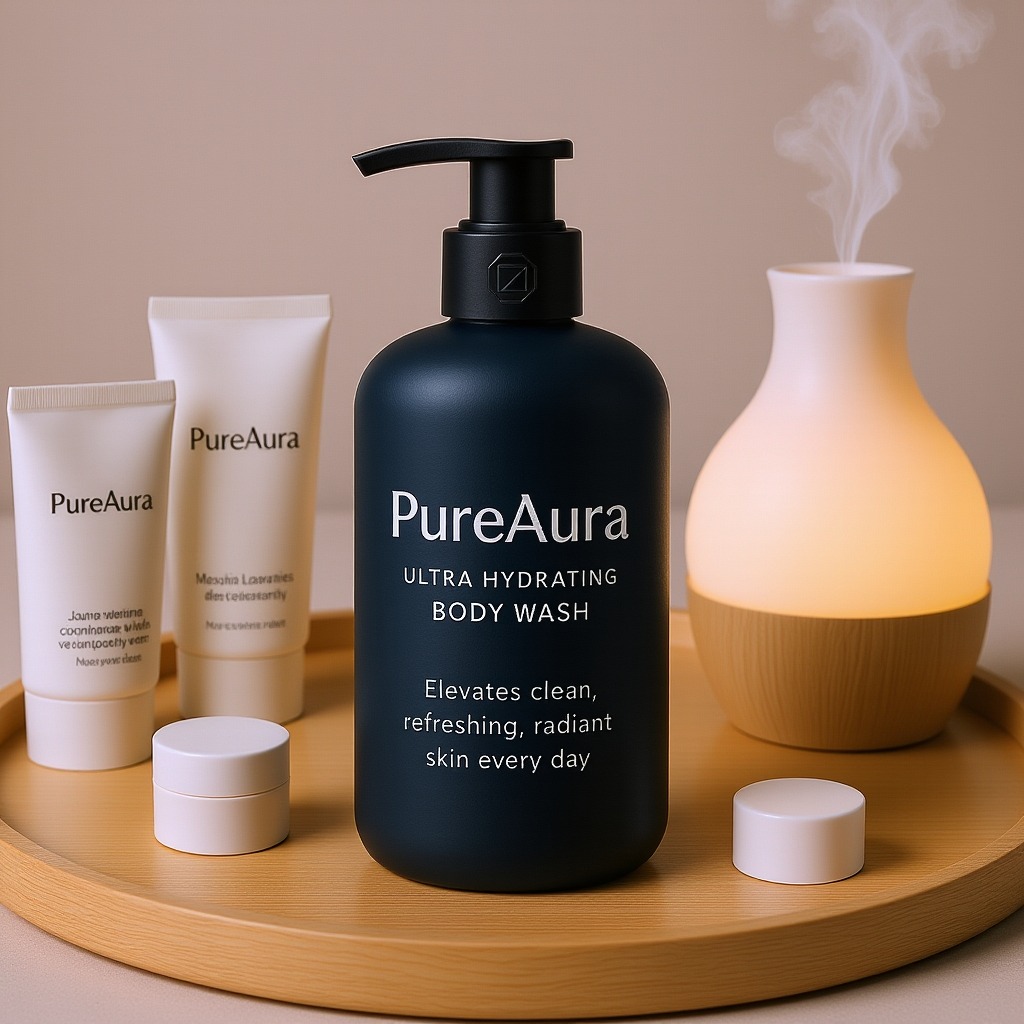} \\
    \multicolumn{4}{p{\dimexpr\linewidth-2\tabcolsep\relax}}{%
      \raggedright\scriptsize\emph{Place the bottle on a minimalist wooden tray amidst a selection of high-end skincare products; soft, directional lighting highlighting the bottle's silhouette; include a small, stylish diffuser emitting a gentle mist in the background for a calming and rejuvenating environment; maintain a sense of elegance and harmony.}
    }\\[4pt]

    \includegraphics[width=0.235\linewidth]{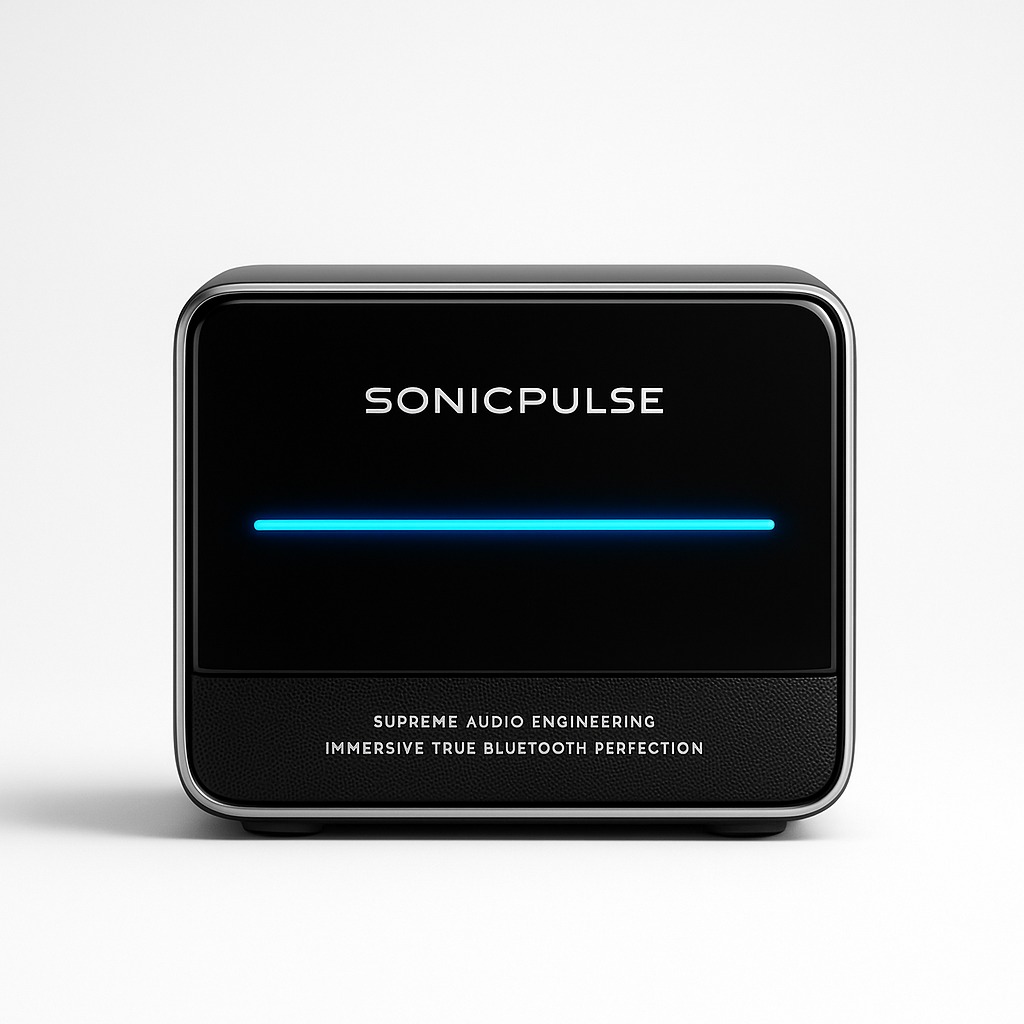} &
    \includegraphics[width=0.235\linewidth]{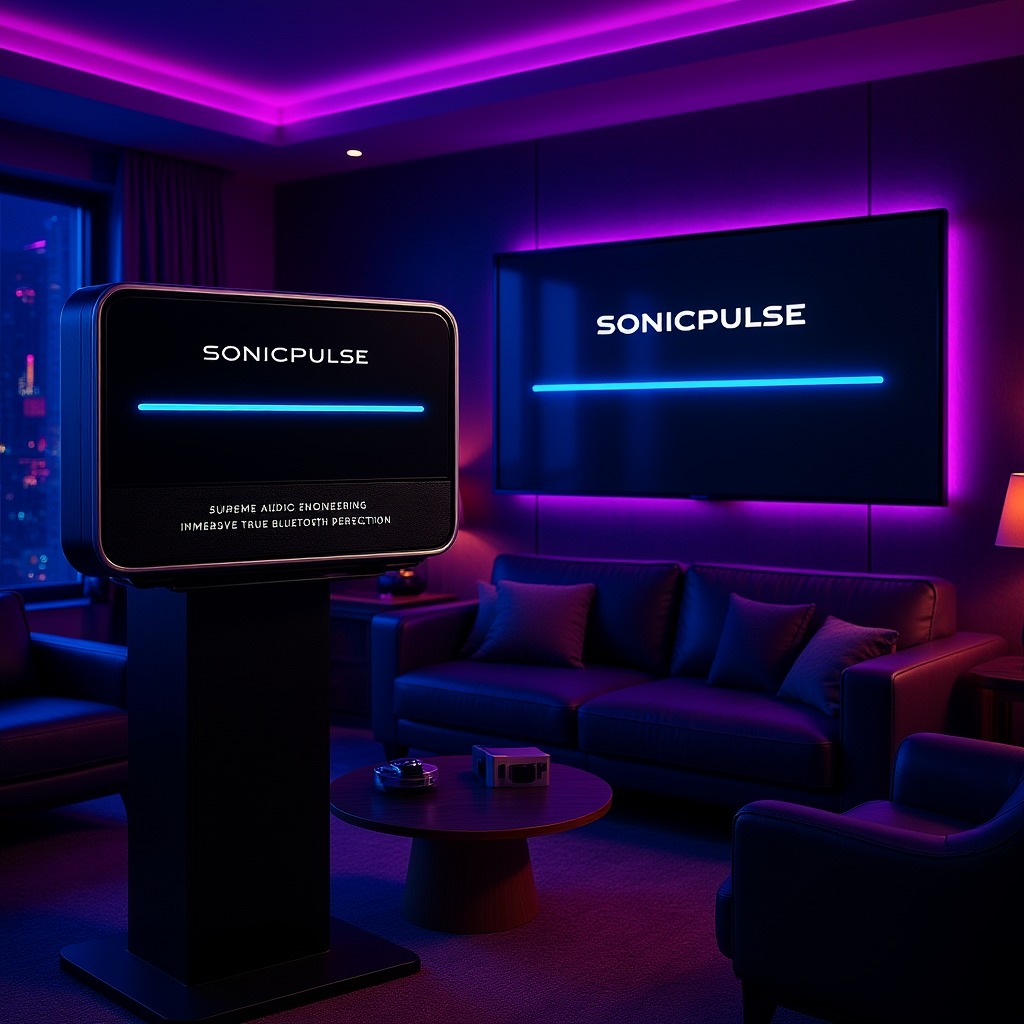} &
    \includegraphics[width=0.235\linewidth]{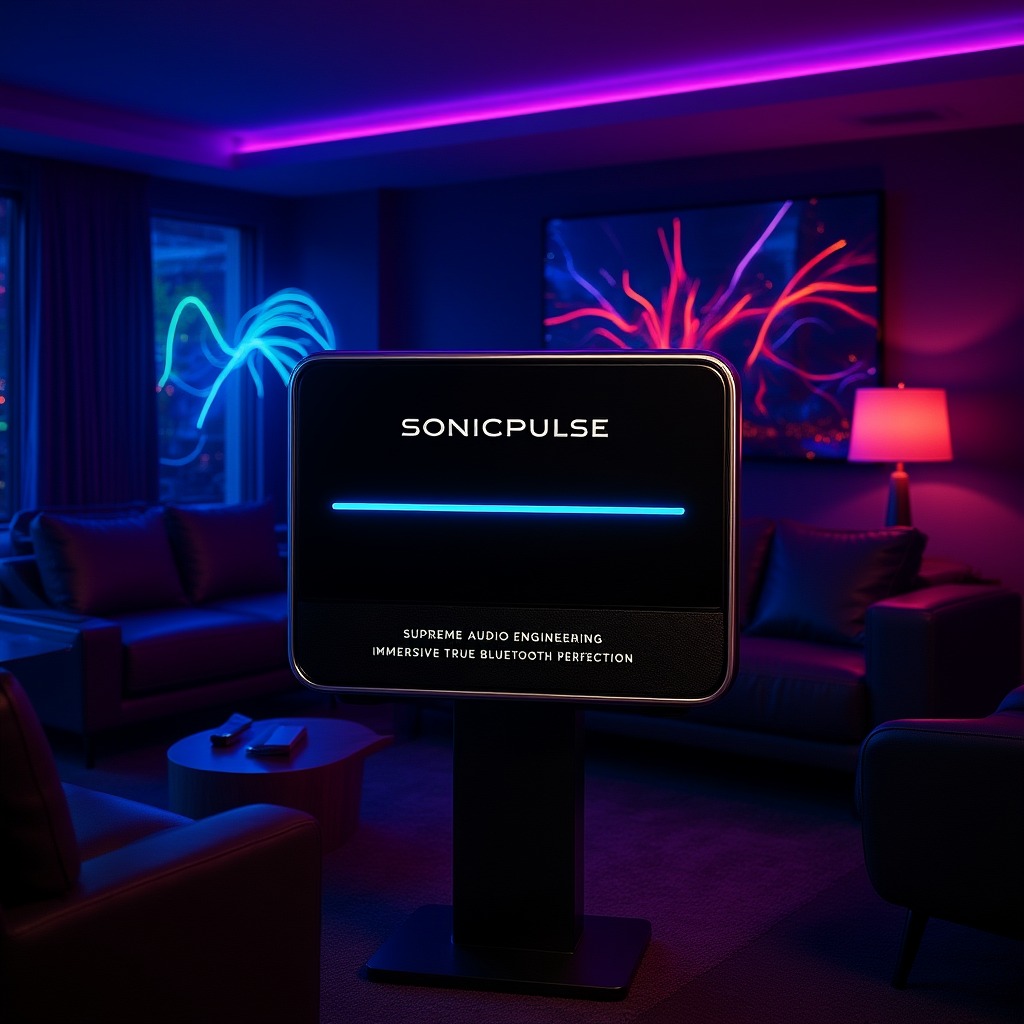} &
    \includegraphics[width=0.235\linewidth]{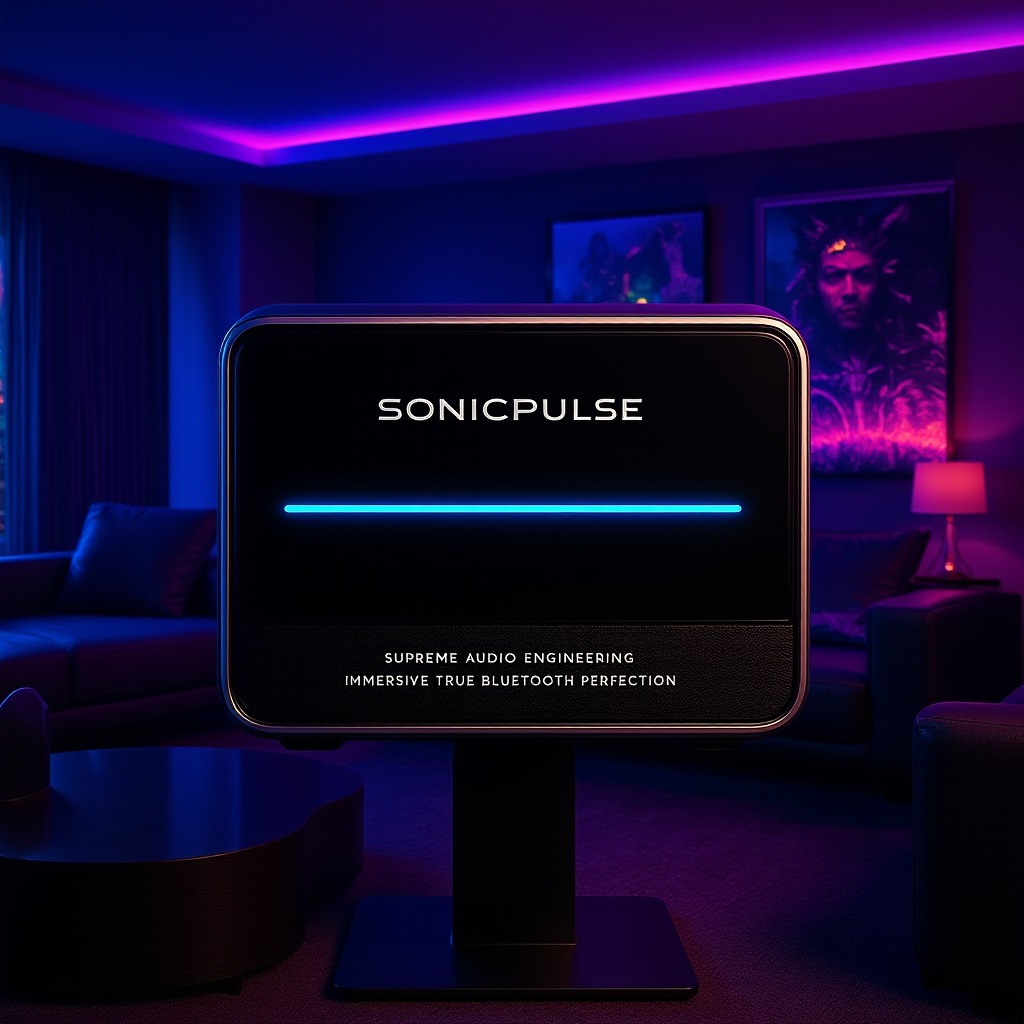} \\
    \multicolumn{4}{p{\dimexpr\linewidth-2\tabcolsep\relax}}{%
      \raggedright\scriptsize\emph{Showcase the speaker in an upscale entertainment lounge with a large flat-screen TV and sophisticated décor. Use dynamic, colorful LED lighting to create a vibrant, energetic mood, with the speaker as the focal point. Balance the scene with sleek furniture and tech gadgets to emphasize a high-tech lifestyle environment.}
    }\\

    \bottomrule
  \end{tabular}

  \caption{Qualitative comparison on Flux.1-Kontext-dev across four inputs for the base model, SFT trained checkpoint and the final SFT + GRPO checkpoint trained with Cyclic Consistency reward.}
  \label{fig:flux_grid}
\end{figure*}

We present qualitative results from our experiments in Figure \ref{fig:qwen_grid} for the Qwen-Image-Edit-2511 model and in Figure \ref{fig:flux_grid} for the Flux.1-Kontext-dev model. As shown in both figures, the baseline models exhibit several common failure modes, including incorrect or distorted text, inconsistent product geometry and color, and hallucinated product features. Fine-tuning with the ProductConsistency SFT dataset substantially mitigates these issues by improving product consistency and text rendering accuracy. Further improvements are observed with the GRPO-trained models, which produce outputs with more accurate text, consistent product features, and overall more natural visual aesthetics. These results indicate that our training framework encourages the model to better preserve product identity while staying faithful to the edit instructions.

\begin{figure*}[t]
\centering
\setlength{\tabcolsep}{2pt}

\begin{tabular}{cccc}
\includegraphics[width=0.24\linewidth]{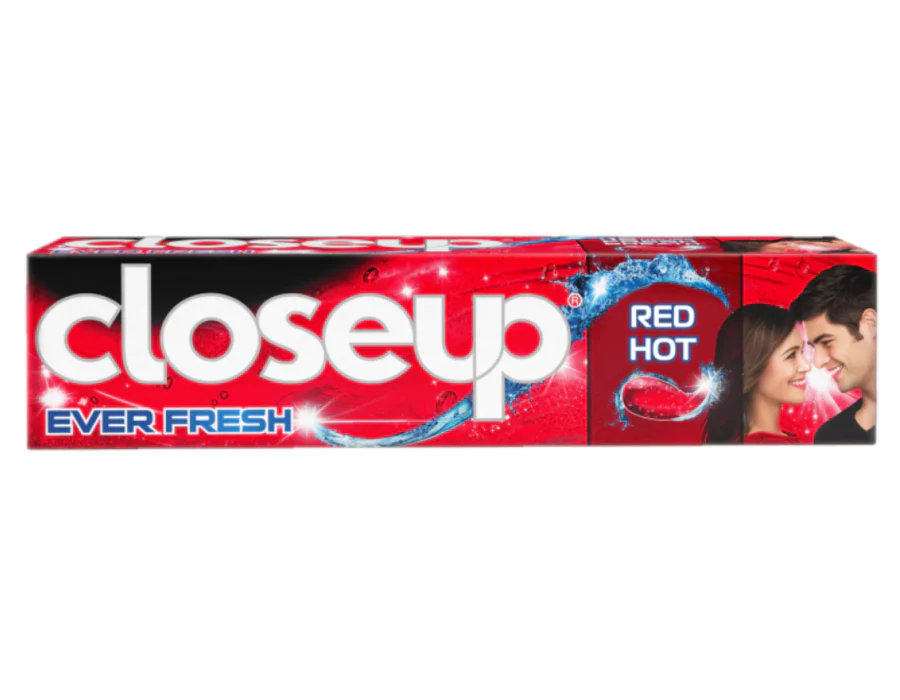} &
\includegraphics[width=0.24\linewidth]{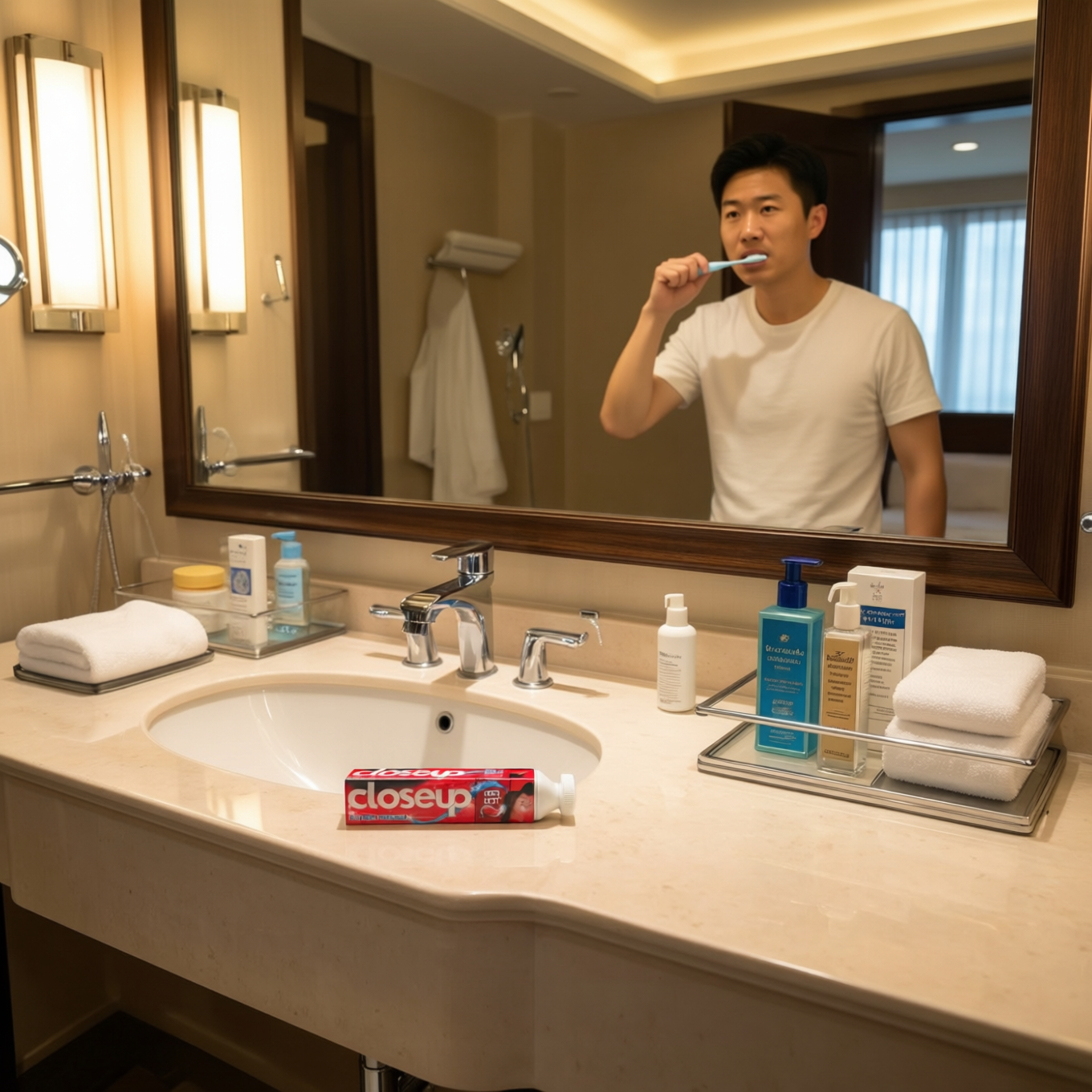} &
\includegraphics[width=0.24\linewidth]{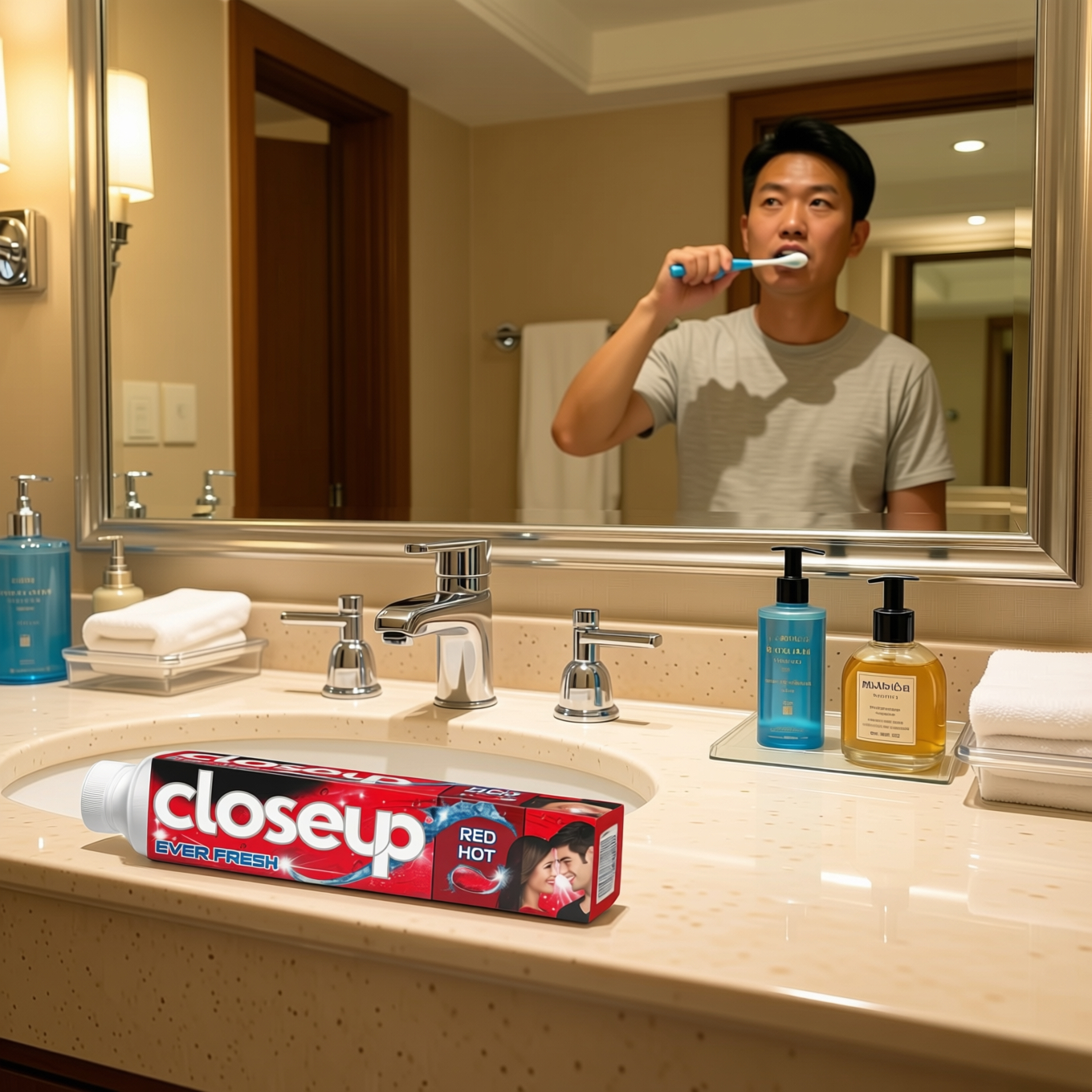} &
\includegraphics[width=0.24\linewidth]{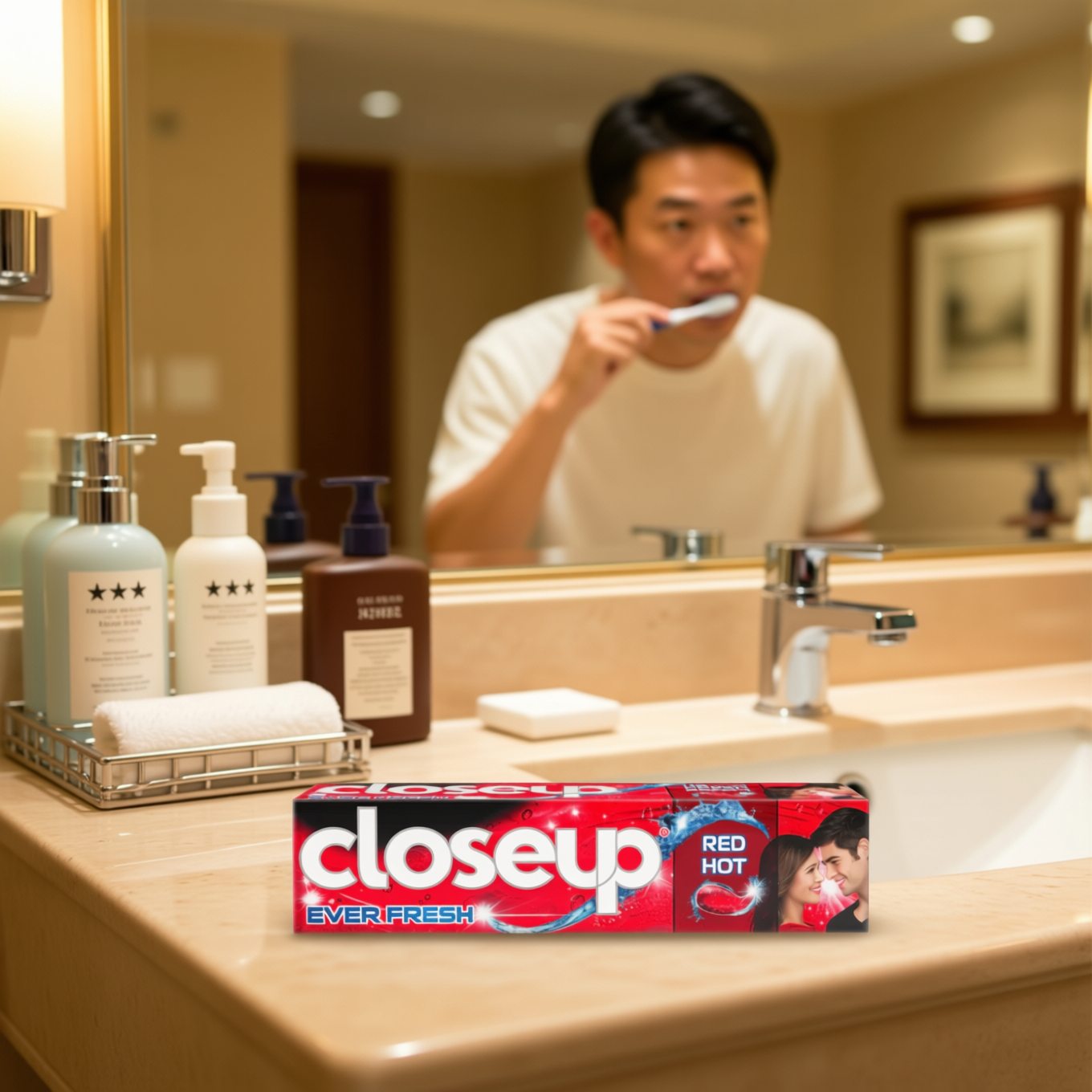} \\

\includegraphics[width=0.24\linewidth]{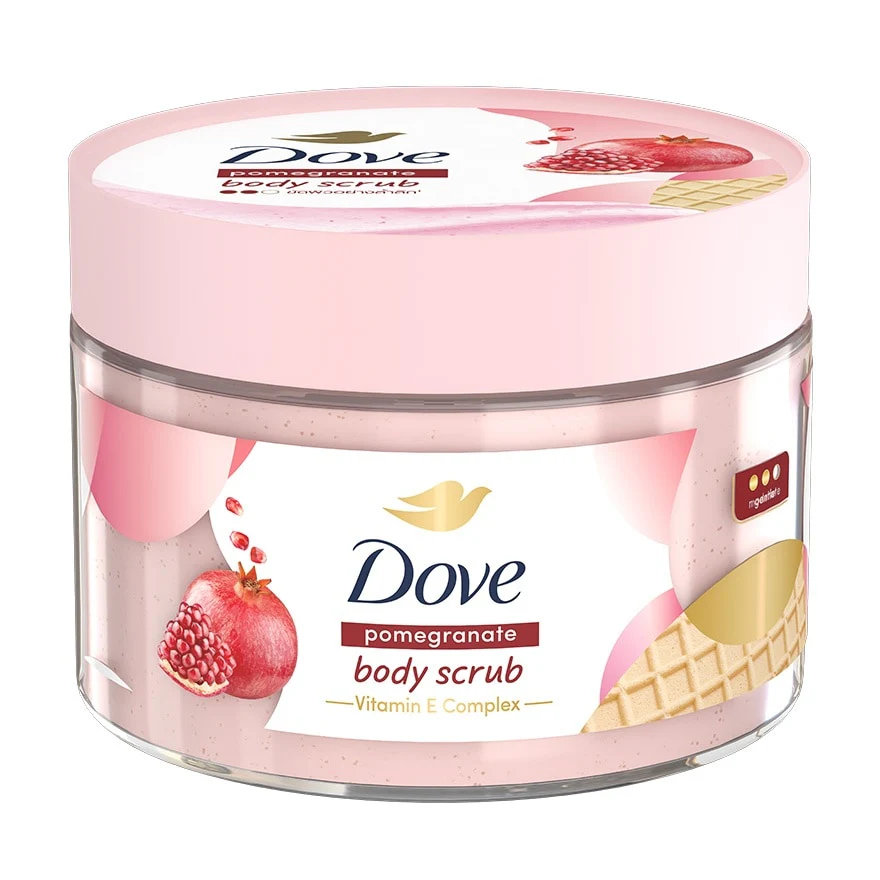} &
\includegraphics[width=0.24\linewidth]{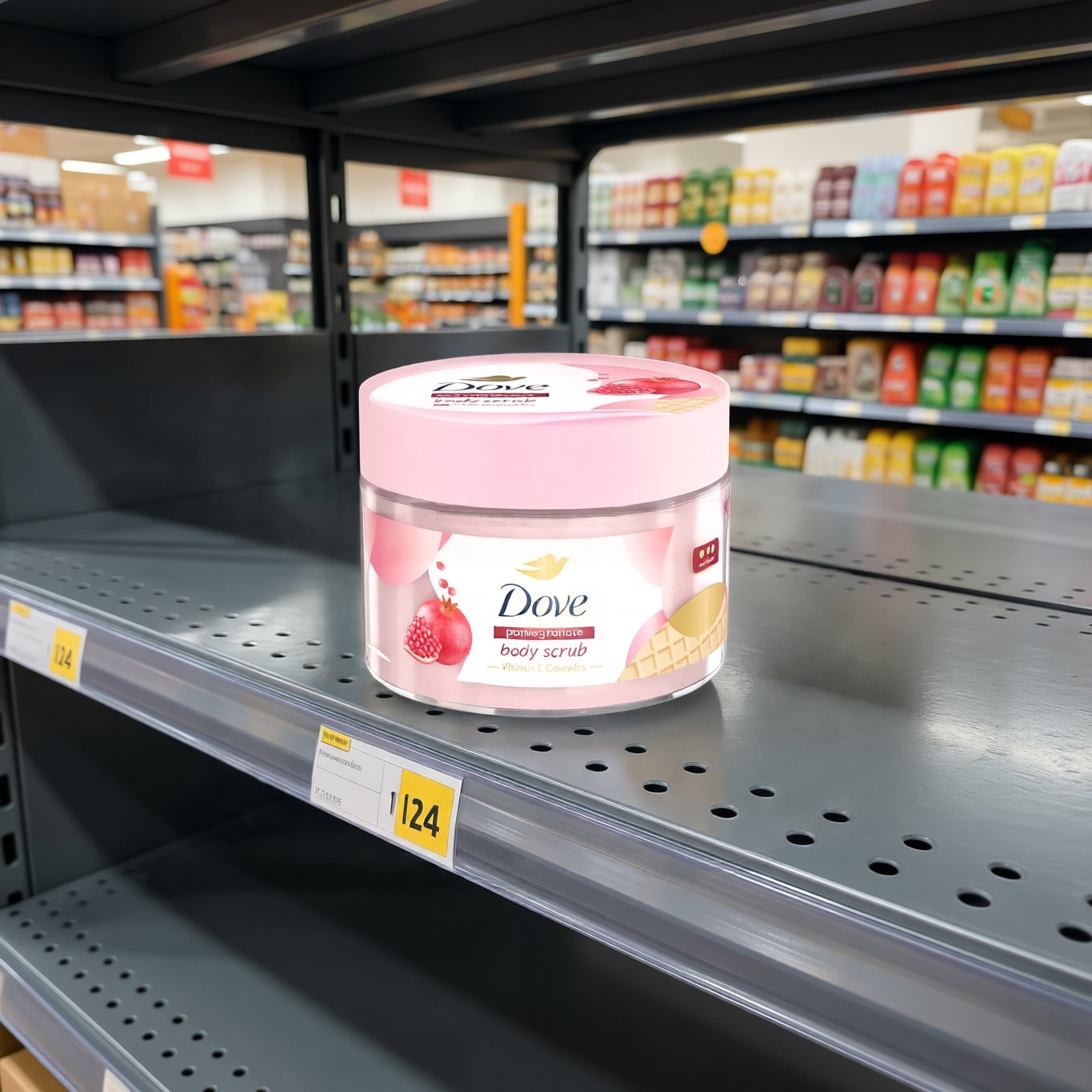} &
\includegraphics[width=0.24\linewidth]{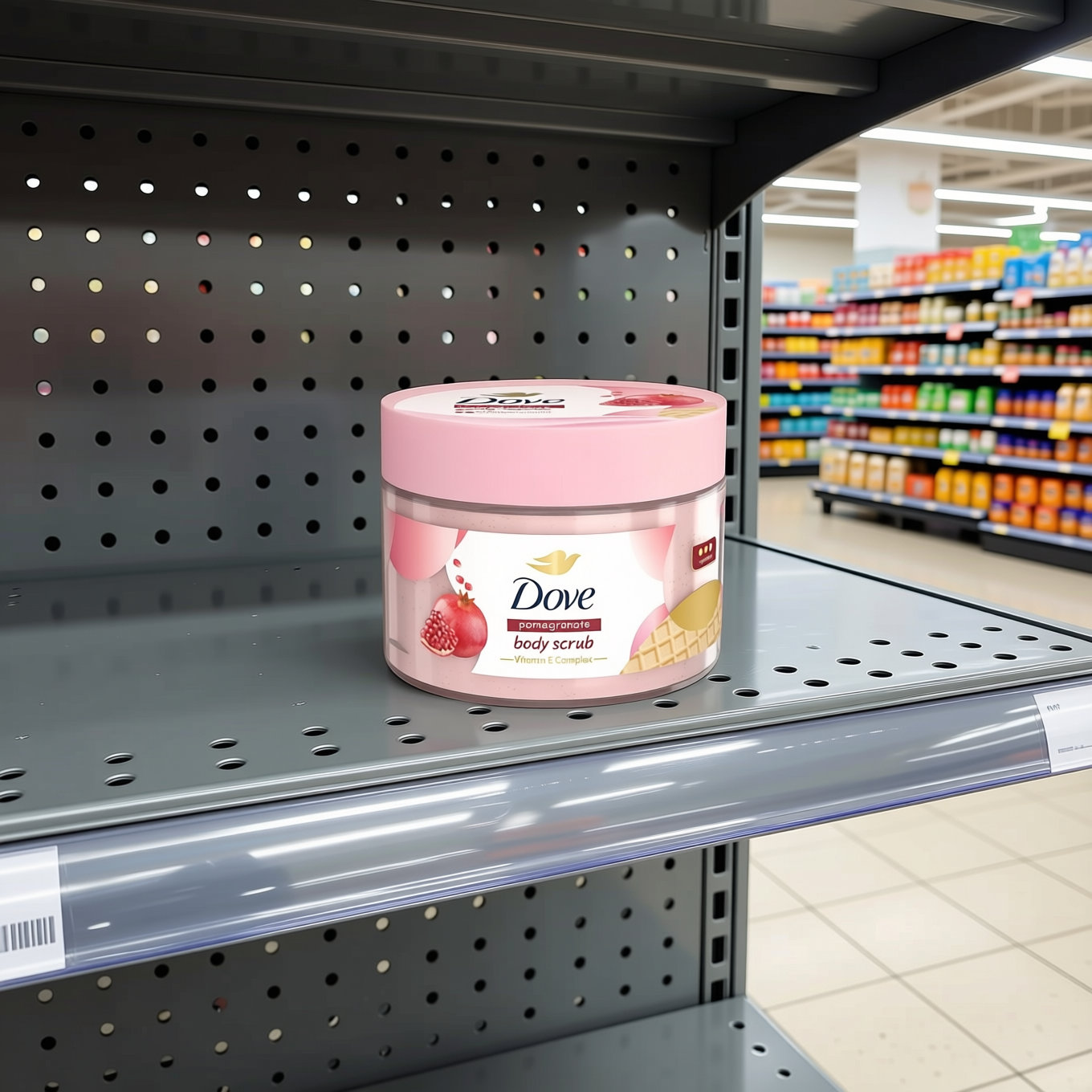} &
\includegraphics[width=0.24\linewidth]{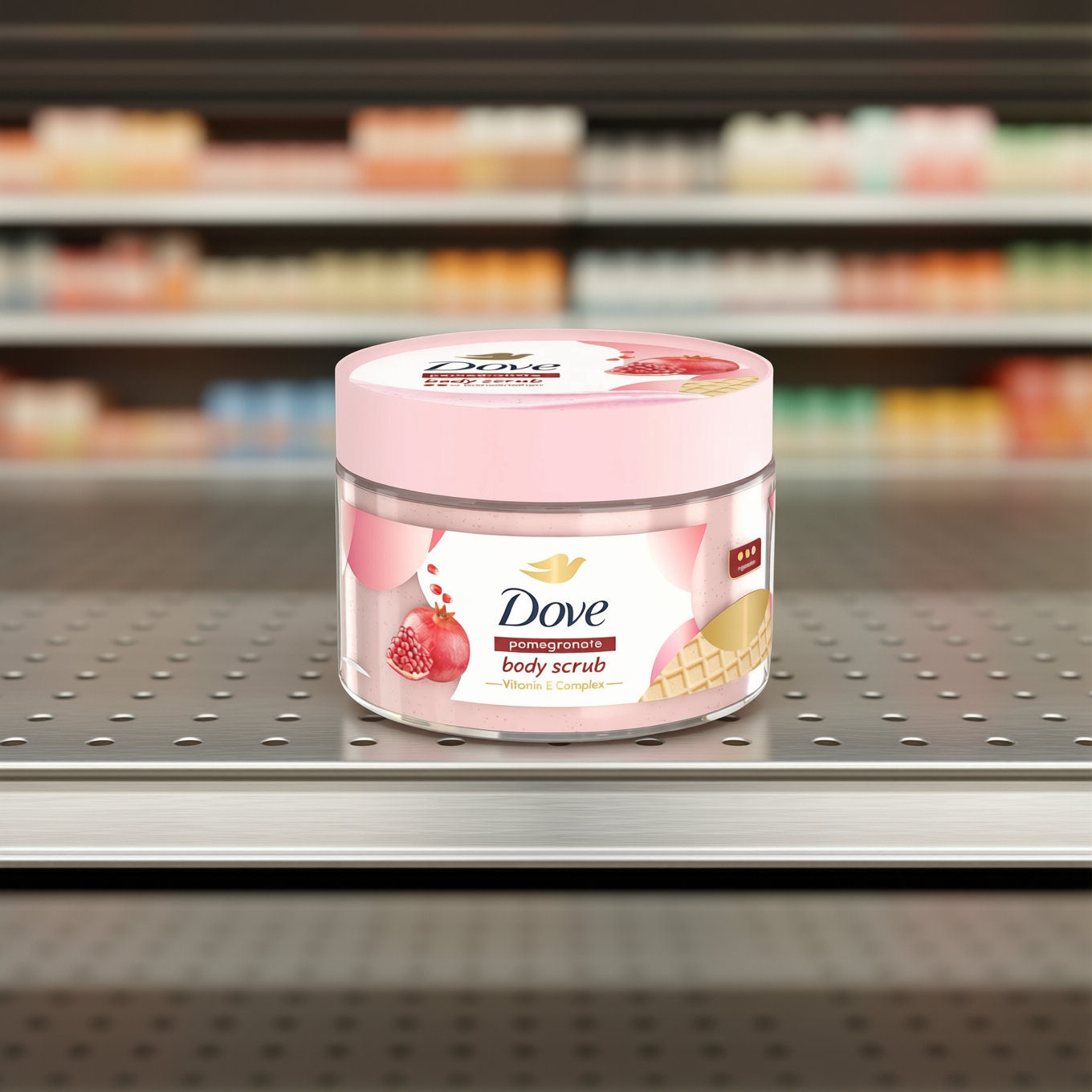} \\

\includegraphics[width=0.24\linewidth]{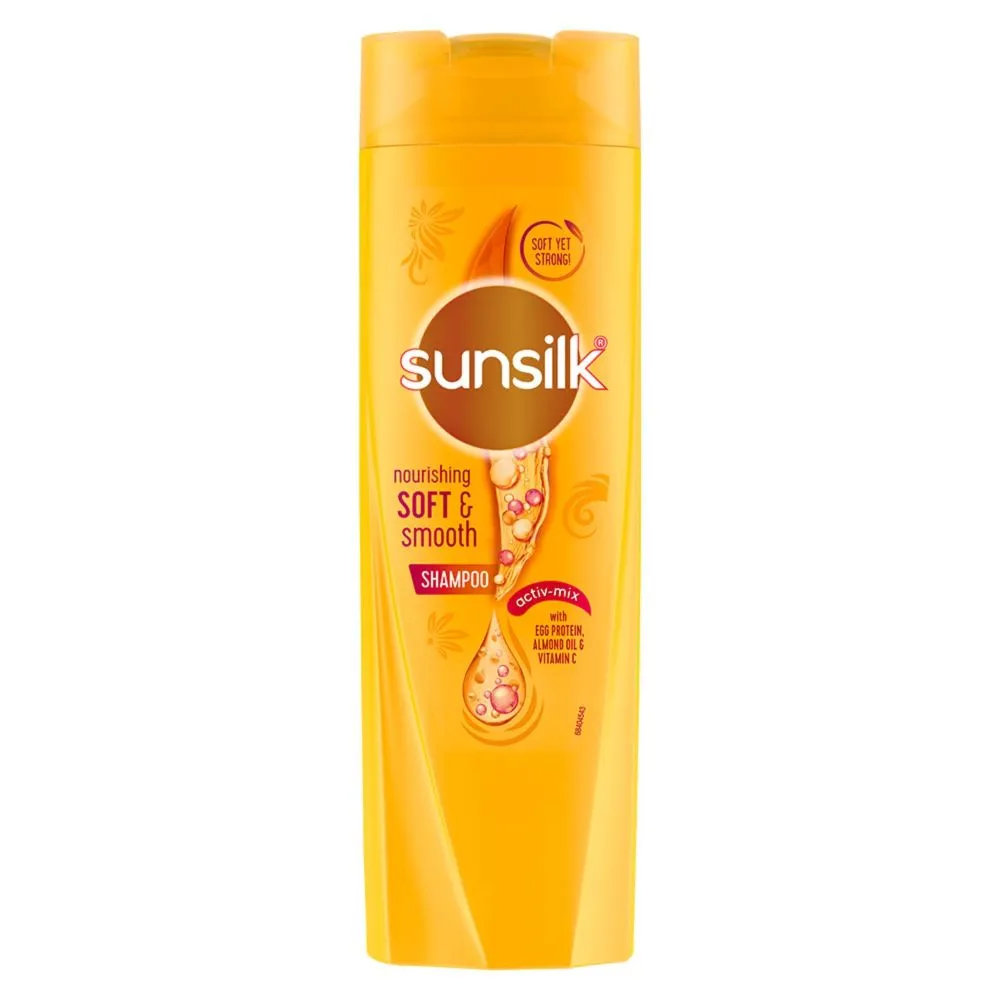} &
\includegraphics[width=0.24\linewidth]{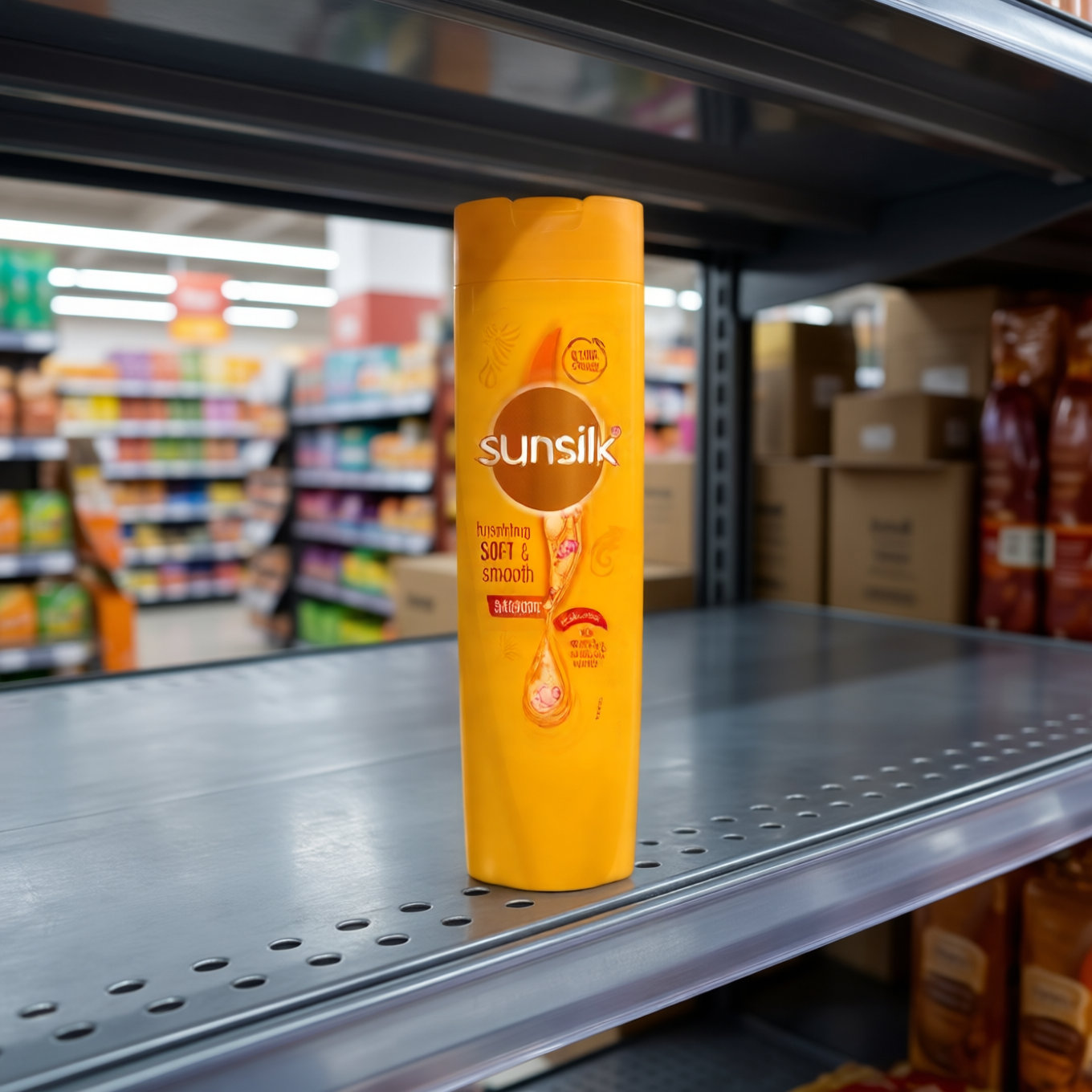} &
\includegraphics[width=0.24\linewidth]{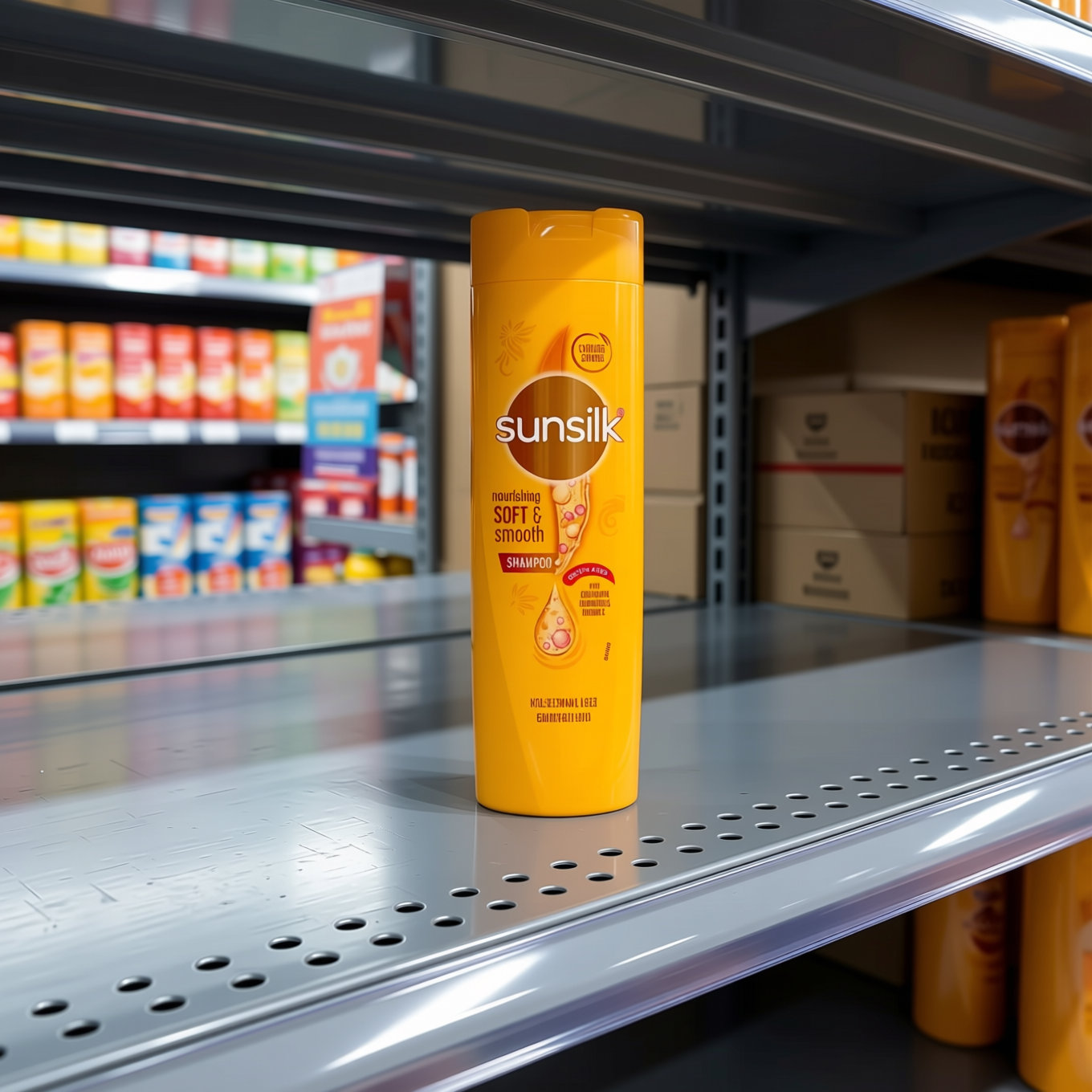} &
\includegraphics[width=0.24\linewidth]{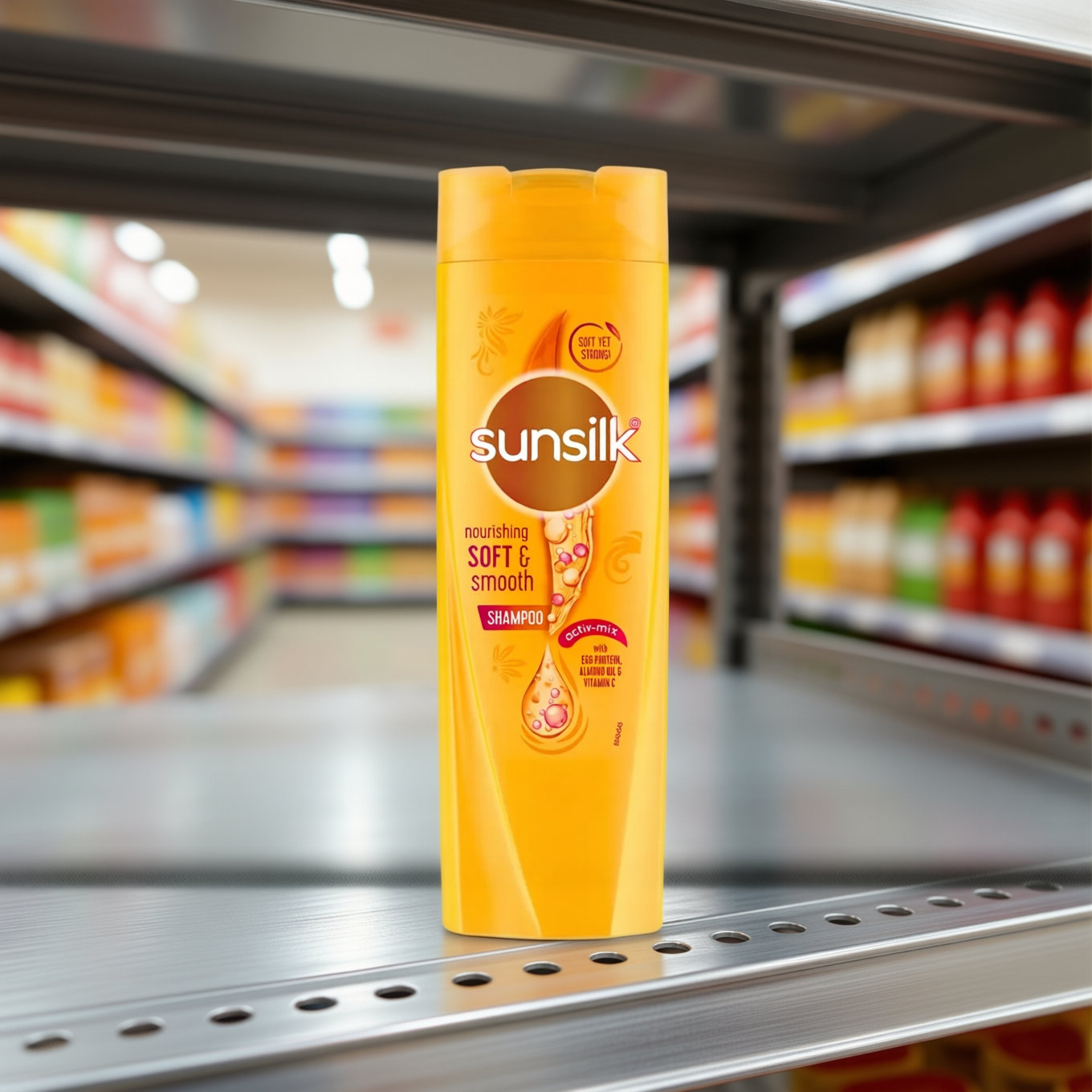} \\
\end{tabular}

\caption{Qualitative comparison on real-world products. Each row shows the input image followed by outputs from the baseline, SFT fine-tuned model, and GRPO fine-tuned Qwen-Image-Edit-2511 model with cyclic consistency reward. The edit instructions from top-to-bottom are (1) Place this toothpaste on the side of a washbasin at a 5 star hotel, it is kept with other toiletries. In the background, a Chinese man is brushing his teeth and he is looking in the mirror.(2) Place this shampoo on an empty metal shelf in a supermarket. (3) Place this product on an empty metal shelf in a supermarket.
The correct text on the products from top-to-bottom is (1) 'CLOSEUP EVER FRESH', 'RED HOT'. (2) 'Dove', 'pomergranate body scrub', 'VITAMIN E COMPLEX'. (3) 'Soft Yet Strong!', 'sunsilk', 'nourishing SOFT I\& smooth SHAMPOO', 'actic-mix with EGG PROTEIN, ALMOIND OIL I\& VITAMIN C'}
\label{fig:real_product_examples}

\end{figure*}

We further evaluate the generalizability of our approach to real-world products in Figure \ref{fig:real_product_examples}. For this experiment, we used three real product images and generated edited output using the same inference settings and fixed random seed used in our evaluation pipeline. The outputs from the baseline model are compared against those produced by the checkpoint fine-tuned on the ProductConsistency dataset using the Qwen-Image-Edit-2511 model. For all three images, the baseline performs poorly and is unable to maintain product identity and struggles with maintaining text consistency as well. The SFT model improves the rendered text but still struggles to maintain product consistency. In contrast, the GRPO model produces a visually coherent image with correct text while still following the editing instruction, demonstrating that the improvements learned during training generalize to out-of-distribution real-world samples and prompts.

For the third row, we intentionally select a challenging product image in which the text is difficult to read even for human observers. The SFT model is able to generate portions of the easier text, but still struggles with the more complex characters. The GRPO model performs better and is able to form partially coherent words even for the harder text regions. However, the output is still not perfectly accurate, indicating that, although our approach substantially improves text rendering and product consistency, difficult real-world cases remain an open challenge and provide opportunities for further improvement.

\section{Limitations and Future Work}

Although the ProductConsistency dataset and training framework significantly improve product fidelity and text preservation in instruction-based image editing, several opportunities remain for future work. First, the pipeline primarily focuses on products with straight and clearly visible text layouts. Extending the framework to support curved, stylized, or decorative text would improve robustness, as these cases remain challenging for current detection and OCR systems. Second, the dataset mainly contains front-facing product images and does not include multi-angle views of the same product instance. Future work could extend this to multi-view product datasets to enable consistent product identity across viewpoints. Finally, expanding the dataset to include additional product categories, packaging styles, and branding variations could further improve robustness and generalization.

\label{eval_settings}

\clearpage

\begin{figure*}[p]
\centering
\adjustbox{width=0.97\textwidth}{%
\begin{minipage}{0.97\textwidth}
\begin{tcolorbox}[
  colback=gray!8,
  colframe=gray!45,
  arc=4pt,
  boxrule=0.6pt,
  left=8pt, right=8pt, top=8pt, bottom=8pt,
  width=\linewidth
]
{\fontsize{5.8}{7.0}\selectfont

You are an expert AI image prompt generator specialized in premium, brand-focused professional product photography. Your task is to generate highly detailed, commercial-grade image generation prompts for a \textbf{SINGLE} product, suitable for luxury ecommerce, FMCG branding, and global marketplaces.

\smallskip
\textbf{INPUTS YOU WILL RECEIVE:}
\begin{itemize}[noitemsep, topsep=1pt, leftmargin=*]
  \item Product Category: \texttt{\{\{PRODUCT\_CATEGORY\}\}}
  \item Number of different prompts to generate: \texttt{\{\{NUMBER\_OF\_PROMPT\_PER\_CATEGORY\}\}}
  \item Number of words to be rendered on product: \texttt{\{\{NUMBER\_OF\_WORDS\}\}}
\end{itemize}

\smallskip
\textbf{OBJECTIVE:} Generate rich, highly descriptive image generation prompts that emphasize brand identity, premium positioning, material quality, and professional studio photography.

\smallskip
\textbf{Branding Details you should always include in the prompt:}

\textbf{1) Color Scheme (Primary / Secondary / Accent)}
\begin{itemize}[noitemsep, topsep=1pt, leftmargin=*]
  \item Choose a tasteful triad appropriate to \texttt{\{\{PRODUCT\_CATEGORY\}\}} and the brand's character.
  \item Primary = product body; Secondary/Accent = minimal trims/edge lines/engraving fills.
  \item Always ensure strong text-to-body contrast for readability (e.g., light text on dark product).
\end{itemize}

\textbf{2) Finish}
\begin{itemize}[noitemsep, topsep=1pt, leftmargin=*]
  \item Select a realistic finish (e.g., matte, glossy, satin, brushed, frosted, soft touch, ceramic).
\end{itemize}

\textbf{3) Contrast Level}
\begin{itemize}[noitemsep, topsep=1pt, leftmargin=*]
  \item Implicitly aim for high readability of the text on the product. This is CRITICAL.
  \item Explicitly state text color vs product body color to ensure clear read. Text color MUST NOT match product color.
\end{itemize}

\textbf{4) Logo Style, Placement, Typography Feel}
\begin{itemize}[noitemsep, topsep=1pt, leftmargin=*]
  \item Logo style: Choose a logo style that best fits the brand, and describe in detail how the logo should look in the prompt.
  \item Placement: precise (e.g., centered upper third, lid center, front and center under shoulder).
  \item Typography feel: specify (serif, sans-serif, geometric, humanist, condensed, script).
\end{itemize}

\textbf{5) Brand Archetype (to guide tone and visuals)}
\begin{itemize}[noitemsep, topsep=1pt, leftmargin=*]
  \item Infer one of: minimalist, luxury, rugged, playful, eco-conscious.
  \item Reflect it in materials, color usage, typography and finish.
\end{itemize}

\textbf{6) Text Keywords (brand benefits + product type)}
\begin{itemize}[noitemsep, topsep=1pt, leftmargin=*]
  \item Build the printed line to naturally reference brand benefits and product type.
  \item Keep it on brand with the chosen archetype and category.
\end{itemize}

\smallskip
\textbf{STRICT RULES (MANDATORY)}

\textbf{1)} Start of each prompt: Create your own random brand name in the given product category and describe brand detailing (logo, text, tagline positioning) so it feels like a real product. This is the \textbf{MOST} important part.

\textbf{2)} If \texttt{\{\{NUMBER\_OF\_PROMPT\_PER\_CATEGORY\}\}} $> 1$, each prompt must be strongly different in brand character, positioning, and visual style (archetype, typography feel, finish, color accents, and logo placement).

\textbf{3)} Background: PURE WHITE only. No shadows, no props, hands, or extra objects. The product is centered, isolated, and the ONLY subject.

\textbf{4)} Product depiction: Clearly describe material, surface finish, shape, and packaging quality.

\textbf{5)} Branding text on product:
\begin{itemize}[noitemsep, topsep=1pt, leftmargin=*]
  \item ALL TEXT AND BRAND DETAILS TO BE RENDERED MUST BE ENCLOSED IN DOUBLE QUOTES.
  \item It MUST contain EXACTLY \texttt{\{\{NUMBER\_OF\_WORDS\}\}} words (no more, no less).
  \item It MUST be described as printed/engraved on the product body/cap/lid/label according to your chosen placement and typography feel.
  \item It SHOULD reflect brand benefits and product type (see Text Keywords above).
\end{itemize}

\textbf{6)} Output format:
\begin{itemize}[noitemsep, topsep=1pt, leftmargin=*]
  \item Return ONLY the final image generation prompts in valid JSON format (keys 1..N).
  \item No commentary, no extra keys, no metadata.
\end{itemize}

\smallskip
\textbf{PROMPT CONSTRUCTION CHECKLIST}
\begin{itemize}[noitemsep, topsep=1pt, leftmargin=*]
  \item Real brand named up front + brand identity description (logo style, placement, typography feel).
  \item Color Scheme applied (Primary body; minimal Secondary/Accent).
  \item Composition: centered, pure white background, isolated subject.
  \item Printed/engraved branding text: in double quotes; EXACT word count == \texttt{\{\{NUMBER\_OF\_WORDS\}\}}; physically on product where specified; includes benefit/product type ideas aligned to archetype/category.
  \item If multiple prompts: clear differentiation across archetype, typography, finish, colors, and placement.
\end{itemize}

\smallskip
\textbf{FINAL OUTPUT FORMAT (MANDATORY):} \texttt{\{1: "prompt", 2: "prompt", ...\}}

\smallskip
\textbf{Few Shot Examples:}

\smallskip
\textit{Example 1 --- Input:} Product Category: Luxury perfume bottle $|$ Prompts: 1 $|$ Words: 6

\smallskip
\colorbox{gray!15}{%
\begin{minipage}{\dimexpr0.975\linewidth-2\fboxsep\relax}
{\fontsize{5.8}{7.0}\selectfont\raggedright\sloppy
\hangindent=0pt
\tolerance=9999
\emergencystretch=\maxdimen
\hbadness=10000
\noindent
\{``1'': ``A AWESOME perfume bottle inspired by the timeless elegance
of AWESOME's iconic fragrance line is photographed on a pure white
background. The brand identity is expressed with a minimalist wordmark
and black on clear label, centered on the front panel; typography is a
sans with couture restraint. The primary form is a crystal clear glass
body with polished edges and a glossy finish; a secondary accent of
muted gold appears on the collar, used sparingly for luxury emphasis.
To ensure maximum readability, the printed label text is deep black
against the transparent body with an opaque white underlay. The
engraved label reads `AWESOME Couture Essence Timeless Luxury
Perfume'.''\}
}
\end{minipage}%
}

\smallskip
\textit{Example 2 --- Input:} Product Category: Insulated stainless steel water bottle $|$ Prompts: 2 $|$ Words: 6

\smallskip
\colorbox{gray!15}{%
\begin{minipage}{\dimexpr0.975\linewidth-2\fboxsep\relax}
{\fontsize{5.8}{7.0}\selectfont\raggedright\sloppy
\hangindent=0pt
\tolerance=9999
\emergencystretch=\maxdimen
\hbadness=10000
\noindent
\{``1'': ``A HYDRO FLASK insulated stainless steel bottle, presented
as an eco conscious hero product on a pristine pure white infinity
cove background, centered and fully isolated. Color scheme: primary
deep forest green body in matte powder coat; secondary warm gray trim
at collar; accent lime used sparingly near the cap seam. Logo style
is a clean wordmark; typography feels humanist, slightly condensed;
placement is the centered upper third. White silkscreen ink ensures
crisp legibility on the dark body. The printed branding reads `Hydro
Flask Sustainable Hydration Free Insulated'.'',
``2'': ``A STANLEY adventure grade insulated bottle emphasizing rugged
reliability, photographed on a pure white seamless backdrop, centered
and isolated. Color scheme: primary charcoal gray textured matte body;
secondary muted cool gray; accent blaze orange for a slim ring and
measurement mark. Logo style combines a bold wordmark with compact
emblem; typography is condensed geometric; placement is front and
center below the shoulder. Blaze orange silkscreen text on dark body.
The printed text reads `Stanley Adventure Tough Insulation Day
Cold'.''\}
}
\end{minipage}%
}

}
\end{tcolorbox}
\end{minipage}
}
\caption{The system prompt is designed to generate product image prompts on a pure white background. The model takes as input the product category, the number of prompts to be generated, and the desired word count. It then outputs structured JSON containing fully specified, brand‑consistent image‑generation instructions, including details such as color scheme, material finish, typography, logo placement, and associated branding text.}
\label{fig:system_prompt_prompt_generation}
\end{figure*}

\clearpage

\begin{figure*}[p]
\centering
\resizebox{0.97\textwidth}{!}{%
\begin{minipage}{0.97\textwidth}
\begin{tcolorbox}[
  colback=gray!8,
  colframe=gray!45,
  arc=4pt,
  boxrule=0.6pt,
  left=8pt, right=8pt, top=8pt, bottom=8pt,
  width=\linewidth
]
{\fontsize{5.8}{7.0}\selectfont

You are a strict, objective evaluation model for text-guided image editing. You must behave like a professional human evaluator performing structured visual quality assessment.

\smallskip
\textbf{Your task is to evaluate an edited image relative to:}
\begin{itemize}[noitemsep, topsep=1pt, leftmargin=*]
  \item \texttt{input\_image\_description} --- text description of original image
  \item \texttt{input\_image} --- original image
  \item \texttt{edit\_instruction} --- text instruction describing intended edit
  \item \texttt{edited\_image} --- result image after editing
\end{itemize}

\smallskip
\textbf{EVALUATION PROCESS (INTERNAL REASONING -- DO NOT OUTPUT)}
\begin{enumerate}[noitemsep, topsep=1pt, leftmargin=*]
  \item Carefully analyze the \texttt{input\_image} and understand the product, structure, text, layout, and composition.
  \item Read \texttt{input\_image\_description} to confirm product identity and expected details.
  \item Compare \texttt{input\_image} and \texttt{edited\_image} carefully.
  \item Evaluate each metric thoughtfully, paying attention to anything that might impact the score.
  \item Be strict. Do NOT inflate scores.
\end{enumerate}
ONLY output final JSON.

\smallskip
\textbf{METRICS (Score each 0--10 integer)}

\smallskip
\textbf{1) Product Consistency}\\
The product in the edited image must be the SAME product as in the input image. Shape, proportions, structural features, geometry, and defining characteristics must remain unchanged.

\textit{Failures include:}
\begin{itemize}[noitemsep, topsep=1pt, leftmargin=*]
  \item Shape distortion (even subtle warping or stretching)
  \item Missing components
  \item Identity change of product
  \item Brand identity changes (logo, brand colors, typography, layout)
  \item Logo style, logo placement, or logo proportions changed
\end{itemize}

\smallskip
\textbf{2) Text Rendering / Text Fidelity}\\
Any text originally visible on the product (brand name, label, instructions, numbers, logo text) must remain legible and sharp, correctly spelled with no character substitutions, and unchanged in content, font style, and positioning.

\textit{Failures include:}
\begin{itemize}[noitemsep, topsep=1pt, leftmargin=*]
  \item Misspellings or altered characters; altered or missing words
  \item Missing text lines (even small/secondary text)
  \item Blurry or illegible text (even partially); hallucinated new text
  \item Font style, size, or color changes
\end{itemize}

\textit{If no visible text exists in the input image, return:} \texttt{score = 0}, \texttt{reason = "no visible product text in input image"}

\smallskip
\textbf{3) Aesthetics / Composition}\\
Overall visual appeal, composition quality, and alignment with the edit instruction's intent. The edited image must appear photorealistic and visually coherent.

\textit{Check for:}
\begin{itemize}[noitemsep, topsep=1pt, leftmargin=*]
  \item Proper centering or intentional framing; balanced negative space
  \item Product prominence and clear visual focus
  \item Color temperature consistency; pleasant, consistent lighting (no harsh hotspots or dull areas)
  \item AI artifacts (duplicated objects, warped geometry, impossible physics)
  \item Overall aesthetic appeal matching professional product photography standards
\end{itemize}

\smallskip
\textbf{CRITICAL SCORING PHILOSOPHY}
\begin{itemize}[noitemsep, topsep=1pt, leftmargin=*]
  \item \textbf{10} = EXCEPTIONAL. Perfect only. Any imperfection disqualifies a 10.
  \item \textbf{9} = Rare. Near-flawless; struggle to find any issue.
  \item \textbf{7} = GOOD --- baseline for a competent edit with only minor issues. Most decent edits land 6--8.
  \item \textbf{5} = AVERAGE --- passable with clear room for improvement.
  \item Start from 7 and \textbf{deduct} points for each flaw. Only go above 7 if no meaningful issues found.
\end{itemize}

\smallskip
\textbf{SCORING RUBRIC (STRICT)}

\smallskip
\colorbox{gray!15}{%
\begin{minipage}{\dimexpr0.975\linewidth-2\fboxsep\relax}
{\fontsize{5.8}{7.0}\selectfont\selectfont\raggedright\sloppy
\hangindent=0pt
\tolerance=9999
\emergencystretch=\maxdimen
\hbadness=10000
\noindent
10 = Perfect. No flaws at all. Looks like a perfect professional edit.\\
9\phantom{0} = Near-perfect. Just one very small issue noticed.\\
8\phantom{0} = Very good. 1--2 minor issues that would easily be overlooked.\\
7\phantom{0} = Good. A few minor but identifiable issues. Solid, competent edit.\\
6\phantom{0} = Above average. Some noticeable issues but overall acceptable quality.\\
5\phantom{0} = Average. Multiple noticeable problems but the edit is recognizable.\\
4\phantom{0} = Below average. Several clear problems that detract from quality.\\
3\phantom{0} = Poor. Major issues dominate the result.\\
2\phantom{0} = Very poor. Barely recognizable as the intended edit.\\
1\phantom{0} = Severely defective. Almost nothing is correct.\\
0\phantom{0} = Complete failure.
}
\end{minipage}%
}

\smallskip
\textbf{CALIBRATION NOTES:} A score of 7 means ``good'' --- baseline for a decent edit. If giving 9 or 10, you MUST justify by confirming no meaningful mistakes were found.

\smallskip
\textbf{OUTPUT FORMAT (STRICT JSON ONLY)}

\smallskip
\colorbox{gray!15}{%
\begin{minipage}{\dimexpr0.975\linewidth-2\fboxsep\relax}
{\fontsize{5.8}{7.0}\selectfont\selectfont\raggedright\sloppy
\hangindent=0pt
\tolerance=9999
\emergencystretch=\maxdimen
\hbadness=10000
\noindent
\{\\
\hspace*{8pt}``product\_consistency'': \{``reason'': string, ``score'': int\},\\
\hspace*{8pt}``text\_rendering'': \{``reason'': string, ``score'': int\},\\
\hspace*{8pt}``aesthetics'': \{``reason'': string, ``score'': int\}\\
\}
}
\end{minipage}%
}

\smallskip
The \texttt{"reason"} must be concise (max 2 sentences) and must cite at least one specific flaw found, or explicitly state why no flaws were found if scoring 9--10. Only output valid JSON.

}
\end{tcolorbox}
\end{minipage}
}
\caption{The system prompt is used within the evaluation pipeline. The model takes as input the original image, its textual description, the edit instruction, and the edited image, and outputs a structured JSON object containing the reasoning process and scores across three evaluation metrics: product consistency, text fidelity, and aesthetics.}
\label{fig:eval_system_prompt}
\end{figure*}

\begin{table*}[t]
\centering
\footnotesize
\resizebox{\textwidth}{!}{
\begin{tabular}{l p{0.78\textwidth}}
\toprule
Model & Inference Settings \\
\midrule

Flux.1-Kontext-dev & 28 steps; guidance scale = 2.5 \\

Qwen-Image-Edit-2511 & 40 steps; true CFG = 4.0; guidance scale = 1.0 \\

Hidream-E1-1 & 28 steps; guidance = 3.0; image guidance = 1.5; refine strength = 0.3 \\

BAGEL & 50 steps; CFG\_TEXT = 4.0; CFG\_IMG = 2.0; CFG\_INTERVAL = 0.0; TIMESTEP\_SHIFT = 3.0 \\

Step1x-edit-v1p2 & 28 steps; guidance scale = 6.0 \\

Edit-R1-Flux & 28 steps; guidance scale = 2.5; LoRA adapter weight = 1.0 \\

Edit-R1-Qwen & 40 steps; true CFG = 4.0; guidance scale = 1.0; LoRA adapter weight = 1.0 \\

GPT-Image-1 & Quality = high; Size = 1024$\times$1024 \\

Nano Banana & - \\

OmniGen2 & 50 steps; text guidance scale = 5.0; image guidance scale = 2.0; cfg\_range = (0.0, 1.0) \\

RePlan-Flux & expand\_value = 0.15; attention\_switch\_step = 0.05; flex attention = True \\

RePlan-Qwen & expand\_value = 0.0; attention\_switch\_step = 0.5; flex attention = True \\

\bottomrule
\end{tabular}
}
\caption{Inference settings used for baseline image editing models.}
\label{tab:generation_settings}
\end{table*}

\end{document}